%% file: main_arxiv.tex
\documentclass{article}

\usepackage[accepted]{include/aistats/aistats2025_mod}

\usepackage[utf8]{inputenc} %
\usepackage[T1]{fontenc}    %
\usepackage{hyperref}       %
\usepackage{url}            %
\usepackage{booktabs}       %
\usepackage{amsfonts}       %
\usepackage{nicefrac}       %
\usepackage{microtype}      %
\usepackage{xcolor}         %
\usepackage{bbm}
\usepackage{subcaption}
\usepackage{caption}
\usepackage{comment}
\usepackage{graphicx}

\usepackage{include/titletoc}
\usepackage{amsmath}
\usepackage{amssymb}
\usepackage{natbib}
\usepackage{cuted}

\usepackage{multirow}
\usepackage{array}
\usepackage{multicol}
\usepackage{tabularx}
\usepackage{tabularray}
\UseTblrLibrary{booktabs}
\usepackage[inline]{enumitem}

\include{include/custom}

\include{include/aistats/custom_aistats2025}

\usepackage[nameinlink]{cleveref} %
\crefname{appsec}{appendix}{appendices}
\Crefname{appsec}{Appendix}{Appendices}

\usepackage{xcolor}

\definecolor{mydarkblue}{rgb}{0,0.08,0.45}
\definecolor{ourmethodblue}{rgb}{0,0.447,0.698}
\definecolor{ourmethodlblue}{rgb}{0.651, 0.808, 0.890}
\definecolor{evenlighterblue}{RGB}{213,241,255}

\hypersetup{ %
    pdftitle={},
    pdfauthor={},
    pdfsubject={},
    pdfkeywords={},
    pdfborder=0 0 0,
    pdfpagemode=UseNone,
    colorlinks=true,
    linkcolor=mydarkblue,
    citecolor=mydarkblue,
    filecolor=mydarkblue,
    urlcolor=mydarkblue,
    pdfview=FitH
}

\usepackage{pifont}%
\newcommand{\cmark}{\ding{51}}%
\newcommand{\xmark}{\ding{55}}%

\newcommand{\TODO}[1]{\textcolor{red}{#1}}

\begin{document}

\runningtitle{A Monte Carlo Framework for Calibrated Uncertainty Estimation in Sequence Prediction}

\twocolumn[

\aistatstitle{
A Monte Carlo Framework for\\Calibrated Uncertainty Estimation in Sequence Prediction
}

\vspace*{-10pt}
\centering
\textbf{Qidong Yang}$^{\ast\,\text{1,2}}$
\; \textbf{Weicheng Zhu}$^{\ast\,\text{2}}$
\; \textbf{Joseph Keslin} $^{\text{3}}$
\; \textbf{Laure Zanna} $^{\text{2}}$
\\
\centering
\textbf{Tim G. J. Rudner} $^{\text{2}}$
\; \textbf{Carlos Fernandez-Granda} $^{\text{2}}$

\vspace{6pt}

\centering
$^{\text{1}}$ MIT 
\; $^{\text{2}}$ New York University
\; $^{\text{3}}$ University of Illinois Urbana-Champaign

\vspace*{20pt}

]

\input{drafts/_aistats2025/main_text}
\input{drafts/_aistats2025/acknowledgements}

\bibliography{references}
\bibliographystyle{plainnat}

\clearpage

\input{drafts/_aistats2025/appendices}

\end{document}

%% file: include/custom.tex
\usepackage[utf8]{inputenc}
\usepackage{microtype}
\usepackage{graphicx}
\usepackage{booktabs} %
\usepackage{titletoc}
\usepackage{hyperref}

\usepackage{amsmath}
\usepackage{amssymb}
\usepackage{bbm} 
\usepackage{bm}
\usepackage{verbatim}
\usepackage{float}
\usepackage{color,soul}
\usepackage{enumitem}
\usepackage{mathtools}
\usepackage{hhline}
\usepackage[title]{appendix}
\usepackage{tabularx}
\usepackage{tikz}
\usepackage{wrapfig,booktabs}
\usepackage{xspace}
\usepackage{cancel}
\usepackage{sidecap}

\graphicspath{ {../figures/} }

\DeclarePairedDelimiterX{\infdivx}[2]{(}{)}{%
  #1\;\delimsize\|\;#2%
}

\definecolor{mydarkblue}{rgb}{0,0.08,0.45}
\hypersetup{ %
    pdftitle={},
    pdfauthor={},
    pdfsubject={Proceedings of the International Conference on Machine Learning 2020},
    pdfkeywords={},
    pdfborder=0 0 0,
    pdfpagemode=UseNone,
    colorlinks=true,
    linkcolor=mydarkblue,
    citecolor=mydarkblue,
    filecolor=mydarkblue,
    urlcolor=mydarkblue,
    pdfview=FitH
}

\usetikzlibrary{shapes.geometric, arrows, bayesnet, calc, positioning}

\newcommand{\calL}{\mathcal{L}}

\newcommand{\R}{\mathbb{R}}

\newcommand{\closer}[3]{{\kern-#1ex{#2}\kern-#3ex}}

\mathchardef\mhyphen="2D

%% file: include/aistats/custom_aistats2025.tex
\usepackage{titletoc}
\usepackage{algorithm}
\usepackage{algorithmic}
\usepackage{colortbl}

\definecolor{azure}{rgb}{0.0, 0.5, 1.0}
\definecolor{airforceblue}{rgb}{0.36, 0.54, 0.66}
\definecolor{darkgreen}{rgb}{0.0, 0.2, 0.13}

\newcommand\defines{\,\dot{=}\,}

\usepackage{standalone}
\usepackage{tikz}
\usepackage{pgfplots}
\usepackage{pgf}
\usetikzlibrary{calc}
\usetikzlibrary{positioning}
\usetikzlibrary{angles,quotes}
\usetikzlibrary{backgrounds}
\usetikzlibrary{fit}
\usetikzlibrary{arrows}
\usetikzlibrary{arrows.meta}
\usetikzlibrary{shapes.symbols}
\usetikzlibrary{shadings}
\usetikzlibrary{shapes}
\usetikzlibrary{fadings}
\usetikzlibrary{bayesnet}
\usetikzlibrary{matrix}
\usetikzlibrary{plotmarks}
\usetikzlibrary{intersections}
\usetikzlibrary{pgfplots.fillbetween}
\pgfplotsset{compat=1.14}
\usepackage{siunitx}

\usepackage{colortbl}
\definecolor{mediumgray}{gray}{0.7}
\definecolor{lightgray}{gray}{0.76}
\definecolor{lightlightgray}{gray}{0.9}
\definecolor{C1}{HTML}{1F77B4}
\definecolor{C2}{HTML}{FF7F0E}
\definecolor{C3}{HTML}{2CA02C}
\definecolor{C4}{HTML}{D62728}
\definecolor{C5}{HTML}{9467BD}
\colorlet{C1light}{C1!70!white}
\colorlet{C2light}{C2!70!white}
\colorlet{C3light}{C3!70!white}
\colorlet{C4light}{C4!70!white}
\colorlet{C5light}{C5!70!white}
\colorlet{C1lighter}{C1!50!white}
\colorlet{C2lighter}{C2!50!white}
\colorlet{C3lighter}{C3!50!white}
\colorlet{C4lighter}{C4!50!white}
\colorlet{C5lighter}{C5!50!white}
\colorlet{C1vlight}{C1!20!white}
\colorlet{C2vlight}{C2!20!white}
\colorlet{C3vlight}{C3!20!white}
\colorlet{C4vlight}{C4!20!white}
\colorlet{C5vlight}{C5!20!white}
\colorlet{linkcolor}{violet}

\usepackage{tcolorbox}

%% file: drafts/_aistats2025/main_text.tex
\begin{abstract}
Probabilistic prediction of sequences from images and other high-dimensional data remains a key challenge, particularly in safety-critical domains. In these settings, it is often desirable to quantify the uncertainty associated with the prediction (instead of just determining the most likely sequence, as in language modeling). In this paper, we propose a Monte Carlo framework to estimate probabilities and confidence intervals associated with sequences. Our framework uses a Monte Carlo simulator, implemented as an autoregressively trained neural network, to sample sequences conditioned on an image input. We then use these samples to estimate  probabilities and confidence intervals. Experiments on synthetic and real data show that the framework produces accurate discriminative predictions, but can suffer from miscalibration. To address this shortcoming, we propose a time-dependent regularization method, which produces calibrated predictions. 
\end{abstract}

\begin{figure}[t!]
    \vspace*{-8pt}
    \centering
    \includegraphics[trim={0.6cm 0cm 0.6cm 0cm}, clip, width=\linewidth]{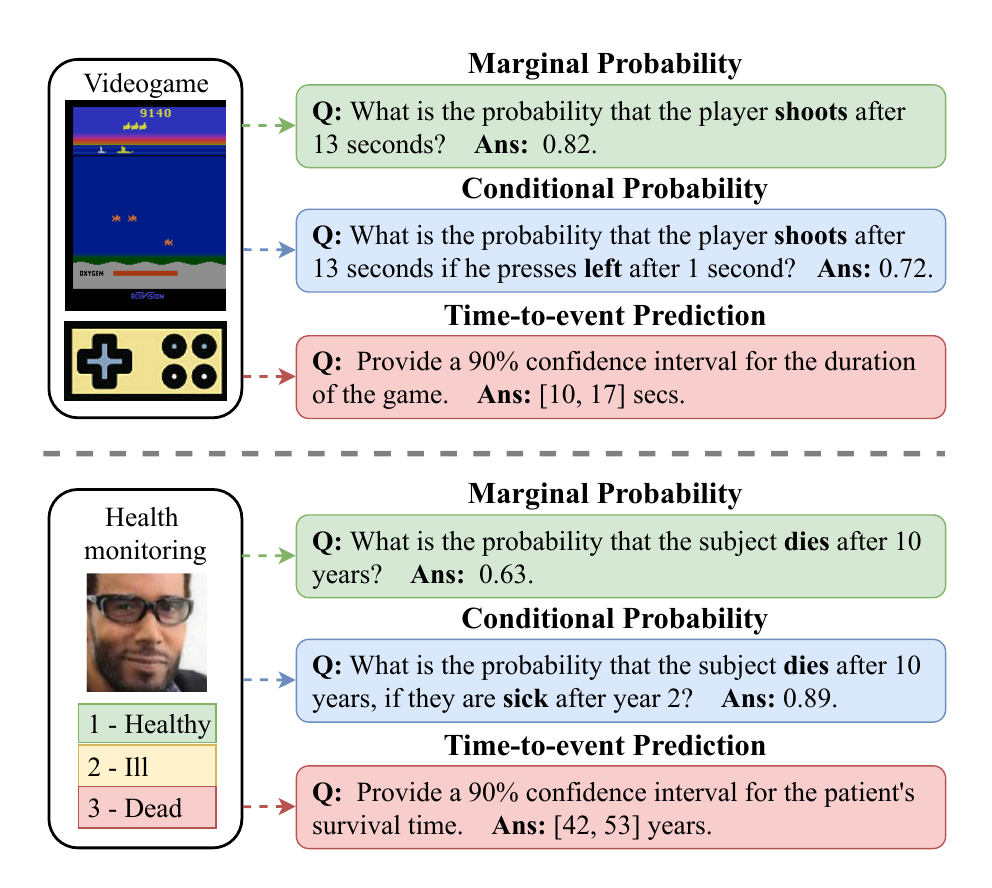}
    \vspace*{-16pt}
    \caption{\textbf{Sequence prediction with uncertainty estimation.} The proposed framework enables estimation of marginal  probabilities, conditional  probabilities, and time-to-event confidence intervals associated with a sequence given an input image. We consider sequential decision-making tasks where the input image is a screenshot from an Atari video game and the sequence to predict is the sequence of actions taken by a user (top), and we consider a synthetic-data forecasting problem where the input image is the face of a person and the sequence to predict is the evolution of their health status (bottom).}
    \label{fig:problem}
    \vspace*{-10pt}
\end{figure} 

\section{\uppercase{Introduction}}

We consider the problem of predicting a sequence of multi-class labels from high-dimensional input data, such as images. Potential applications include patient prognostics from medical imaging data~\citep{pham2017predicting}, weather forecasting~\citep{andrychowicz2023deeplearningdayforecasts}, and modeling of human behavior~\citep{thakkar2024adaptivehumantrajectoryprediction}. Our focus is on {\it probabilistic} prediction, which requires uncertainty quantification in the form of probabilities or confidence intervals.

\Cref{fig:problem} shows two examples of sequence prediction problems: modeling player behavior in a video game and health monitoring. The data consist of ``starting state'' images and subsequent sequences of player actions (e.g., {\it move left}, {\it move right}, {\it shoot}, etc.) and patients' health statuses ({\it healthy}, {\it ill} or {\it dead}), respectively.
The goal is to predict the sequence of actions/health statuses from the starting state image. 

There has been a large volume of work on estimating the {\it most likely} sequence from high-dimensional data, for example, in language modeling~\citep{you2016image,herdade2019image,li2022trocrtransformerbasedopticalcharacter,NEURIPS2020_1457c0d6}.
However, predicting a most likely sequence is {\it not sufficient} when the evolution of the sequence is uncertain. 
In this case, multiple states are possible at a future time given a fixed input, and therefore a single deterministic estimate is not acceptable. Instead, we need to predict the {\it probability} that each state occurs at a given point in the future.
Similarly, in prediction of time-to-event values (e.g., the time until the game ends, or the survival time of a subject), predicting the most likely value is also insufficient, as there is a range of possible times given a specific input.
Instead, we wish to produce  confidence intervals that are likely to contain the time of interest with high probability.

Predicting the probability of the different possible states of a sequence at a fixed future time and survival modeling are standard classification and probability-estimation problems~\citep{liu2022deep}.
However, existing approaches require separate models for each time and type of event. 
In this paper, we propose a framework that uses a single model to generate probabilities and confidence intervals associated with any possible future time and any type of event. 

The proposed {\bf f}ramework for M{\bf o}nte {\bf C}arlo {\bf u}ncertainty quantification of {\bf s}equences ({\bf foCus}) combines autoregressive models---which are the state of the art for many sequence-based tasks~\citep{bahdanau2014neural, gehring2017convolutional, vaswani2017attention,NEURIPS2020_1457c0d6}, with Monte Carlo estimation to produce useful uncertainty estimates for sequence prediction tasks.
However, when studying this framework, we find that autoregressive simulators trained via maximum likelihood estimation are severely miscalibrated, meaning that the associated probabilities or confidence intervals do not provide accurate uncertainty quantification for the tasks at hand.
To address this shortcoming, we develop a time-dependent regularizer for autoregressive simulators training and show that it enables foCus to generate better calibrated probability estimates.

To summarize, our main contributions are as follows:\vspace*{-10pt}
\begin{enumerate}[leftmargin=15pt]
\setlength\itemsep{0pt}
    \item
    We propose a Monte Carlo framework for probabilistic prediction of sequences from high-dimensional input data using autoregressive models.
    \item
    We perform an empirical evaluation of the proposed framework on a hand-tailored synthetic benchmarking task for sequence prediction and on non-synthetic sequential decision-making tasks and find that neural network-based autoregressive simulators are prone to time-dependent miscalibration.
    \item
    We develop a time-dependent regularization method and show that it allows learning simulators that produce calibrated uncertainty estimates under the proposed framework.
\end{enumerate}
\vspace*{-6pt}

\input{drafts/_aistats2025/related_work}

\section{\uppercase{A Monte Carlo Framework}}

\label{sec:problem_statement}

We consider the problem of predicting a sequence of multi-class labels from high-dimensional data.
More formally, our goal is to estimate the conditional distribution of a sequence ${{Y}}$ consisting of $\ell$ discrete random variables each with $c$ possible states given an observed input $x$, interpreted as a sample of a random vector ${{X}}$.
In our health-monitoring example, the $c:=3$ states are {\it healthy}, {\it ill} and {\it dead}, and ${{X}}$ represents an input image. 

Even for short sequence lengths, directly estimating the joint conditional probability mass function of ${{Y}}$ given ${{X}}=x$ is intractable due to the combinatorial explosion of possible sequences (e.g., for $c:=3$ and $\ell:=100$ there are $3^{100} > 10^{47}$ possible sequences!), which is an instance of the notorious curse of dimensionality.
Instead, we propose to estimate the following probabilities and confidence intervals characterizing the sequence, which are illustrated in \Cref{fig:problem}:\vspace*{-8pt}
\begin{enumerate}[leftmargin=15pt, topsep=7pt, itemsep=2pt, parsep=2pt, partopsep=0pt]
\item
The {\it marginal} probability $\operatorname{P}({{Y}}_i=a \, | \, {{X}}=x)$ that the $i$th entry ${{Y}}_i$ of the sequence is equal to $a\in \{1,...,c\}$. We refer to this as a marginal probability, but---strictly speaking---it is a conditional probability, as it is conditioned on ${{X}}=x$.
In health monitoring, this is the probability that a subject is healthy, ill, or dead at time $i$.
\item
The {\it conditional}  probability $\operatorname{P}({{Y}}_i=a \, | \, {{Y}}_j=b, {{X}}=x)$ that the $i$th entry ${{Y}}_i$ of the sequence is equal to $a\in \{1,...,c\}$ given that the $j$th entry is equal to $b\in \{1,...,c\}$.
In health monitoring, this could be the conditional probability that a subject is dead at time $i$ given that they are ill at time $j$. 
\item
The $\alpha$ {\it confidence interval} $I_{\alpha}$ for the time $\widetilde{T}$ until a certain event associated with the sequence occurs (e.g. $\widetilde{T}:= \min \{i \; : \; {{Y}}_i = a\}$ for some $a\in \{1,...,c\}$), which should satisfy $\operatorname{P}(\widetilde{T} \in I_{\alpha}) = \alpha,$
where $\alpha$ is typically set to 0.9 or 0.95.
For example, in health monitoring, $\widetilde{T}$ can represent the time of death or recovery. 
\end{enumerate}

\begin{figure*}[t]
    (a) Training of the autoregressive simulator \hspace{1.8cm} (b) Sampling from the simulator given an input image \\
    \includegraphics[width=0.98\linewidth]{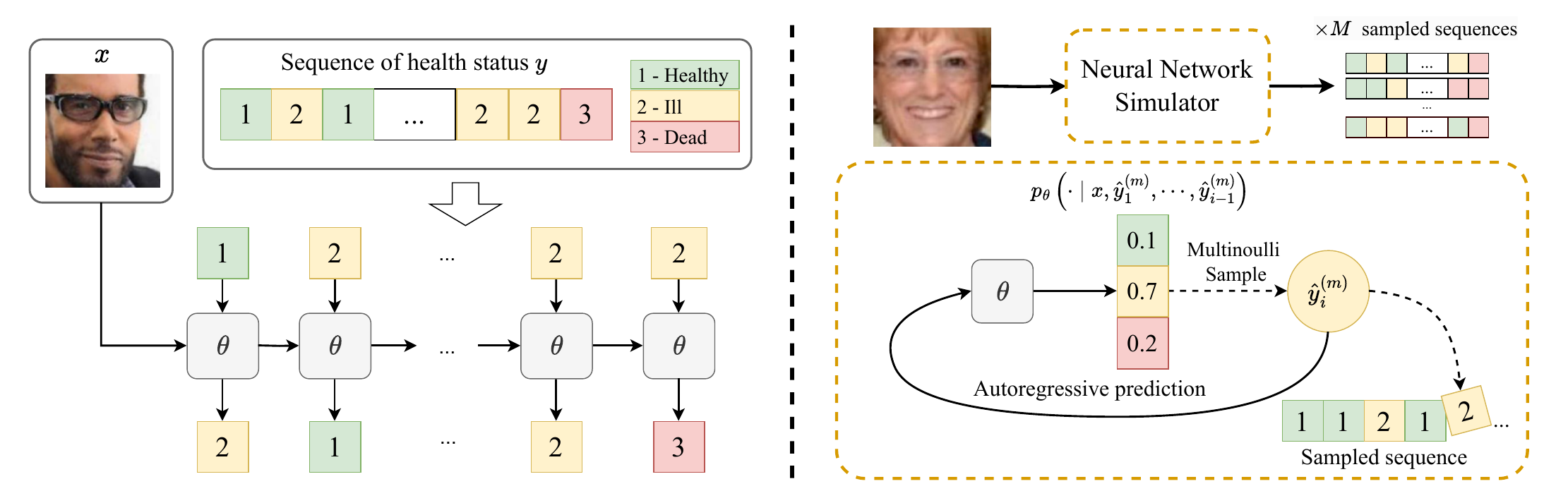} \\
    \vspace{3mm}
    (c) Probabilistic prediction via the Monte Carlo method \\
    \vspace{1mm}
    \includegraphics[trim={7mm -7mm 8mm 12mm}, clip, height=0.215\linewidth]{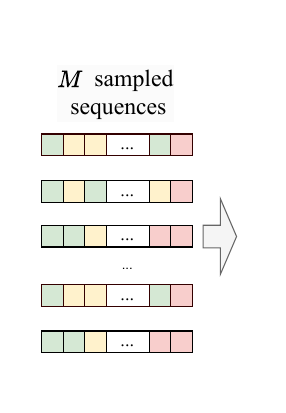}
    \includegraphics[height=0.228\linewidth]{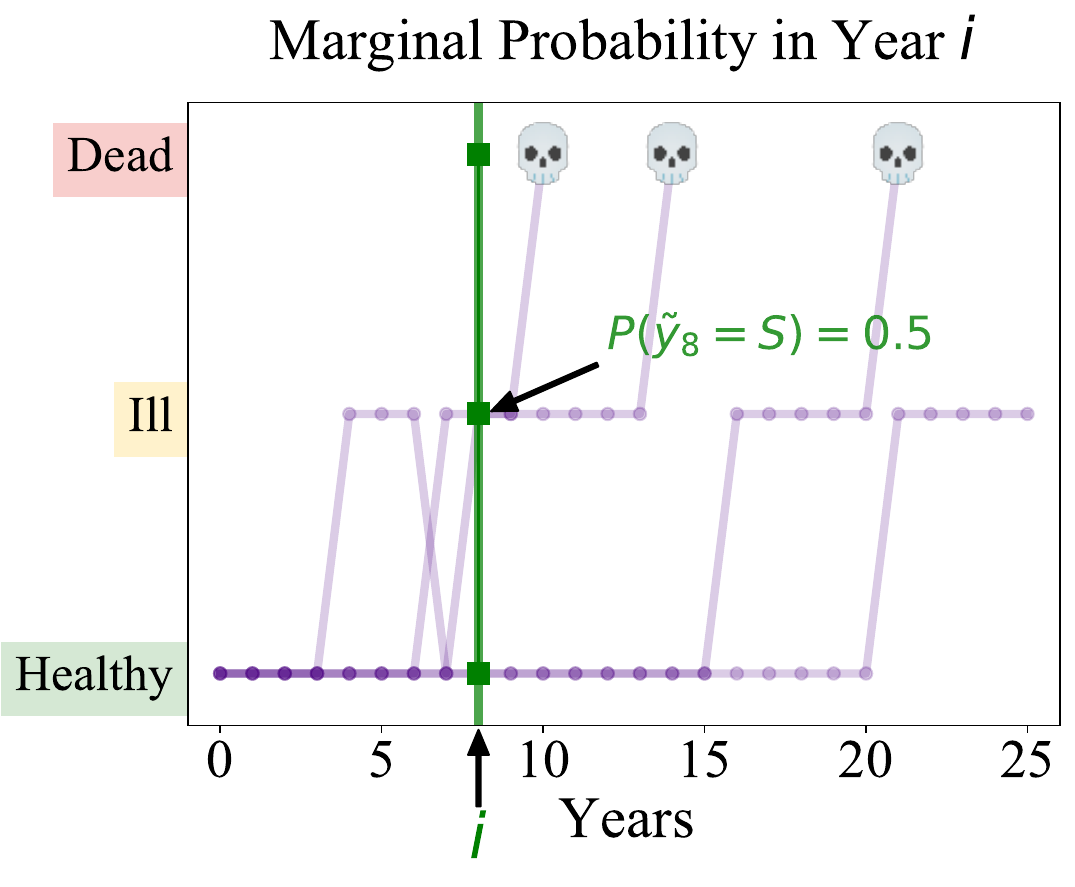}
    \includegraphics[height=0.228\linewidth]{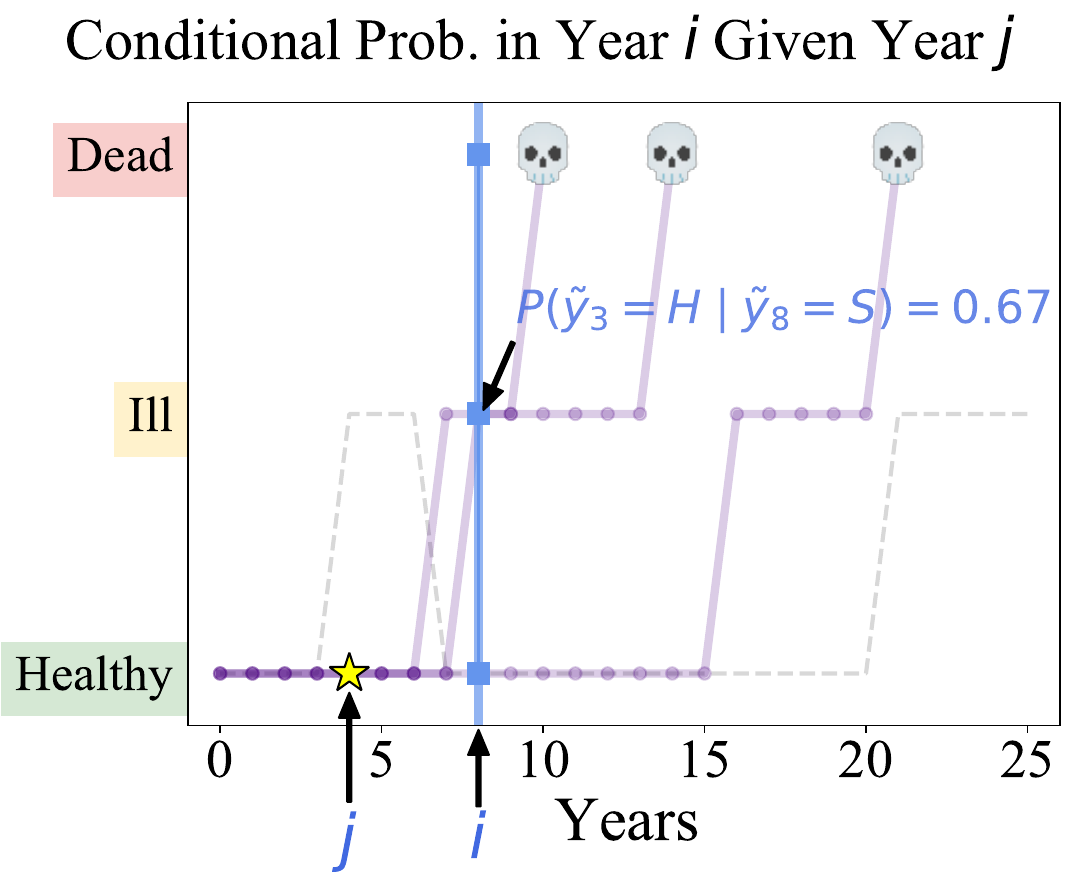} 
    \includegraphics[height=0.228\linewidth]{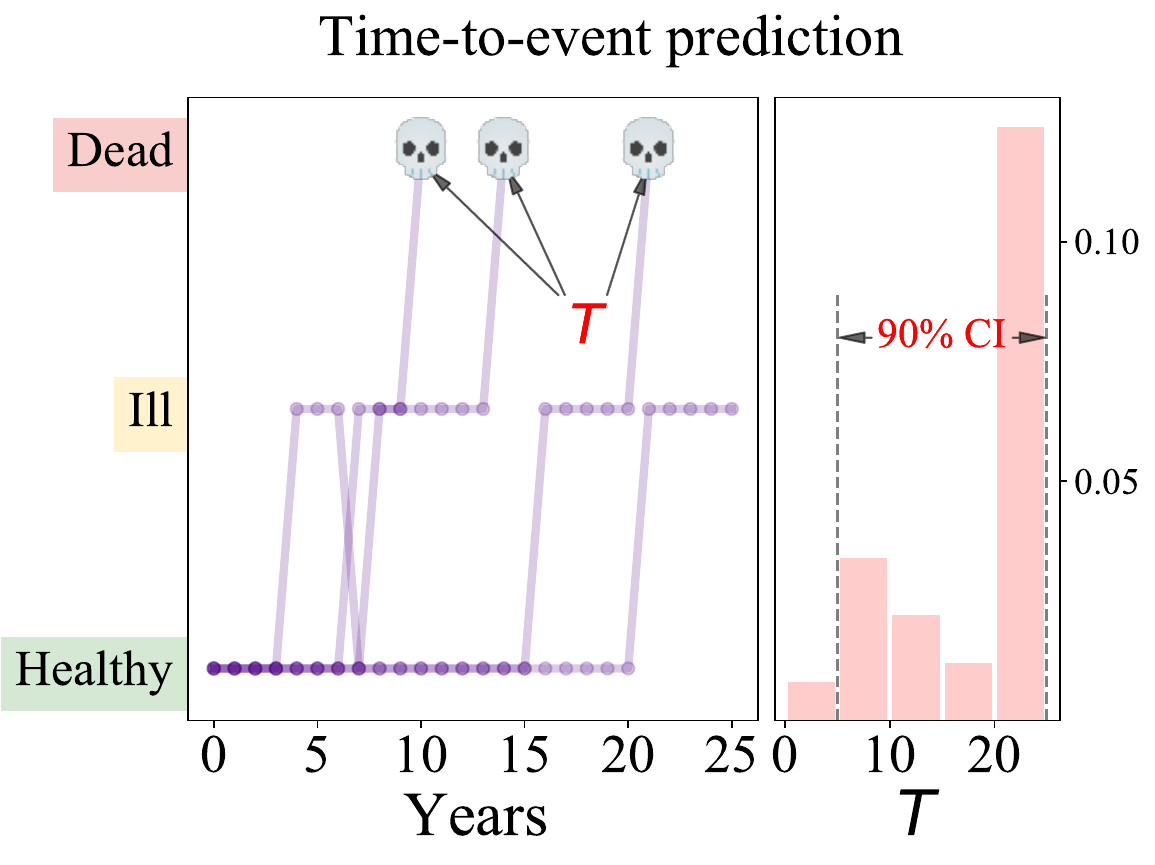}
    \vspace*{-5pt}
    \caption{\textbf{Monte Carlo framework for uncertainty estimation in sequence prediction.}
    (a) A neural network simulator is trained to autoregressively predict the conditional distribution of each entry in a sequence given an input image and the preceding states. (b) The simulator is used to generate multiple sample sequences, by iteratively sampling from the estimated conditional distribution. (c) The Monte Carlo method is applied to estimate marginal probabilities, conditional probabilities, and time-to-event confidence intervals from the samples.}
    \label{fig:methodology}
\end{figure*}

\subsection{Monte Carlo Estimation}
\label{sec:monte_carlo_estimation}

As noted in the previous section, a key challenge in estimating arbitrary probabilities and confidence intervals associated with a sequence of random variables ${{Y}}$ is that it is intractable to \textit{explicitly} estimate the sequence joint distribution given the input ${{X}}$ (unless we make highly simplifying modeling assumptions, such that the sequence forms a Markov chain).
Our proposed {\bf f}ramework for M{\bf o}nte {\bf C}arlo {\bf u}ncertainty quantification of {\bf s}equences (\textbf{foCus}) addresses this challenge by instead {\it implicitly sampling} from the conditional distribution using a neural network simulator, described in \Cref{sec:nn_simulator}.

Given an input $x$, we apply the simulator to generate $M$ sequences $\{(\hat{y}^{(m)}_1$, ..., $\hat{y}^{(m)}_l)\}_{m=1}^{M}$. %
As shown in panel (c) of \Cref{fig:methodology}, the sequences are then used to estimate any desired probability or confidence interval. The marginal probability of state $a$ at time $i$ is estimated by the fraction of sequences in state $a$ at time $i$:
\begin{align}
    \operatorname{P}({{Y}}_i=a \, | \, {{X}}=x) = \frac{1}{M} \sum_{m=1}^M \mathbbm{1}_{\{\hat{y}^{(m)}_i = a\}}.
\end{align}
The conditional probability of state $a$ at time $i$ given state $b$ at time $j$ is estimated by the fraction of sequences in state $a$ at time $i$ out of sequences in state $b$ at time $j$:
\begin{align}
\operatorname{P}({{Y}}_i=a \, | \, {{Y}}_j=b, {{X}}=x) = \frac{\sum_{m=1}^M \mathbbm{1}_{\{\hat{y}^{(m)}_i = a, \hat{y}^{(m)}_j = b\}}}{\sum_{m=1}^M \mathbbm{1}_{\{\hat{y}^{(m)}_j = b\}}}.
\end{align}
To estimate the $\alpha$ confidence interval $I_\alpha$ of the time-to-event $\widetilde{T}$, we first compute the value of $\widetilde{T}$ associated with each simulated sequence. These yields $M$ times $\{T^{(1)}$, ..., $T^{(M)}\}$, which can be sorted to calculate the $(1-\alpha)/2$ and $(1+\alpha)/2$ percentiles $q_{(1-\alpha)/2}$ and $q_{(1+\alpha)/2}$. The confidence interval $I_\alpha$ is set to $[q_{(1-\alpha)/2},q_{(1+\alpha)/2}]$. 
\begin{table*}[t!]
\setlength{\tabcolsep}{6.5pt}
\small
\caption{
\textbf{Marginal probability estimation.} The table reports sequence-level metrics evaluating the performance of the proposed foCus framework for estimation of marginal probabilities (see \Cref{sec:problem_statement}). We compare versions of foCus without regularization (see \Cref{sec:pathology}) and with constant and time-dependent regularization (see \Cref{sec:regularization}).
Results are presented as mean ± standard error from three independent model realizations. Time-dependent regularization improves calibration substantially (lower ECE), while maintaining a comparable AUC, which results in superior probability estimates (lower cross entropy and Brier score). Similar results are obtained for conditional probability estimation, as reported in %
\Cref{tab:last_epoch_score_condition}.} 

\label{tab:last_epoch_score_uncondition}
\begin{center}
\begin{tabular}{lccccc}
\toprule
Scenario & Regularization & ECE $(\downarrow)$ & AUC $(\uparrow)$ & CE $(\downarrow)$ & BS $(\downarrow)$ \\
\midrule
\multirow{3}{*}{Seaquest} & \xmark & 0.0435 $\pm$ 0.0004 & 0.8671 $\pm$ 0.0035 & 1.0577 $\pm$ 0.0241 & 0.1247 $\pm$ 0.0012 \\
 & time-dependent & \bf{0.0277 $\pm$ 0.0023} & \bf{0.8678 $\pm$ 0.0028} & \bf{0.6705 $\pm$ 0.0285} & \bf{0.1144 $\pm$ 0.0007}\\
 & constant & 0.0365 $\pm$ 0.0002 & 0.8625 $\pm$ 0.0020 & 0.8173 $\pm$ 0.0068 & 0.1177 $\pm$ 0.0008 \\
\midrule
\multirow{3}{*}{River Raid} & \xmark & 0.0583 $\pm$ 0.0016 & \bf{0.6453 $\pm$ 0.0009} & 1.2034 $\pm$ 0.0281 & 0.1750 $\pm$ 0.0016 \\
 & time-dependent & \bf{0.0388 $\pm$ 0.0013} & 0.6346 $\pm$ 0.0035 & \bf{0.8585 $\pm$ 0.0132} & \bf{0.1671 $\pm$ 0.0012}\\
 & constant & 0.0474 $\pm$ 0.0004 & 0.6280 $\pm$ 0.0020 & 1.0274 $\pm$ 0.0158 & 0.1686 $\pm$ 0.0005 \\
\midrule
\multirow{3}{*}{Bank Heist} & \xmark & 0.0559 $\pm$ 0.0032 & \bf{0.6938 $\pm$ 0.0028} & 1.1874 $\pm$ 0.0540 & 0.2340 $\pm$ 0.0020 \\
 & time-dependent & \bf{0.0148 $\pm$ 0.0014} & 0.6782 $\pm$ 0.0016 & \bf{0.7647 $\pm$ 0.0130} & \bf{0.2166 $\pm$ 0.0005} \\
 & constant & 0.0399 $\pm$ 0.0016 & 0.6928 $\pm$ 0.0046 & 0.8894 $\pm$ 0.0112 & 0.2211 $\pm$ 0.0007 \\
\midrule
\multirow{3}{*}{H.E.R.O.} & \xmark & 0.0947 $\pm$ 0.0014 & 0.6785 $\pm$ 0.0061 & 1.1310 $\pm$ 0.0225 & 0.1261 $\pm$ 0.0009 \\
 & time-dependent & 0.0481 $\pm$ 0.0034 & \bf{0.7159 $\pm$ 0.0105} & \bf{0.6940 $\pm$ 0.0246} & \bf{0.1170 $\pm$ 0.0008} \\
 & constant & \bf{0.0352 $\pm$ 0.0001} & 0.7041 $\pm$ 0.0102 & 0.7218 $\pm$ 0.0391 & 0.1212 $\pm$ 0.0011 \\
\midrule
\multirow{3}{*}{Road Runner} & \xmark & 0.0779 $\pm$ 0.0035 & \bf{0.6913 $\pm$ 0.0100} & 1.1586 $\pm$ 0.0291 & 0.1575 $\pm$ 0.0034 \\
 & time-dependent & \bf{0.0204 $\pm$ 0.0012} & 0.6823 $\pm$ 0.0084 & \bf{0.5255 $\pm$ 0.0077} & \bf{0.1382 $\pm$ 0.0003} \\
 & constant & 0.0303 $\pm$ 0.0027 & 0.6898 $\pm$ 0.0140 & 0.6275 $\pm$ 0.0250 & 0.1394 $\pm$ 0.0010 \\
\midrule
\multirow{3}{*}{FaceMed} & \xmark & 0.1503 $\pm$ 0.0048 & 0.7534 $\pm$ 0.0079 & 1.6932 $\pm$ 0.0188 & 0.3464 $\pm$ 0.0012 \\
& time-dependent & \bf{0.0757 $\pm$ 0.0068} & \bf{0.7614 $\pm$ 0.0024} & \bf{0.9085 $\pm$ 0.0303} & \bf{0.3328 $\pm$ 0.0008} \\
& constant & 0.0974 $\pm$ 0.0045 & 0.7613 $\pm$ 0.0028 & 1.0071 $\pm$ 0.0499 & 0.3356 $\pm$ 0.0030 \\
\bottomrule
\end{tabular}
\end{center}
\end{table*}
\subsection{Autoregressive  Simulation}
\label{sec:nn_simulator}
In order to produce the simulated sequences required by our Monte Carlo framework, we
employ a neural-network simulator. The simulator consists of a convolutional neural network (CNN) that encodes the input image $x$, and a recurrent neural network (RNN) that iteratively estimates the conditional distribution of the $i$th entry ${{Y}}_i$ of the sequence given ${{X}}=x$ and the values of all previous entries, i.e. \(\operatorname{P}\left({{Y}}_i = y_i \mid {{X}}=x, {{Y}}_1 = y_{1}, ..., {{Y}}_{i-1} = y_{i-1} \right)\). 

\Cref{fig:methodology}(b) illustrates how the simulator is used to obtain a sample sequence $(\hat{y}_1, \dots, \hat{y}_{\ell})$. The input image $x$ is fed into the CNN, producing a hidden vector $h_0$ that is fed into the RNN to then generate the simulated sequence iteratively. At each iteration $i \in \{1,...,\ell\}$, the input of the RNN is the value $\hat{y}_{i-1}$ of the previous entry (except for $i=1$) and the hidden vector $h_{i-1}$. The outputs are an estimate of the conditional distribution of ${{Y}}_{i}$ given the previous entries and ${{X}}=x$, and an updated hidden vector $h_i$. The $i$th entry $\hat{y}_i$ of the sample sequence is sampled from this conditional distribution. %

The simulator is trained using a dataset of image-sequence pairs, as illustrated in \Cref{fig:methodology}(a).
During training, the model uses the ground truth value $y_{i-1}$ of the previous entry along with the hidden vector $h_{i-1}$ to predict the subsequent $i$th entry ${{Y}}_i$.
As explained in \Cref{sec:pathology,sec:regularization} the design of the training loss plays a crucial role in avoiding miscalibration in the downstream probabilities and confidence intervals computed using the simulator.  

\section{\uppercase{Evaluating Uncertainty}}
\label{sec:evaluation}

\begin{figure*}[t]
    \centering

    \includegraphics[height=0.24\linewidth]{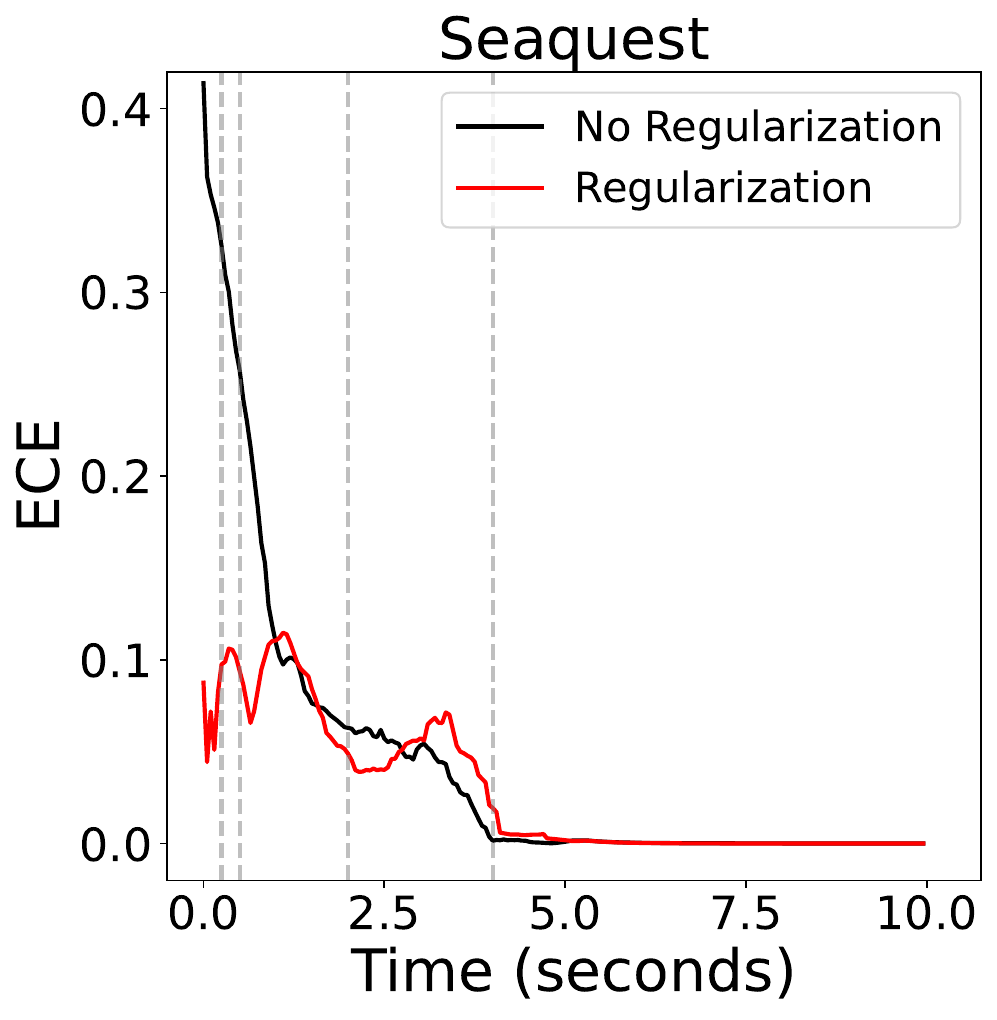}
    \includegraphics[height=0.24\linewidth]{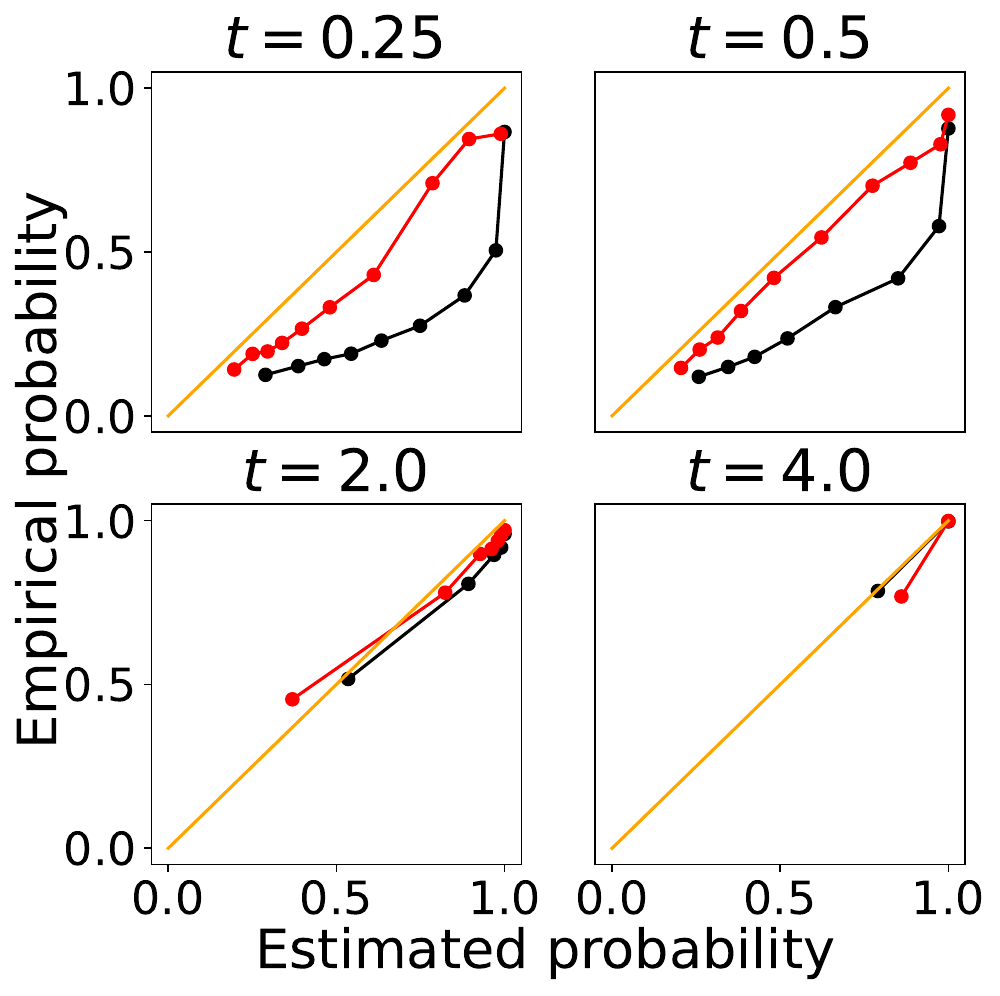}
    \includegraphics[height=0.24\linewidth]{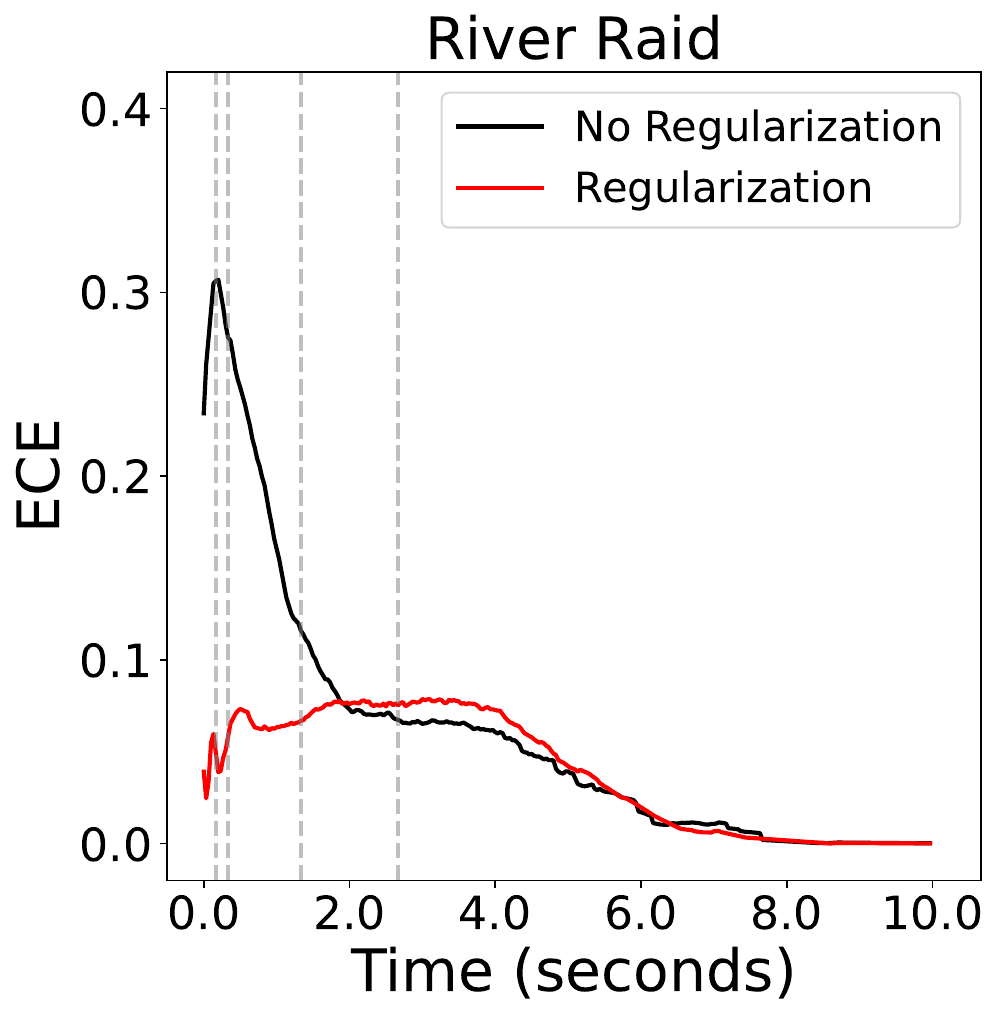}
    \includegraphics[height=0.24\linewidth]{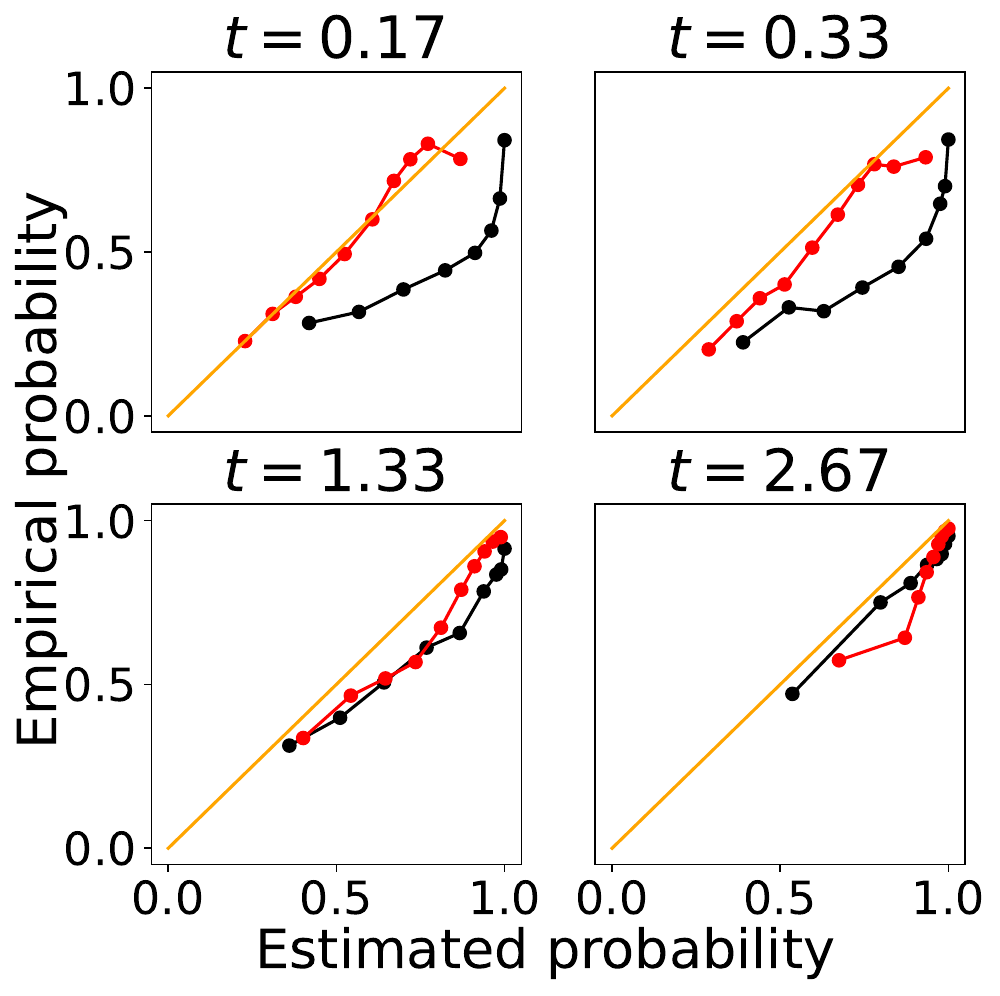} \\

    \includegraphics[height=0.24\linewidth]{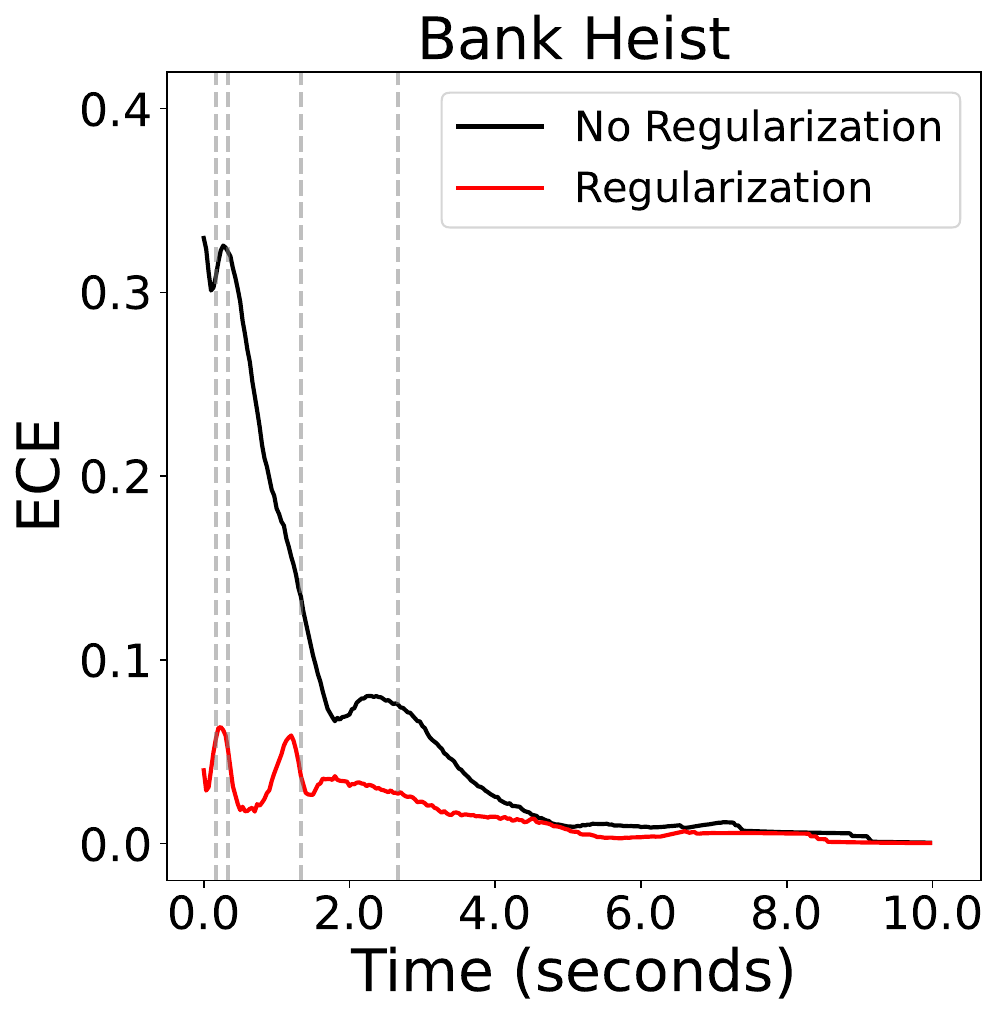}
    \includegraphics[height=0.24\linewidth]{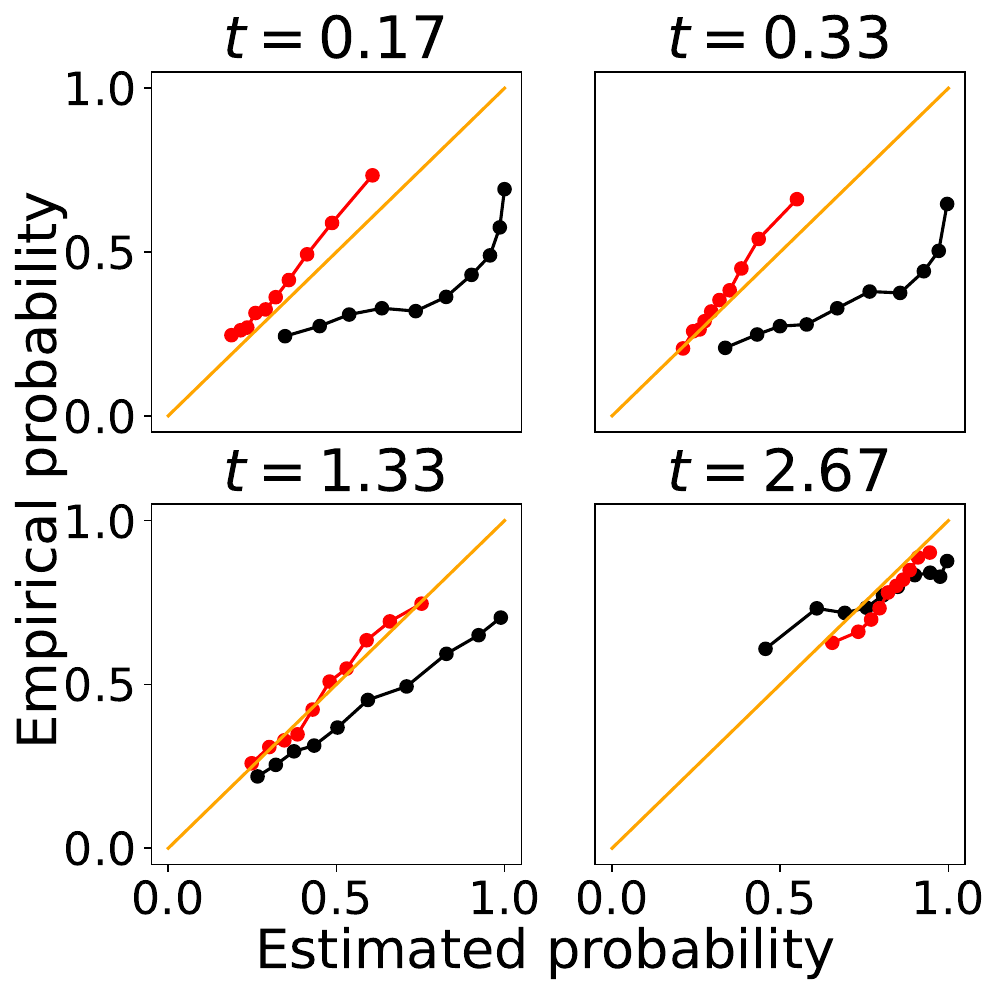}
    \includegraphics[height=0.24\linewidth]{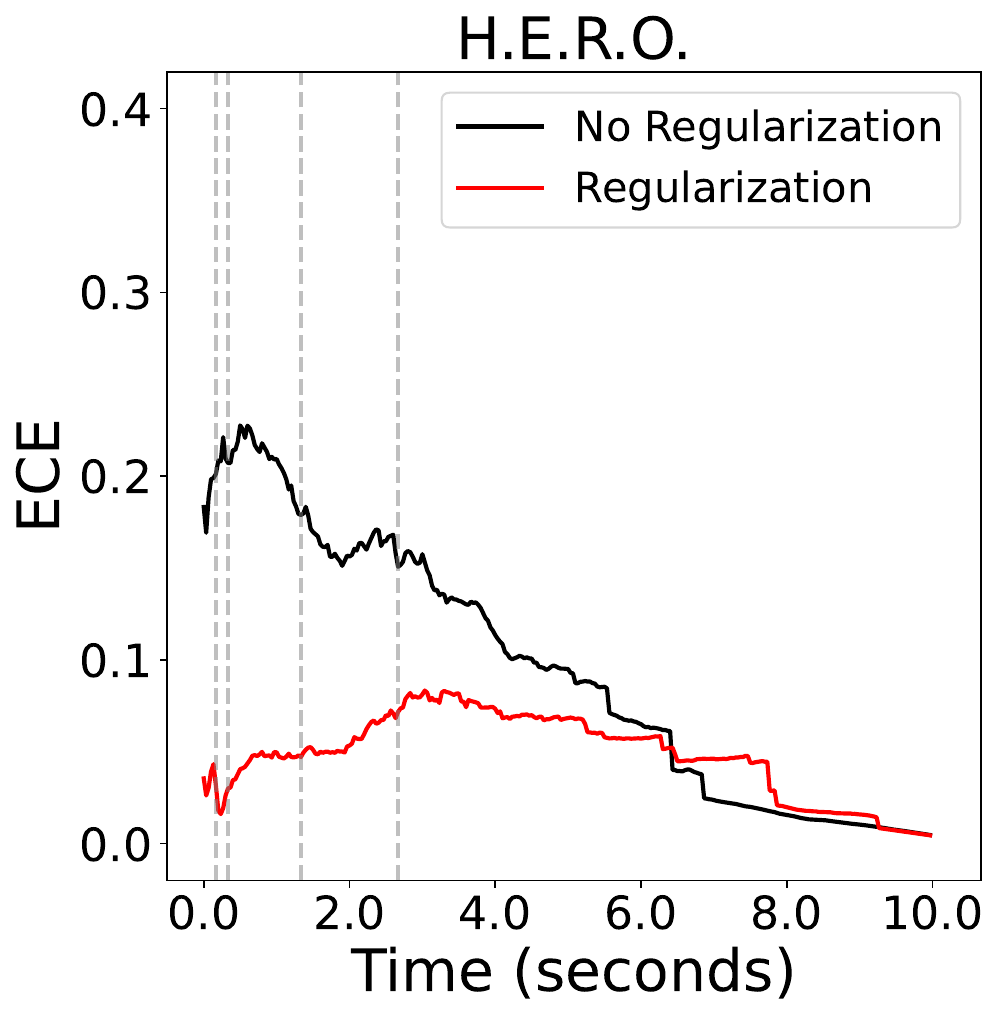}
    \includegraphics[height=0.24\linewidth]{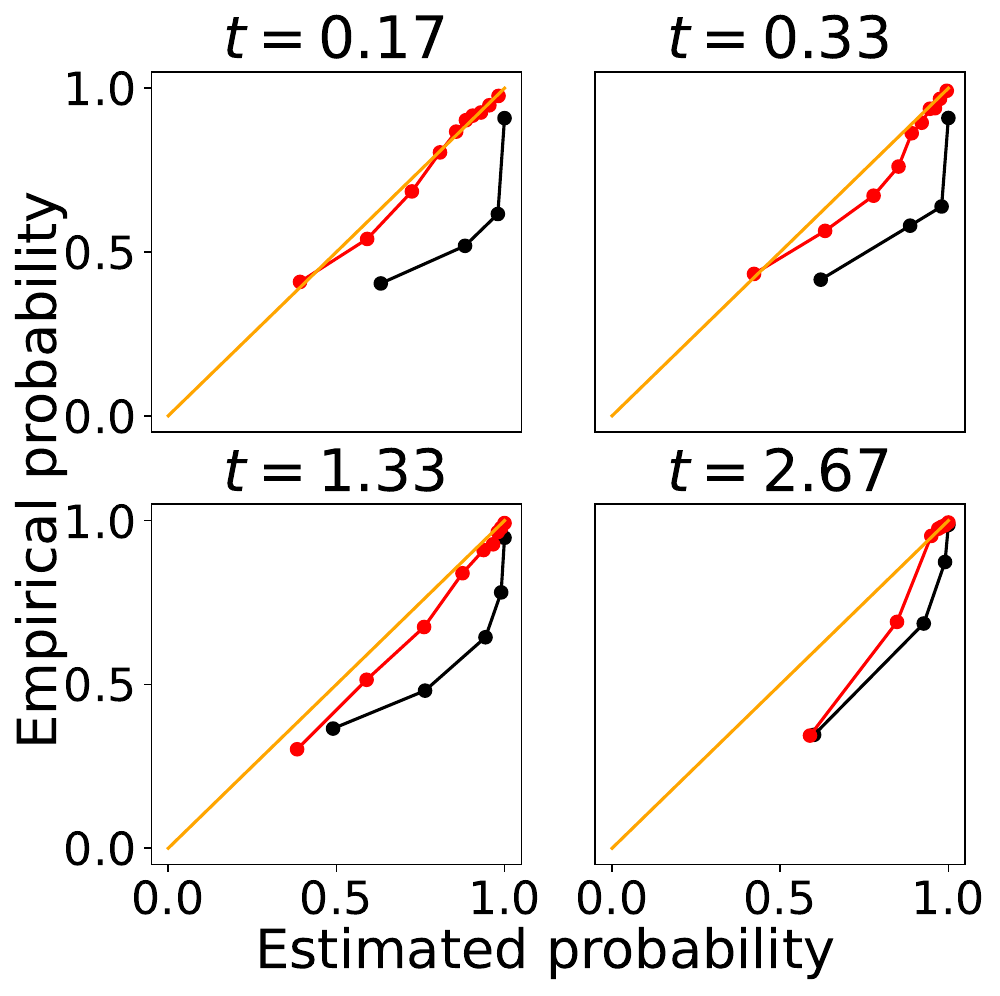} \\

    \includegraphics[height=0.24\linewidth]{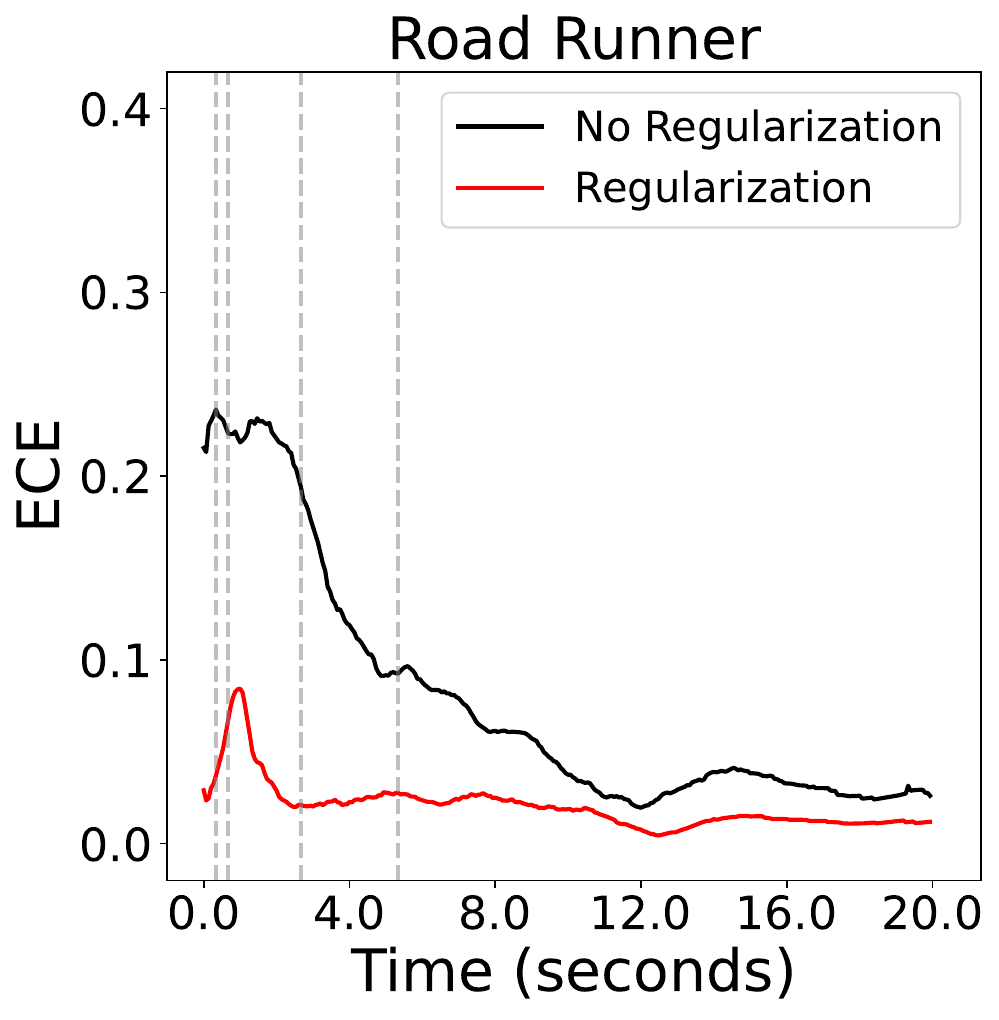}
    \includegraphics[height=0.24\linewidth]{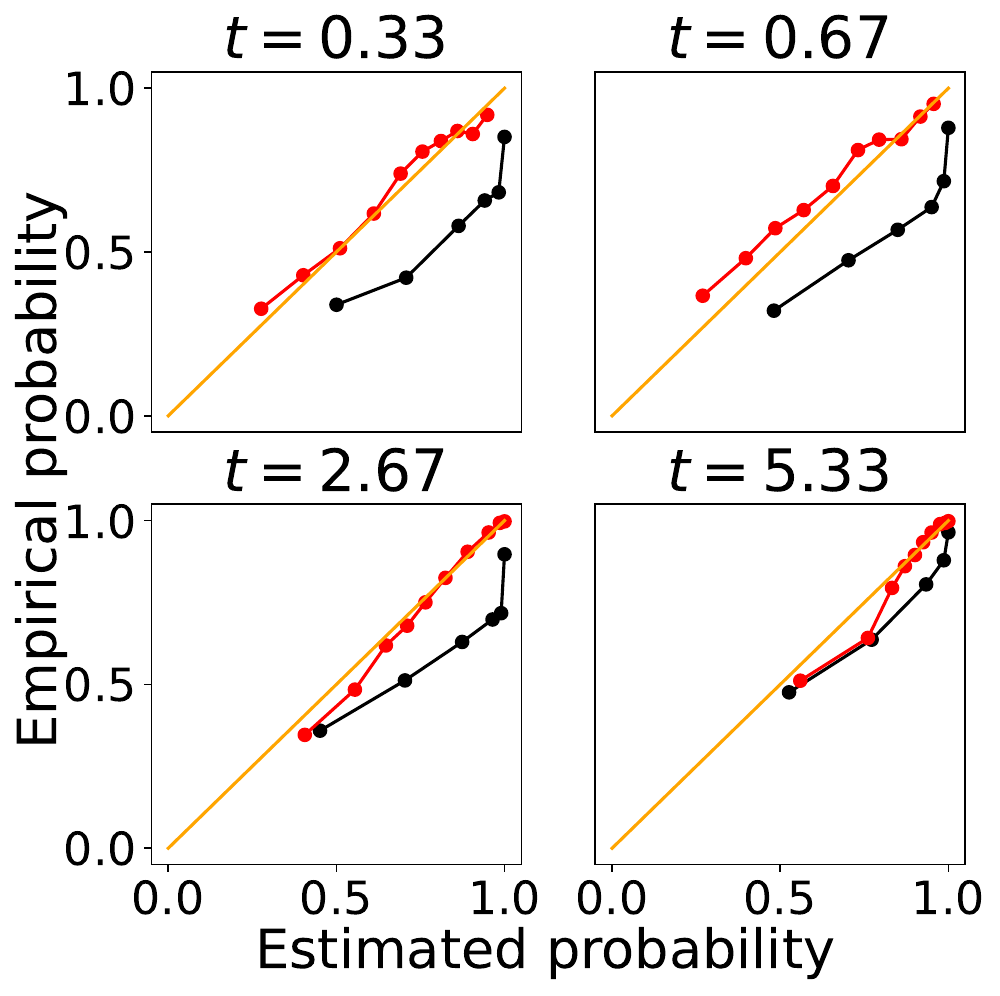}
    \includegraphics[height=0.24\linewidth]{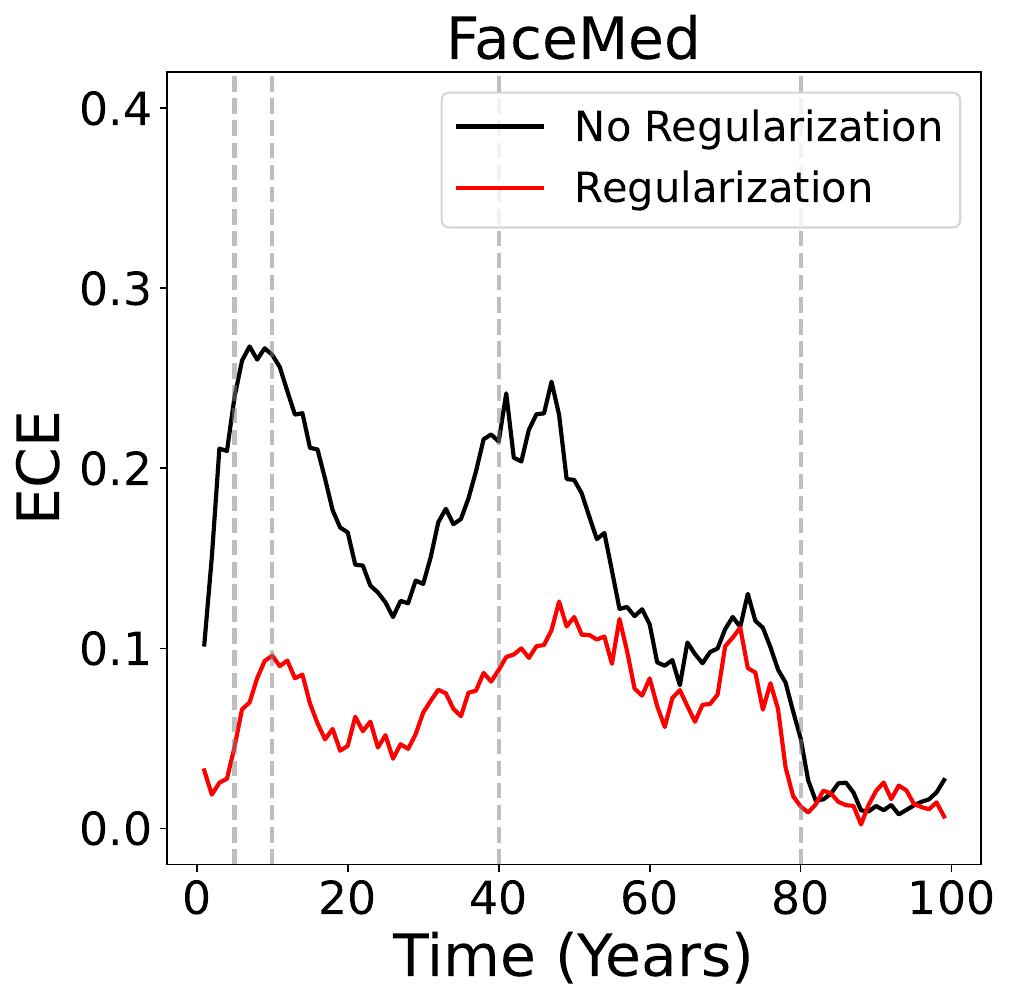}
    \includegraphics[height=0.24\linewidth]{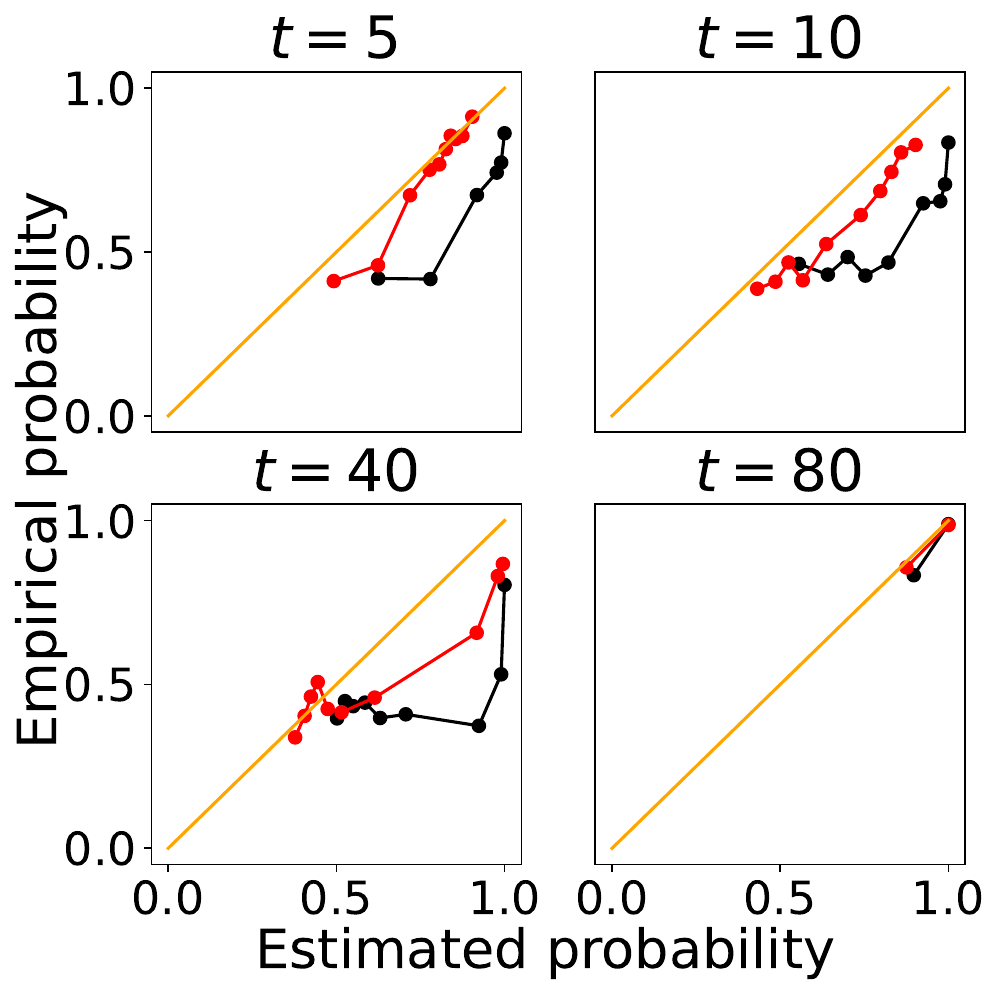}
    \caption{\textbf{Entry-wise calibration error and reliability diagrams for marginal probability estimation.}
    The large graphs plot the entry-level ECE of the proposed foCus framework for estimation of marginal probabilities (see \Cref{sec:problem_statement}) without regularization (black line, see \Cref{sec:pathology}) and time-dependent regularization (red line, see \Cref{sec:regularization}). Unregularized foCus produces miscalibrated estimates, particularly in the earlier entries, which are dramatically improved by time-dependent regularization for all datasets. The small graph show reliability diagrams for some of the steps, which confirm the improvement in calibration. 
    Additional reliability diagrams and results for constant regularization are shown in \Cref{app:additional_results_marginal}. %
    }
    \label{fig:reliability_diagram}
    \vspace{-8pt}
\end{figure*}

\begin{figure*}[ht]
\begin{center}
\includegraphics[width=0.495\textwidth]{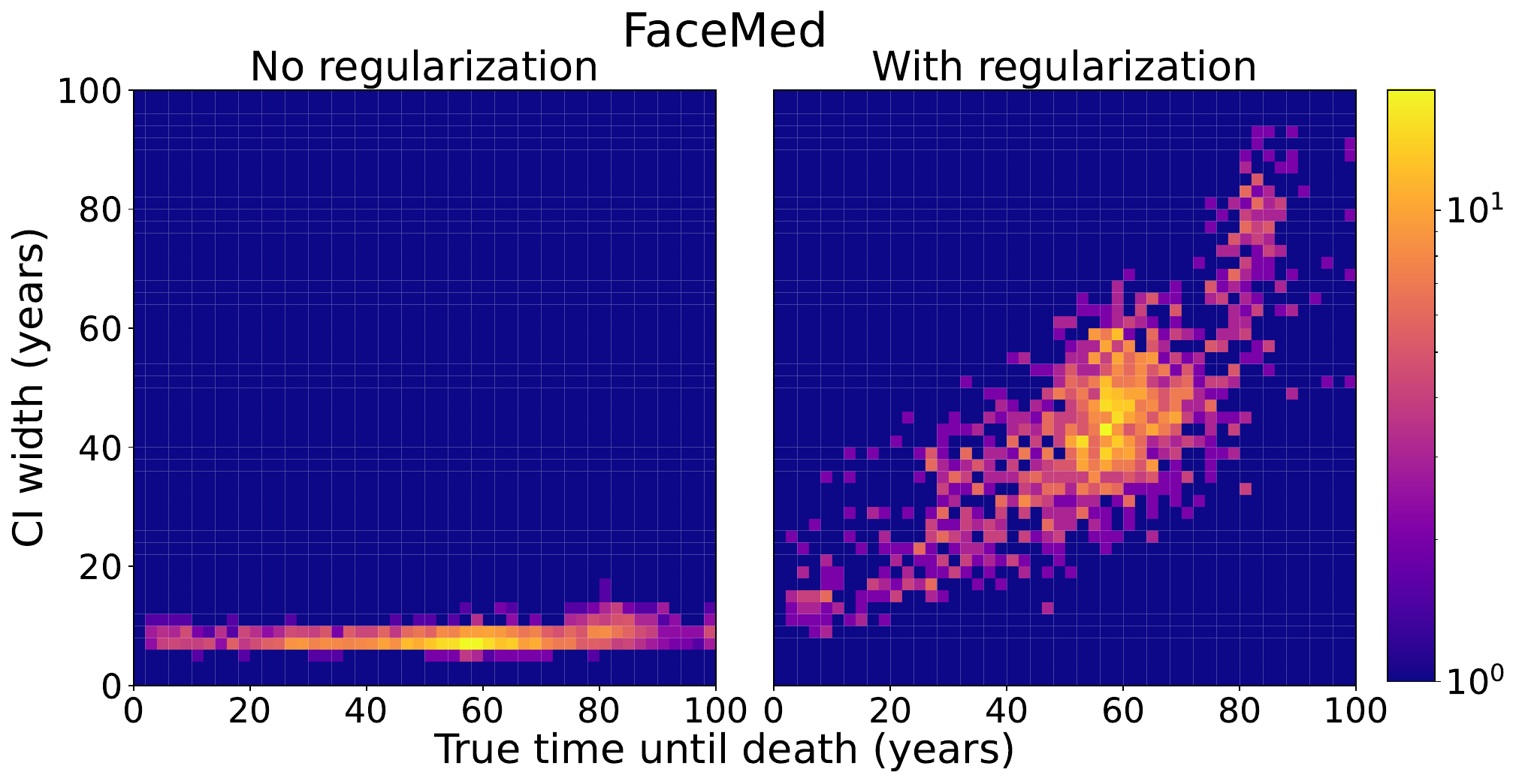} 
\includegraphics[width=0.495\textwidth]{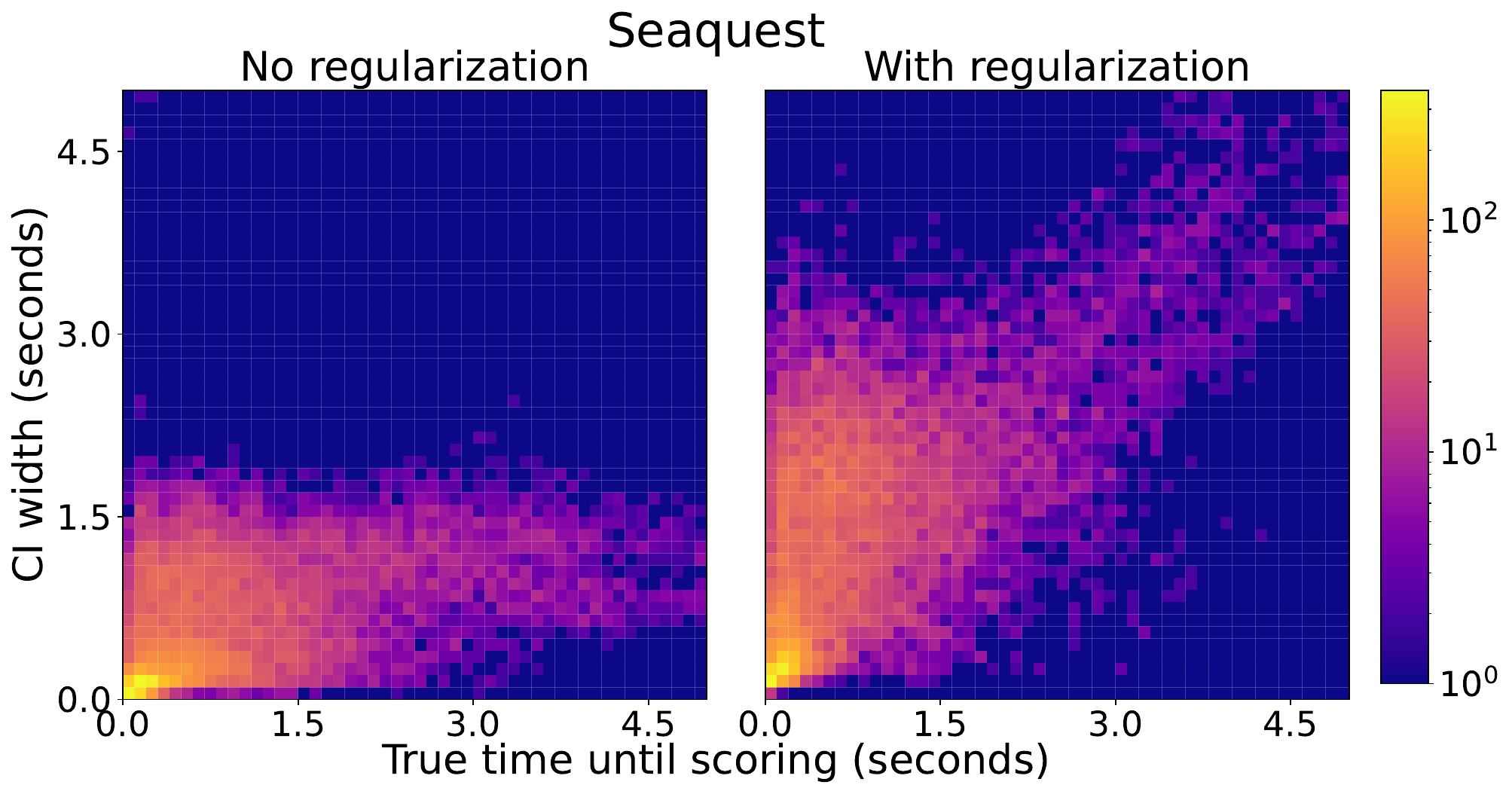} \vspace{0.05cm}\\%[-6pt]

\hspace{0mm}
\includegraphics[width=0.451\textwidth]{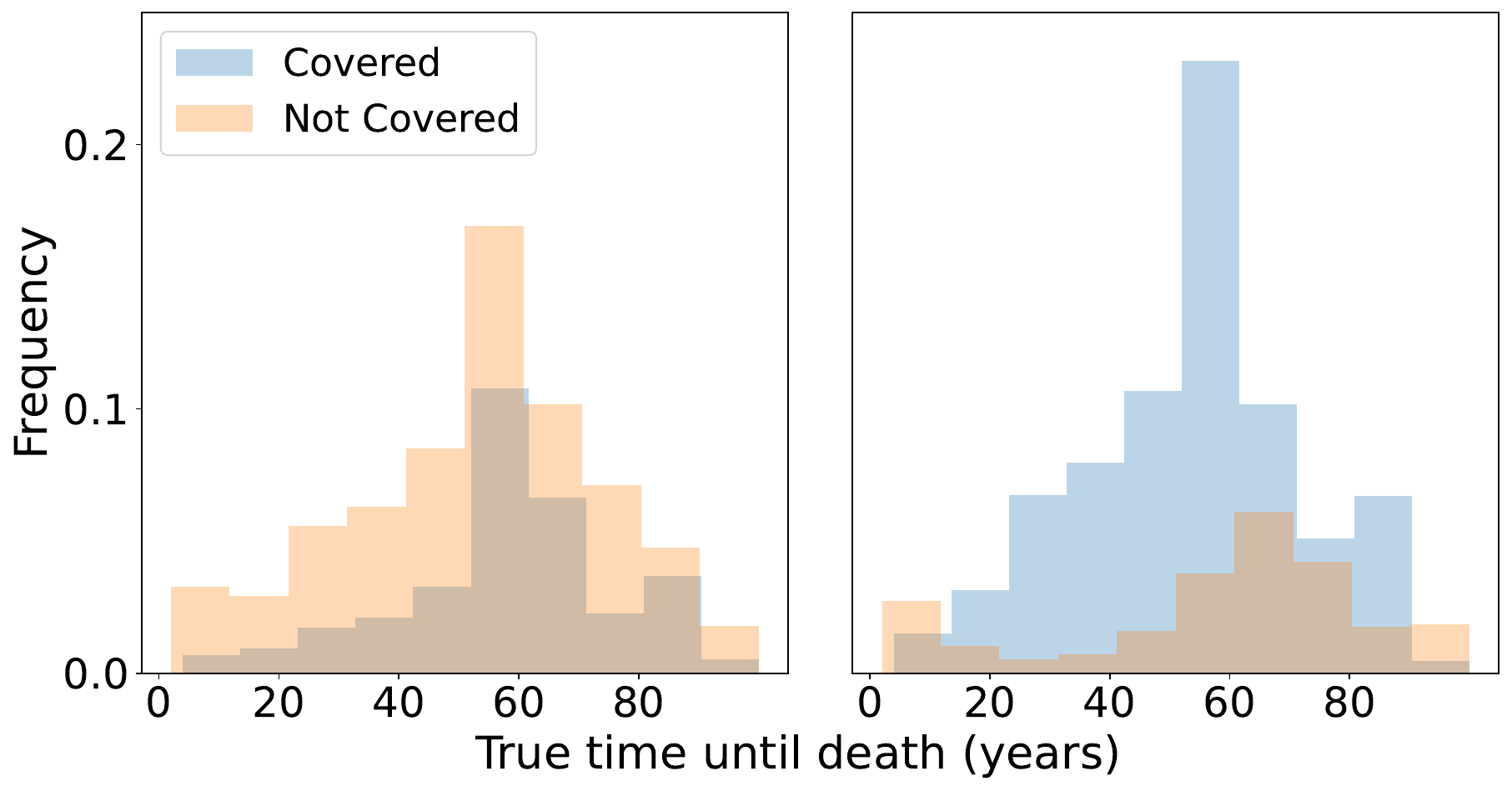}
\hspace{6.8mm}
\includegraphics[width=0.451\textwidth]{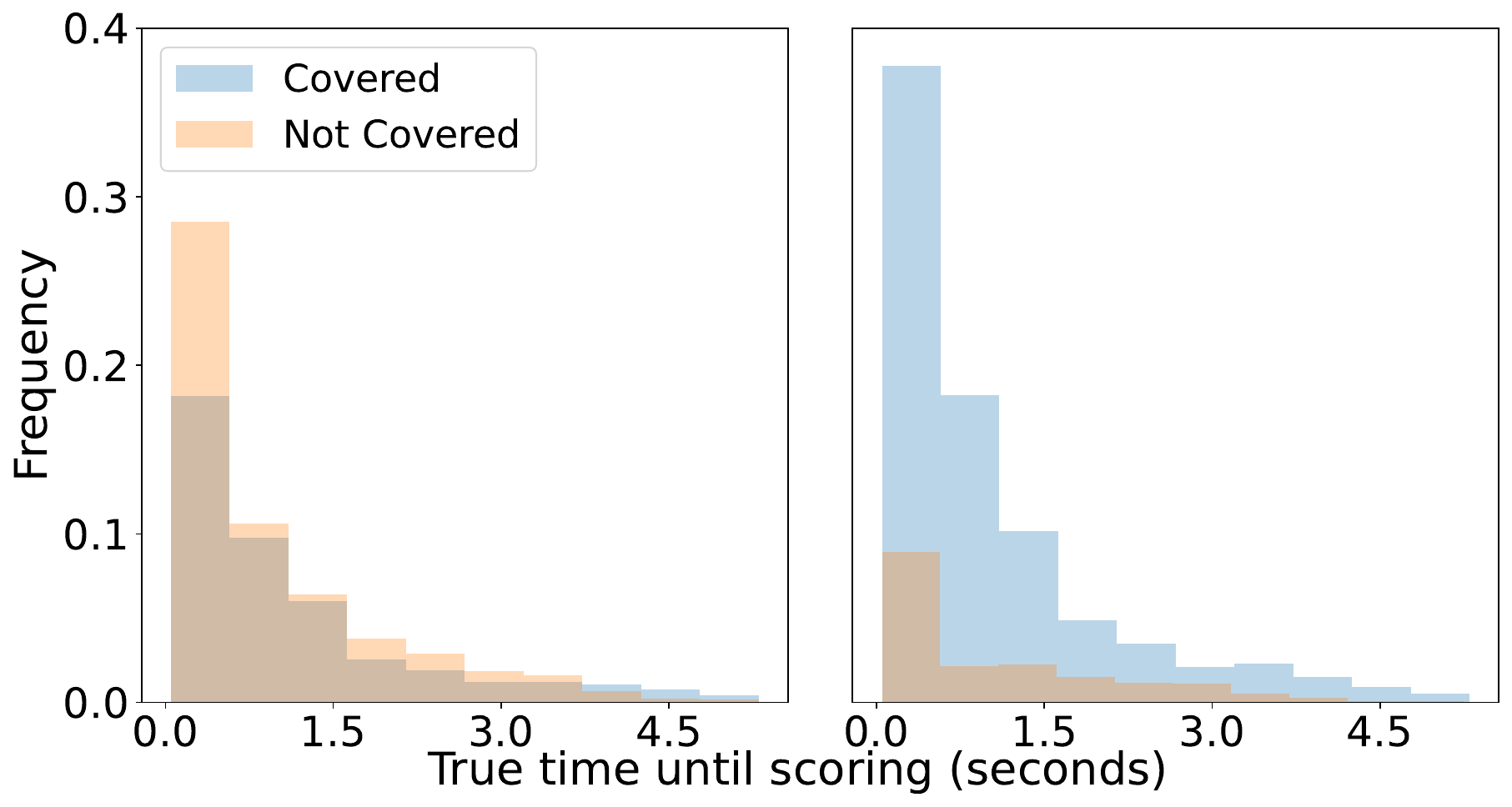}
\hspace{6.8mm}
\caption{\textbf{Confidence intervals for time-to-event prediction and coverage probability.}
The upper panel shows heatmaps of the length of 0.9 confidence intervals for time-to-event prediction using the proposed foCus framework without and with time-dependent regularization. The histograms below show the frequency of intervals containing the true times, as a function of the true time. Unregularized foCus produces short intervals with poor coverage, whereas regularization yields intervals that tend to be larger when the ground-truth times are larger, and are much better calibrated. 
The plots correspond to the FaceMed (left) and Seaquest (right) datasets. \Cref{app:additional_results_ci} shows analogous plots for the remaining datasets.}
\label{fig:CI_analysis}
\end{center}
\end{figure*}

\subsection{Evaluation metrics}
\label{sec:evaluation_metrics}

We assess marginal and conditional probability estimates with a set of complementary metrics. Macro Area Under the ROC Curve (AUC) quantifies discriminative ability.  Expected Calibration Error (ECE) quantifies calibration. Brier Score (BS) and Cross Entropy (CE) provide a more holistic evaluation. These metrics are computed for each entry and then aggregated via averaging to obtain sequence-level metrics.  

To assess confidence intervals, %
we evaluate discriminability via the Mean Absolute Error (MAE) (normalized by the average sequence length). 
Calibration is evaluated by computing the coverage probability of the intervals, which should be close to the target confidence level $\alpha$.  We also measure the average confidence interval width relative to the average sequence length. Additional details are provided in \Cref{app:metrics}.

\vspace*{-3pt}
\subsection{Datasets}
\vspace*{-3pt}
\label{sec:exp_env}

\paragraph{FaceMed} is a synthetic dataset designed to predict individual health status trajectories based UTKFace \citep{utkface}, which contains of real face images of subjects with different ages. For each image, we simulate a sequence of health states using a Markov chain model that depends on the age of the subject (see \Cref{app:dataset_details}). The health states are {\it healthy}, {\it ill}, and {\it dead}. The goal is to predict the future marginal and conditional probability distribution of health status in the future, as well as confidence intervals for subject survival. %
\vspace{-2mm}    
\paragraph{Atari games} is a non-synthetic benchmark consisting of five real human gameplay datasets from Atari-HEAD~\citep{Zhang_2020}, consisting of screenshots associated with subsequent sequences of player actions. 
There are 19 actions, including {\it move left}, {\it fire}, and an {\it end} action, which indicates the end of a game sequence. The games are:%
\begin{itemize}[leftmargin=15pt, topsep=0pt, itemsep=0pt, parsep=0pt]
    \item Seaquest: Control a submarine to rescue divers while shooting sharks and enemy submarines;
    \item River Raid: Navigate a fighter jet to destroy enemy targets while managing fuel;
    \item Bank Heist: Drive through a maze-like city to rob banks while avoiding police;
    \item H.E.R.O.: Traverse a mineshaft to rescue trapped miners while avoiding enemies and hazards;
    \item Road Runner: Guide a bird to collect seeds while evading a chasing coyote and obstacles.
\end{itemize}
The goal is to estimate the marginal and conditional probabilities of player actions, and confidence intervals of the time until the player scores. %

All datasets are split into training, validation and test sets following a 7:2:1 ratio. Additional details about datasets are provided in \Cref{app:dataset_details}.
\vspace*{-3pt}
\section{MISCALIBRATION}
\vspace*{-3pt}
\label{sec:pathology}

In this section, we report the results of applying foCus when we train the simulator described in \Cref{sec:nn_simulator} using a standard unregularized cross-entropy loss
\begin{align}
\label{eq:MLE}
    \hspace*{-3pt}\operatorname{CE}(\theta)
    =
    \underset{(x, y)\sim \mathcal{D}}{\mathbb{E}} \left[ - \sum_{i=1}^{\ell} \log p_\theta\left(y_i \mid x, y_1, ..., y_{i-1}\right) \right] \hspace*{-2pt},\hspace*{-3pt}
\end{align}
where $x$ is an input image, $y$ is the corresponding sequence and $\mathcal{D}$ is the training set of image-sequence pairs. Here $\theta$ represents the parameters in the neural-network simulator and $p_{\theta}\left(y_i \mid x, y_1, ..., y_{i-1}\right)$ denotes the corresponding estimate of the conditional probability \(\operatorname{P}\left({{Y}}_i = y_i \mid {{X}}=x, {{Y}}_1 = y_{1}, ..., {{Y}}_{i-1} = y_{i-1} \right)\). Further details about the training procedure are provided in \Cref{app:model_training}.

\Cref{tab:last_epoch_score_uncondition,tab:last_epoch_score_condition} show that this version of foCus has strong discriminative performance across all our datasets when estimating marginal and conditional probabilities, respectively, indicated by the high sequence-level AUC values. However, the sequence-level ECE values are also high, suggesting that the probability estimates are not well calibrated. This is corroborated by 
\Cref{fig:reliability_diagram}, which shows entry-level ECE values for all datasets. Miscalibration is time-dependent and particularly severe at the beginning of the sequence. The reliability diagrams in \Cref{fig:reliability_diagram} show that the model suffer from {\it overconfidence} (the estimated probabilities are more extreme than the empirical probabilities), which is typical of deep learning models, as they tend to overfit the training labels~\citep{liu2022deep}. 

We also observe miscalibration in the confidence intervals estimated by this version of foCus. \Cref{tab:last_epoch_length_score} in \Cref{app:additional_results_ci} reports the coverage probabilities computed on test data for $\alpha = 0.9$, which are below 0.5 in all cases except one! Furthermore, \Cref{fig:CI_analysis} reveals that the widths of the confidence intervals remain invariant over time, which is problematic. Given that uncertainty should naturally increase with longer time horizons, this lack of variation further indicates miscalibration.
\section{TIME-DEPENDENT LOGIT REGULARIZATION}
\label{sec:regularization}

To address simulator-induced miscalibration in foCus, we incorporate a regularization term that penalizes the $\ell_2$-norm of the logits within the simulator.
This regularizer is motivated by recent work on function-space regularization~\citep{rudner2023fseb,rudner2024uap,rudner2024gap,klarner2024guided}, which interprets regularized objectives as performing maximum a posteriori (MAP) estimation under data-driven-priors over neural network parameters.
Intuitively, the $\ell_2$-norm penalty prevents the logits from becoming very large during training.
This mitigates overconfidence, since large logits result in more extreme probability estimates.
We can think of the $\ell_2$-norm penalty as promoting neural network parameters that induce predictive functions with higher predictive entropy.
As in \Cref{sec:pathology}, let $p_{\theta}(y_{i} \vert x, y_1, ..., y_{i-1})$ denote the estimate of the conditional probability \(\operatorname{P}\left({{Y}}_i = y_i \mid {{X}}=x, {{Y}}_1 = y_{1}, ..., {{Y}}_{i-1} = y_{i-1} \right)\) produced by the simulator for $i \in \{1,...,\ell\}$, which is obtained by feeding a logit vector $\mathbf{z} (x, y_{1}, ..., y_{i-1}) \in \R^{c}$ into a softmax function.
The training loss is 
\begin{align}
\begin{split}
\label{eq:L2_Rg}
\calL(\theta)
\defines
\underset{(x, y)\sim \mathcal{D}}{\mathbb{E}} \bigg[ \sum\nolimits_{i=1}^{\ell} & - \log p_{\theta}(y_i \vert x, y_1, ..., y_{i-1}) \\ 
& + \lambda_i \| \mathbf{z} (x, y_{1}, ..., y_{i-1}) \|_2 \bigg],
\end{split}
\end{align}
where $\lambda_i$ is a regularization coefficient that governs the regularization strength when predicting the $i$th entry of the sequence. 
The regularization $\lambda_i$ is designed to be {\it time dependent}, motivated by our observation that the baseline version of foCus suffers from different degrees of miscalibration at different entries.

A crucial challenge is how to select the value of this hyperparameter, given the large dimensionality of the hyperparameter space. 
We propose a selection procedure, based on the observation that miscalibration in the initial entries is propagated by the autoregressive structure of the simulator (see \Cref{sec:nn_simulator}). Consequently, optimizing the regularization parameter at the beginning of the sequence has more impact on the overall calibration performance of foCus (see \Cref{app:ablation} for additional analysis). The procedure is as follows:
\begin{enumerate}[leftmargin=5mm, topsep=0pt, itemsep=0pt, parsep=0pt]
    \item For $1 \leq i  \leq k_1$ (where $k_1$ is a hyperparameter) we use the sequence-level ECE of marginal probabilities (see \Cref{sec:evaluation_metrics}) computed over validation set to iteratively select $\lambda_i$, setting $\lambda_j=0$ for all $j > i$.
    \item For $k_1 < i  \leq k_2$ (where $k_2$ is a hyperparameter) we constrain all the parameters to equal the same constant, $\lambda_i = \lambda_{\operatorname{all}}$, selected also based on the validation ECE. 
    \item For $i>k_2$ we set $\lambda_i=0$.
\end{enumerate}
We set $k_1=3$ and performed hyperparameter optimization to select $k_2$, which typically resulted in small values (see \Cref{tab:hyperparameters}).
Details about the optimal regularization parameters and a comprehensive overview of the hyperparameter search process are provided in \Cref{app:hyperparameter_search}.

\vspace*{-3pt}
\subsection{Results}
\vspace*{-3pt}
\label{sec:results}

\paragraph{Marginal and conditional probabilities.}

\Cref{tab:last_epoch_score_uncondition} compares the sequence-level evaluation metrics for marginal probability estimation of foCus with (1) no regularization as described in \Cref{sec:pathology}, (2) our proposed time-dependent regularization described in \Cref{sec:regularization}, and (3) constant regularization where the regularization parameter in \Cref{eq:L2_Rg} is set to a single constant $\lambda_i=\lambda_{\text{const}}$ (determined based on validation ECE). 
All methods achieve similar AUCs in each dataset, indicating a similar discriminative ability. In contrast, the ECE is significantly lower for time-dependent regularization for all datasets, indicating better calibration performance. 
This results in better probability estimates, as evinced by the lower cross entropy (CE) and Brier scores (BS).
For FaceMed this is confirmed by comparing the estimated probabilities to the ground-truth marginal probabilities.
The unregularized baseline method and constant regularization yield RMSEs of $0.1847 \pm 0.0014$ and $0.1754 \pm 0.0028$, respectively, while time-dependent regularization reduces the RMSE to $0.1720 \pm 0.0009$ (see \Cref{app:additional_results_marginal}).
The same holds for conditional probability predictions, as reported in \Cref{tab:last_epoch_score_condition}:
Time-dependent regularization again significantly improves calibration, and as a result the overall probability estimates. \Cref{app:case_study} provides a detailed description of the conditional probability estimates for one of the video games.

\Cref{fig:reliability_diagram} further demonstrates the improvement in calibration provided by time-dependent regularization.
Interestingly, we observe a \emph{calibration propagation} phenomenon, where regularizing a small number of early entries produces improved calibration across the whole sequence. For example, for H.E.R.O. regularization is applied to the first 6 entries (0.2 seconds), yet the ECE improvement is evident up until entry 150 (5 seconds). %

\textbf{Time-to-event confidence intervals.}
\Cref{tab:last_epoch_length_score} compares the evaluation metrics for the confidence intervals produced by foCus, again with (1) no regularization, (2) time-dependent regularization, and (3) constant regularization.
In this case, we observe a certain trade-off between discriminative performance, quantified by the relative MAE, and calibration, quantified by coverage probabilities.
The MAE for models trained without or with constant regularization are consistently lower than those of time-dependent regularization, but the coverage probabilities of time-dependent regularization are a lot closer to 90\% (between 69\% and 92\%, compared to at most 70\% for the other two methods). 

\Cref{fig:CI_analysis} shows heatmaps of the confidence-interval widths for different ground-truth times (upper panel) for FaceMed and Seaquest (see \Cref{app:supp_figs} for additional plots), as well as a histogram with the fraction of intervals containing the ground-truth times.
We observe that unregularized simulator training, foCus produces narrow confidence intervals with poor coverage across the board, whereas time-dependent regularization yields intervals that tend to be larger when the true times are larger, achieving much better coverage.

%

%

%
%

%

%
\begin{comment}
    
\subsubsection{Comparison of marginal and conditional probability estimates}
\Cref{tab:last_epoch_score_uncondition} reveals that the constant regularization model achieves AUC scores for marginal probability prediction that are comparable to those of the vanilla model. Additionally, the ECE, BS, and CE metrics are all more favorable for the constant regularization model, indicating that it preserves discriminative power while improving calibration, thereby enhancing the overall quality of marginal probability prediction. However, the calibration and overall performance of the constant regularization model still fall short when compared to the time-dependent regularization model. Similar trends are observed for conditional probability prediction, as shown by Table~\ref{tab:last_epoch_score_condition}.
\end{comment}

%

%
\begin{comment}

\subsubsection{Comparison of time-to-event confidence intervals}
\Cref{tab:last_epoch_length_score} also shows that the constant regularization model achieves similar discriminative power in time-to-event prediction to the vanilla model, as indicated by the comparable relative MAE. Additionally, the constant regularization model demonstrates better calibration performance, reflected in its higher CI coverage probability. However, its time-to-event calibration remains inferior to that of the time-dependent regularization model, which achieves even higher CI coverage probability than the constant regularization model.
\end{comment}

\vspace*{-3pt}
\section{\uppercase{Discussion and Limitations}}
\vspace*{-3pt}

In this paper, we studied an important, yet underexplored topic: how to achieve reliable uncertainty quantification when predicting sequences from high-dimensional data.
We proposed a Monte Carlo framework based on learned autoregressive simulators that enables flexible estimation of probabilities and confidence intervals.
Our experiments on sequential decision-making tasks revealed that simulator models learned via maximum likelihood estimation can lead to severely miscalibrated uncertainty estimates.
We showed that this shortcoming can be addressed by training the autoregressive simulator model using a time-dependent regularizer, which we find consistently leads to well-calibrated uncertainty estimates.

Our proposed regularization is conceptually and mathematically simple but requires choosing a set of regularization coefficients $\{ \lambda_i \}_{i=1}^{l}$ from a combinatorially large space, making an exhaustive search infeasible in practice.
This is not unique to our approach: Real-world sequences often display non-stationary statistical properties that are difficult to model in a data-driven fashion.
Nevertheless, we find that our simple coefficient selection protocol leads to significant improvement in calibration, although more sophisticated strategies could well result in further gains.
Other potentially fruitful directions for future research are to perform uncertainty quantification of continuous-valued and spatiotemporal sequences in weather and climate applications---areas in which neural-network simulators are rapidly gaining popularity~\citep{pathak2022fourcastnetglobaldatadrivenhighresolution,Kochkov_2024,subel2024building}.

%% file: drafts/_aistats2025/related_work.tex
\section{\uppercase{Related Work}}
\label{sec:related_work}
\paragraph{Sequence generation.}
Discrete sequence generation plays a fundamental role in many natural language processing applications, such as language modeling~\citep{NEURIPS2020_1457c0d6,touvron2023llamaopenefficientfoundation}, image captioning~\citep{ghandi2023deep}, language translation~\citep{bahdanau2014neural, gehring2017convolutional, vaswani2017attention}, and text summarization~\citep{dong2018survey}. In these examples,  sequence generation is typically performed in an autoregressive manner, where each token is generated based on the tokens previously generated. Alternatively, sequences can also be generated non-autoregressively~\citep{sun2020approach, gu2017non, shu2020latent}, where tokens are produced simultaneously or with fewer dependencies on earlier tokens. In both paradigms, the primary objective is to generate the most likely sequence. However, these approaches typically do not focus on quantifying the uncertainty associated with generated sequences, which is the goal of this paper.

\paragraph{Imitation and reinforcement learning.}
Our focus is on \emph{predicting} sequences from high-dimensional data, which is fundamentally different from imitation learning~\citep{Hussein2017}, which seeks to replicate human behavior, and  reinforcement learning~\citep{sutton2018reinforcement}, which seeks to determine an optimal policy 
by allowing the agent to interact with the environment, guided by a reward function~\citep{schulman2017proximalpolicyoptimizationalgorithms,lillicrap2019continuouscontroldeepreinforcement}.

\paragraph{Calibration.}

Miscalibration is a well-known challenge in classification models based on deep learning~\citep{guo2017calibration, wang2024calibrationdeeplearningsurvey, wang2021rethinking}, particularly when the goal is to provide accurate uncertainty quantification~\citep{liu2022deep}. 
While significant progress has been made in calibrating unitask classification models, calibration in sequence prediction tasks remains underexplored and even lacks a clear definition. While \citet{marx2024calibratedprobabilisticforecastsarbitrary} explores calibration in sequences, their focus is on step-wise calibration, where each sequence step has an input-output pair, which makes the problem step-wise calibration within sequences, whereas our work focuses on calibration for the entire sequence. 
\citet{kuleshov2015calibrated} offers a framework for measuring calibration in structured high-dimensional random vectors via event pooling, which inspires our approach to uncertainty estimation in sequence prediction. 

Various methods have been developed to enhance calibration in classification, including post-processing the logits~\citep{gupta2021calibration,kull2017beyond,kull2019beyond}, ensembling methods~\citep{zhang2020mix, lakshminarayanan2017simple, maddox2019simple}, soft labeling~\citep{mukhoti2020calibrating, szegedy2016rethinking, zhang2017mixup, thulasidasan2019mixup, liu2022deep} and training with regularization~\citep{pereyra2017regularizingneuralnetworkspenalizing, kumar2018trainable, rudner2023fseb}. In this work, we propose a time-dependent regularization method to improve calibration specifically in sequence prediction tasks.

\vspace{-1mm}
\paragraph{The Monte Carlo method in deep learning.}
The Monte Carlo method is used for generating ensembles in Bayesian deep learning from samples of the model parameters \citep{blundell2015weight, gal2016dropoutbayesianapproximationrepresenting}. In contrast, in our proposed framework we sample sequences using a neural-network simulator with fixed parameters, and apply the Monte Carlo method to the sampled sequences. 
Monte Carlo methods are also used for uncertainty estimation on large language model outputs \citep{malinin2021uncertainty, jiang-etal-2021-know, kuhn2023semantic, xiong2024can}, where multiple sequences are generated to assess the confidence of factual outputs and mitigate hallucinations. Unlike these works, our goal here is not to determine the most likely sequence, but rather to provide a probabilistically-accurate characterization of possible future sequences.

%% file: drafts/_aistats2025/acknowledgements.tex
\section*{Acknowledgments}
W.Z. is supported by the National Institute On Aging of the National Institutes of Health under Award R01AG079175, Award R01AG085617, and NSF grant NRT-1922658. C.F.G. was partially supported by NSF grant DMS 2009752.

%% file: drafts/_aistats2025/appendices.tex
\begin{appendix}
    
\crefalias{section}{appsec}
\crefalias{subsection}{appsec}
\crefalias{subsubsection}{appsec}

\onecolumn

\section*{\LARGE Appendix}

The appendix is organized as follows:
\begin{itemize}
    \item In \Cref{app:metrics}, we define the metrics we used to evaluate  marginal and conditional probability estimation (\Cref{app:metrics_prob}) and confidence interval estimation (\Cref{app:metrics_ci}).
    \item In \Cref{app:dataset_details}, we provide additional information about the datasets utilized in this study.
    \item In \Cref{app:technical_details}, we include additional details regarding the model and hyperparameter search. 
    \item In \Cref{app:results}, we include additional results. \Cref{app:additional_results_marginal} reports results for marginal probability estimation with comprehensive metrics and more reliability diagrams. We also include comparisons with the ground truth probabilities for the synthetic FaceMed dataset. In \Cref{app:additional_results_conditional}, we report results for conditional probability estimation. In \Cref{app:additional_results_ci}, we report results for confidence interval estimation.
    \item In \Cref{app:ablation},  we provide a sensitivity analysis of the effect of regularization applied at different entries in a sequence.
    \item In \Cref{app:case_study}, 
    we present an example illustrating how the probabilities estimated by foCus change when we condition on a certain event. 
    \item In \Cref{app:supp_figs}, we show the evolution of the performance of foCus as we train the simulator with different types of regularization.
\end{itemize}

\section{Evaluation metrics}
\label{app:metrics}

\subsection{Marginal and conditional probability}
\label{app:metrics_prob}
For each class $a\in \{1, \cdots, c\}$ and each entry $i$ of the sequence, we evaluate the estimated probabilities $\operatorname{P}({{Y}}_i=a \, | \, {{X}}=x)$ or $\operatorname{P}({{Y}}_i=a \, | \, {{Y}}_j=b, {{X}}=x)$ using the relevant data (for marginal probabilities, these are all sequences; for conditional probabilities, all sequences such that the $j$th entry equals $b$). 
The entry-level metrics we propose are defined below. The sequence-level metrics are computed by averaging the entry-level metrics across all entries.

\paragraph{Macro AUC.} The Area Under the Curve (AUC) per class is computed separately for each class $a$ at each entry $i$ using a one-vs-all approach. We aggregate all the class AUCs via averaging to obtain the overall macro AUC.

\paragraph{Brier Score}
The Brier Score (BS) evaluates both calibration and discriminative ability. The Brier Score per class is the mean-squared error between the predicted probabilities and binarized label per class. The entry-level BS is the mean of BS per class, averaged over all the classes.
\paragraph{Cross Entropy} The cross-entropy (CE) loss is computed following \eqref{eq:MLE}.

\paragraph{Expected Calibration Error.}

 We use confidence expected calibration error (ECE) \citep{guo2017calibration} to assess calibration. The \textit{confidence} is defined as the predicted probability of the class $a$ with the highest estimated probability. These confidences are grouped into $B$ bins, based on $B$-quantiles. ECE is the mean absolute difference between the accuracy (empirical probability of correct predictions) and the average confidence within each bin. A lower ECE indicates better calibration. 

 To provide further insight, we also plot reliability diagrams. These diagrams compare the empirical probability (accuracy) with the estimated probability (confidence) in each bin. A well-calibrated model will produce a reliability diagram that is close to the diagonal.

 %
 \begin{comment}
 all predicted probabilities and labels are concatenated across classes, yielding $\Pi_i = \text{CONCAT}(\Pi_{1,i}, \cdots, \Pi_{c,i})$ and  $Y_i = \text{CONCAT}(Y_{1,i}, \cdots, Y_{c,i})$. We partition $\Pi_i$ into $B$ bins, where $Q_1$, \ldots, $Q_{B-1}$ are the $B$-quantiles. For each bin  $I_b:=[Q_{b-1}, Q_b] \cap \Pi_i$ (setting $Q_0=0$), we compute the mean empirical and predicted probabilities, denoted as $p_\text{emp}^{(b)}$ and $p^{(b)}$, respectively:
\begin{align}
    p_\text{emp}^{(b)} & = \frac{1}{|I_b|}\sum\limits_{k\in \text{Index}(I_b)} Y_i[k],\\
    p^{(b)}& =\frac{1}{|I_b|}\sum\limits_{k\in \text{Index}(I_b)} \Pi_i[k],
\end{align}
where $\text{Index}(I_b) = \{k \mid \Pi_i[k] \in I_b \}$. The pairs $(p^{(b)},p_\text{emp}^{(b)})$ are then plotted as a reliability diagram. Expected calibration error (ECE) is then calculated as: 
\begin{equation}
    \text{ECE} = \frac{1}{B} \sum_{b=1}^{B} \left\vert p_\text{emp}^{(b)} -  p^{(b)}\right\vert.
\end{equation}
\end{comment}

\subsection{Confidence intervals}
\label{app:metrics_ci}

Let the ground truth time-to-event for the $k$th data point be denoted as $T[k]$, and the estimated confidence interval as ${I_\alpha}[k] = \left[{q_{(1-\alpha)/2}}[k], {q_{(1+\alpha)/2}}[k]\right]$.

\paragraph{Coverage Probability} The coverage probability measures the proportion of samples where the true time-to-event $T[k]$ lies within the estimated confidence interval ${I_\alpha}[k]$. 
This metric reflects how well calibrated the estimated confidence intervals are. Ideally, for a confidence level $\alpha$, the coverage probability should equal $\alpha$.

\paragraph{Relative Confidence Interval Width}
The width of the confidence interval quantifies the uncertainty in the model estimates. A wider confidence interval indicates higher uncertainty. To account for different sequence lengths across datasets, we normalize the confidence interval width by the mean of the true time-to-event averaged over each dataset. The relative confidence interval width is defined as:
\begin{equation}
    \frac{ \frac{1}{N}\sum_{k=1}^N \left({q_{(1+\alpha)/2}}[k] - {q_{(1-\alpha)/2}}[k]\right)}{\frac{1}{N}\sum_{k=1}^N T[k]},
\end{equation}
where $N$ is the number of data.
\paragraph{Relative Mean Absolute Error (MAE)} The relative MAE is the mean of the absolute difference between the estimated and ground truth time-to-event, normalized by the mean of the ground truth times-to-event. We estimate the time-to-event by averaging over the time-to-event values $\widehat{T}^{(1)}$, ..., $\widehat{T}^{(m)}$  corresponding to the $m$ Monte Carlo simulations:
\begin{equation}
\text{Relative MAE} = \frac{\sum_{k=1}^N\left\vert T[k] - \frac{1}{M}\sum_{m=1}^M \widehat{T}^{(m)}[k]\right\vert}{\sum_{k=1}^N T[k]}.
\end{equation}

\section{Datasets}
\label{app:dataset_details}

\paragraph{FaceMed}
FaceMed is a synthetic dataset based on the UTKFace dataset \citep{utkface}, which contains face images along with corresponding ages. We simulate health status transitions between three distinct states: 1 for healthy, 2 for ill, and 3 for dead. The simulated health states per each year form a sequence for each patient. The underlying transition probabilities among these health states are determined by the individual’s age. The goal of the sequence prediction task is to forecast a patient’s health status trajectory using their facial images.

\begin{figure}[htbp]
    \centering
    \begin{subfigure}[b]{0.3\textwidth}
        \centering
        \includegraphics[width=\textwidth]{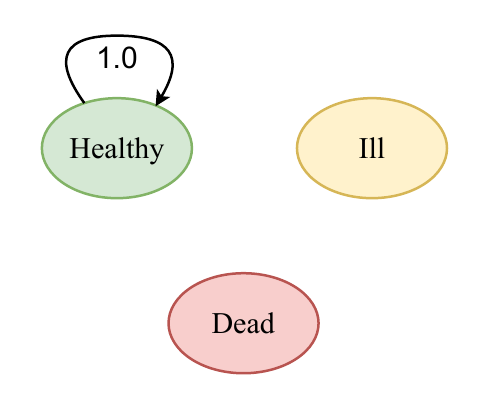}
        \caption{$0 - 40$ years old}
    \end{subfigure}
    \hfill
    \begin{subfigure}[b]{0.3\textwidth}
        \centering
        \includegraphics[width=\textwidth]{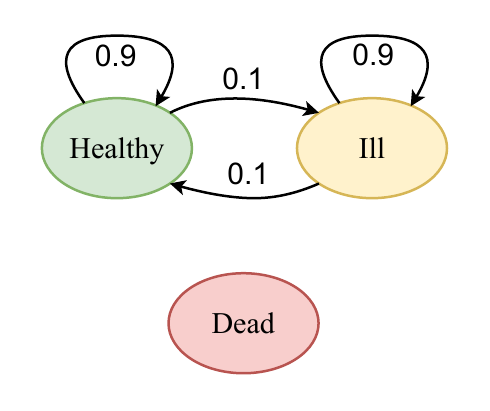}
        \caption{$40 - 80$ years old}
    \end{subfigure}
    \hfill
    \begin{subfigure}[b]{0.3\textwidth}
        \centering
        \includegraphics[width=\textwidth]{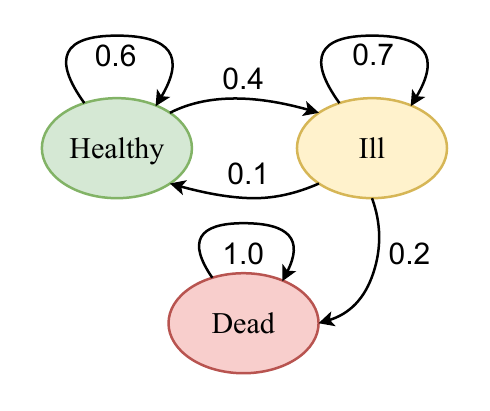}
        \caption{$> 80$ years old}
    \end{subfigure}

    \caption{\textbf{Markov process used to simulate health-status transitions in FaceMed.}}
    \label{fig:mc_diagram}
\end{figure}

To simulate the dynamics of health status, we use an age-dependent Markov process, where the health status at the $i$th entry, ${{Y}}_i$, only depends on the previous state ${{Y}}_{i-1}$, for any $i>1$. The conditional probability between states is given by:
\begin{equation}
    \operatorname{P}({{Y}}_i=a \, | \, {{Y}}_{i-1}=b) = p_{b,a}.
\end{equation}
The transitions among health statuses are illustrated in \Cref{fig:mc_diagram}. Every individual is healthy as the initial state. The transition probabilities for the simulation are defined as follows: For individuals younger than 40, the health status never changes (\Cref{fig:mc_diagram} (a)). For those aged 40 to 80, the health status can transition between healthy and ill, with transition probabilities $p_{1,1} = p_{2,2} = 0.9, \ p_{1,2} = p_{2,1} = 0.1$ (\Cref{fig:mc_diagram} (b)). For individuals older than 80, the likelihood of becoming ill increases, with transition probabilities $p_{0, \cdot} = (0.6, 0.4, 0)$; for ill individuals in this age group, there is a chance of death, reflected in the transition probabilities $p_{1, \cdot} = (0.1, 0.7, 0.2)$ (\Cref{fig:mc_diagram} (c)). These probabilities vary as the individual becomes ``older'' in the simulation, reflecting the increasing risks associated with aging. The average survival time of the simulated sequences is 45.45 years. In the experiments, For individuals who die before 100 years from the beginning, their sequences are padded to cover 100 years. The dataset is split into training, validation, and test sets with 16641, 4738, and 2329 samples, respectively.

\paragraph{Atari games}
Our real-data experiments are based on the Atari-HEAD dataset \citep{Zhang_2020}, a large-scale, high-quality imitation learning dataset that captures human actions alongside eye movements and game frames while playing Atari video games. 
The dataset employs a unique semi-frame-by-frame gameplay format, where the game pauses at each frame until the player performs a keyboard action. This ensures that each frame in the video of game and the corresponding human action are  aligned.

In this work, the sequence prediction task aims to predict a player’s action trajectory based on a given game frame. Each frame serves as a high-dimensional input ${{X}}$. The subsequent actions until the next scoring event are treated as a sequence ${{Y}}$. Since actions are recorded at a high frame-by-frame frequency, they often repeat several times before transitioning to a new action, yielding sequences with redundant information. To reduce this redundancy, we sample actions at a constant frequency determined by the number of frames per sequence entry for each game. The corresponding time $t$ in the game at the $i$th entry of the sequence can be recovered from the entry as follows: $t(\text{second}) = \text{sampling freq.} / 60 (\text{Hz}) \times i$. We pad the sequences with an \emph{end-of-game} value to ensure that all sequences have the same length.

Experiments are conducted on five games from Atari-HEAD: \textbf{Seaquest}, \textbf{River Raid}, \textbf{Bank Heist}, \textbf{H.E.R.O.}, and \textbf{Road Runner}. They represent a broad category of video games available in Atari-HEAD. We use 70 percent of gameplay as training set, 20 percent as the validation set, and 10 percent as the test set. See \Cref{tab:data_config_game} for more detailed information about the data corresponding to each game. 

\begin{table*}[t]
\caption{\textbf{Additional information about video game data.} The table summarizes each video game dataset's training set size, validation set size, test set size, average sequence length until the next scoring point, sequence length after padding, and action sampling frequency.}
\label{tab:data_config_game}
\begin{center}
\small
\begin{tabular}{lcccccc}
\toprule
Game Name & Training & Validation & Testing & Ave. Length (seconds) & Padded Length & Sampling freq.  \\
\midrule
Seaquest & 104,806 & 17,556 & 17,650 & 1.0071  & 200 & 3\\
River Raid & 104,252 & 17,584 & 17,592 & 1.4082 & 300 & 2\\
Bank Heist & 105,765 & 17,553 & 17,548 & 1.7645 & 300 & 2\\
H.E.R.O. & 98,553 & 16,769 & 16,885 & 1.4586 & 300 & 2\\
Road Runner & 239,428 & 68,332 & 34,177 & 2.5796 & 300 & 4\\
\bottomrule
\end{tabular}
\end{center}
\end{table*}

\begin{comment}
\begin{table*}[t]
\caption{\textbf{Technical details for video game datasets.} The table summarizes each video game dataset's training set size, validation set size, test set size, average sequence length until the next scoring point, sequence length after padding, and action sampling frequency.}
%
\label{tab:data_config_game}
\begin{center}
\small
\begin{tabular}{lcccccc}
\toprule
Game Name & Training & Validation & Testing & Ave. Length & Padded Length & Sampling freq.  \\
\midrule
Seaquest & 104,806 & 17,556 & 17,650 & 20.1429  & 200 & 3\\
River Raid & 104,252 & 17,584 & 17,592 & 42.2457 & 300 & 2\\
Bank Heist & 105,765 & 17,553 & 17,548 & 52.9349 & 300 & 2\\
H.E.R.O. & 98,553 & 16,769 & 16,885 & 43.7571 & 300 & 2\\
Road Runner & 239,428 & 68,332 & 34,177 & 38.6935 & 300 & 4\\
\bottomrule
\end{tabular}
\end{center}
\end{table*}
\end{comment}

\section{Technical Details}
\label{app:technical_details}

\subsection{Model Architecture}
\label{app:model_architecture}
As illustrated in \Cref{fig:methodology} (a), we employ a neural-network simulator. The simulator consists of a convolutional neural network (CNN) that encodes the input image $x$, and a recurrent neural network (RNN). The CNN encoder is a Resnet-18~\citep{he2015deepresiduallearningimage}, which produces an image embedding of dimension $256$. The RNN decoder is implemented as a single-layer LSTM with a hidden vector size $256$. The image embedding from the CNN encoder is fed to the RNN decoder as the initial hidden vector $h_0$. The RNN iteratively updates its hidden state $h_i$ based on the previous hidden state $h_{i-1}$ and the preceding input $y_{i-1}$ from the $(i-1)$th entry of the sequence. For each step $i > 1$, the RNN outputs a logit vector $\mathbf{z}(x, y_1, \dots, y_{i-1}) \in \mathbb{R}^c$ through a linear layer with input $h_i$. The logit vector is normalized with a softmax function and used to estimate the class probabilities of a multinoulli distribution:
\begin{equation}
    p_{\theta}\left(a \, | \, x, y_1, ..., y_{i-1}\right) = \frac{\exp\left(\mathbf{z} (x, y_{1}, ..., y_{i-1})[a]\right)}{\sum_{k=1}^{c}\exp(\mathbf{z} \left(x, y_{1}, ..., y_{i-1})[k]\right)},
\end{equation}
where $a \in \{1, \dots, c\}$ and $\mathbf{z} (x, y_{1}, ..., y_{i-1})[k]$ is the $k$th entry of the logit vector.

For the health-status prediction task, the RNN decoder outputs a 3-dimensional logit vector corresponding to the three possible health states ($c=3$). In the case of Atari games, where each action can belong to one of 19 possible classes, the RNN generates a 19-dimensional logit vector ($c=19$).

\subsection{Training}
\label{app:model_training}
As explained in \Cref{sec:regularization}, the neural network simulator is trained by minimizing the cross-entropy loss between the predicted distribution and the one-hot encoded ground truth for each variable, with an additional regularization term that penalizes the $\ell_2$-norm of each entry in the logit vector. 

We train each model for 200 epochs for each scenario, with a batch size of 256, using the Adam optimizer without weight decay. The learning rates are kept constant for each scenario: $1\times 10^{-5}$ for Seaquest, River Raid, Bank Heist, and Road Runner, and $5\times 10^{-5}$ for H.E.R.O.

Model selection during training is challenging due to the numerous metrics involved in probability estimation tasks.  Figures~\ref{fig:length_score_game}, \ref{fig:step_average_score_uncondition_game}, and \ref{fig:step_average_score_condition_game} show the evolution of different metrics during training. We observe that the models are most discriminative (lower relative MAE and higher AUC) toward the end of training in most scenarios. %

\subsection{Inference}
\label{app:model_inference}
\Cref{fig:methodology}(b) illustrates how the neural-network simulator is used to obtain a sample sequence $(\hat{y}_1, \dots, \hat{y}_{\ell})$. The input image $x$ is fed into the CNN encoder, producing a hidden vector $h_0$ that is fed into the RNN decoder to then generate the simulated sequence iteratively. At each iteration $i \in \{1,...,\ell\}$, the input of the RNN decoder is the value $\hat{y}_{i-1}$ of the previous entry (except for $i=1$) and the hidden vector $h_{i-1}$. The outputs are an estimate of the conditional distribution of ${{Y}}_{i}$ given the previous entries and ${{X}}=x$, and an updated hidden vector $h_i$. The $i$th entry $\hat{y}_i$ of the sample sequence is sampled from this conditional distribution. 
Since each entry is drawn randomly from the predicted distribution, the simulator is capable of generating multiple different sequences from the same input image, acting as a simulator. When performing Monte Carlo estimation, we generate $m=100$ sampled sequences for each input image. 

\subsection{Hyperparameter search}
\label{app:hyperparameter_search}
The hyperparameters for time-dependent regularization were determined via the following procedure:
\begin{enumerate}[leftmargin=5mm, topsep=0pt, itemsep=0pt, parsep=0pt]
    \item For $1 \leq i  \leq k_1$ (where $k_1$ is a hyperparameter) we use the sequence-level ECE of marginal probabilities (see \Cref{sec:evaluation_metrics}) computed over validation set to iteratively select $\lambda_i$, setting $\lambda_j=0$ for all $j > i$.
    \item For $k_1 < i  \leq k_2$ (where $k_2$ is a hyperparameter) we constrain all the parameters to equal the same constant, $\lambda_i = \lambda_{\operatorname{all}}$, selected also based on the validation ECE. 
    \item For $i>k_2$ we set $\lambda_i=0$.
\end{enumerate}
We set $k_1=3$. To determine each $\lambda_i$ and $\lambda_{\operatorname{all}}$ we performed a search on the fixed grid $\{0.001, 0.005, 0.01, 0.05\}$ based on the validation ECE for the marginal probability estimation task. For $k_2$ we used the grid $\{1, 11, 21, 51, 101\}$.

For constant regularization, we constrain all the parameters to be the same, $\lambda_i=\lambda_{\text{const}}$. Then $\lambda_{\text{const}}$ we performed a search on the grid $\{0.001, 0.005, 0.01, 0.05\}$, also based on validation ECE. The hyperparameters chosen for both regularization methods are listed in \Cref{tab:hyperparameters}.

\begin{table*}
\caption{\textbf{Chosen regularization hyperparameters.} This table shows the chosen regularization parameters in all scenarios for both time-dependent and constant regularization.}
\label{tab:hyperparameters}
\begin{center}
\small
\begin{tabular}{lll}
\toprule
Scenarios & Time-dependent $\lambda$'s & Constant $\lambda$'s\\
\midrule
Seaquest & $\lambda_{1:3}=0.05, 0.01, 0.05$ \  $\lambda_{4:200}=0$ & $\lambda_{1:200}=0.001$\\
River Raid & $\lambda_{1:6}=0.01$ \  $\lambda_{7:300}=0$ & $\lambda_{1:300}=0.001$ \\
Bank Heist & $\lambda_{1}=0.05$ \  $\lambda_{2:11}=0.01$ \  $\lambda_{12:300}=0$ & $\lambda_{1:300}=0.001$ \\
H.E.R.O. & $\lambda_{1}=0.01$ \  $\lambda_{2:6}=0.005$ \  $\lambda_{7:300}=0$ & $\lambda_{1:300}=0.001$\\
Road Runner & $\lambda_{1}=0.01$ \  $\lambda_{2:21}=0.005$ \  $\lambda_{22:300}=0$ & $\lambda_{1:300}=0.001$\\
FaceMed &  $\lambda_{1:3}=0.01$ \ $\lambda_{4:5} = 0.005$ \ $\lambda_{5:50} = 0.001$ & $\lambda_{1:300}=0.001$ \\ 
\bottomrule
\end{tabular}
\end{center}
\end{table*}

\section{Supplementary Experimental Results}
\label{app:results}
\subsection{Marginal Probability}
\label{app:additional_results_marginal}
\begin{figure*}
    \centering
    \includegraphics[height=0.2\linewidth]{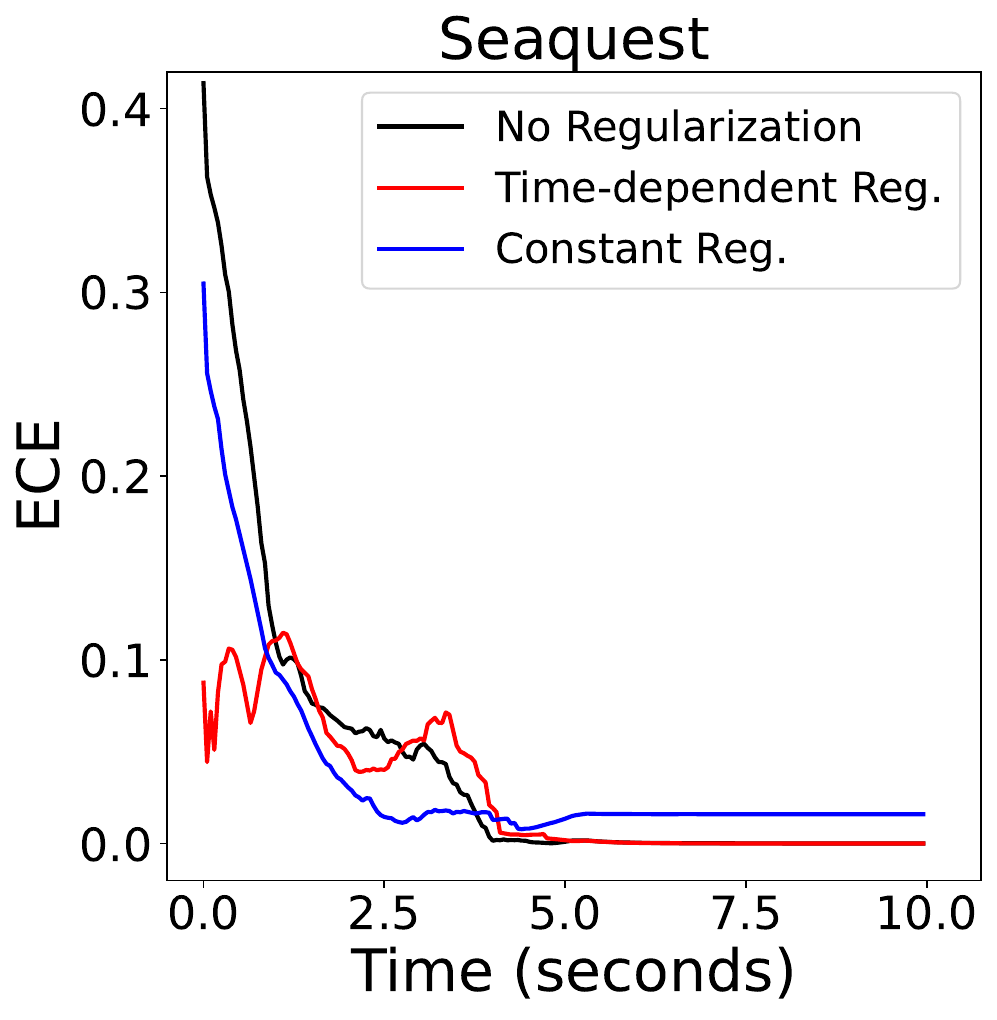}
    \includegraphics[height=0.2\linewidth]{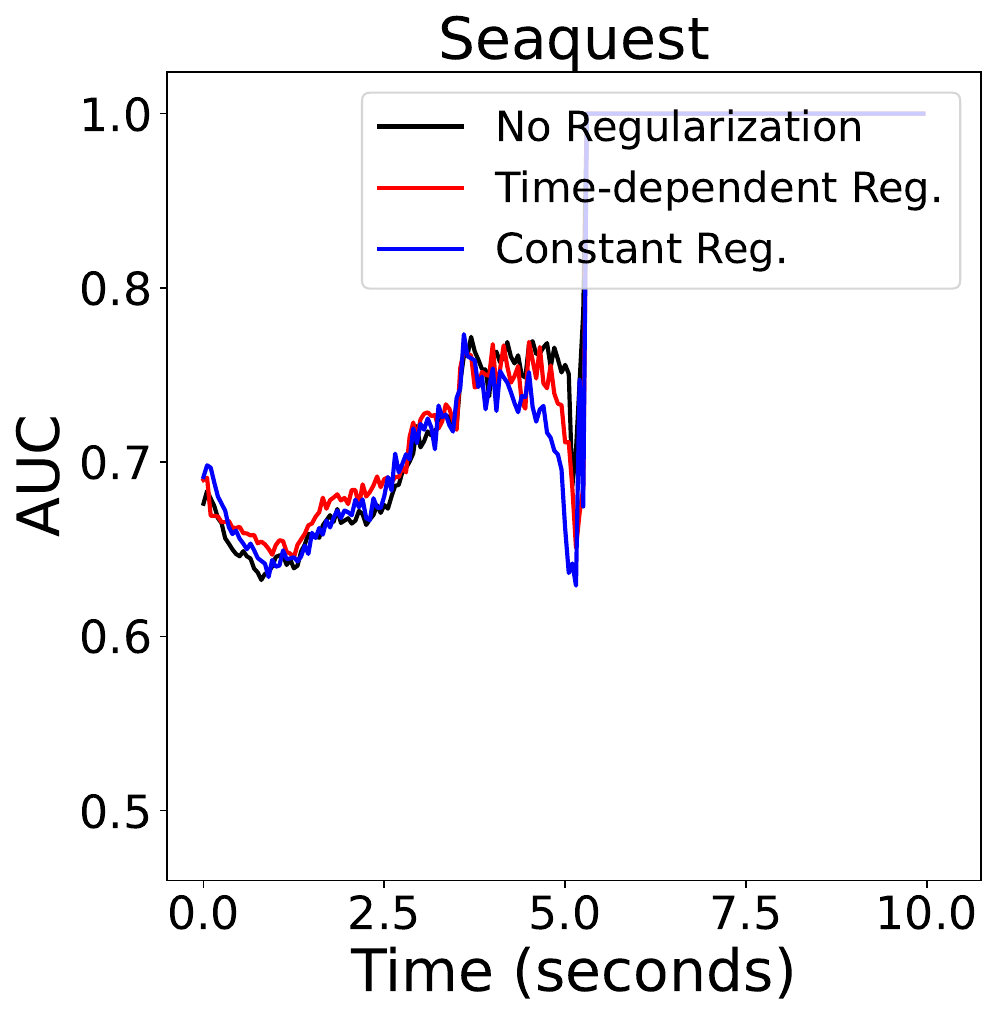}
    \includegraphics[height=0.2\linewidth]{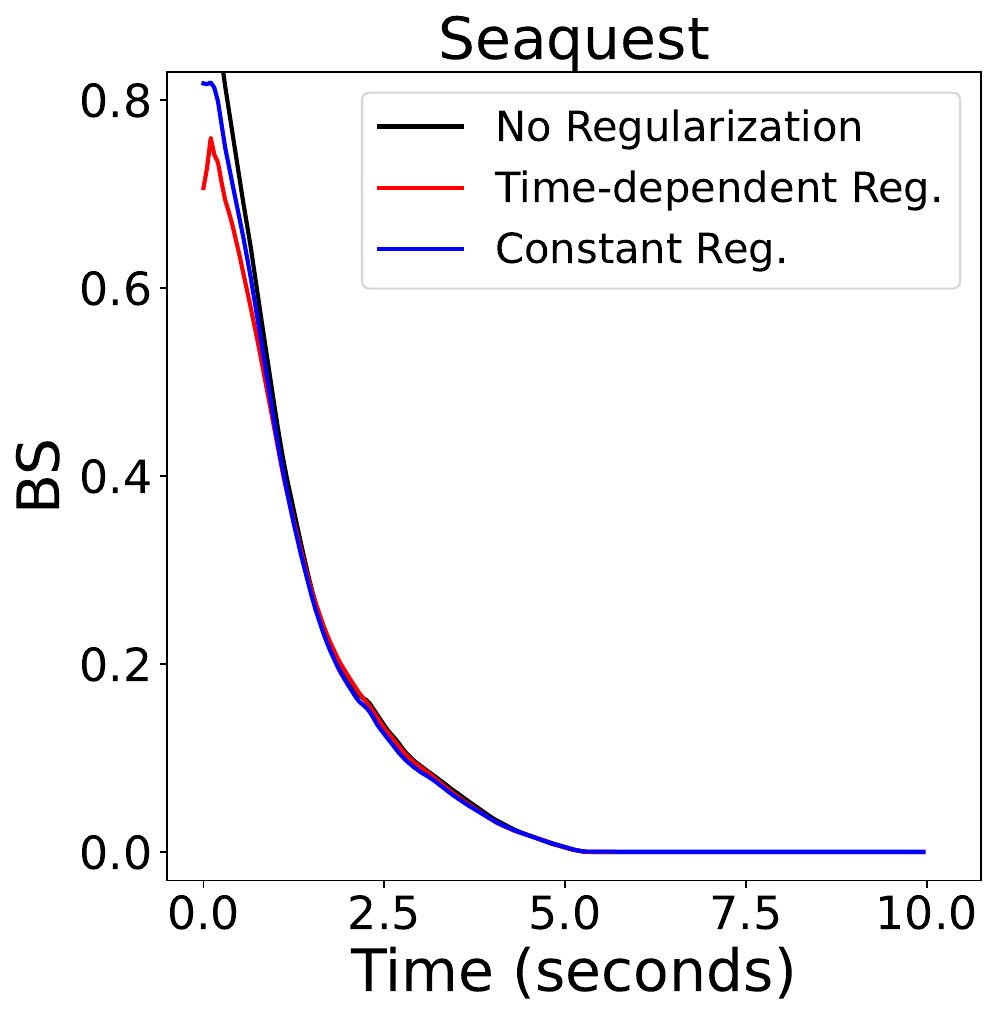}
    \includegraphics[height=0.2\linewidth]{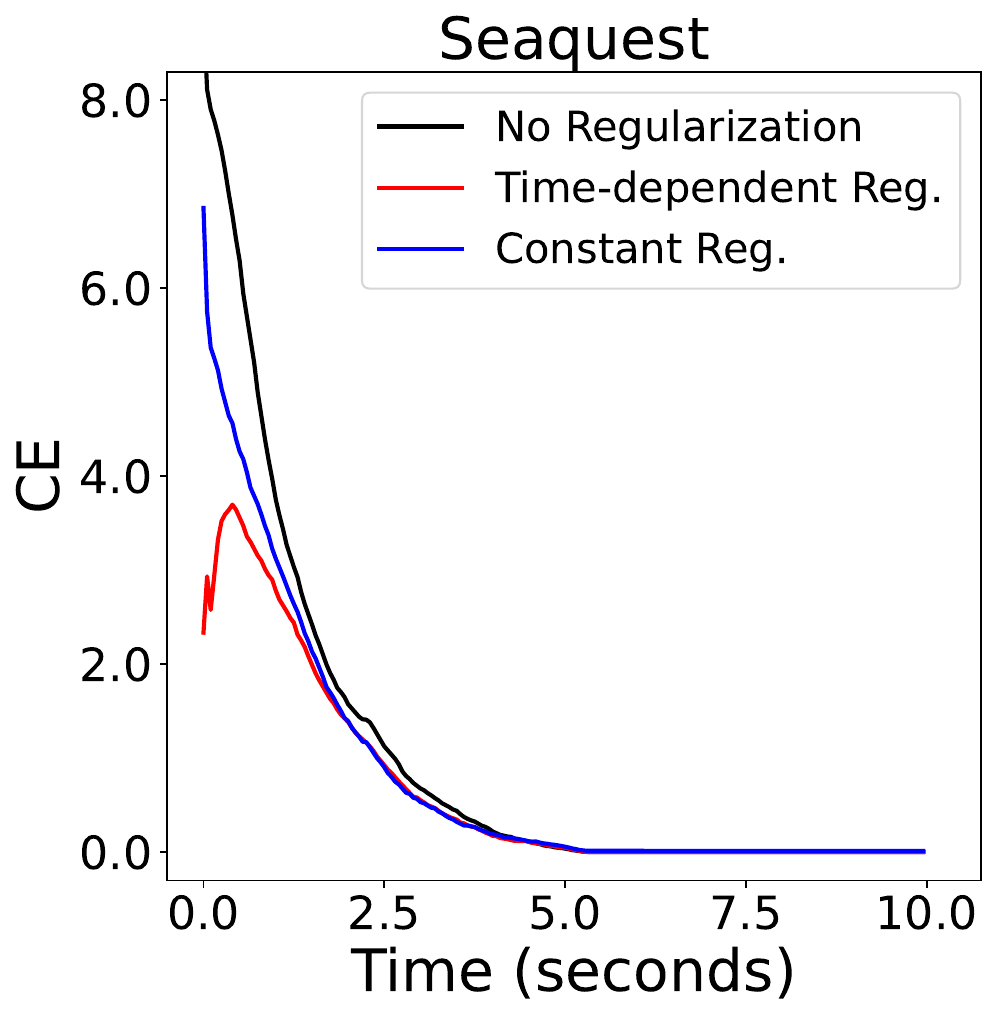} \\

    \includegraphics[height=0.2\linewidth]{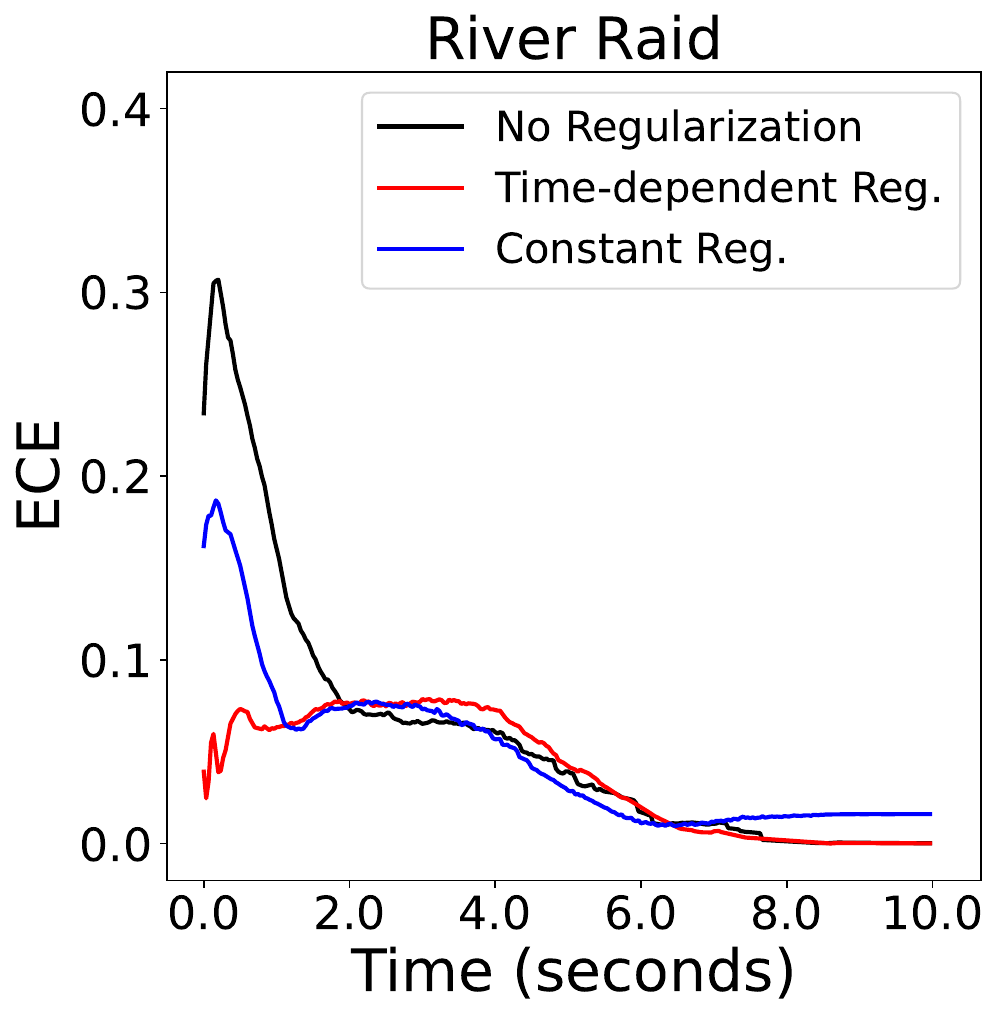}
    \includegraphics[height=0.2\linewidth]{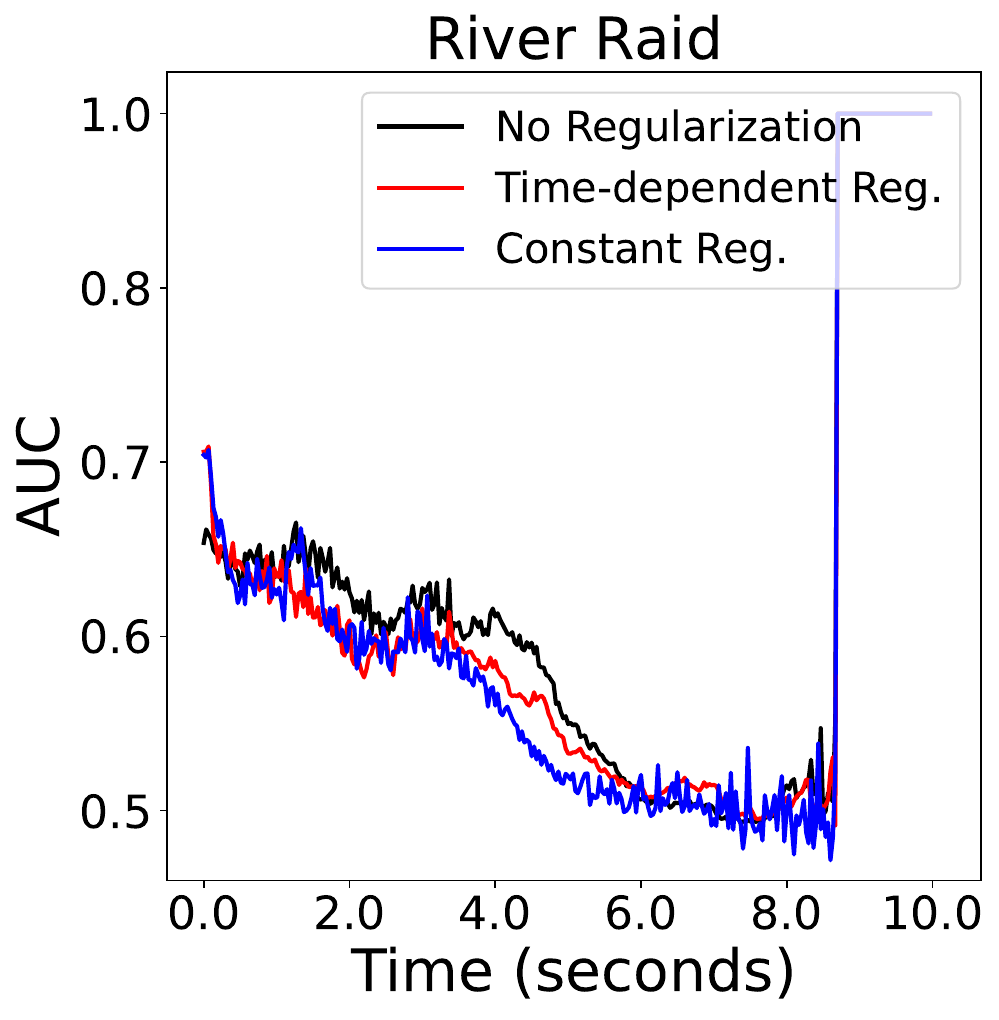}
    \includegraphics[height=0.2\linewidth]{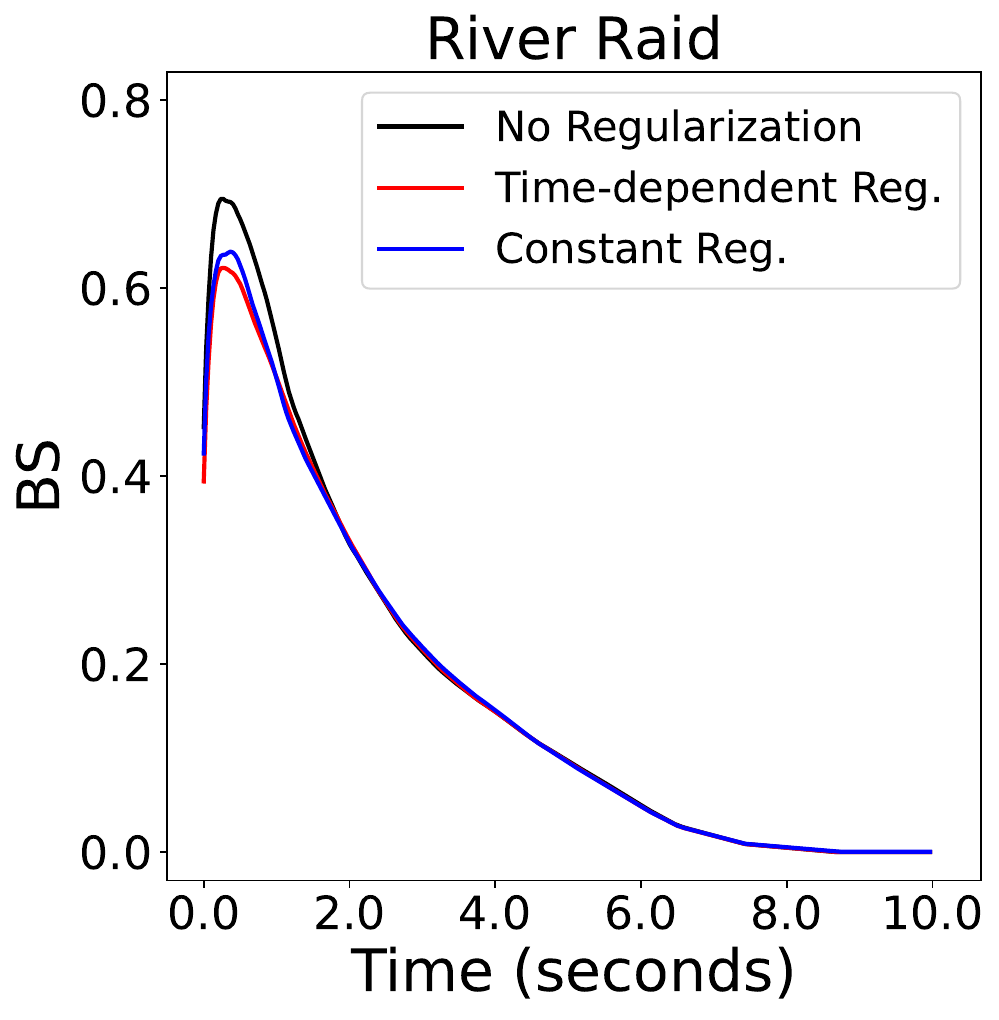}
    \includegraphics[height=0.2\linewidth]{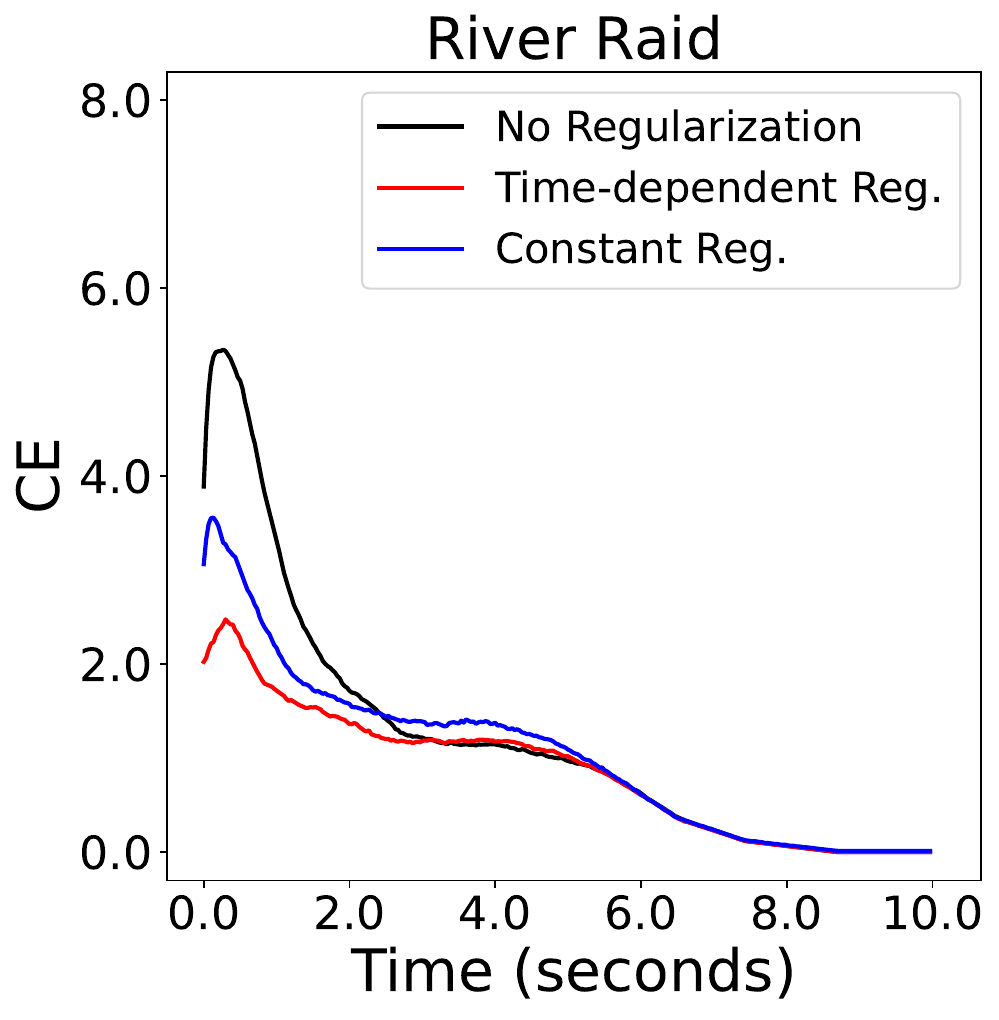} \\

    \includegraphics[height=0.2\linewidth]{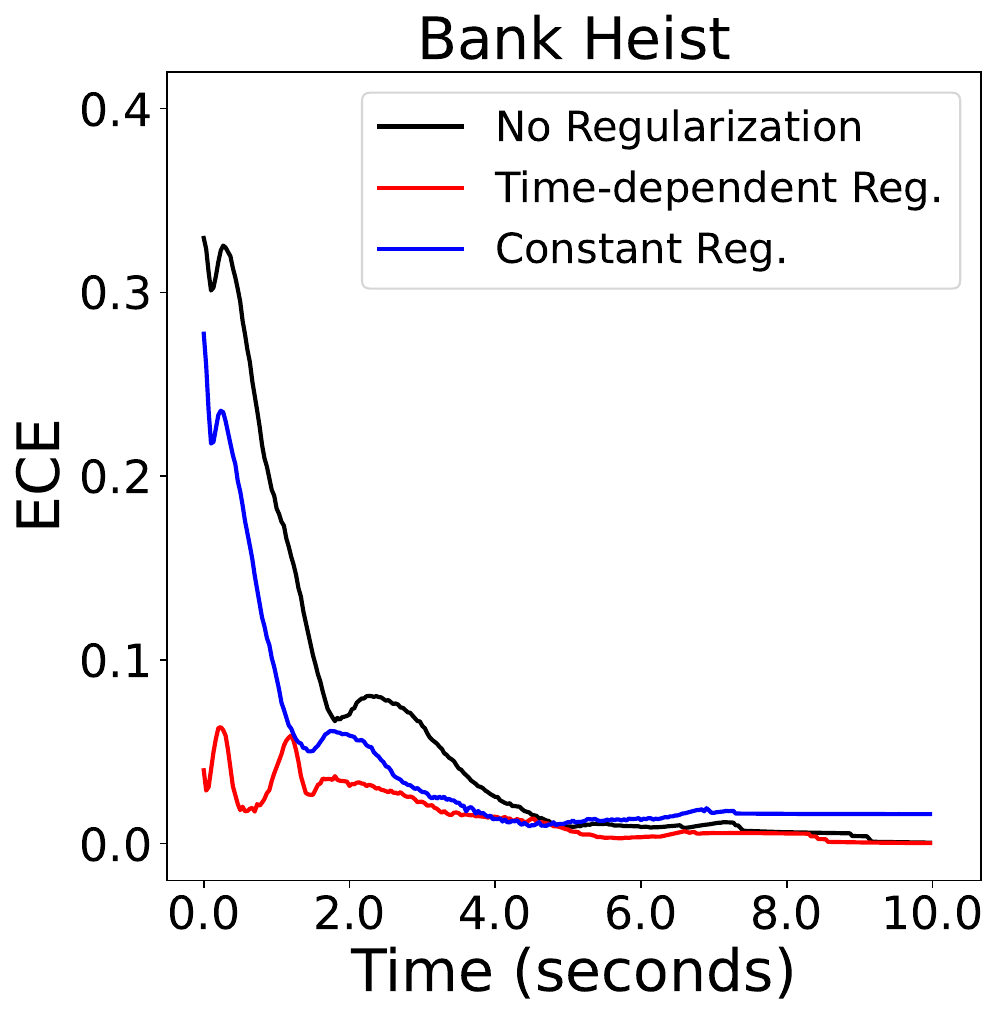}
    \includegraphics[height=0.2\linewidth]{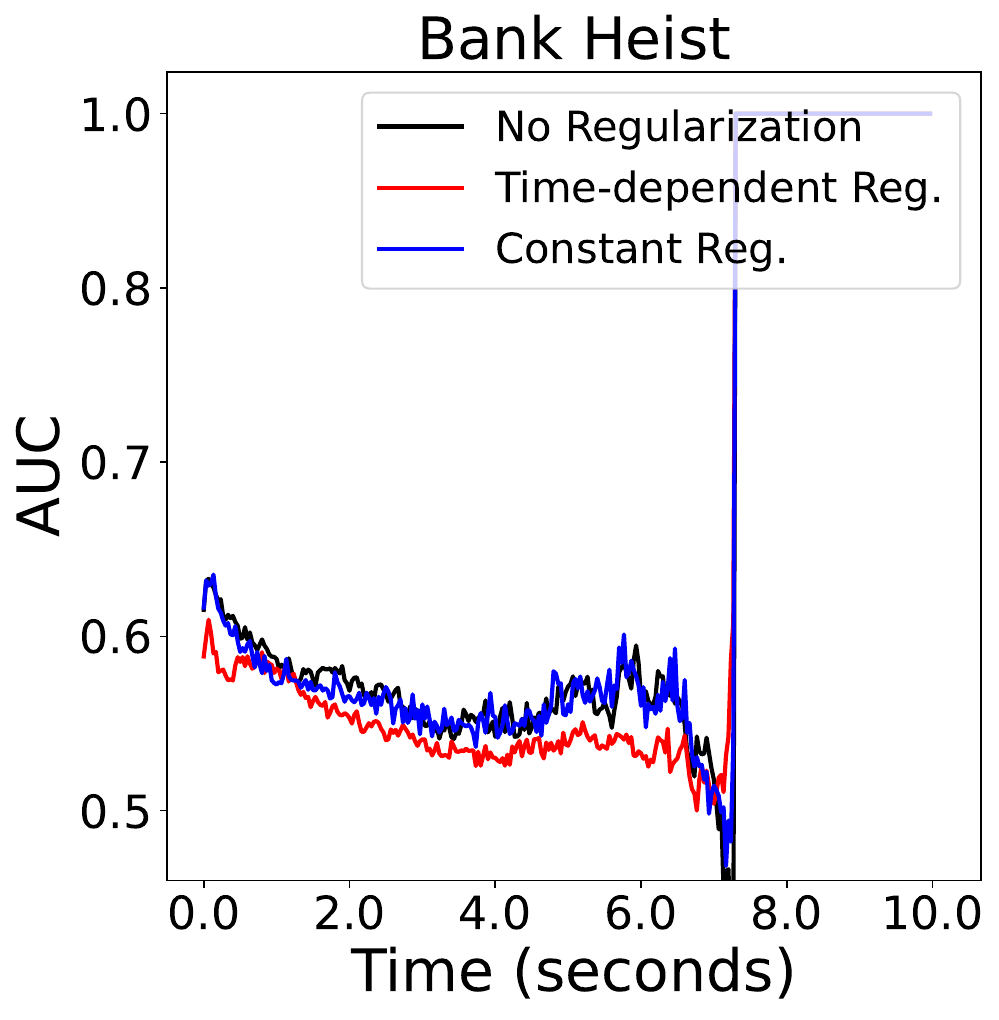}
    \includegraphics[height=0.2\linewidth]{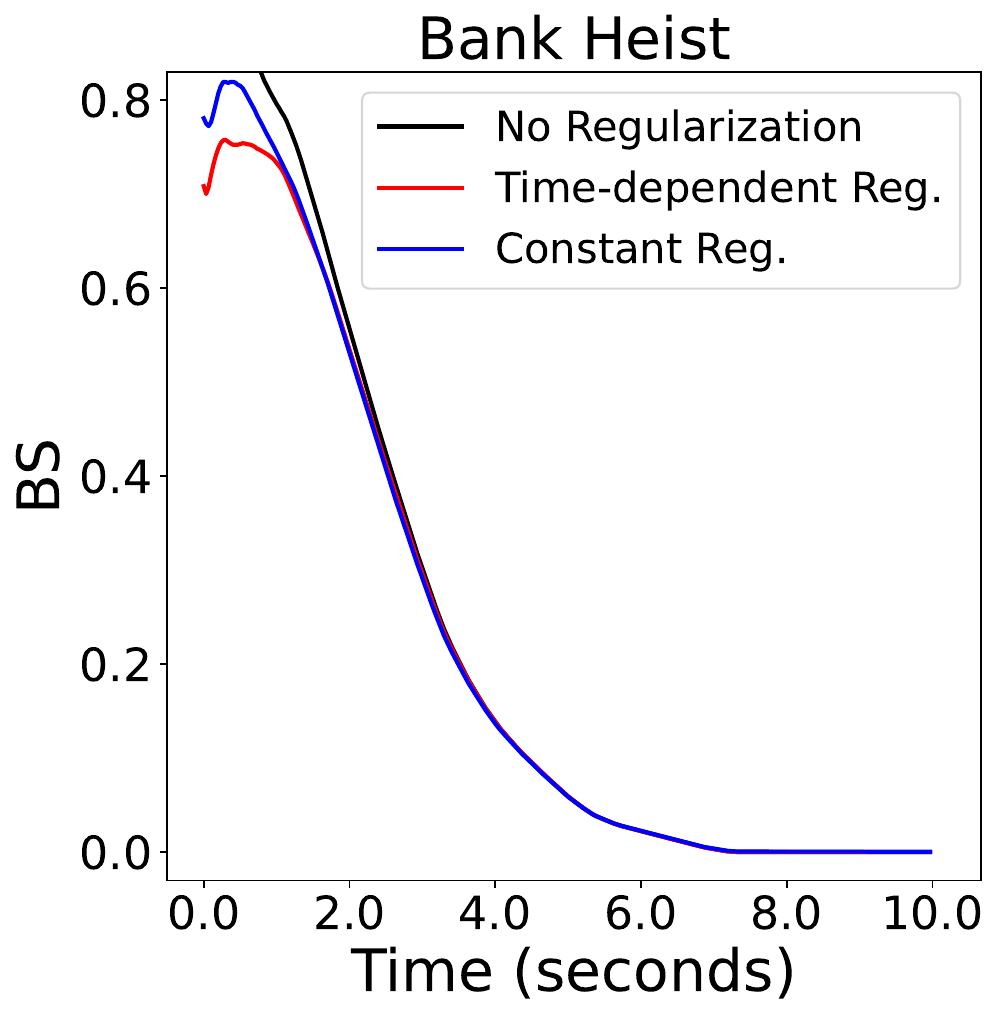}
    \includegraphics[height=0.2\linewidth]{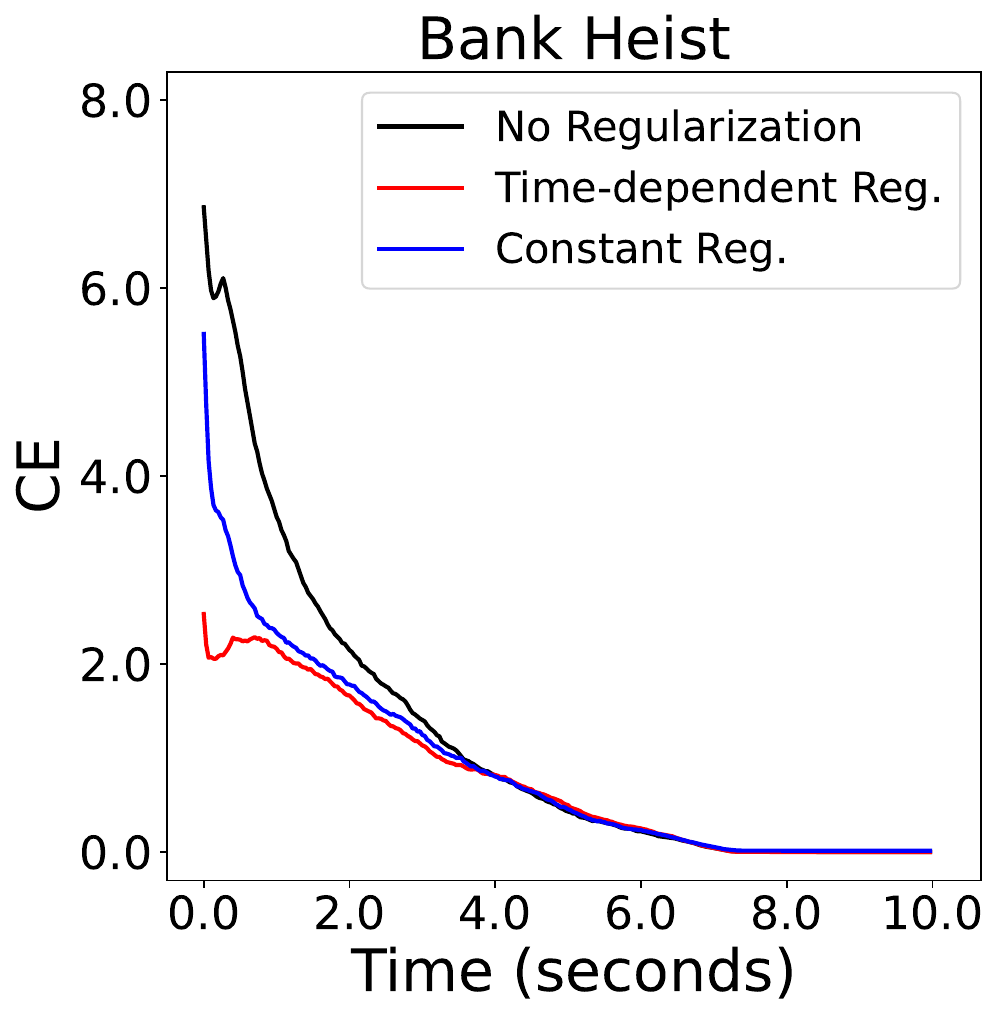} \\

    \includegraphics[height=0.2\linewidth]{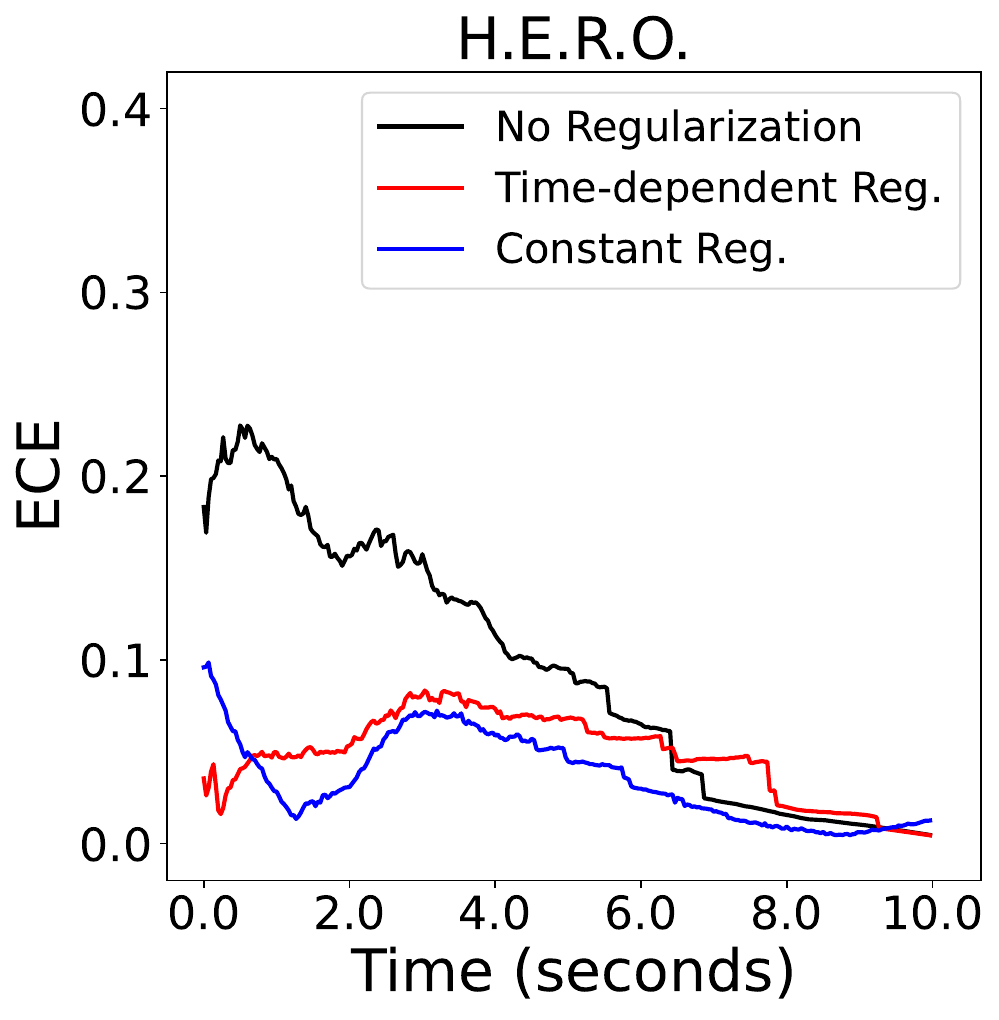}
    \includegraphics[height=0.2\linewidth]{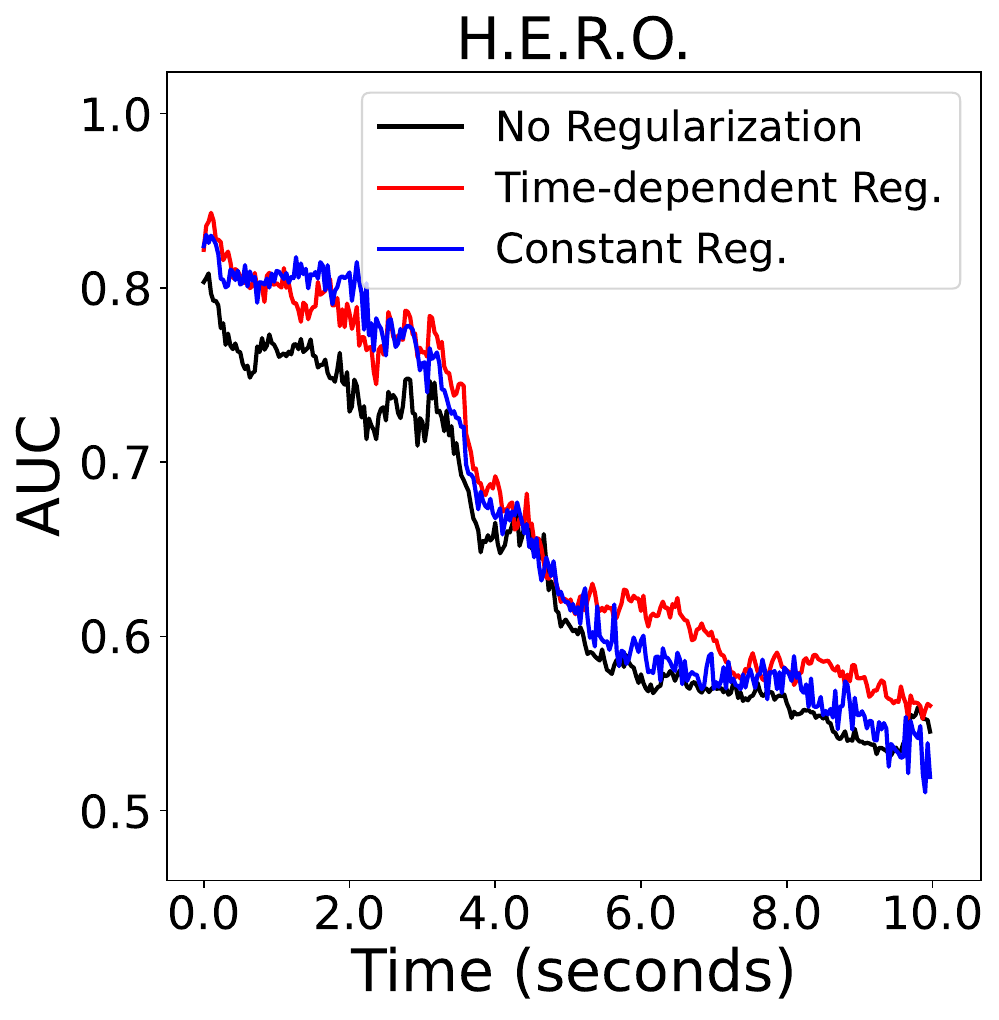}
    \includegraphics[height=0.2\linewidth]{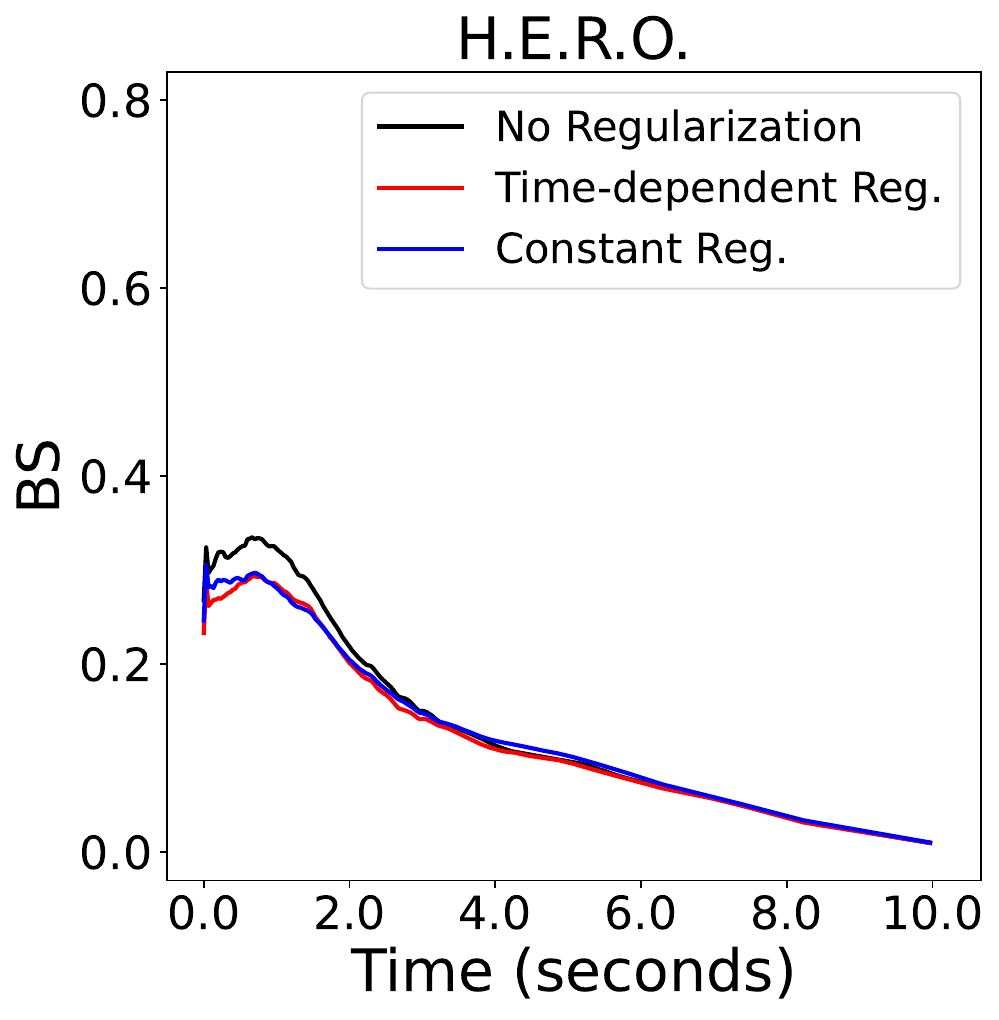}
    \includegraphics[height=0.2\linewidth]{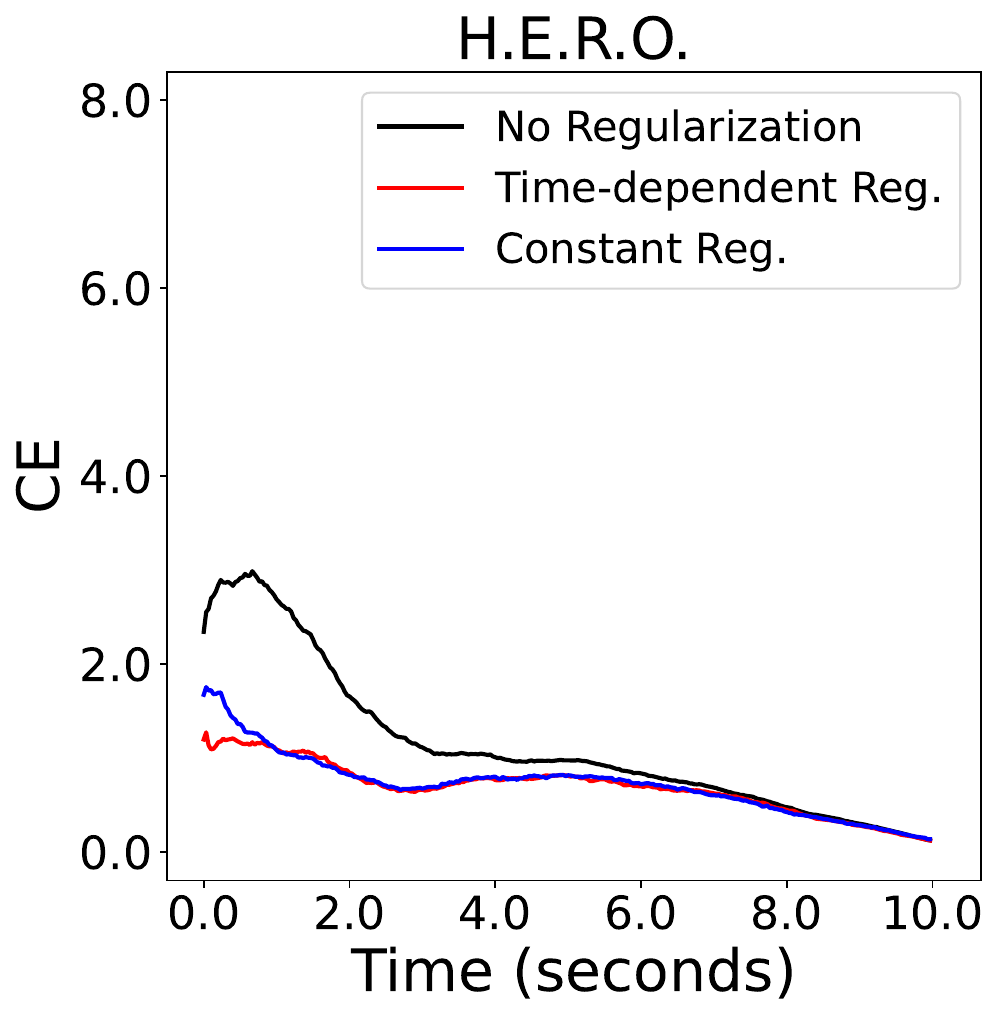} \\

    \includegraphics[height=0.2\linewidth]{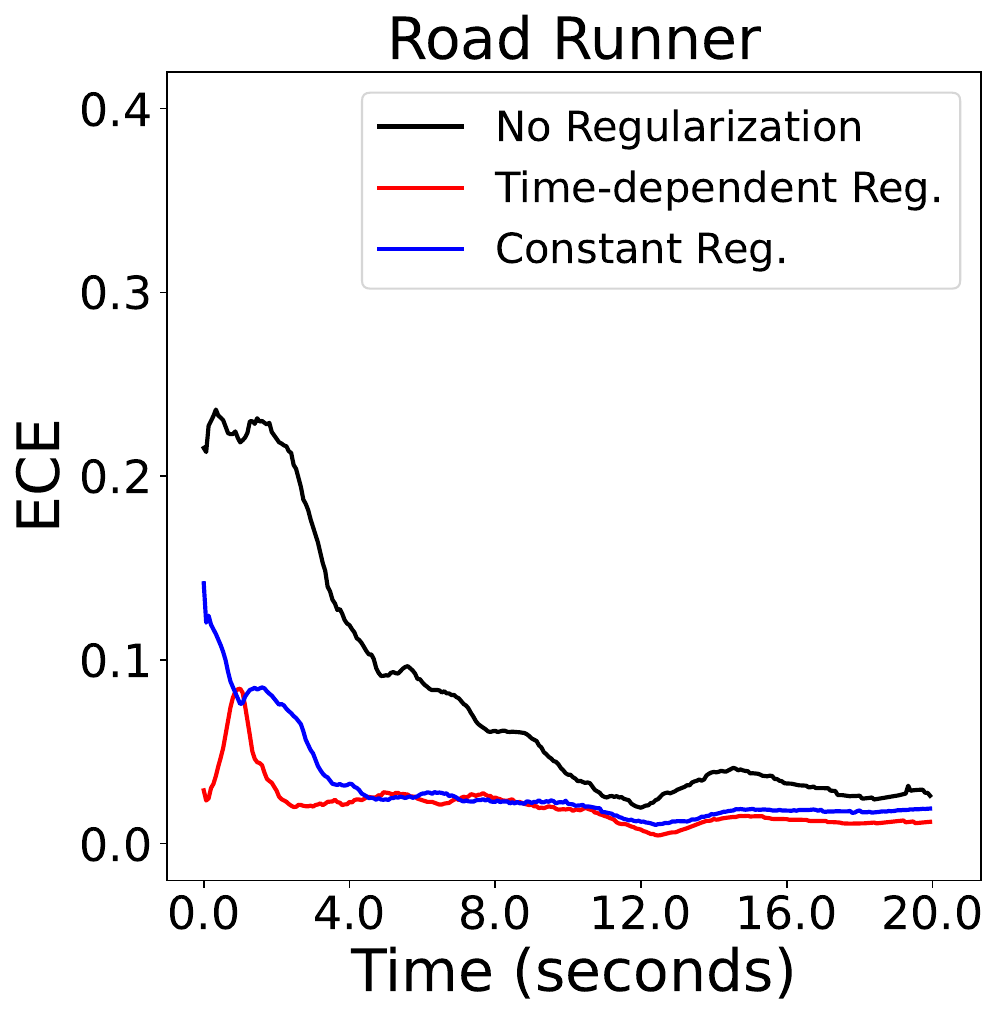}
    \includegraphics[height=0.2\linewidth]{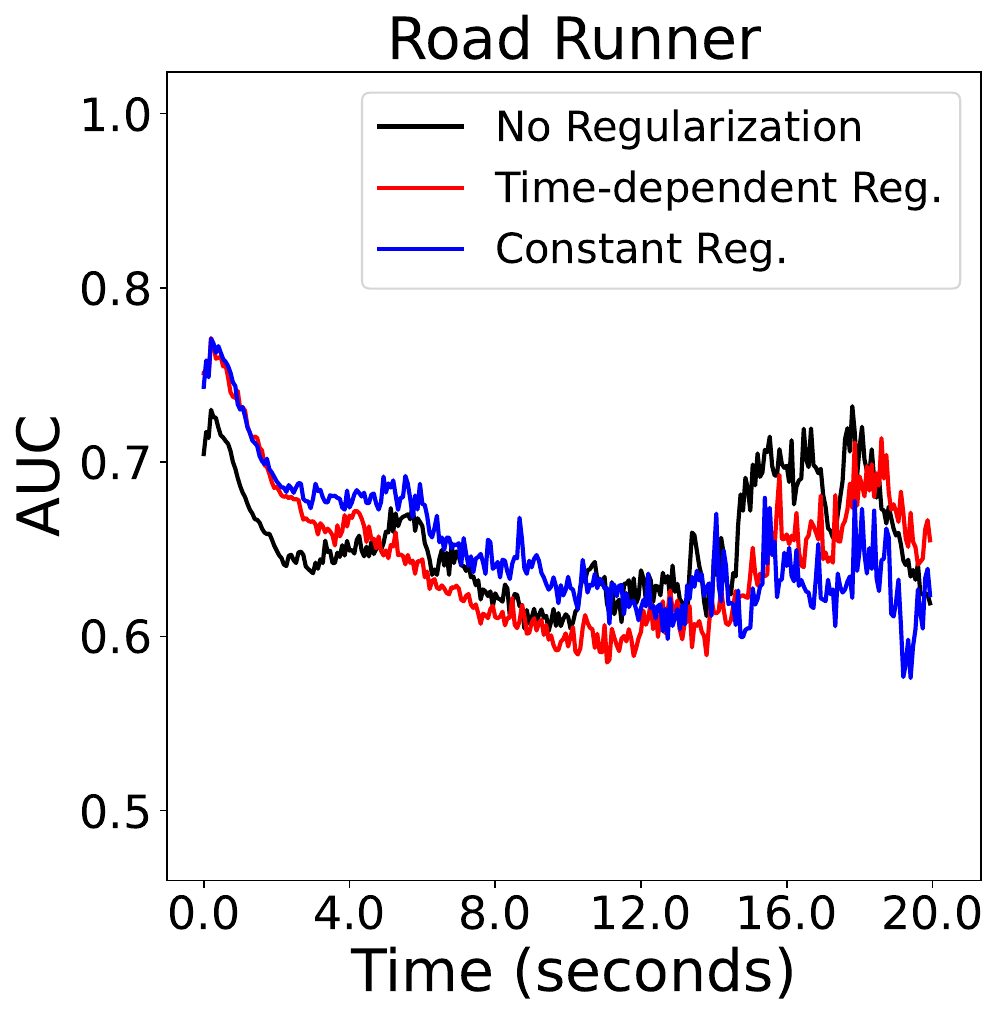}
    \includegraphics[height=0.2\linewidth]{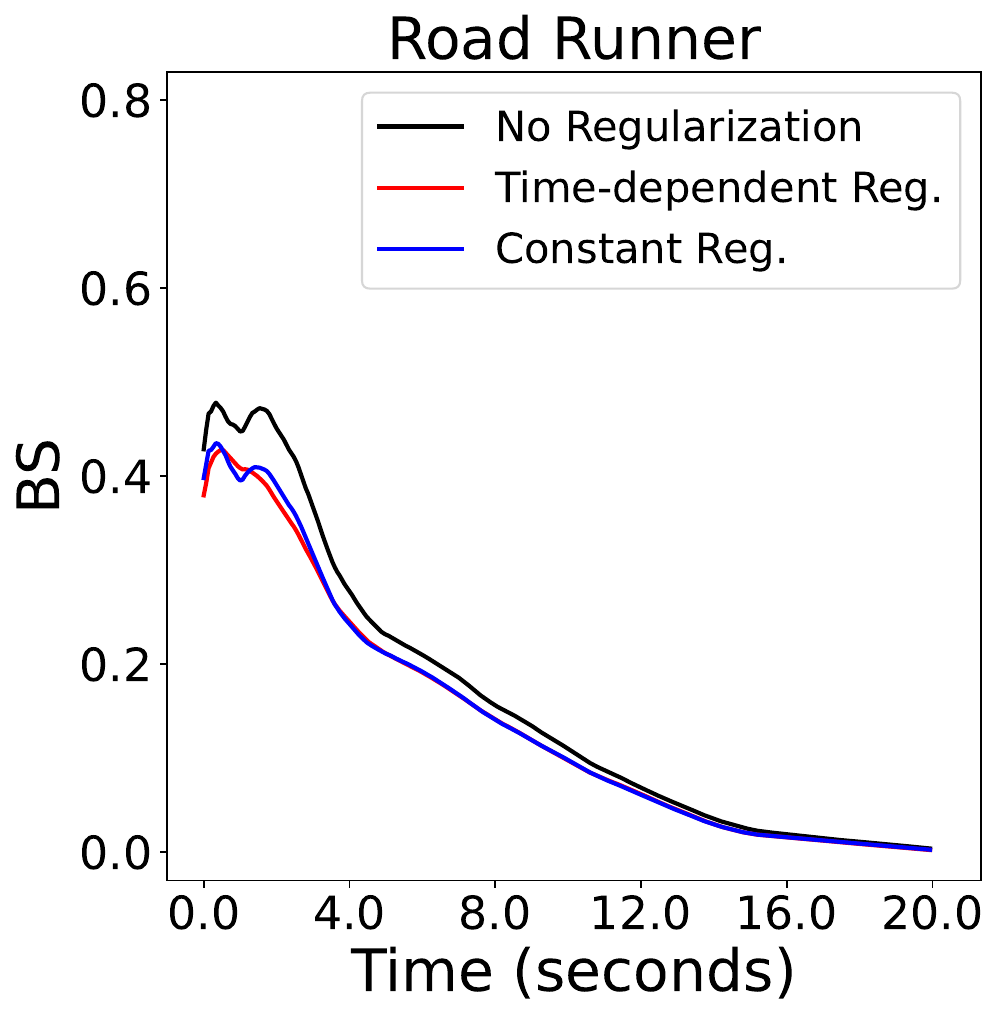}
    \includegraphics[height=0.2\linewidth]{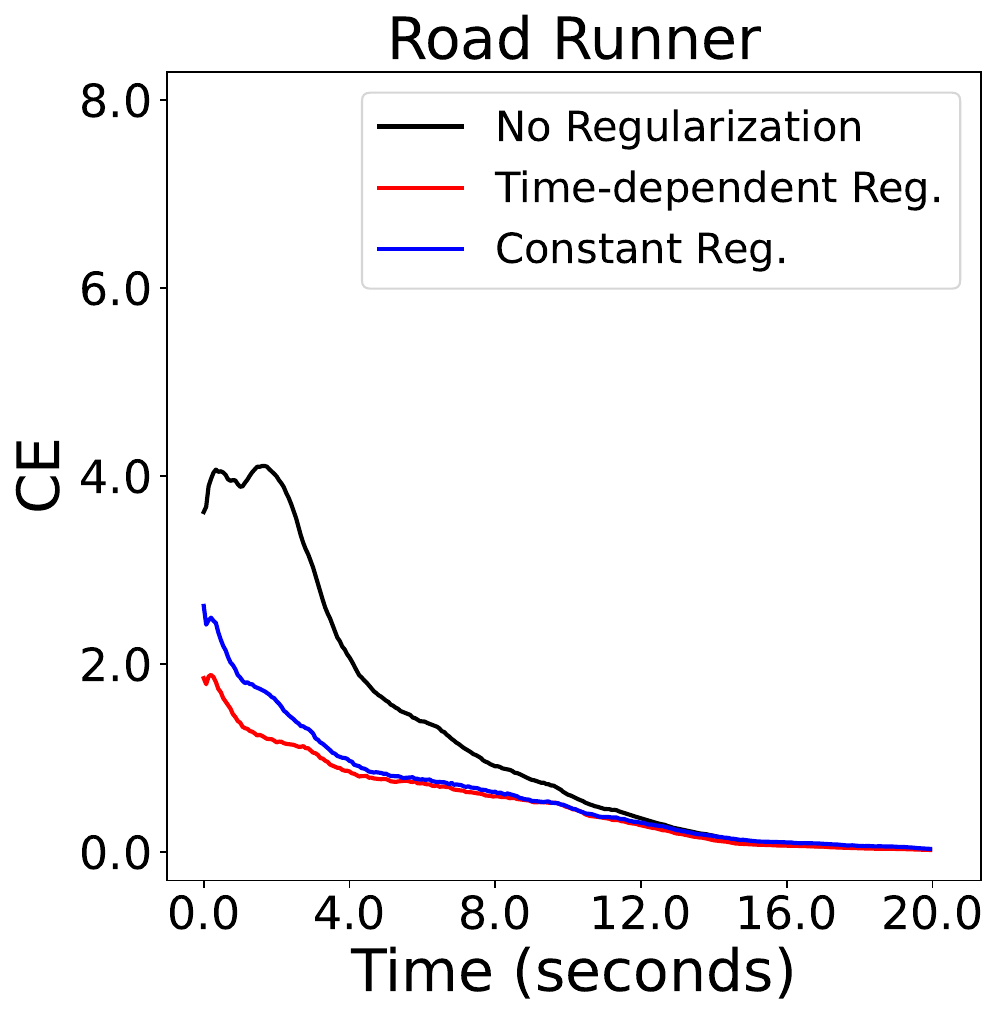} \\
    \includegraphics[height=0.2\linewidth]{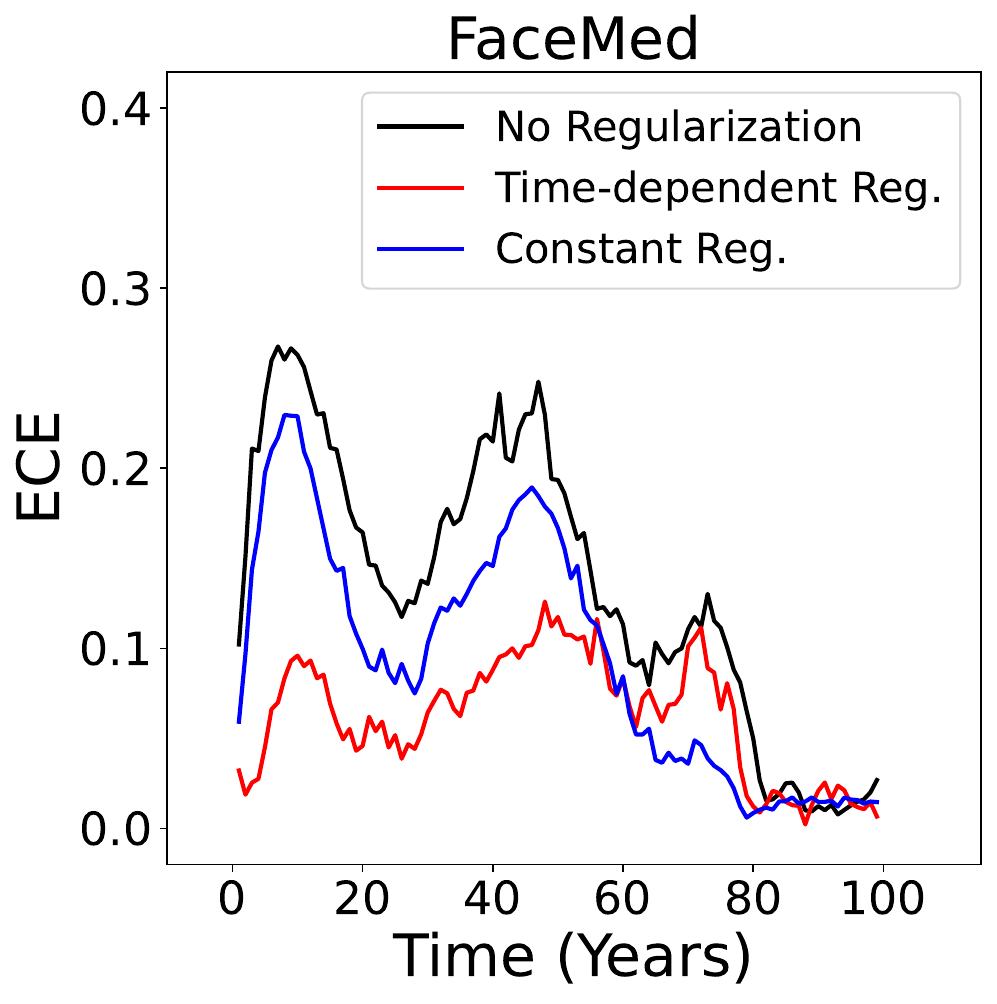}
    \includegraphics[height=0.2\linewidth]{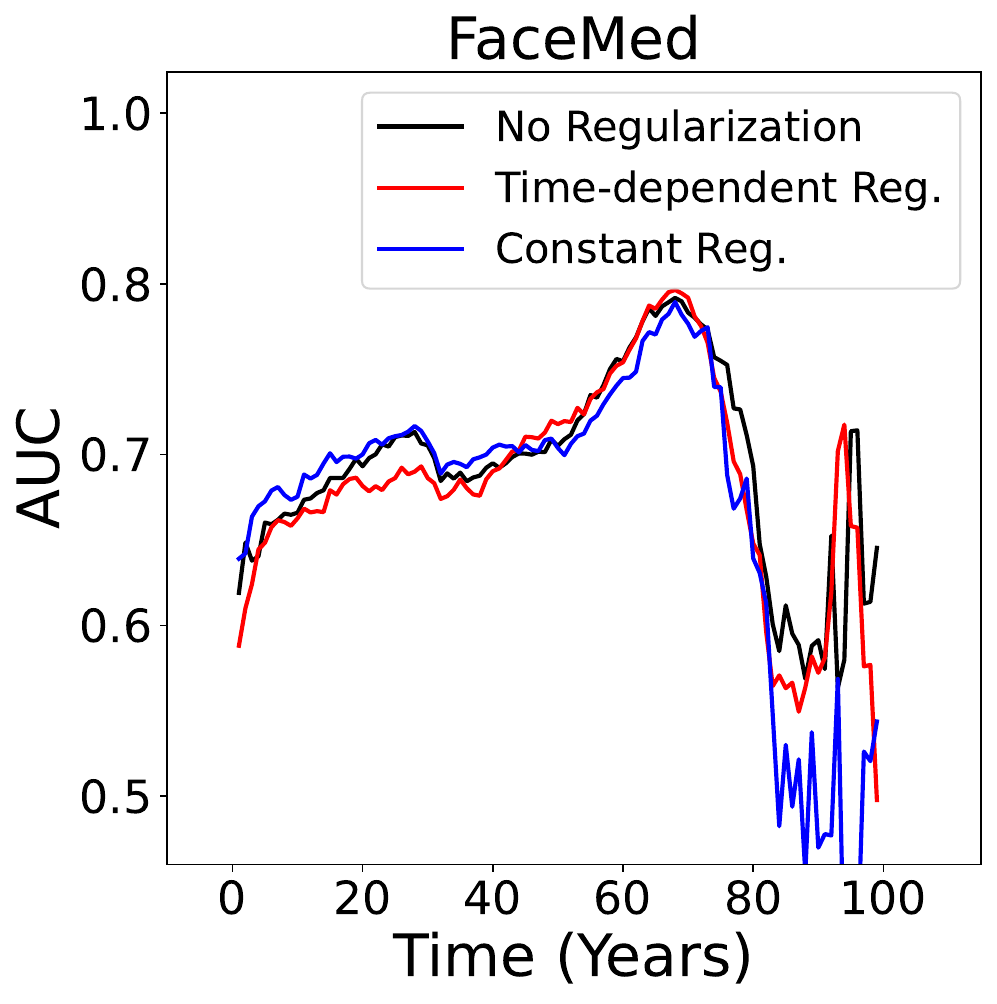}
    \includegraphics[height=0.2\linewidth]{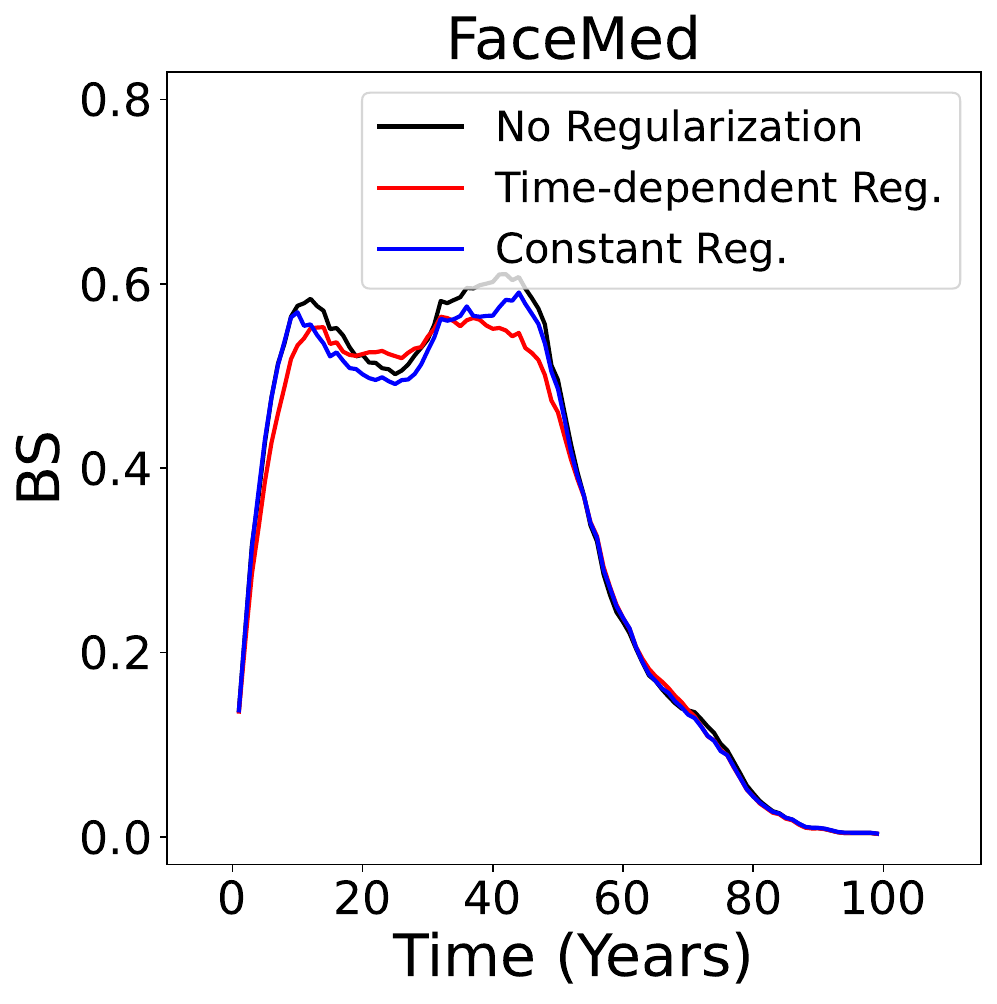}
    \includegraphics[height=0.2\linewidth]{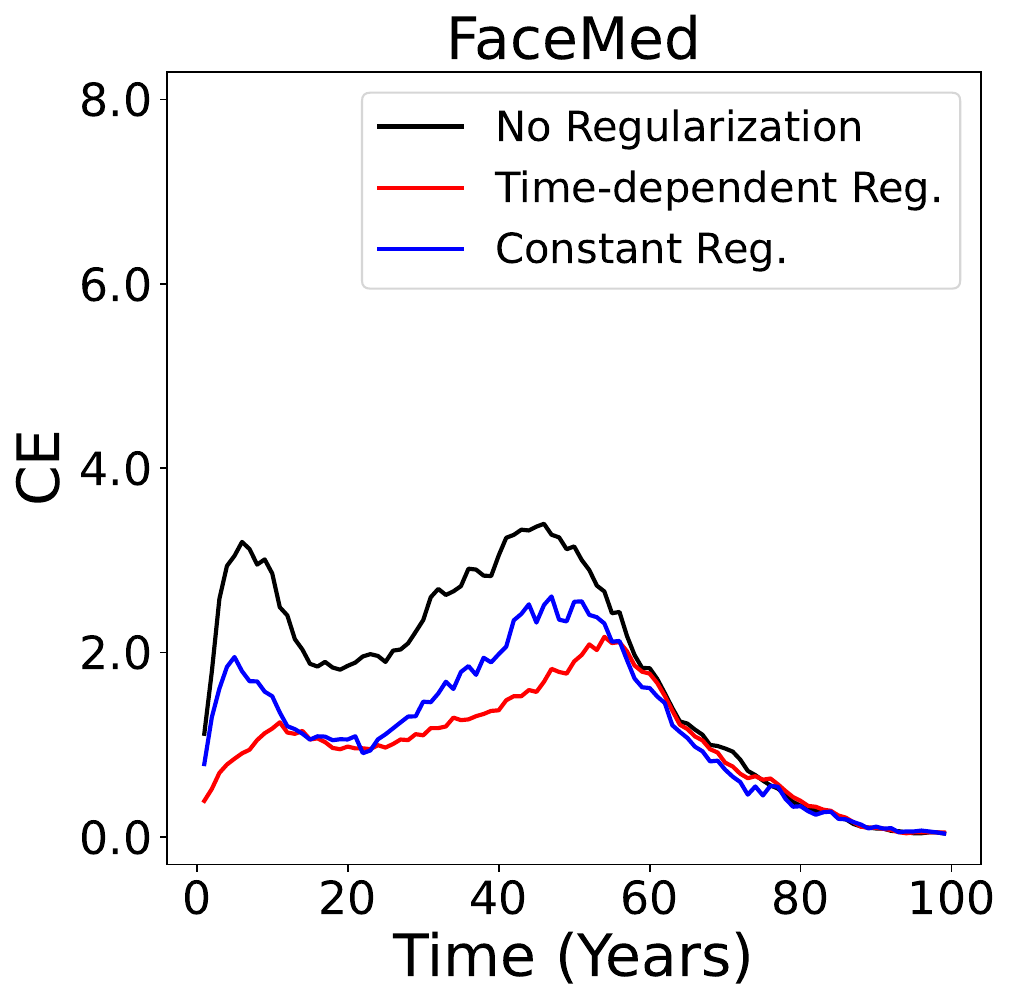}
    
    \caption{\textbf{Entry-wise metrics for marginal probability estimation.} The figure plots the entry-level ECE, AUC, BS, and CE of the proposed foCus framework for estimation of marginal probabilities. It compares versions of foCus: without regularization (black), time-dependent regularization (red), and constant regularization (blue). Both constant regularization and time-dependent regularization improve calibration and overall estimation quality compared to without regularization. Time-dependent regularization's improvement is more significant than constant regularization.
    }
    \label{fig:last_epoch_score_uncondition_game}
    \vspace{-8pt}
\end{figure*}

\begin{figure*}
\begin{center}
 \includegraphics[height=0.20\linewidth]{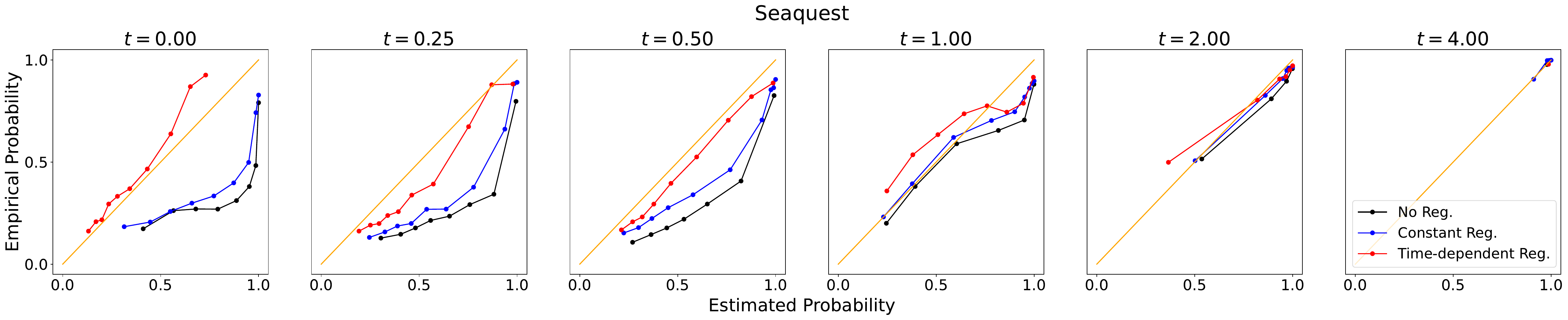}
 \includegraphics[height=0.20\linewidth]{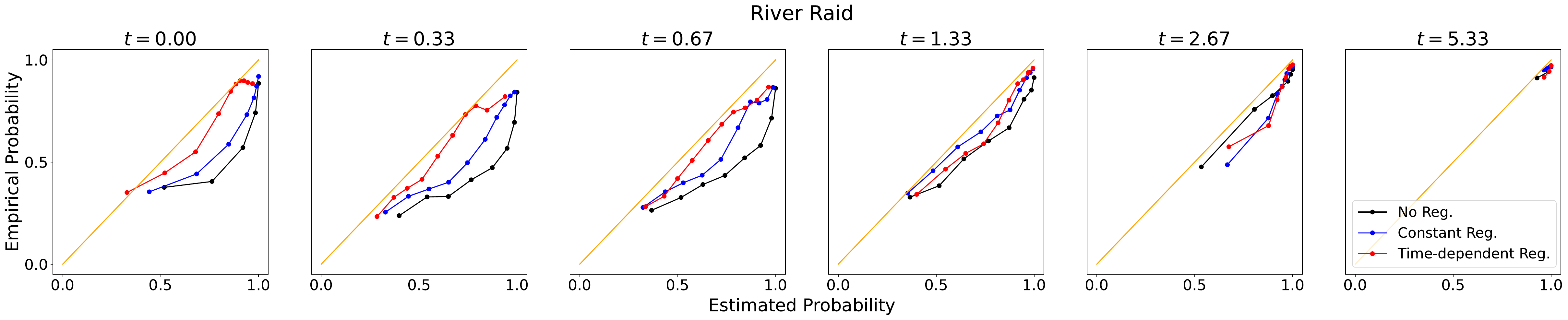}
 \includegraphics[height=0.20\linewidth]{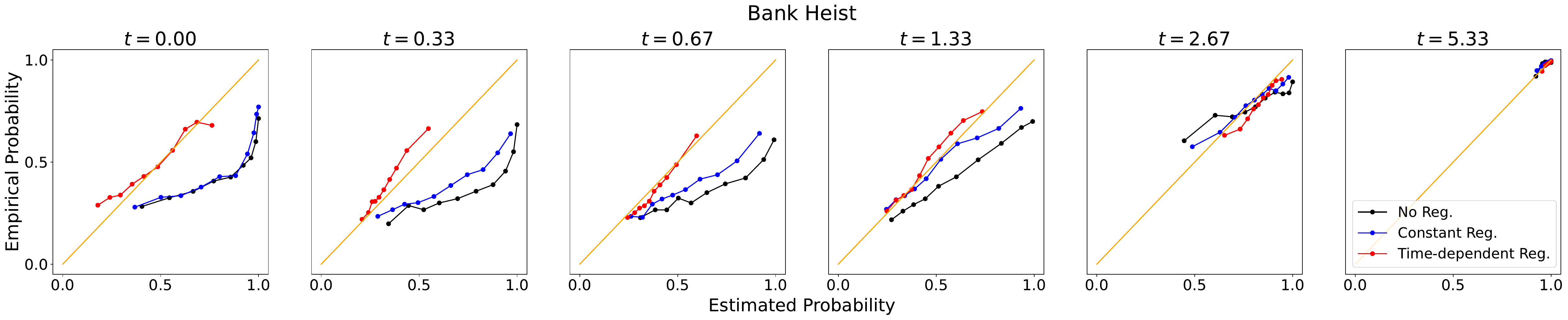}
 \includegraphics[height=0.20\linewidth]{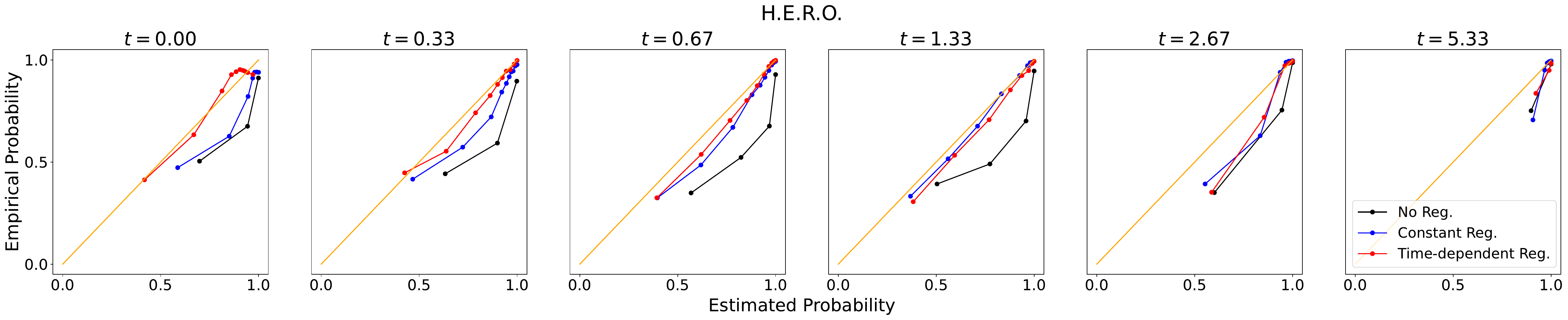}
 \includegraphics[height=0.20\linewidth]{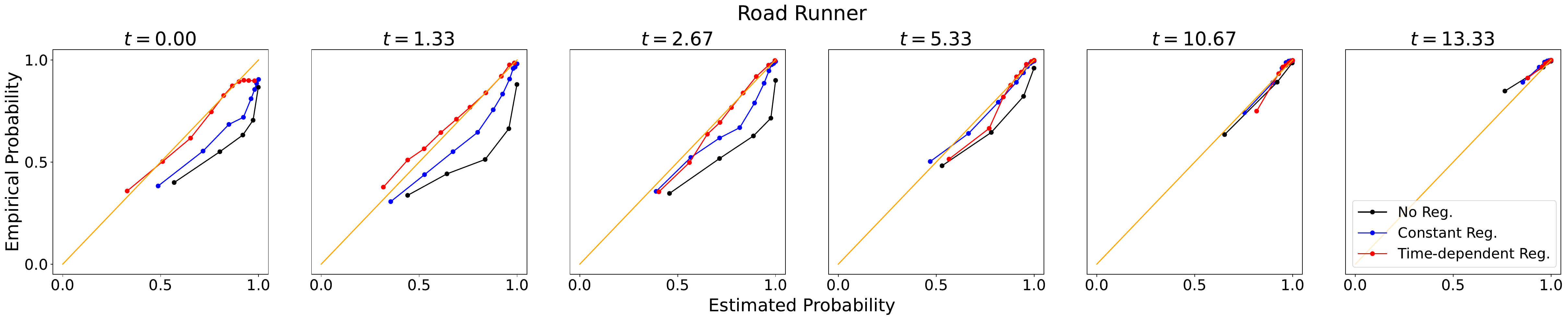}
 \includegraphics[height=0.20\linewidth]{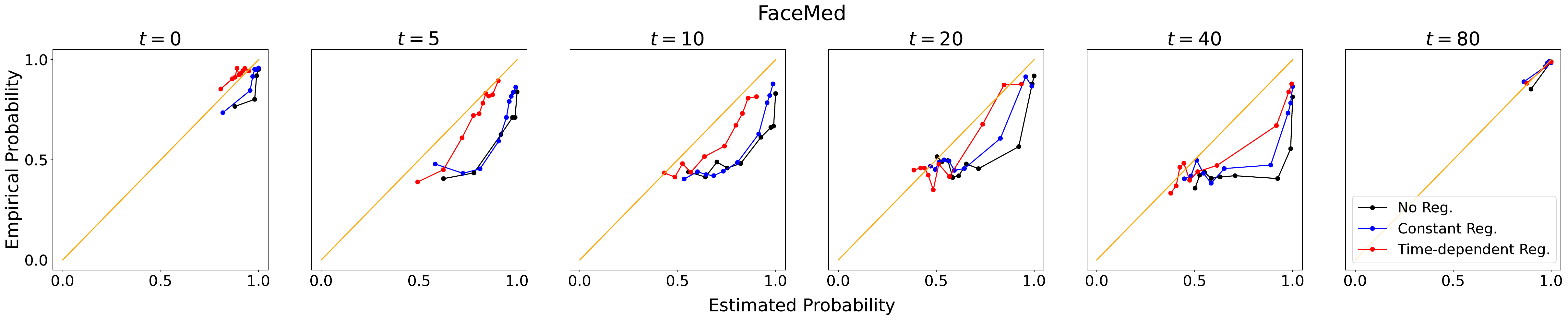}
\caption{
\textbf{Entry-wise reliability diagrams for marginal probability estimation.}
    The figure, supplementing \Cref{fig:reliability_diagram} presents reliability diagrams at additional entries in the sequence. Each diagram compares three versions of the proposed foCus framework: without regularization (black), time-dependent regularization (red), and constant regularization (blue). The plots demonstrate that both regularization methods improve calibration at each sequence entry, with time-dependent regularization showing the most substantial improvements.}
\label{fig:extra_reliability_diagram}
\end{center}
\end{figure*}

\Cref{fig:last_epoch_score_uncondition_game} shows plots of the entry-level metrics (ECE, AUC, BS, and CE) for marginal probability estimation, complementing \Cref{fig:reliability_diagram}. Time-dependent regularization leads to a substantial improvement in calibration, as demonstrated by the significantly lower ECE of the time-regularized model, which has a comparable AUC to the model without regularization. As a result, the cross-entropy and Brier Score metrics are also improved. Constant regularization also improves the probability estimates, but not as much as time-dependent regularization.
\Cref{fig:extra_reliability_diagram} shows additional reliability diagrams like the ones in \Cref{fig:reliability_diagram} and includes a comparison with constant regularization, confirming that time-dependent regularization consistently improves calibration for individual entries.

In \Cref{sec:results}, we also compare the estimated marginal probability with underlying data generating distribution on the synthetic FaceMed data. Since the ground truth sequences in FaceMed are generated using an assumed transitional probability model, as detailed in \Cref{app:dataset_details}, we can analytically compute the \textit{ground truth} marginal probability of ${{Y}}_i$ based on the transitional probabilities and the marginal probability of the previous entry, ${{Y}}_{i-1}$: 
\begin{align}
\operatorname{P}({{Y}}_i = a \mid {{X}} = x) = \sum_{b=1}^c \operatorname{P}({{Y}}_{i-1} = b \mid {{X}} = x) \operatorname{P}({{Y}}_{i} = a \mid {{Y}}_{i-1} = b)
\end{align}
Starting from an initial state of “healthy,” we compute the ground truth marginal probabilities for each entry in the sequence and compare them against the estimates obtained via foCus. We calculate the root mean square error (RMSE) between the ground truth and the estimated marginal probabilities for each entry, then aggregate these values to derive a sequence-level RMSE by averaging. The unregularized baseline method and constant regularization yield RMSE of $0.1847 \pm 0.0014$ and $0.1754 \pm 0.0028$, respectively, while time-dependent regularization reduces the RMSE to $0.1720 \pm 0.0009$. This result further demonstrates the improved performance achieved by time-dependent regularization.

\begin{comment}
\begin{figure*}
\centering
\includegraphics[width=0.5\columnwidth]{drafts/_aistats2025/Figures/Last_Epoch_ECE_Uncondition.pdf}
\includegraphics[width=0.5\columnwidth]{drafts/_aistats2025/Figures/Last_Epoch_Score_Uncondition.pdf}
\caption{This figure shows marginal metrics at each step for all five games from the last training epoch. Marginal BS, AUC, and CE are plotted in column 1, 2, and 3. The black curve is the marginal metric average across baseline models initialized with three different random seeds; the red (blue) curve is the average from the models with step varying (constant) L2 logit regularization also initialized with three different random seeds. The legend also shows the step averaged marginal metric mean $\pm$ the standard deviation from model instantiations.}
\label{fig:last_epoch_score_uncondition_game}
\end{figure*}
\end{comment}

\begin{comment}
\begin{figure*}
\centering
\includegraphics[height=0.22\linewidth]{drafts/_aistats2025/Figures/fig3/reliability_diagram_seaquest.pdf}
\caption{Additional reliability diagram.}
\label{fig:uncondition_reliability}
\end{figure*}
\end{comment}

\subsection{Conditional Proabability}
\label{app:additional_results_conditional}
\begin{figure*}
    \centering
    \includegraphics[height=0.2\linewidth]{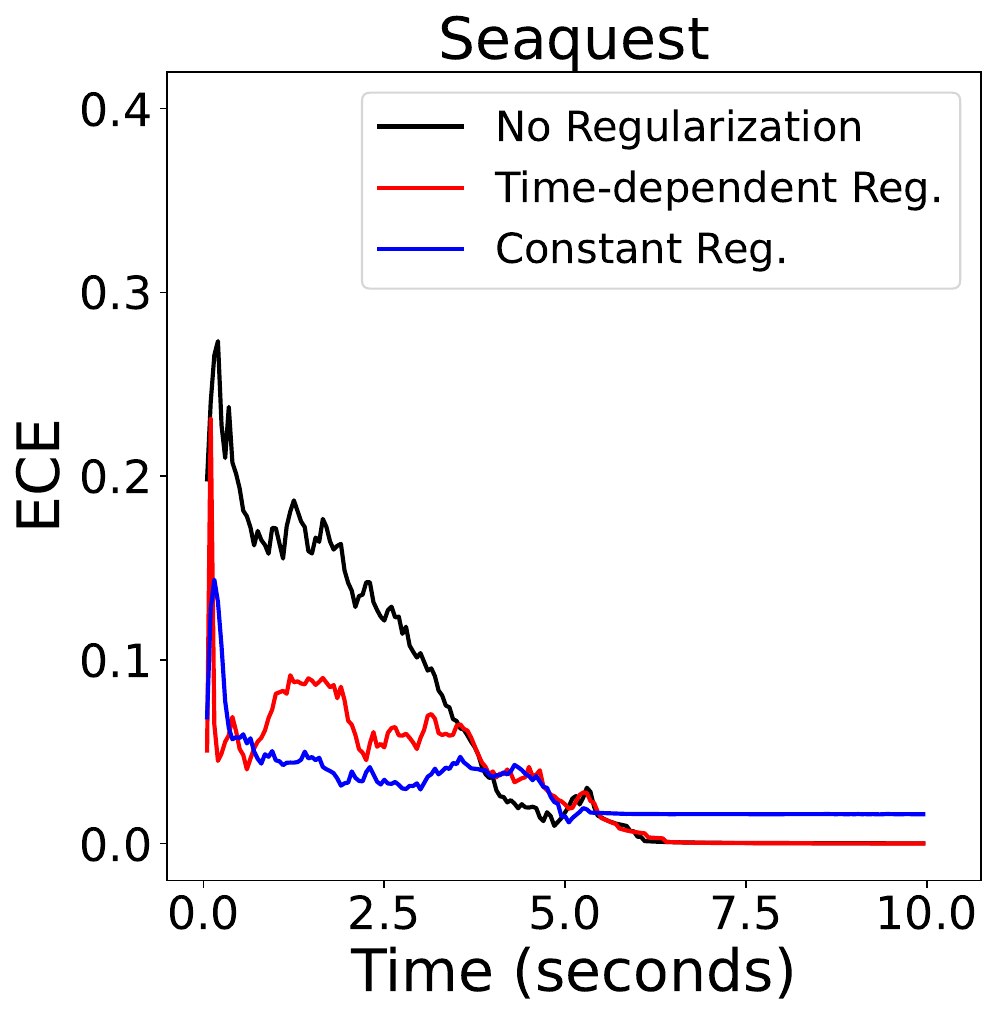}
    \includegraphics[height=0.2\linewidth]{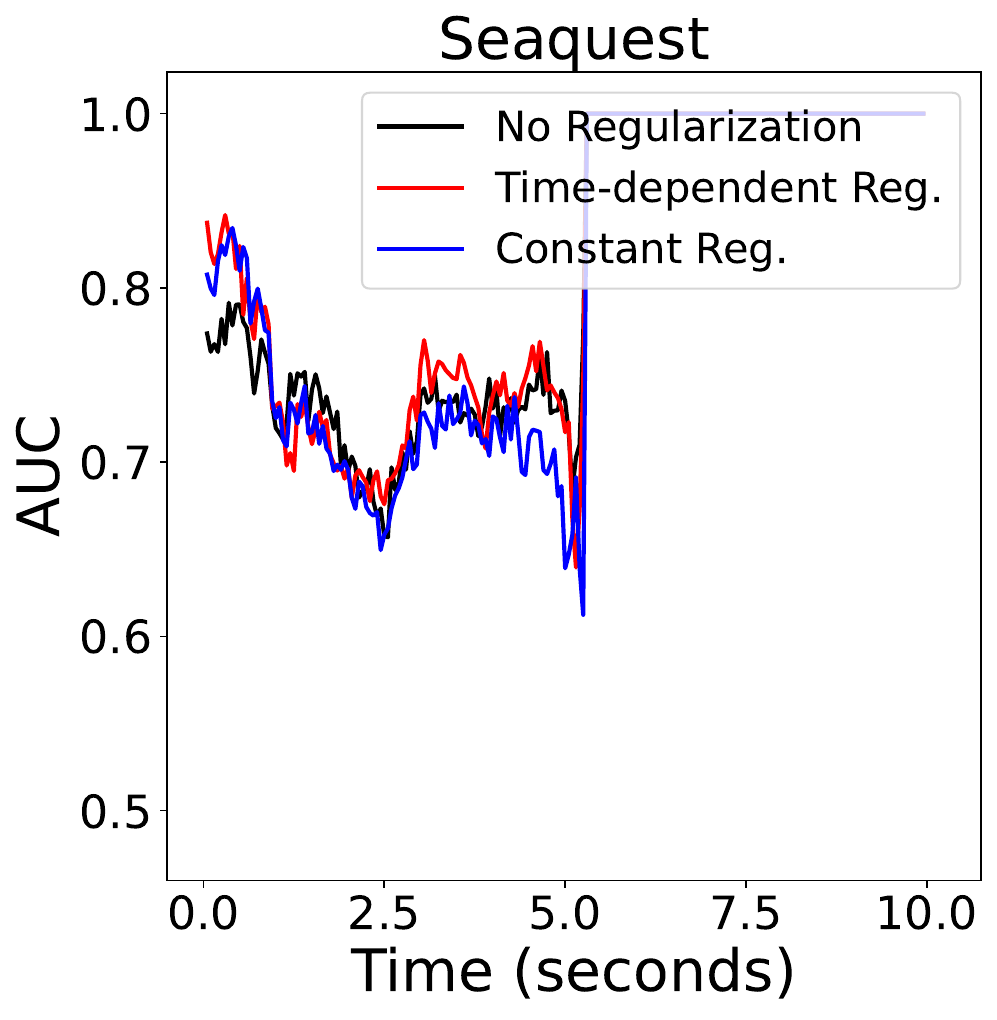}
    \includegraphics[height=0.2\linewidth]{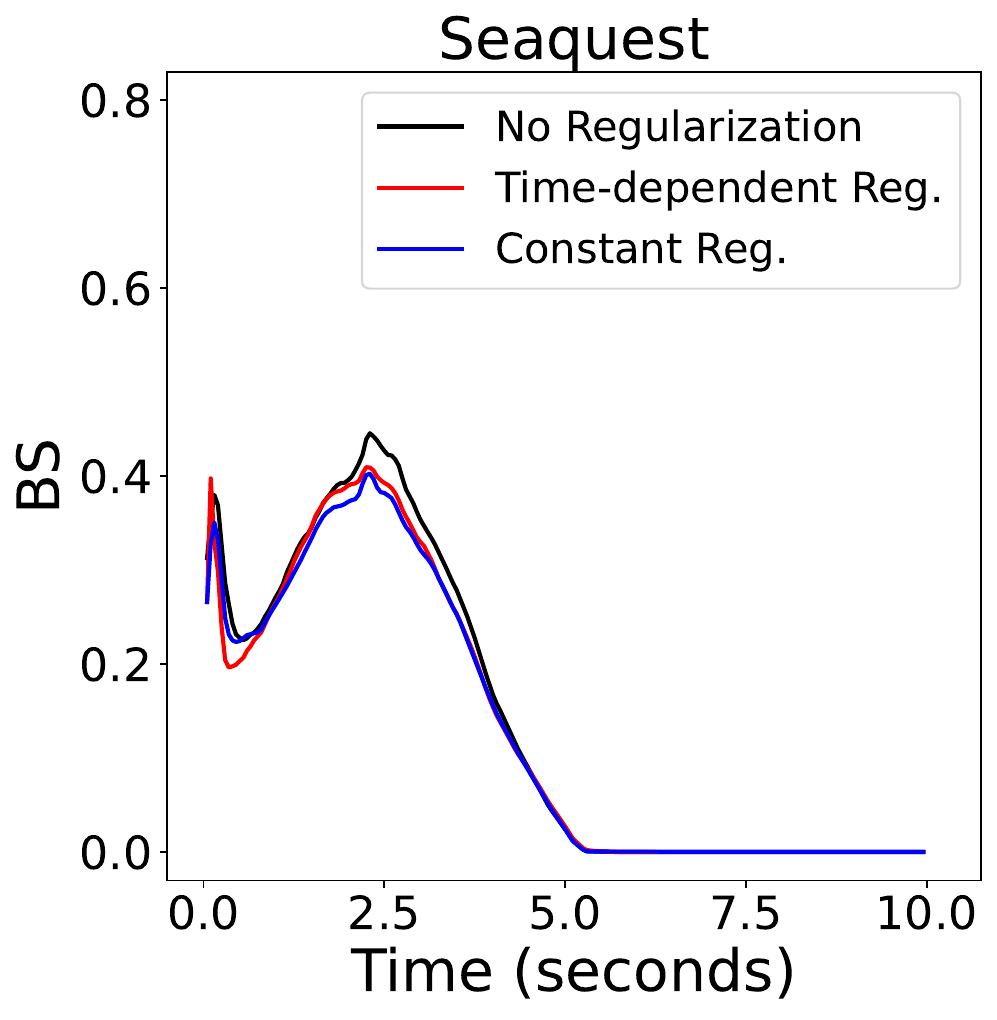}
    \includegraphics[height=0.2\linewidth]{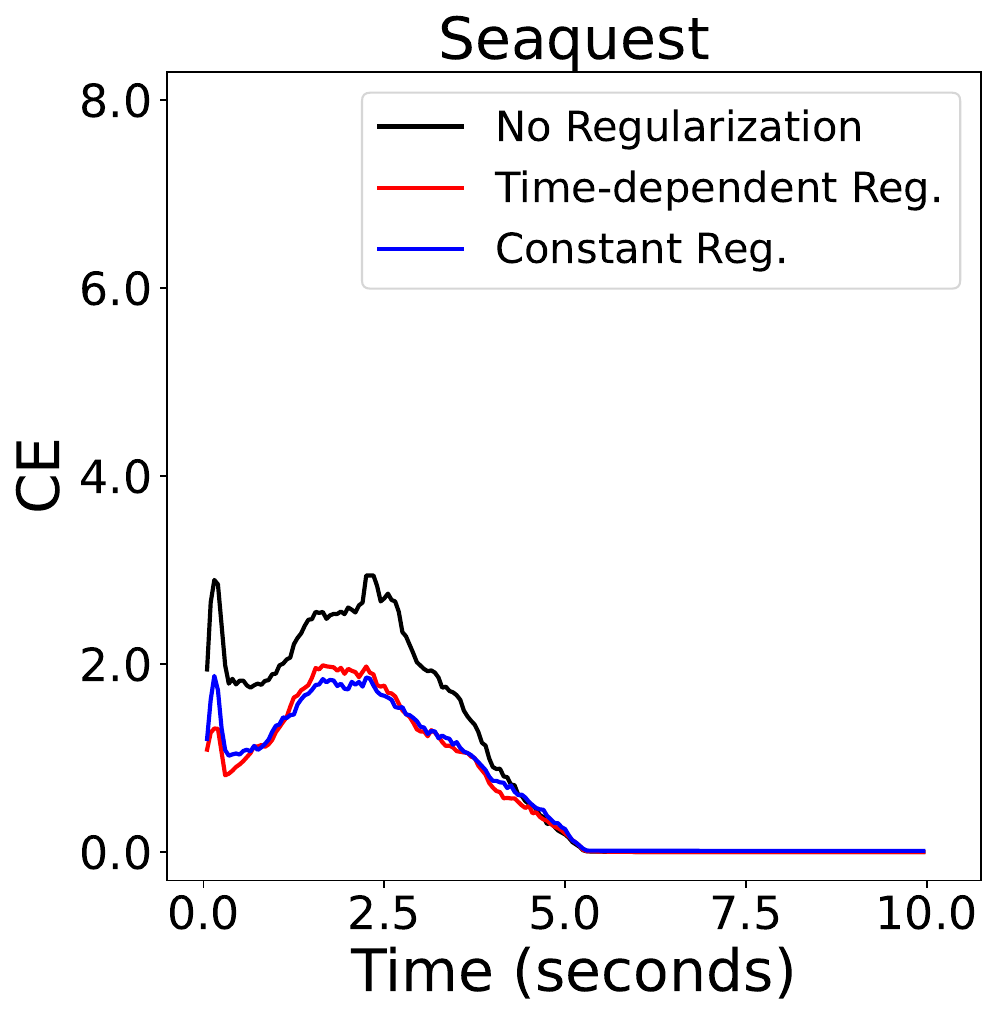} \\

    \includegraphics[height=0.2\linewidth]{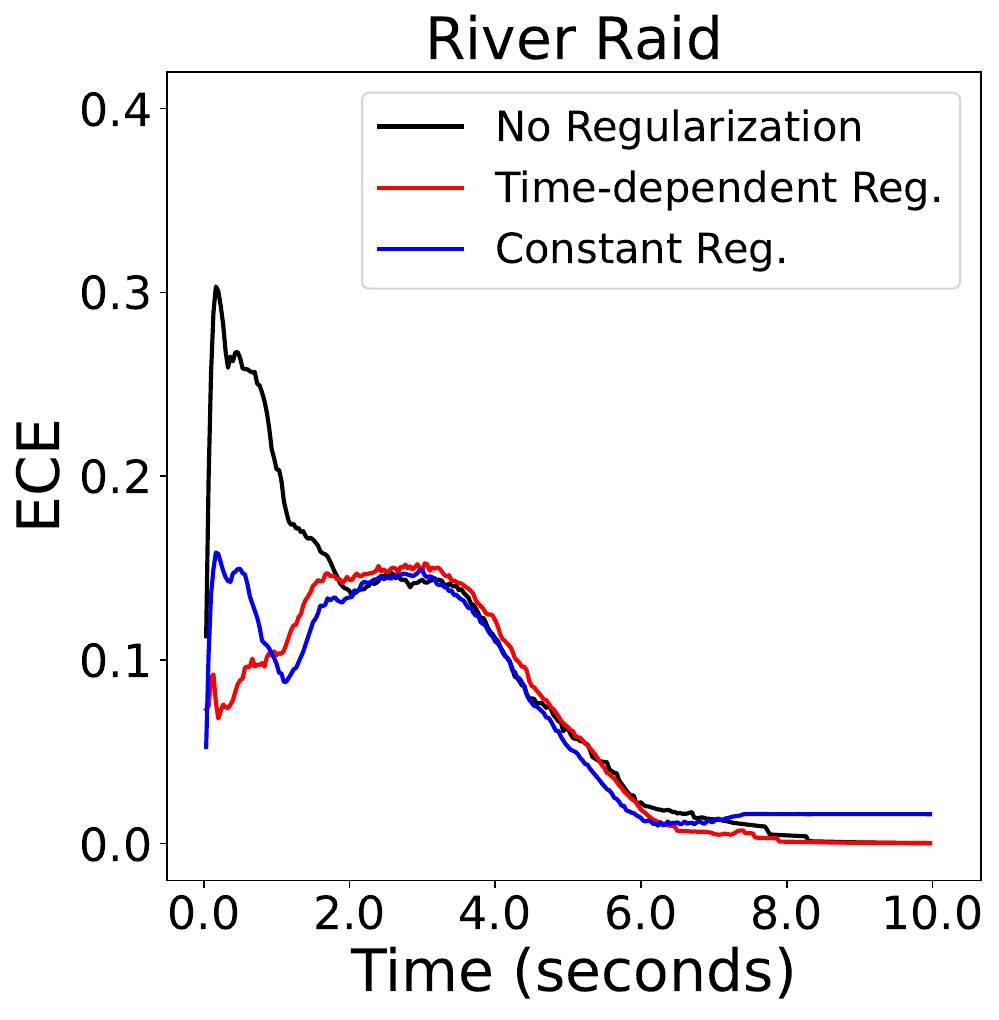}
    \includegraphics[height=0.2\linewidth]{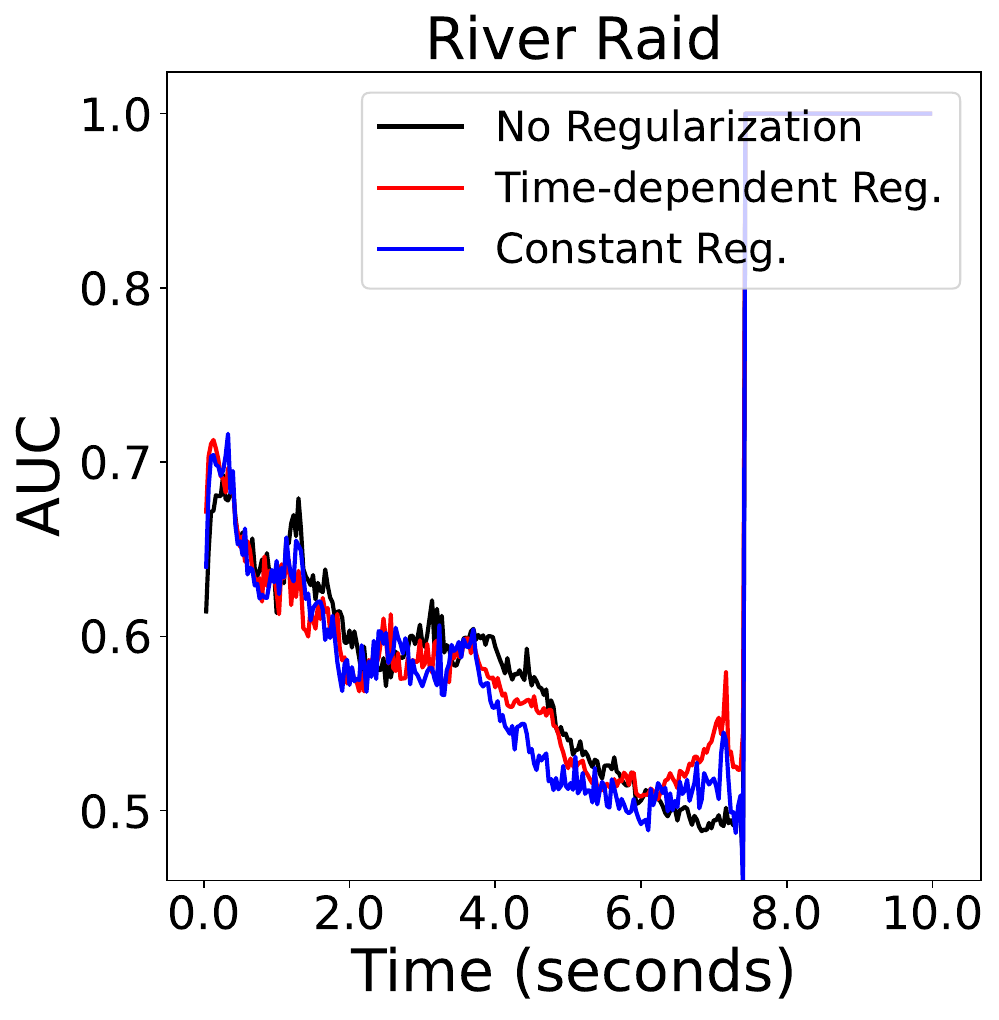}
    \includegraphics[height=0.2\linewidth]{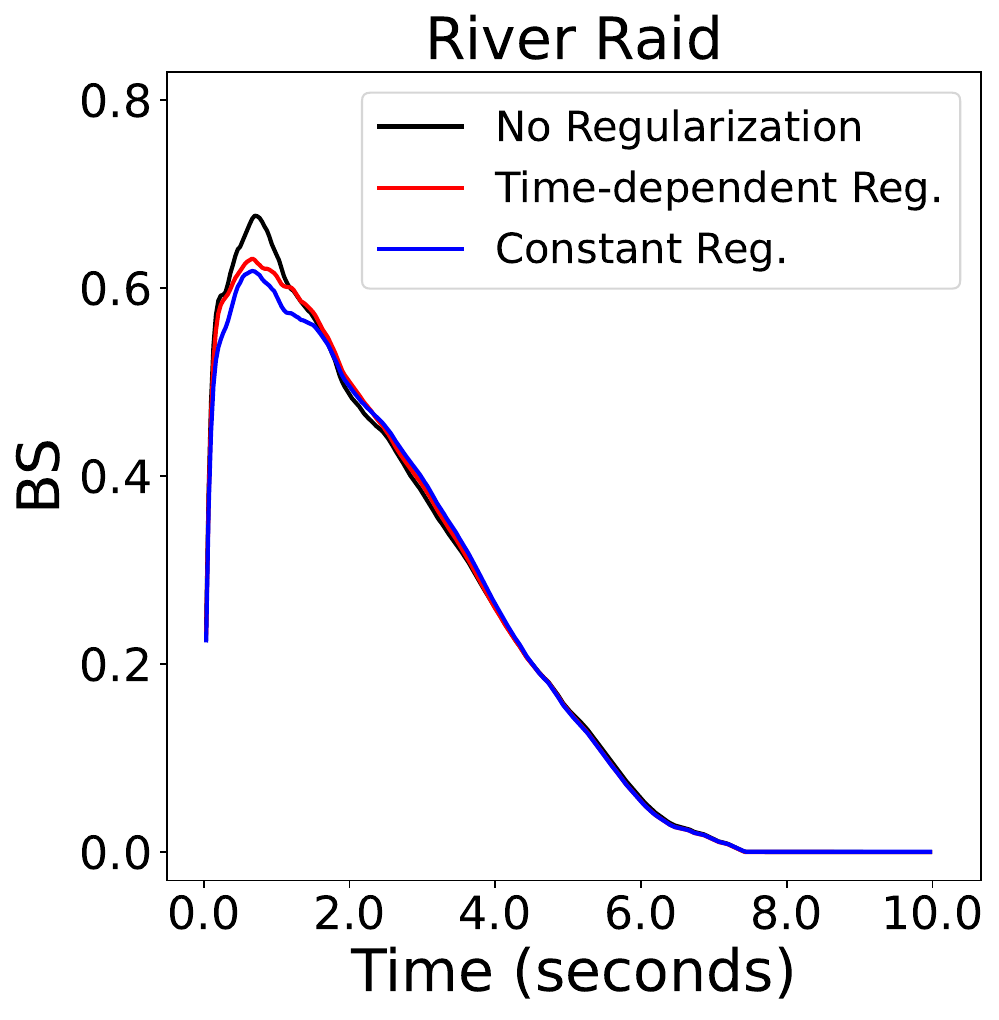}
    \includegraphics[height=0.2\linewidth]{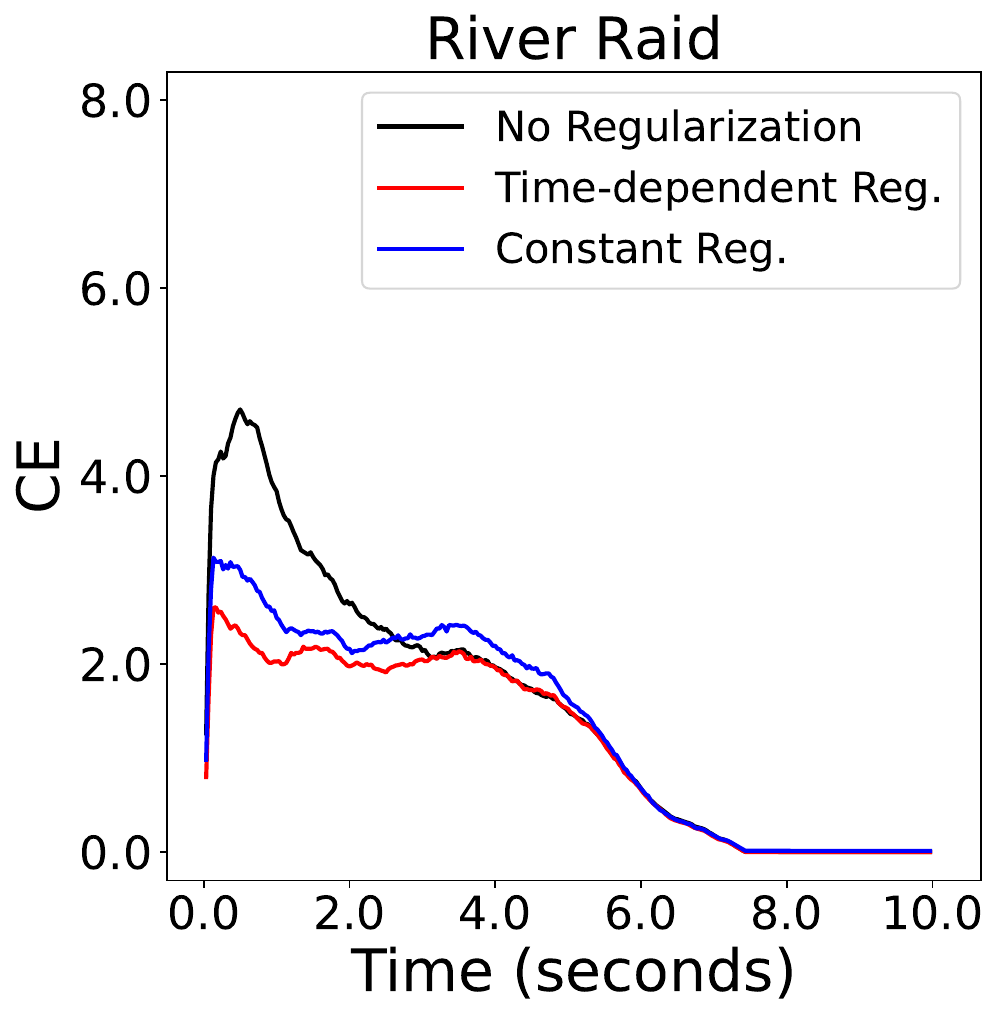} \\

    \includegraphics[height=0.2\linewidth]{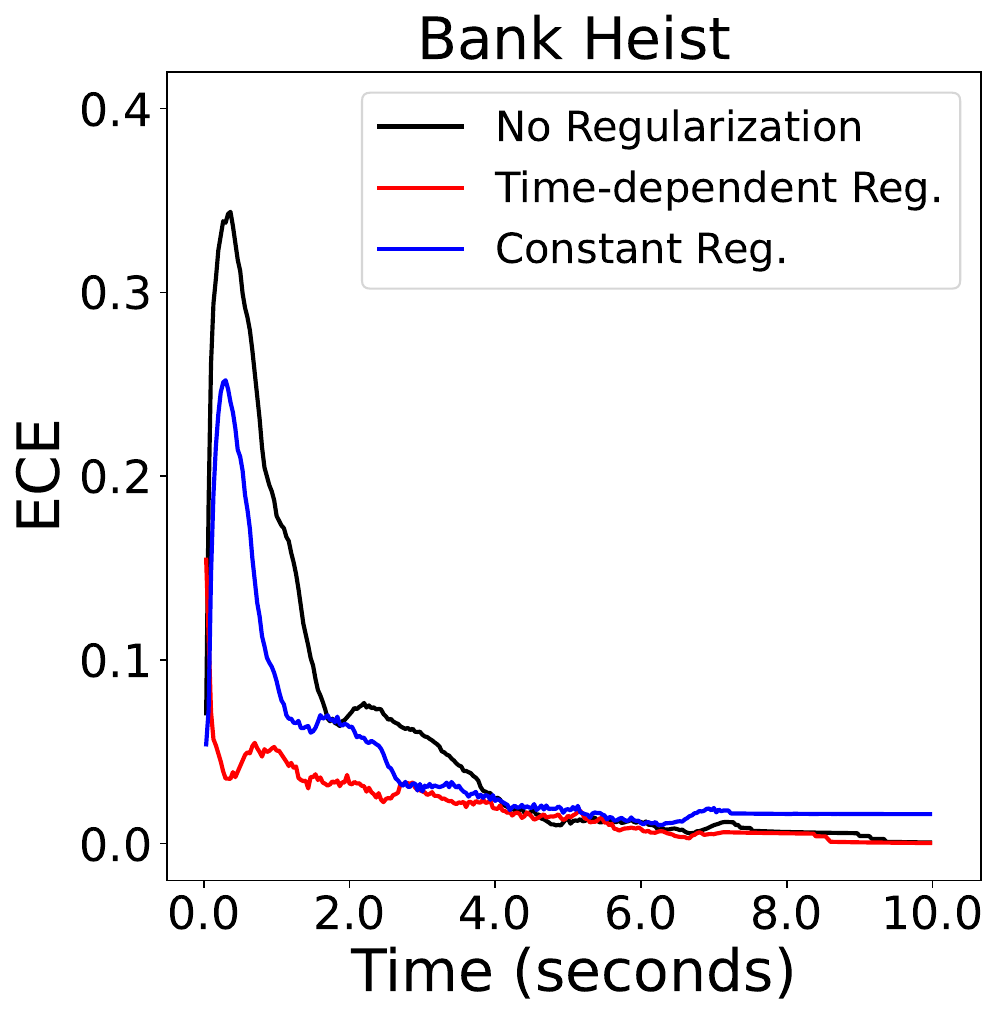}
    \includegraphics[height=0.2\linewidth]{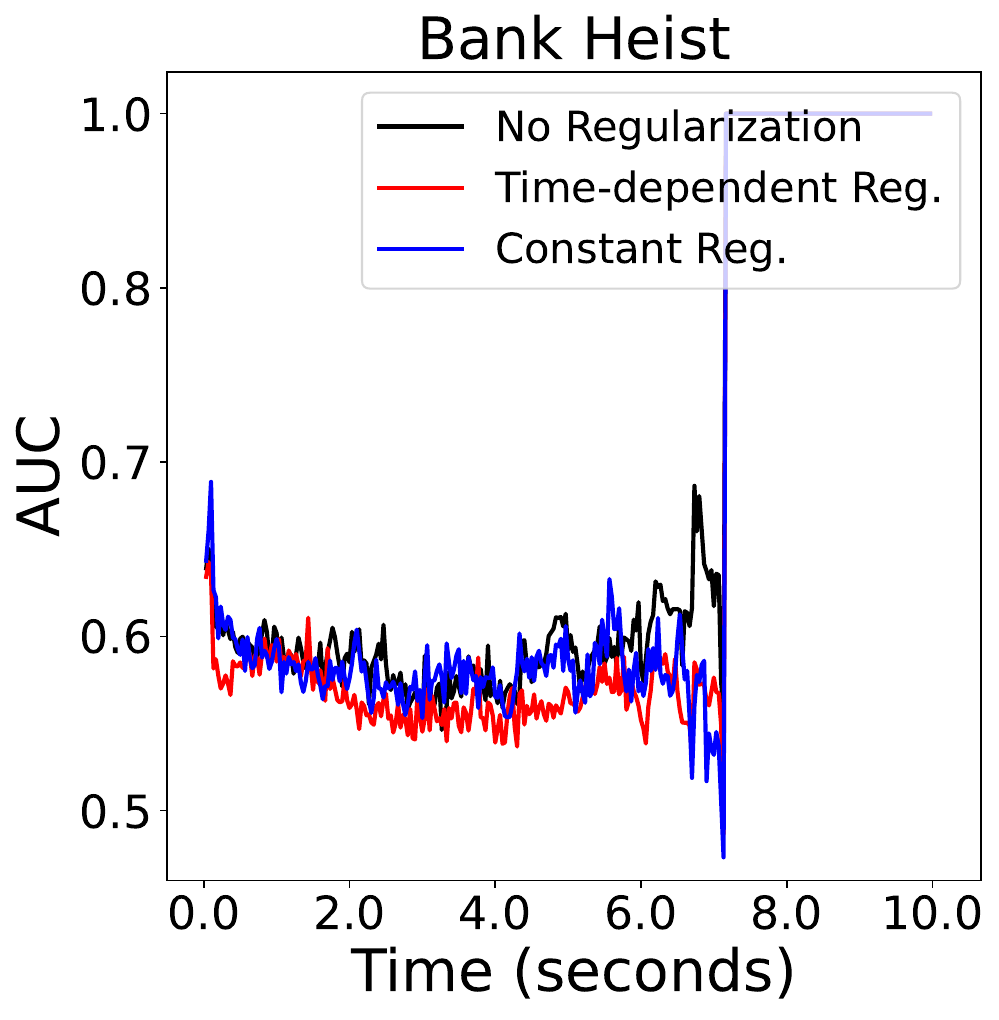}
    \includegraphics[height=0.2\linewidth]{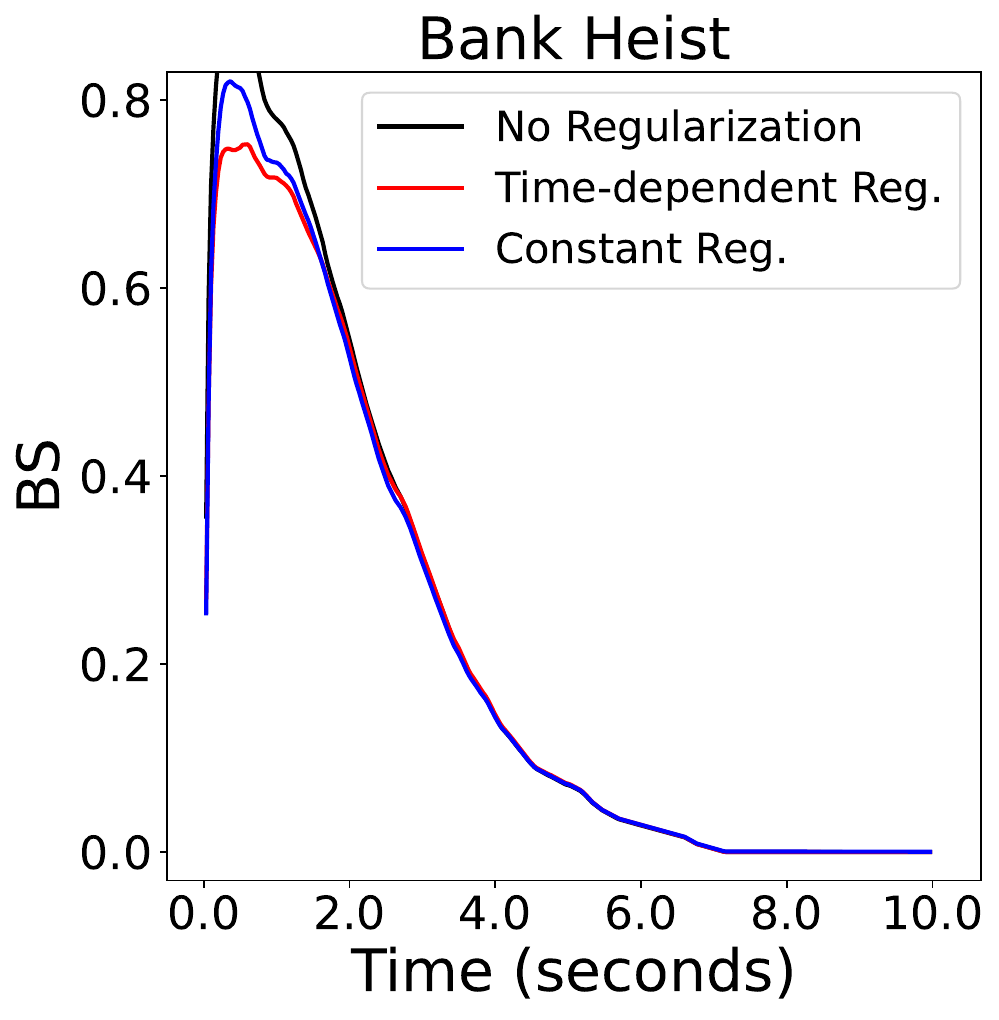}
    \includegraphics[height=0.2\linewidth]{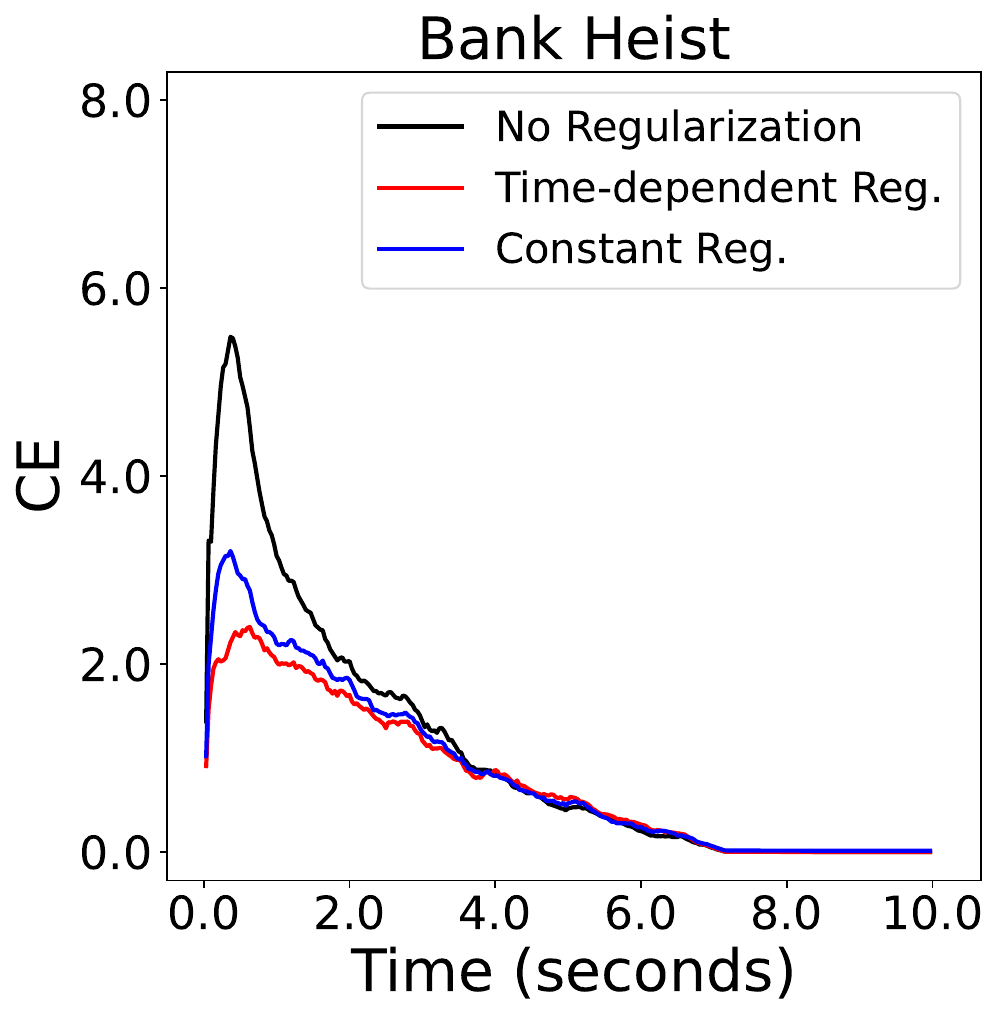} \\

    \includegraphics[height=0.2\linewidth]{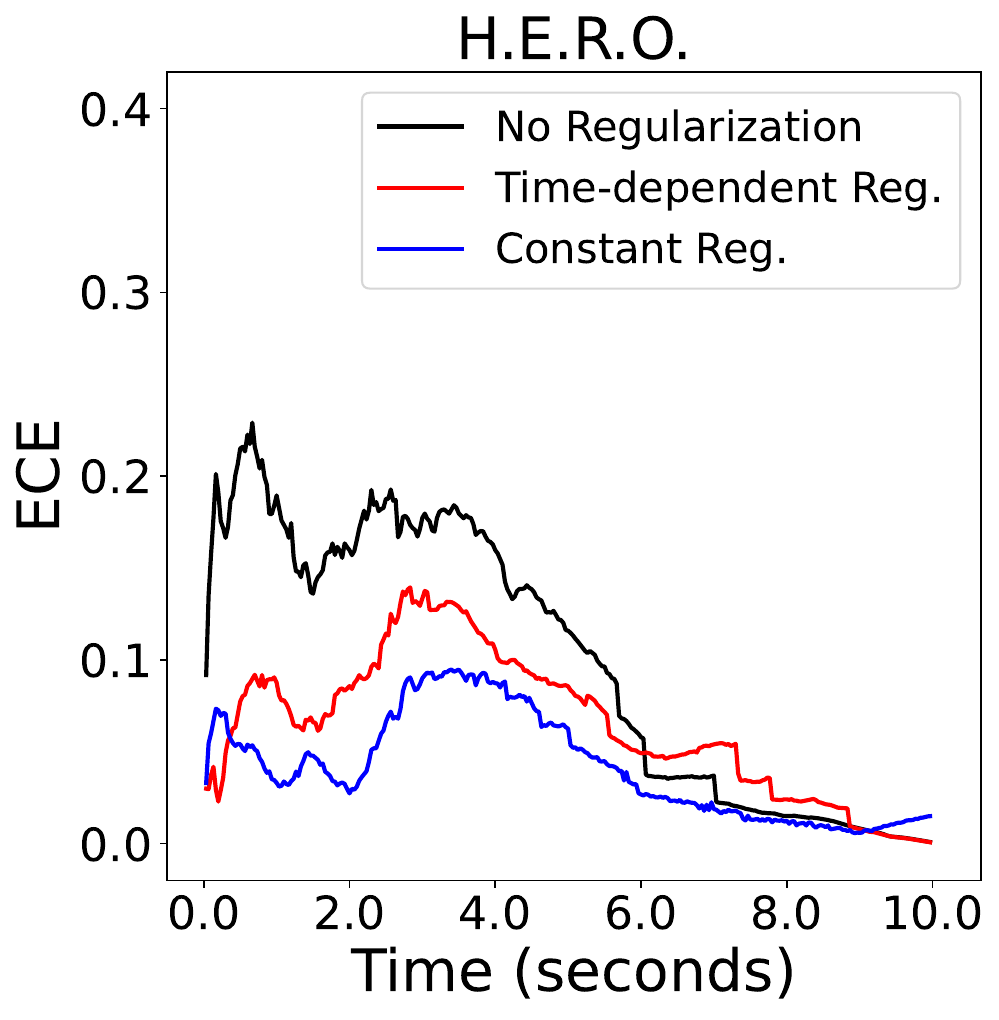}
    \includegraphics[height=0.2\linewidth]{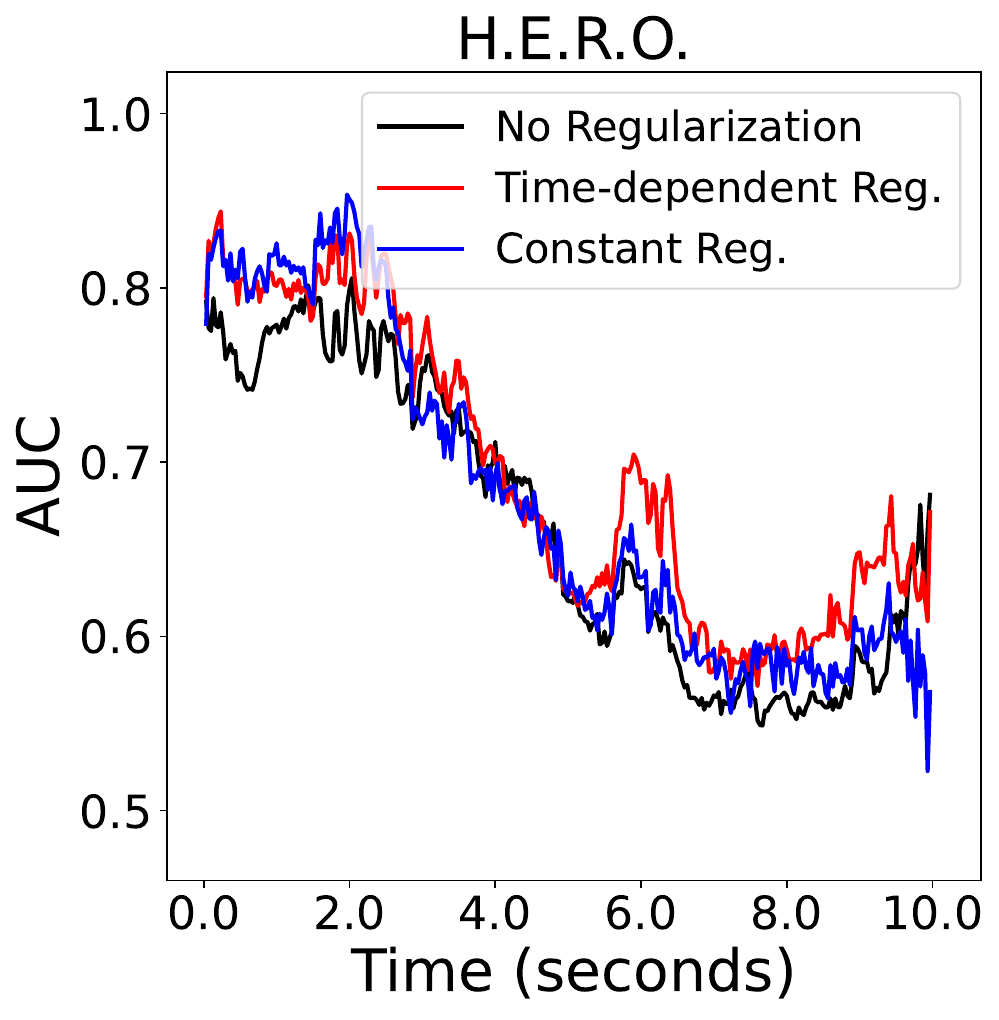}
    \includegraphics[height=0.2\linewidth]{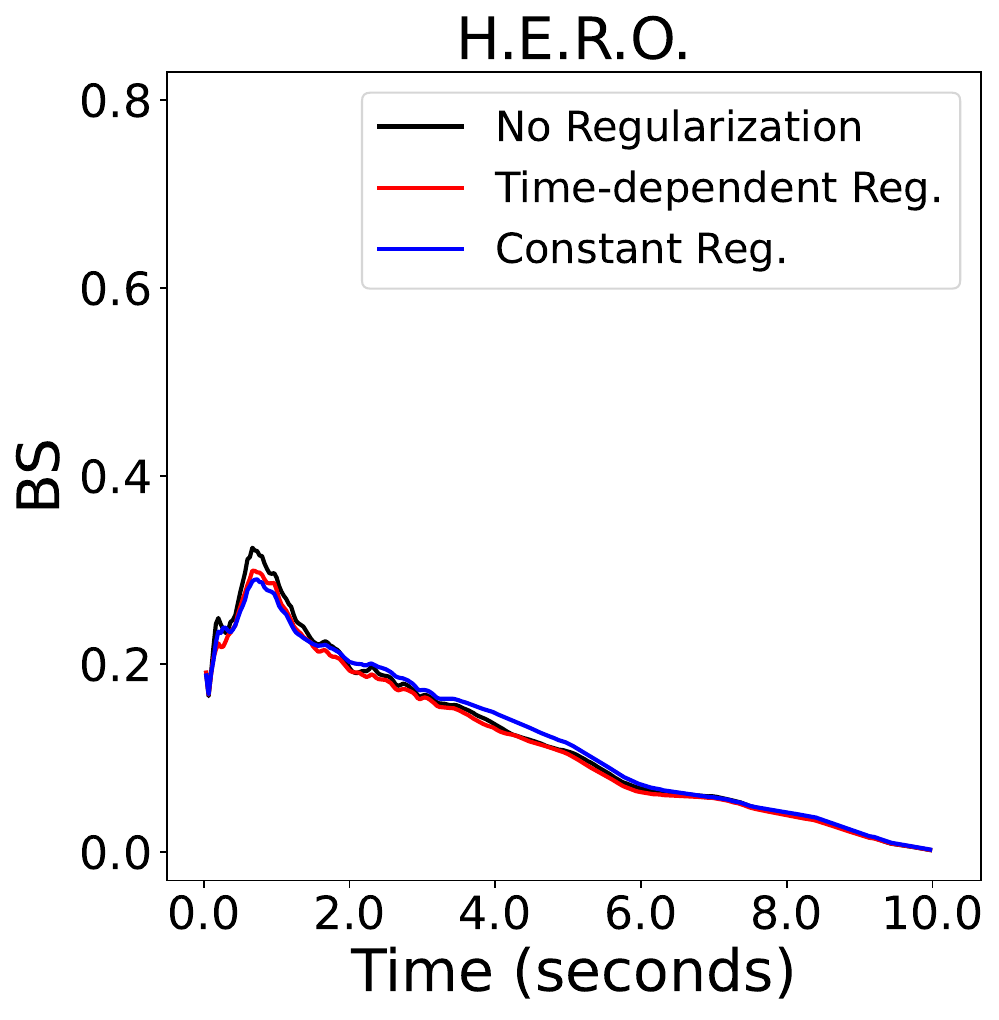}
    \includegraphics[height=0.2\linewidth]{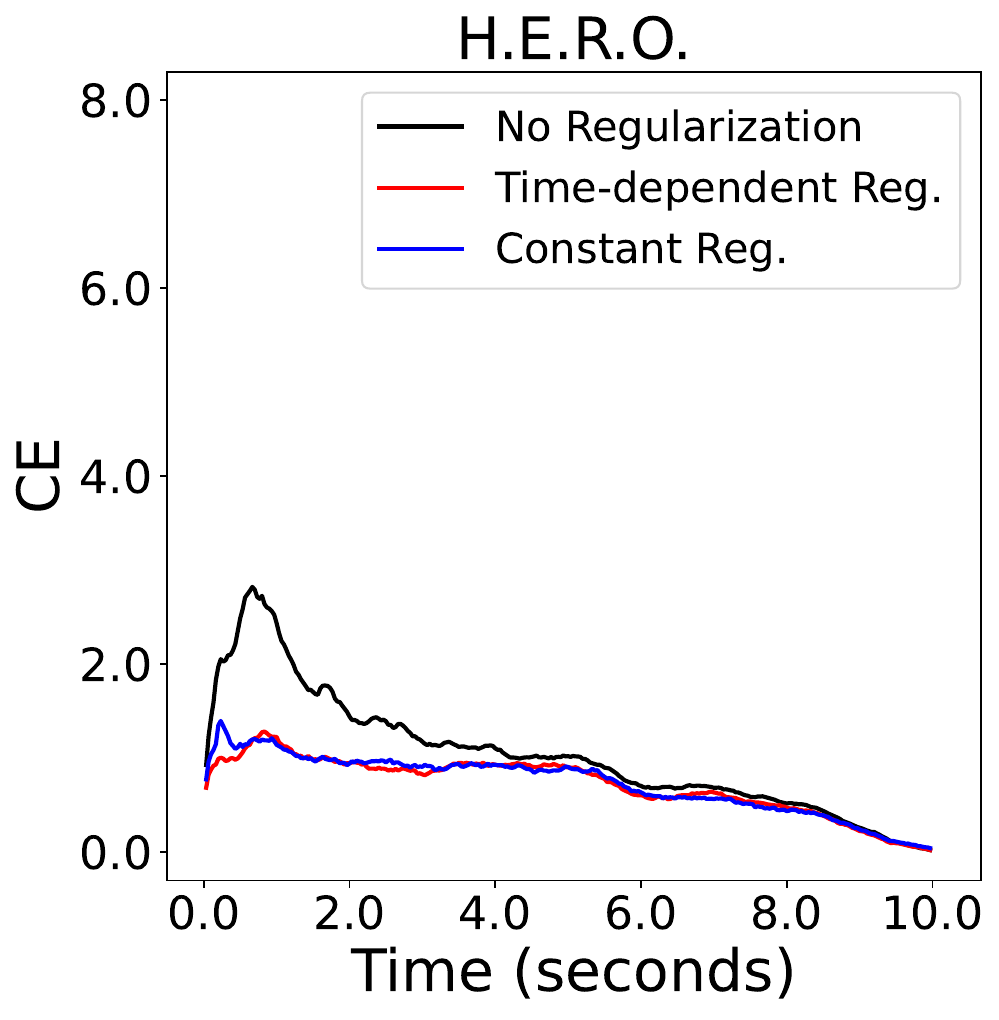} \\

    \includegraphics[height=0.2\linewidth]{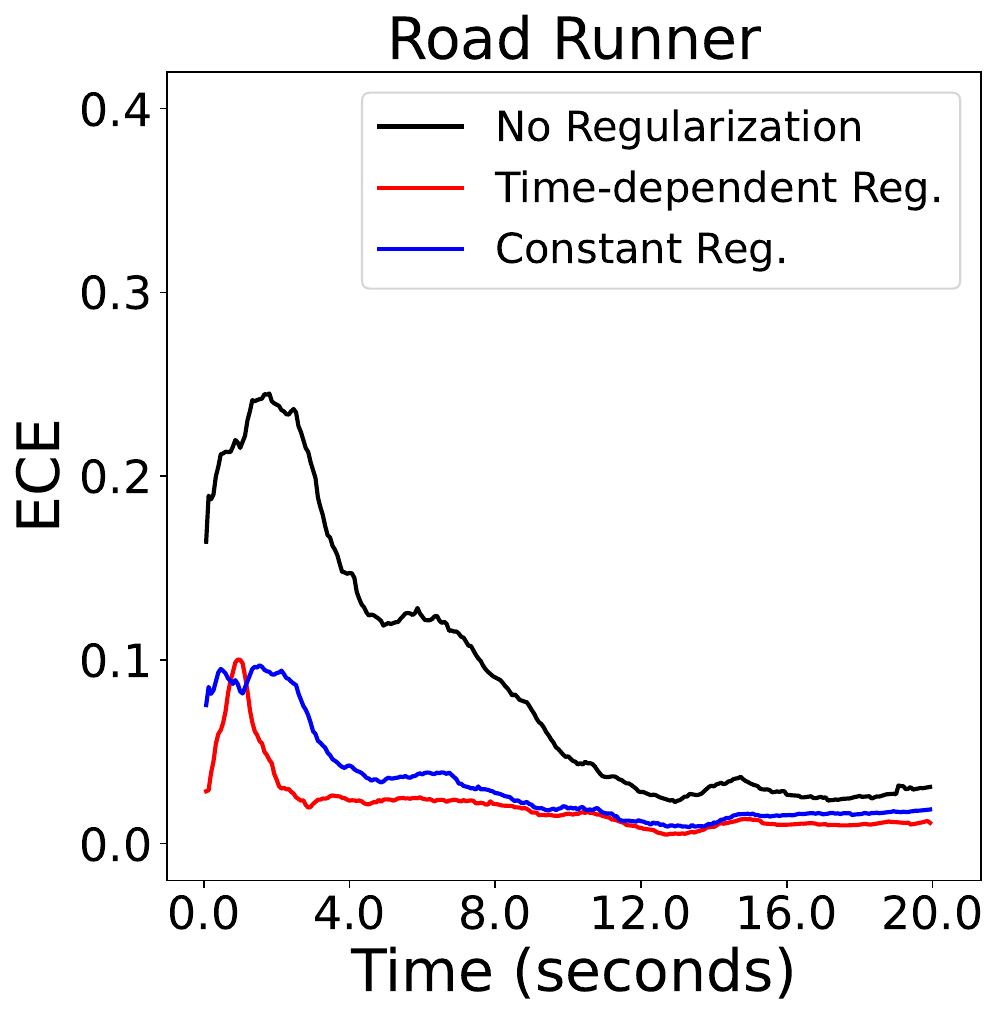}
    \includegraphics[height=0.2\linewidth]{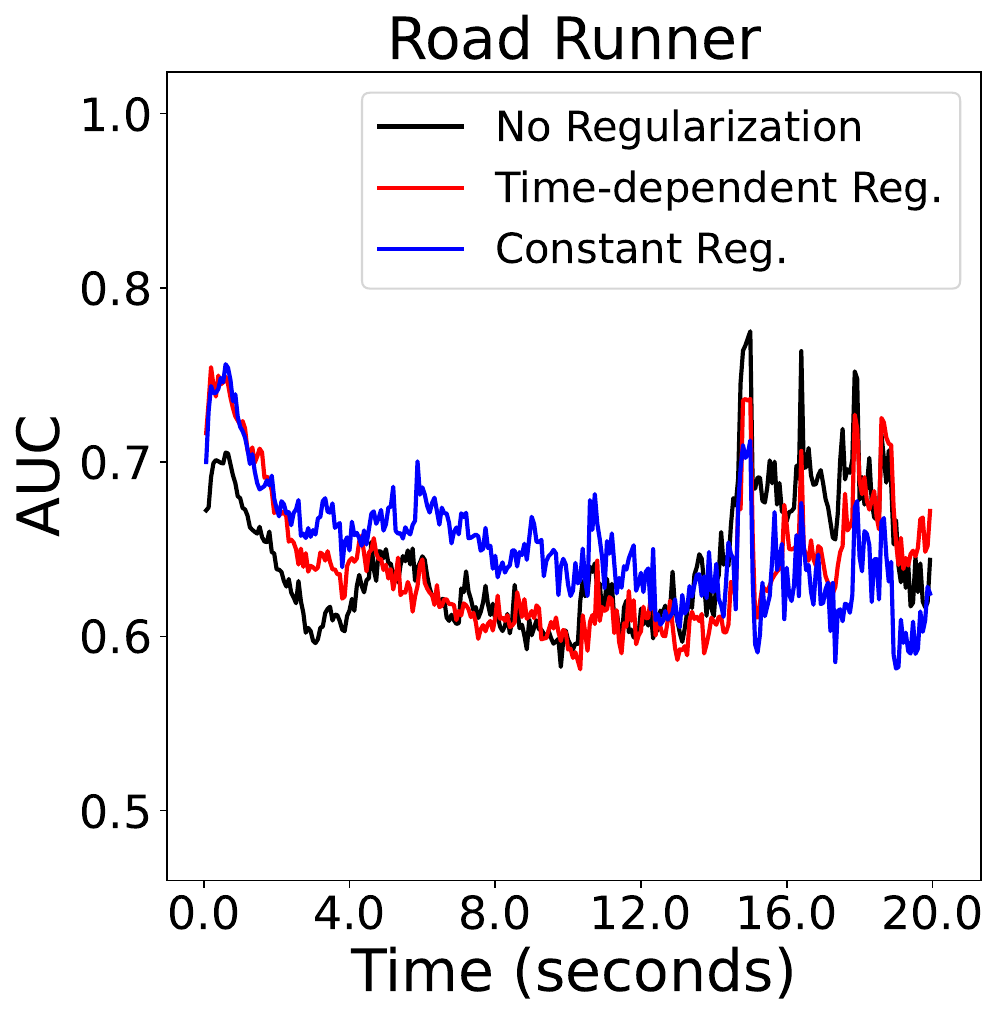}
    \includegraphics[height=0.2\linewidth]{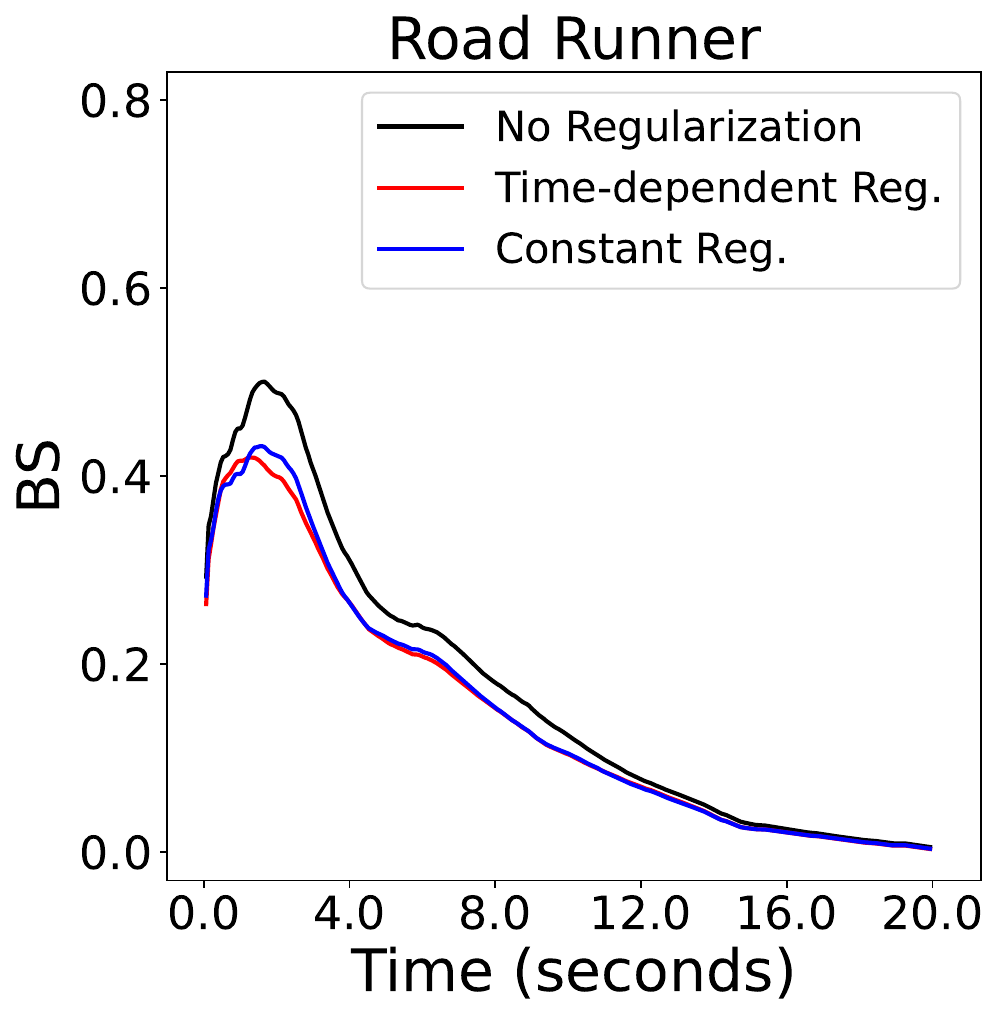}
    \includegraphics[height=0.2\linewidth]{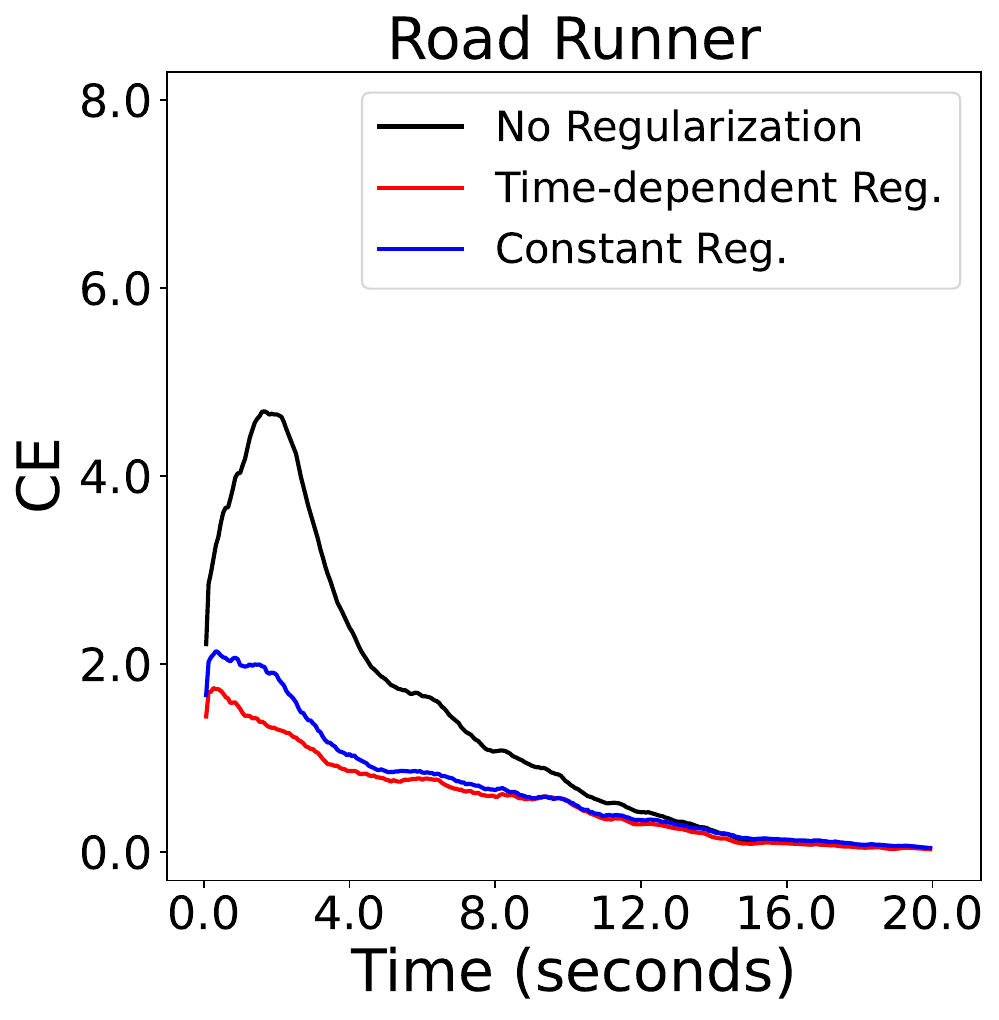} \\

    \includegraphics[height=0.2\linewidth]{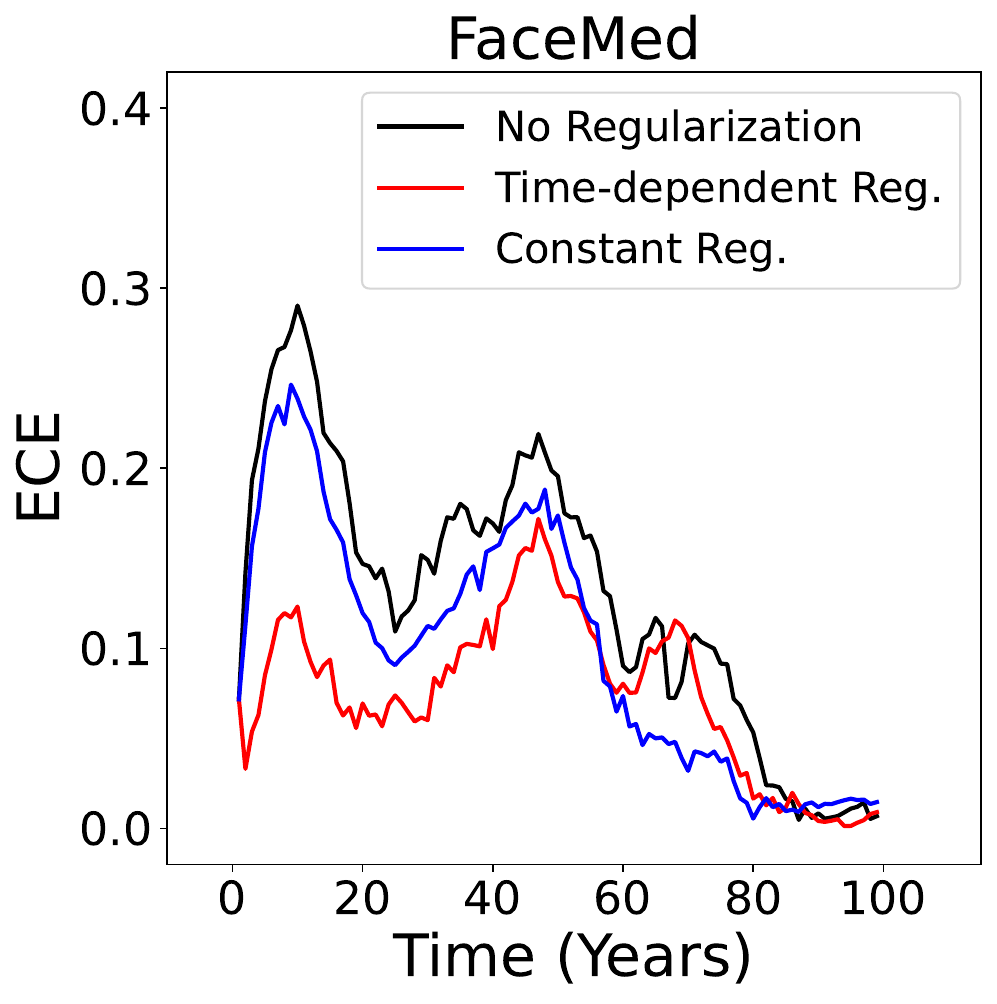}
    \includegraphics[height=0.2\linewidth]{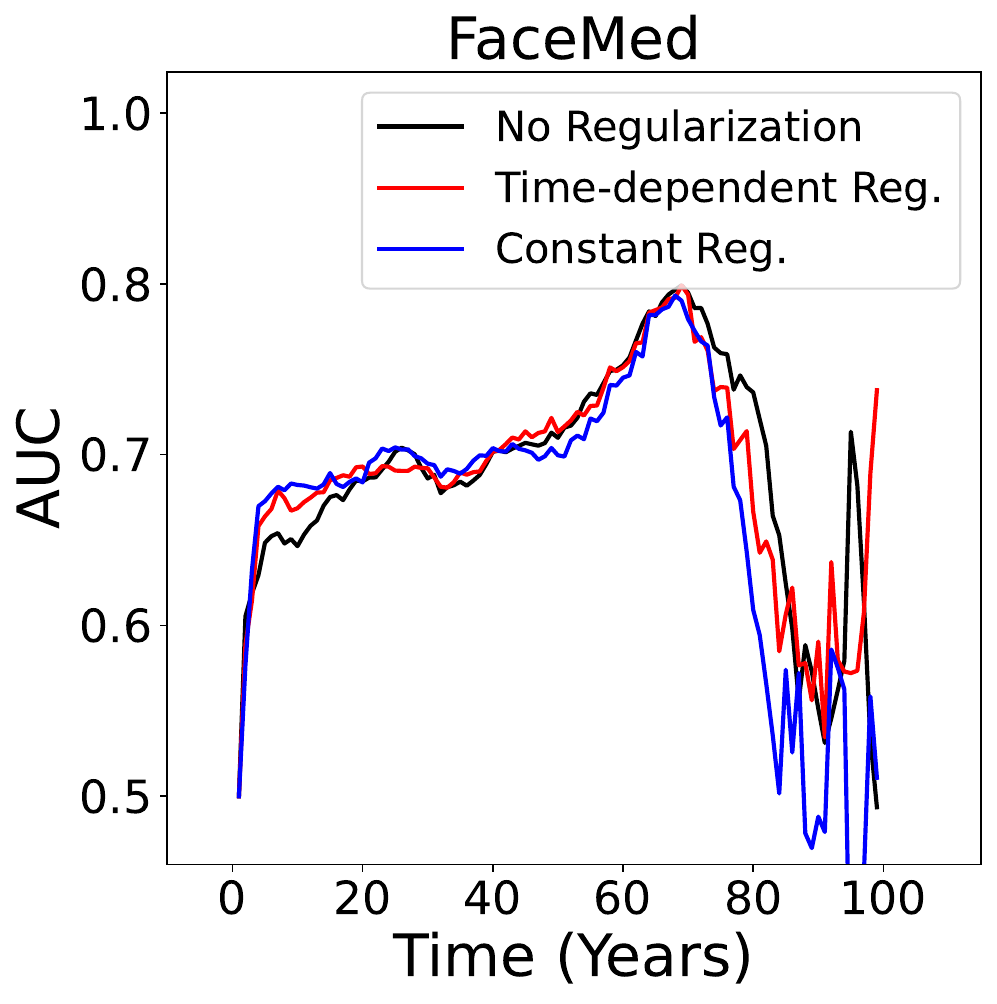}
    \includegraphics[height=0.2\linewidth]{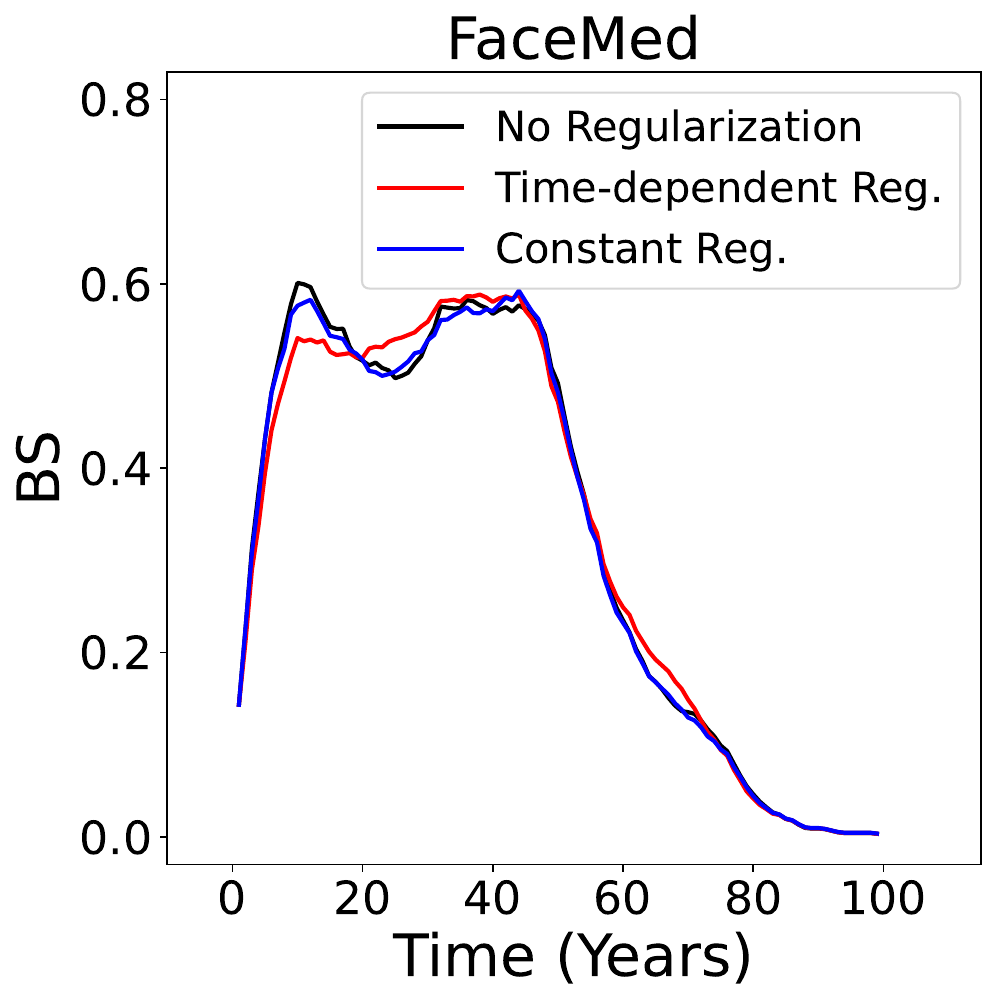}
    \includegraphics[height=0.2\linewidth]{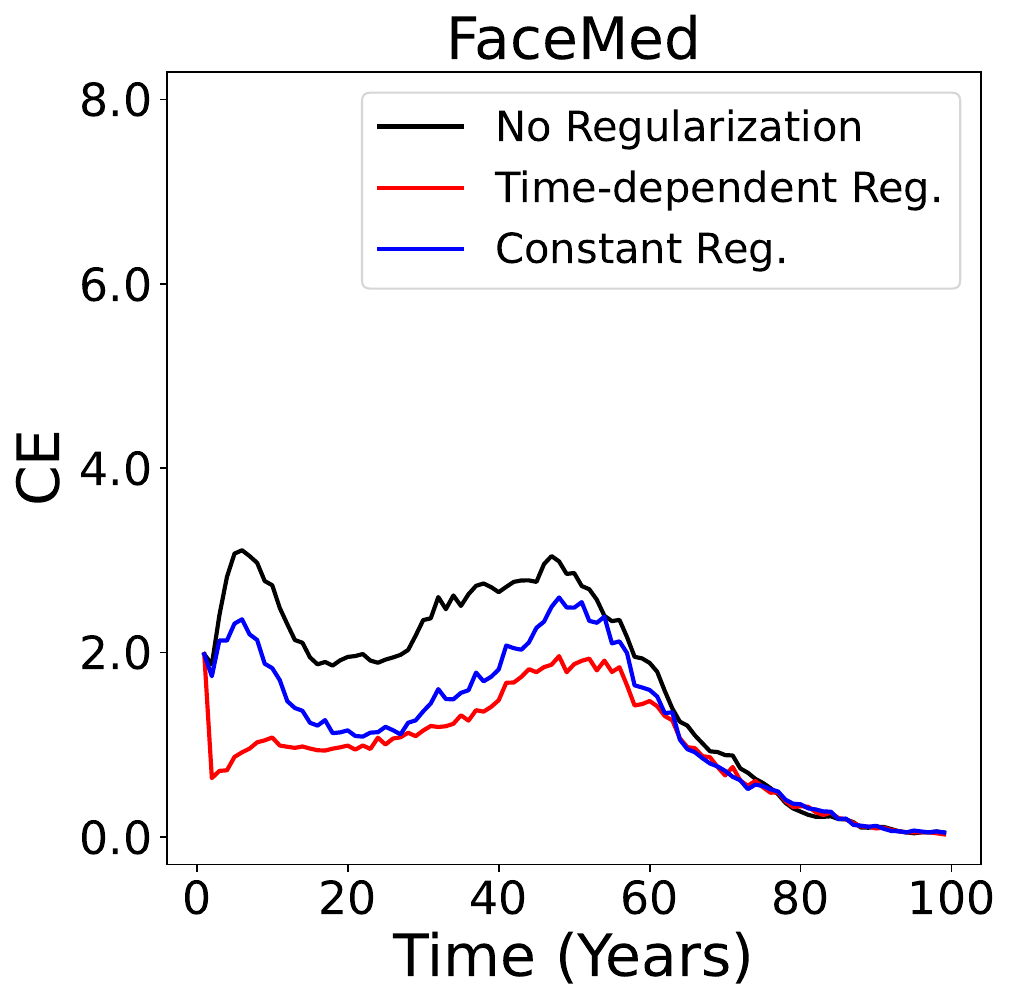}
    
    \caption{\textbf{Entry-wise metrics for conditional probability estimation.} The figure plots the entry-level ECE, AUC, BS, and CE of the proposed foCus framework for estimation of conditional probabilities. It compares versions of foCus: without regularization (black), time-dependent regularization (red), and constant regularization (blue). Both constant regularization and time-dependent regularization improve calibration and overall estimation quality compared to without regularization. Time-dependent regularization's improvement is more significant than constant regularization.
    }
    \label{fig:last_epoch_score_condition_game}
\end{figure*}

We evaluate the conditional probability estimation, given a fixed event in each scenario. For conditional probability estimation, FaceMed is conditioned on the status of the first year being Healthy. Video games are conditioned on the first entry being equal to the most frequent first action in the training set. Specifically, Seaquest is conditioned on first action as \textit{NOOP} (no operation), River Raid as \textit{NOOP}, Bank Heist as \textit{Right}, H.E.R.O. as \textit{NOOP}, and Road Runner as \textit{Left}. \Cref{tab:last_epoch_score_condition} reports the sequence-level metrics for conditional probability estimation. \Cref{fig:last_epoch_score_condition_game} shows entry-level metrics (ECE, AUC, BS, and CE) for conditional probability estimation. Similar to the case of marginal probability estimation \Cref{app:additional_results_marginal}, time-dependent regularization improves calibration and the overall quality of probability estimation. 

\begin{table*}
\small
\caption{
\textbf{Conditional probability estimation.} The table, discussed in \Cref{sec:results}, reports sequence-level metrics evaluating the performance of the proposed foCus framework for estimation of conditional probabilities (see \Cref{sec:problem_statement}). We compare versions of foCus without regularization (see \Cref{sec:pathology}) and with constant and time-dependent regularization (see \Cref{sec:regularization}).
Results are presented as mean ± standard error from three independent model realizations. Time-dependent regularization improves calibration substantially (lower ECE), while maintaining a comparable AUC, which results in superior probability estimates (lower CE and BS).}
\label{tab:last_epoch_score_condition}
\begin{center}
\begin{tabular}{cccccc}
\toprule
Scenario & Regularization & ECE $(\downarrow)$ & AUC $(\uparrow)$ & CE $(\downarrow)$ & BS $(\downarrow)$ \\
\midrule
\multirow{3}{*}{Seaquest} & \xmark & 0.0590 $\pm$ 0.0035 & 0.8887 $\pm$ 0.0045 & 0.9032 $\pm$ 0.0779 & 0.1386 $\pm$ 0.0016 \\
& time-dependent & 0.0316 $\pm$ 0.0017 & \bf{0.8938 $\pm$ 0.0009} & \bf{0.6070 $\pm$ 0.0110} & 0.1301 $\pm$ 0.0053 \\
& constant & \bf{0.0298 $\pm$ 0.0017} & 0.8827 $\pm$ 0.0021 & 0.6227 $\pm$ 0.0278 & \bf{0.1289 $\pm$ 0.0007} \\
\midrule
\multirow{3}{*}{River Raid} & \xmark & 0.0841 $\pm$ 0.0008 & \bf{0.7007 $\pm$ 0.0046} & 1.4968 $\pm$ 0.0497 & 0.2264 $\pm$ 0.0031 \\
 & time-dependent & \bf{0.0652 $\pm$ 0.0012} & 0.6997 $\pm$ 0.0036 & \bf{1.1587 $\pm$ 0.0117} & 0.2248 $\pm$ 0.0018 \\
 & constant & 0.0689 $\pm$ 0.0005 & 0.6926 $\pm$ 0.0026 & 1.3235 $\pm$ 0.0434 & \bf{0.2225 $\pm$ 0.0015} \\
\midrule
\multirow{3}{*}{Bank Heist} & \xmark & 0.0534 $\pm$ 0.0035 & \bf{0.7254 $\pm$ 0.0036} & 1.0618 $\pm$ 0.0563 & 0.2287 $\pm$ 0.0030 \\
 & time-dependent & \bf{0.0184 $\pm$ 0.0005} & 0.7033 $\pm$ 0.0021 & \bf{0.7616 $\pm$ 0.0046} & \bf{0.2141 $\pm$ 0.0008} \\
 & constant & 0.0405 $\pm$ 0.0022 & 0.7149 $\pm$ 0.0124 & 0.8431 $\pm$ 0.0080 & 0.2168 $\pm$ 0.0013 \\
\midrule
\multirow{3}{*}{H.E.R.O.} & \xmark & 0.0993 $\pm$ 0.0039 & 0.7063 $\pm$ 0.0002 & 1.0290 $\pm$ 0.0371 & 0.1195 $\pm$ 0.0021 \\
 & time-dependent & 0.0645 $\pm$ 0.0087 & \bf{0.7420 $\pm$ 0.0132} & \bf{0.7189 $\pm$ 0.0289} & \bf{0.1151 $\pm$ 0.0029} \\
 & constant & \bf{0.0418 $\pm$ 0.0019} & 0.7249 $\pm$ 0.0141 & 0.7283 $\pm$ 0.0572 & 0.1208 $\pm$ 0.0024 \\
\midrule
\multirow{3}{*}{Road Runner} & \xmark & 0.0870 $\pm$ 0.0061 & 0.6790 $\pm$ 0.0091 & 1.2813 $\pm$ 0.0244 & 0.1693 $\pm$ 0.0039 \\
 & time-dependent & \bf{0.0207 $\pm$ 0.0012} & 0.6772 $\pm$ 0.0107 & \bf{0.5489 $\pm$ 0.0089} & \bf{0.1449 $\pm$ 0.0003} \\
 & constant & 0.0321 $\pm$ 0.0032 & \bf{0.6905 $\pm$ 0.0135} & 0.6654 $\pm$ 0.0304 & 0.1472 $\pm$ 0.0012 \\
\midrule
\multirow{3}{*}{FaceMed} & \xmark & 0.1488 $\pm$ 0.0026 & 0.7533 $\pm$ 0.0066 & 1.6428 $\pm$ 0.0147 & 0.3465 $\pm$ 0.0012 \\
 & time-dependent & \bf{0.0888 $\pm$ 0.0065} & 0.7603 $\pm$ 0.0040 & \bf{1.0224 $\pm$ 0.0348} & \bf{0.3350 $\pm$ 0.0018} \\
 & constant & 0.0984 $\pm$ 0.0042 & \bf{0.7652 $\pm$ 0.0050} & 0.9875 $\pm$ 0.0372 & 0.3354 $\pm$ 0.0016 \\
\bottomrule
\end{tabular}
\end{center}
\end{table*}

\begin{comment}
\begin{figure*}
\centering
\includegraphics[width=0.5\columnwidth]{drafts/_aistats2025/Figures/Last_Epoch_ECE_Condition.pdf}
\includegraphics[width=0.5\columnwidth]{drafts/_aistats2025/Figures/Last_Epoch_Score_Condition.pdf}
\caption{Similar to Figure~\ref{fig:last_epoch_score_uncondition_game} but all games have a fixed first step action condition.}
\label{fig:last_epoch_score_condition_game}
\end{figure*}
\end{comment}

\subsection{Confidence Interval for Time-to-event}
\label{app:additional_results_ci}

\begin{table*}
\caption{\textbf{Performance comparison of time-to-event prediction confidence intervals (CI) across six scenarios.} The table presents metrics from the same experiments as in \Cref{tab:last_epoch_score_uncondition}. The time-dependent regularization model achieves significantly better 90\% CI ($I_{0.9}$) coverage, indicating improved calibration of uncertainty estimation.}
\label{tab:last_epoch_length_score}
\begin{center}
\begin{tabular}{lcccc}
\toprule
Scenario & Regularization &  Coverage Prob. of $I_{0.9}$ & Relative Width of $I_{0.9}$ & Relative MAE $(\downarrow)$  \\
\midrule
\multirow{3}{*}{Seaquest} & \xmark & 0.3420 $\pm$ 0.0177 & 0.7401 $\pm$ 0.0149 & 0.4622 $\pm$ 0.0112 \\
 & time-dependent & \bf{0.7065 $\pm$ 0.0121} & 2.1718 $\pm$ 0.0521 & 0.5297 $\pm$ 0.0079 \\
 & constant & 0.4440 $\pm$ 0.0162 & 0.9409 $\pm$ 0.0266 & \bf{0.4538 $\pm$ 0.0069} \\
\midrule
\multirow{3}{*}{River Raid} & \xmark & 0.4463 $\pm$ 0.0105 & 1.0734 $\pm$ 0.0398 & 0.5679 $\pm$ 0.0044 \\
 & time-dependent & \bf{0.8345 $\pm$ 0.0090} & 1.8720 $\pm$ 0.0520 & 0.5674 $\pm$ 0.0037 \\
 & constant & 0.6241 $\pm$ 0.0018 & 1.2079 $\pm$ 0.0241 & \bf{0.5580 $\pm$ 0.0012} \\
\midrule
\multirow{3}{*}{Bank Heist} & \xmark & 0.7066 $\pm$ 0.0134 & 1.8289 $\pm$ 0.0330 & 0.5811 $\pm$ 0.0052 \\
 & time-dependent & \bf{0.9243 $\pm$ 0.0048} & 2.8099 $\pm$ 0.0535 & 0.5991 $\pm$ 0.0080 \\
 & constant & 0.8133 $\pm$ 0.0083 & 1.9123 $\pm$ 0.0335 & \bf{0.5464 $\pm$ 0.0025} \\
\midrule
\multirow{3}{*}{H.E.R.O.} & \xmark & 0.2252 $\pm$ 0.0009 & 0.3712 $\pm$ 0.0122 & \bf{0.2849 $\pm$ 0.0057} \\
 & time-dependent & \bf{0.6850 $\pm$ 0.0121} & 0.9035 $\pm$ 0.0266 & 0.3316 $\pm$ 0.0041 \\
 & constant & 0.5827 $\pm$ 0.0102 & 0.7158 $\pm$ 0.0330 & 0.3337 $\pm$ 0.0025 \\
\midrule
\multirow{3}{*}{Road Runner} & \xmark & 0.2646 $\pm$ 0.0128 & 1.2432 $\pm$ 0.1125 & 0.6079 $\pm$ 0.0277 \\
 & time-dependent & \bf{0.7507 $\pm$ 0.0086} & 2.9315 $\pm$ 0.0277 & 0.6419 $\pm$ 0.0012 \\
 & constant & 0.4971 $\pm$ 0.0073 & 1.6267 $\pm$ 0.1351 & \bf{0.5652 $\pm$ 0.0078} \\
\midrule
\multirow{3}{*}{FaceMed} & \xmark & 0.2789 $\pm$ 0.0039 & 0.1329 $\pm$ 0.0035 & \bf{0.1594 $\pm$ 0.0005} \\
 & time-dependent & \bf{0.7169 $\pm$ 0.0322} & 1.0421 $\pm$ 0.0232 & 0.2311 $\pm$ 0.0066 \\
 & constant & 0.5897 $\pm$ 0.0227 & 0.7632 $\pm$ 0.0212 & 0.4753 $\pm$ 0.0032 \\
\bottomrule
\end{tabular}
\end{center}
\end{table*}

\Cref{tab:last_epoch_length_score} reports the  results for time-to-event confident interval estimation. In this case, we observe a certain trade-off between discriminative performance, quantified by the relative MAE, and calibration, quantified by coverage probabilities. The MAE of no regularization and constant regularization are consistently lower than those of time-dependent regularization, but the coverage probabilities of time-dependent regularization are a lot closer to 90\% (between 69\% and 92\%, compared to at most 70\% for the other two methods). 

\begin{figure}
\centering
\includegraphics[width=0.45\columnwidth]{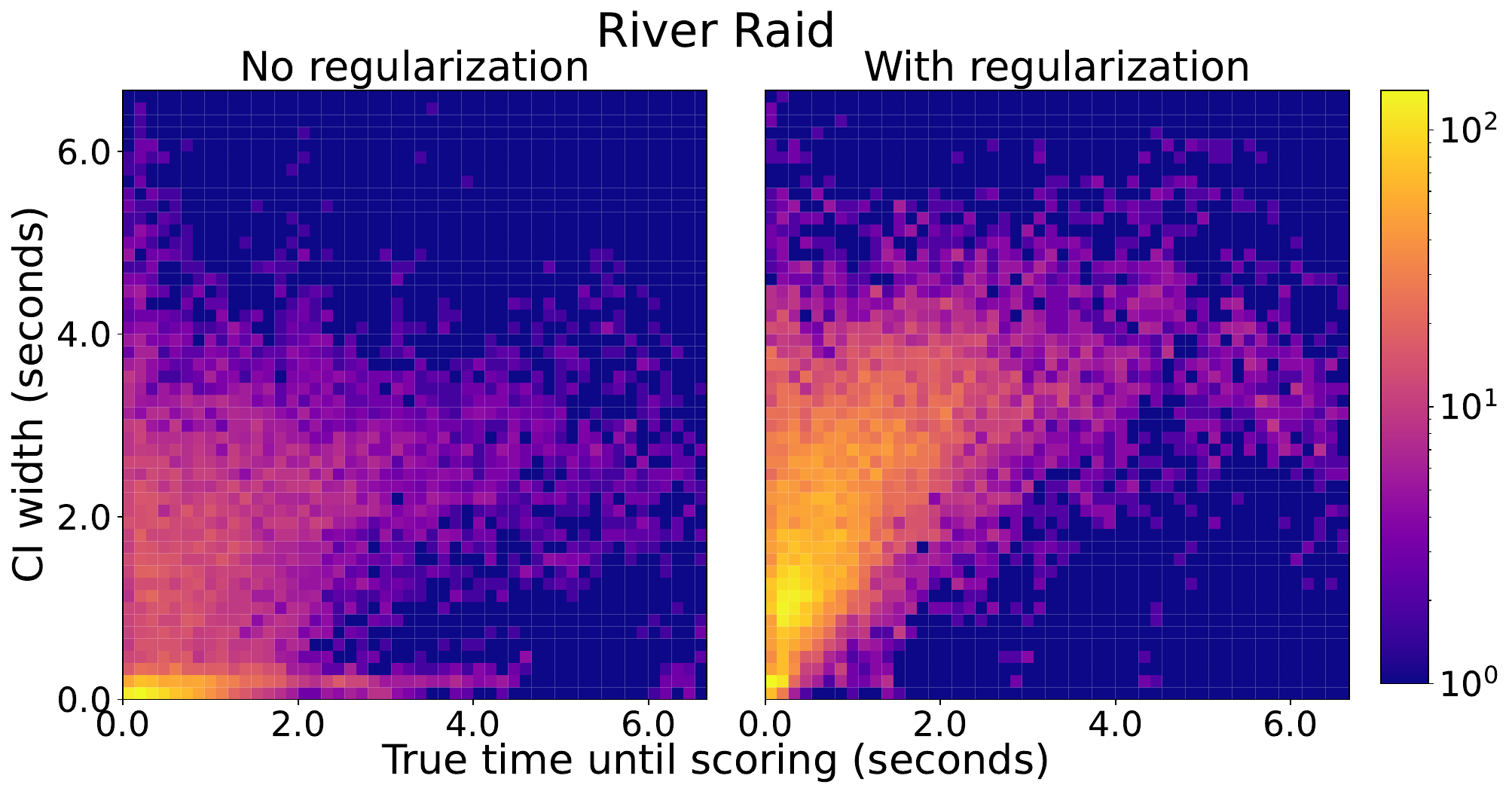}
\includegraphics[width=0.45\columnwidth]{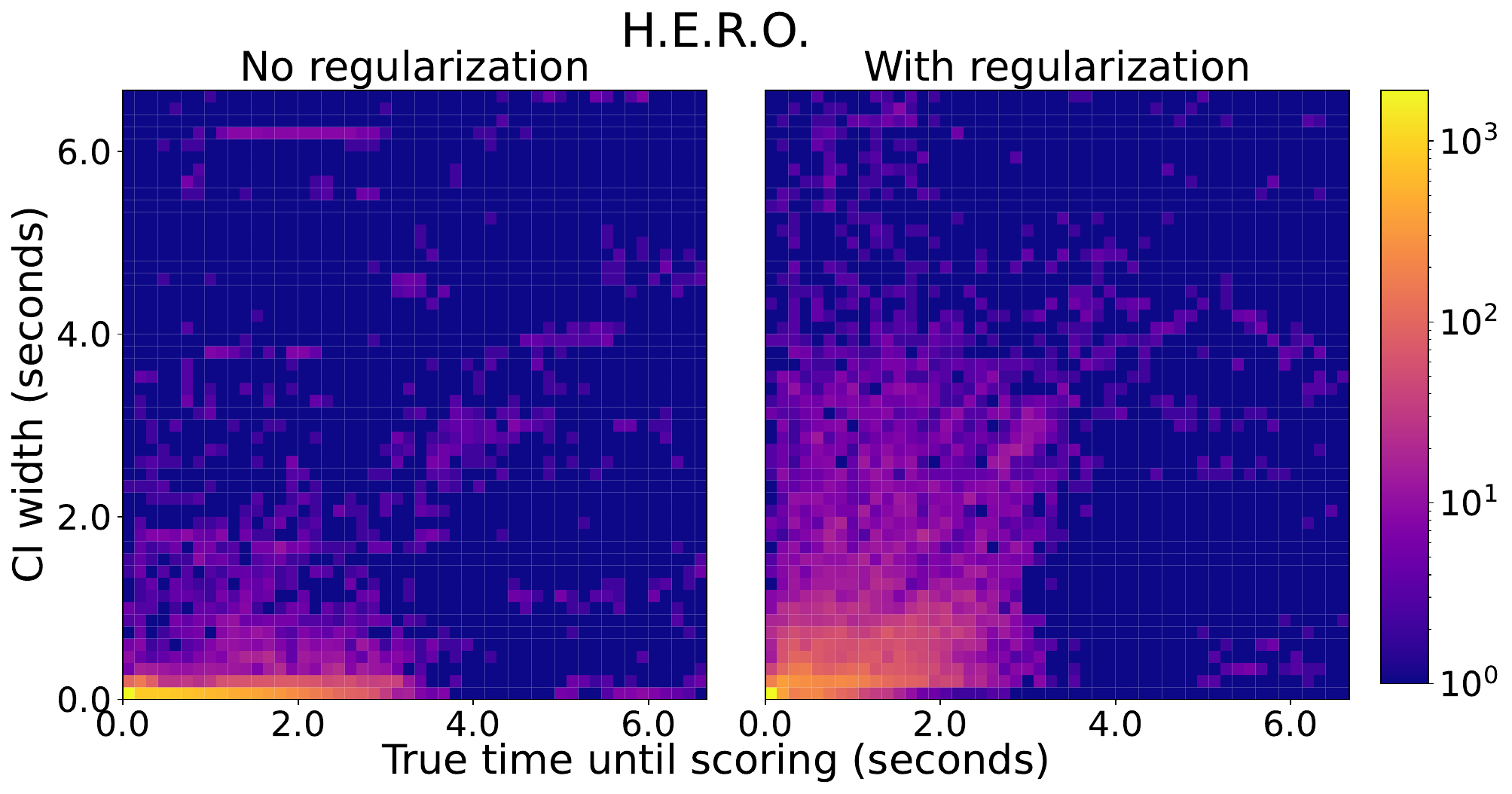}\\

\includegraphics[width=0.401\columnwidth]{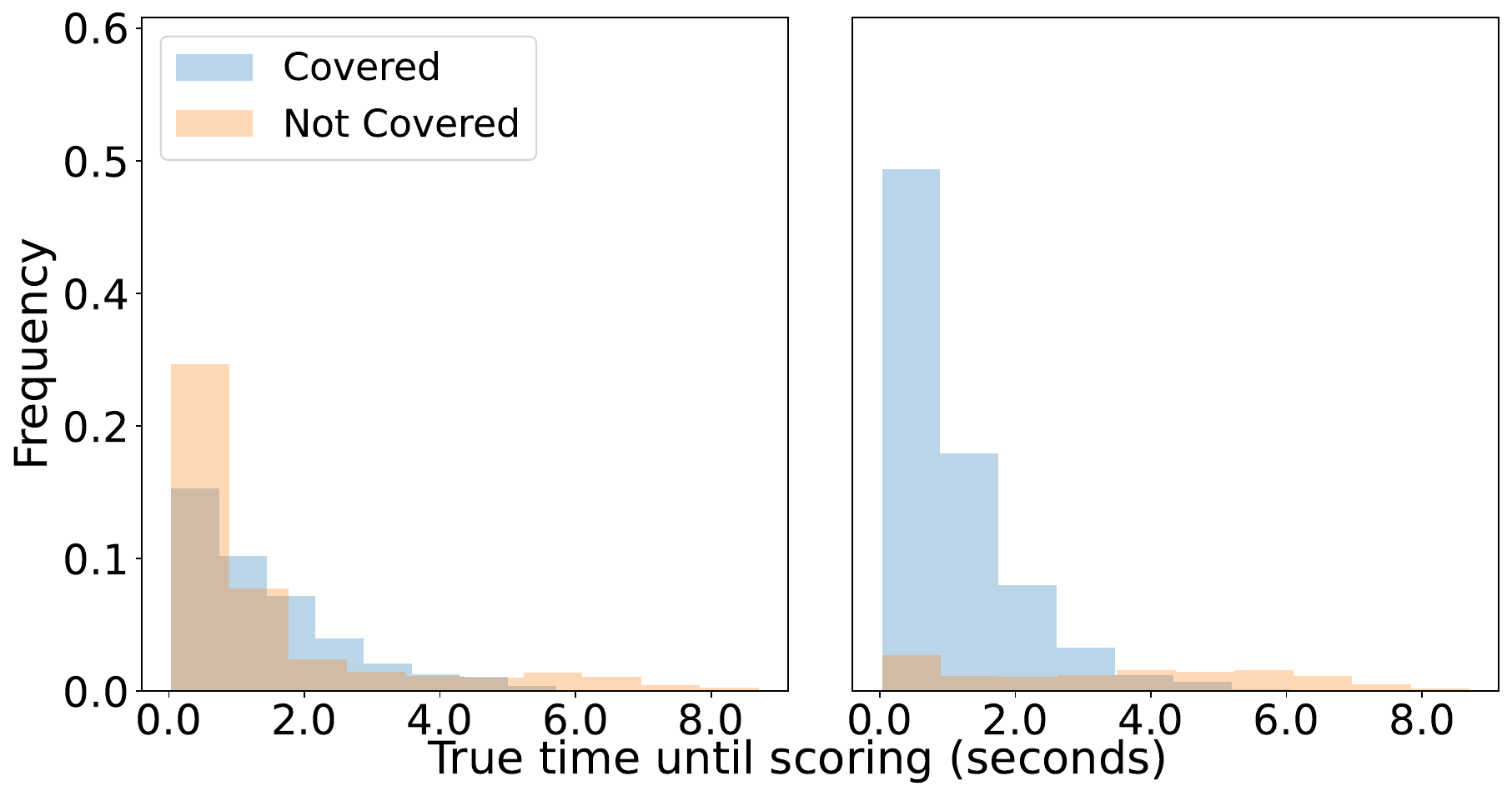}
\hspace{6.8mm}
\includegraphics[width=0.401\columnwidth]{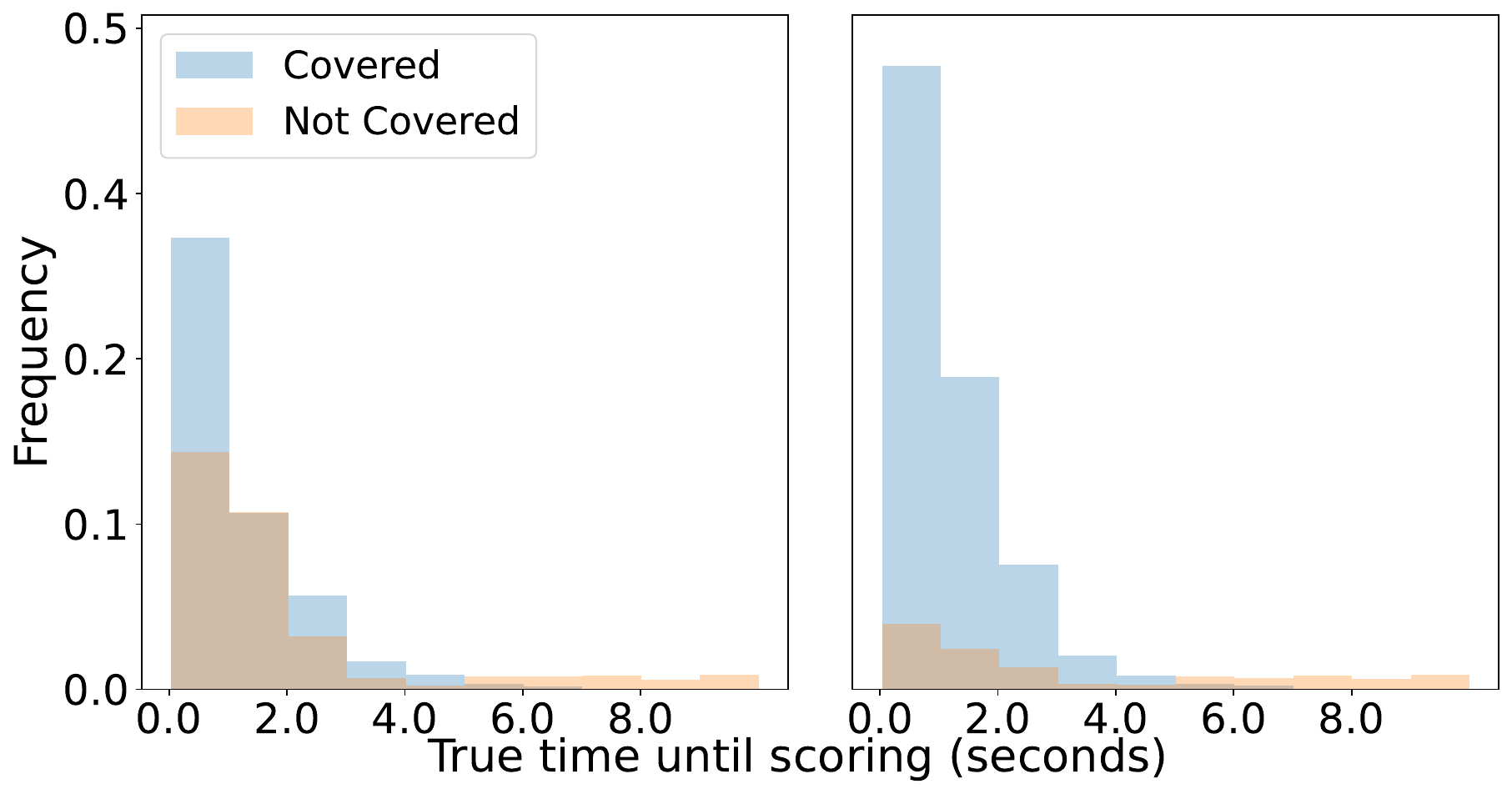}
\hspace{6.8mm}

\vspace{5mm}

\includegraphics[width=0.45\columnwidth]{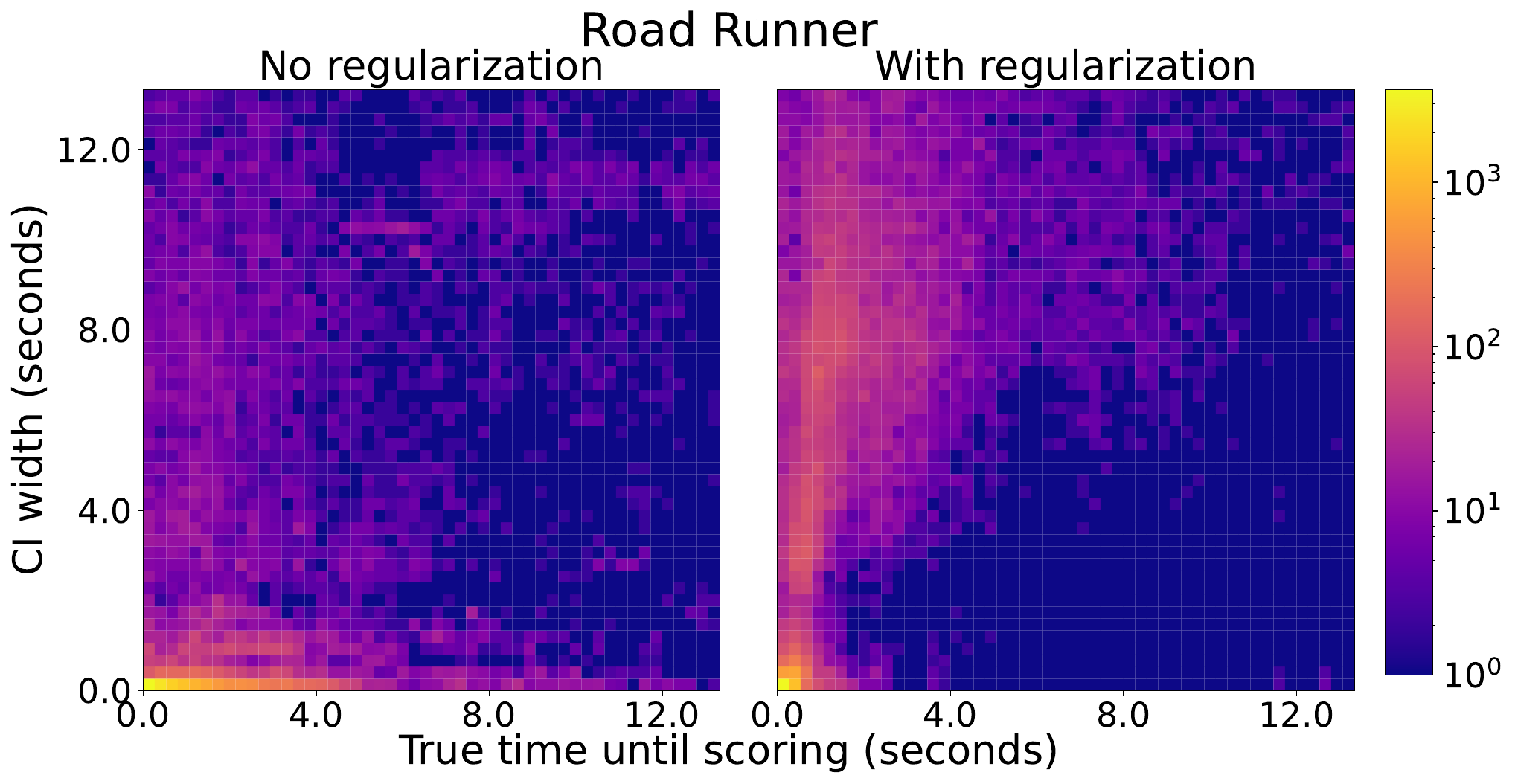}
\includegraphics[width=0.45\columnwidth]{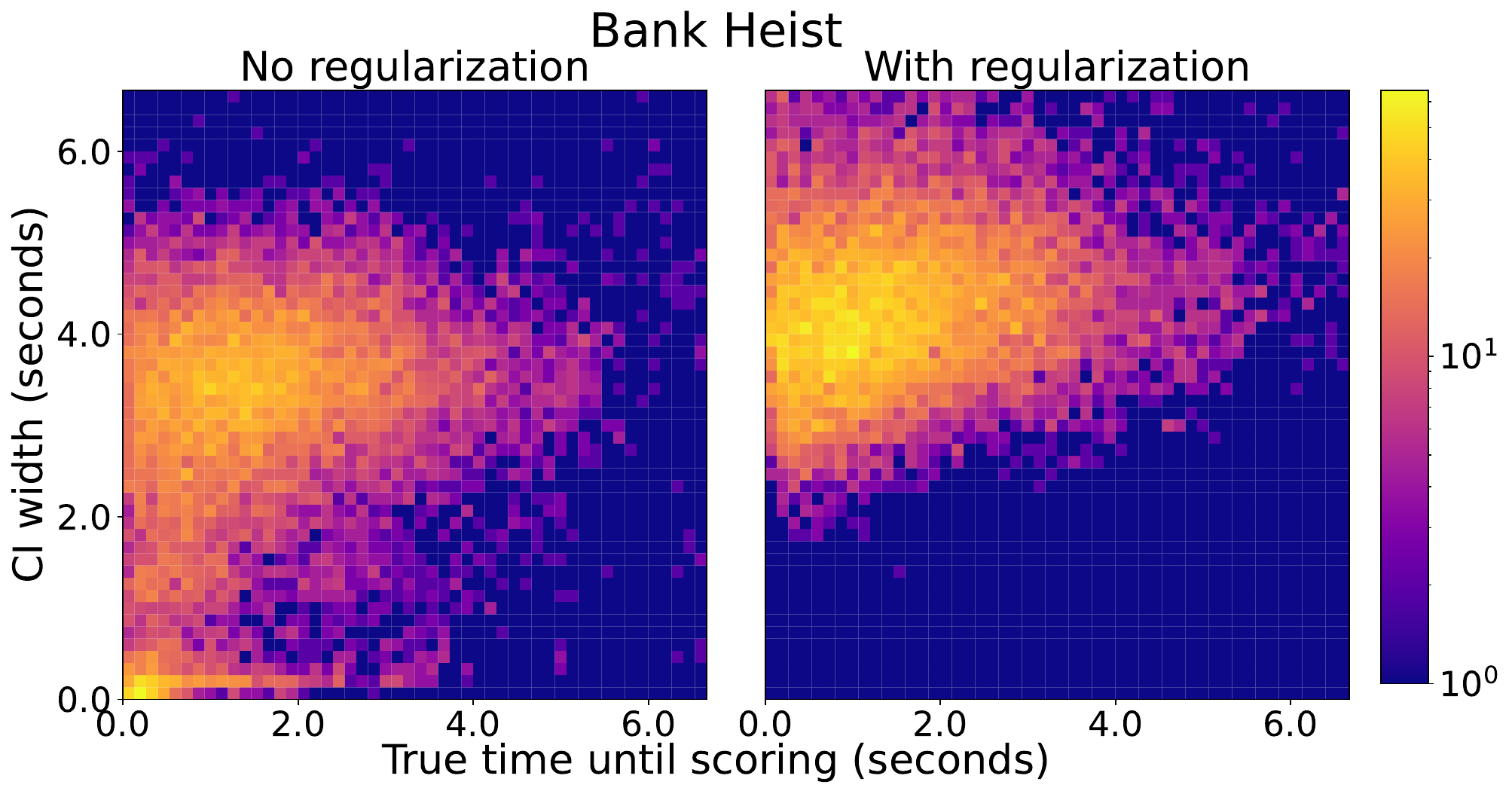}\\

\includegraphics[width=0.401\columnwidth]{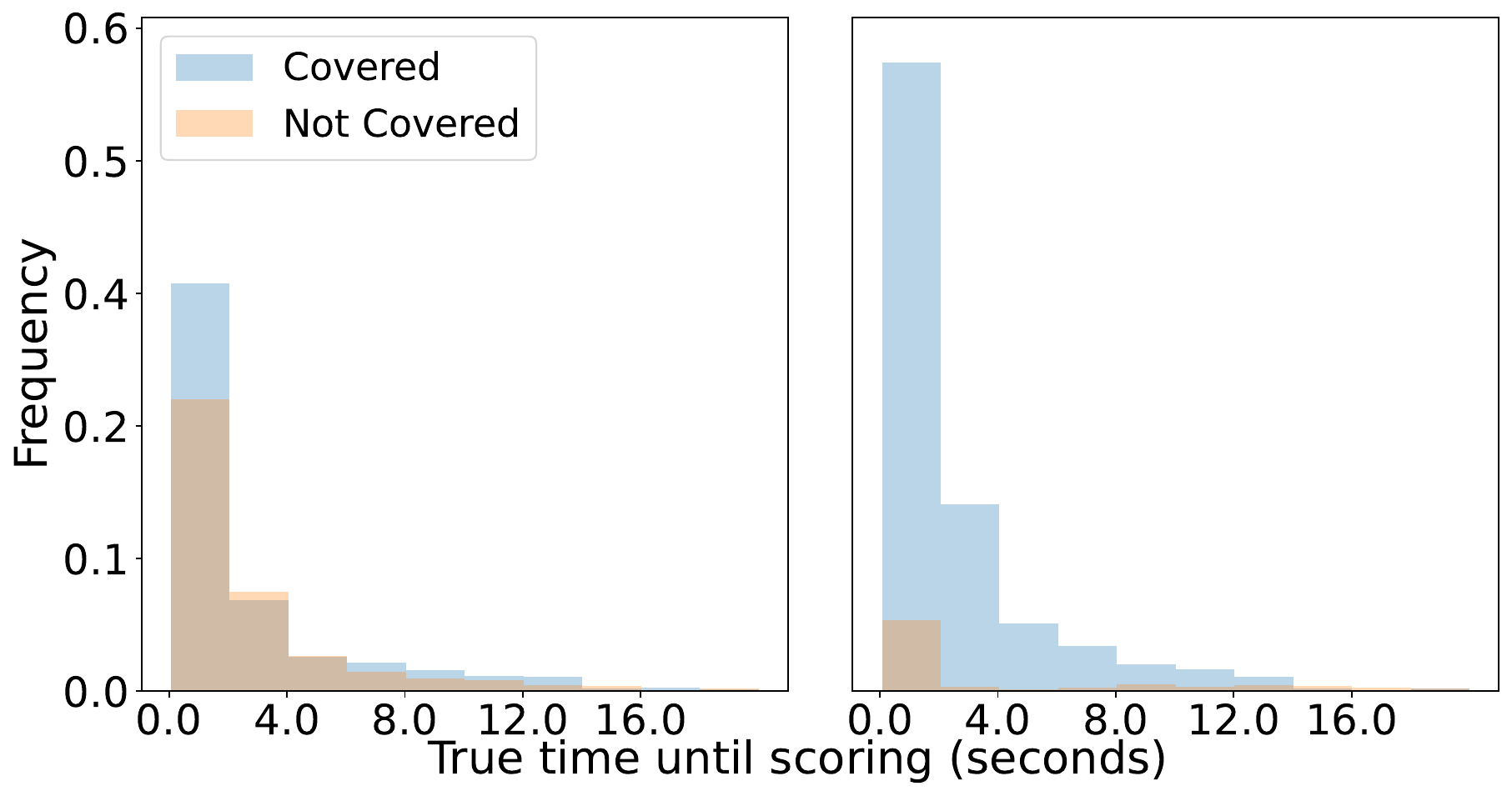}
\hspace{6.8mm}
\includegraphics[width=0.401\columnwidth]{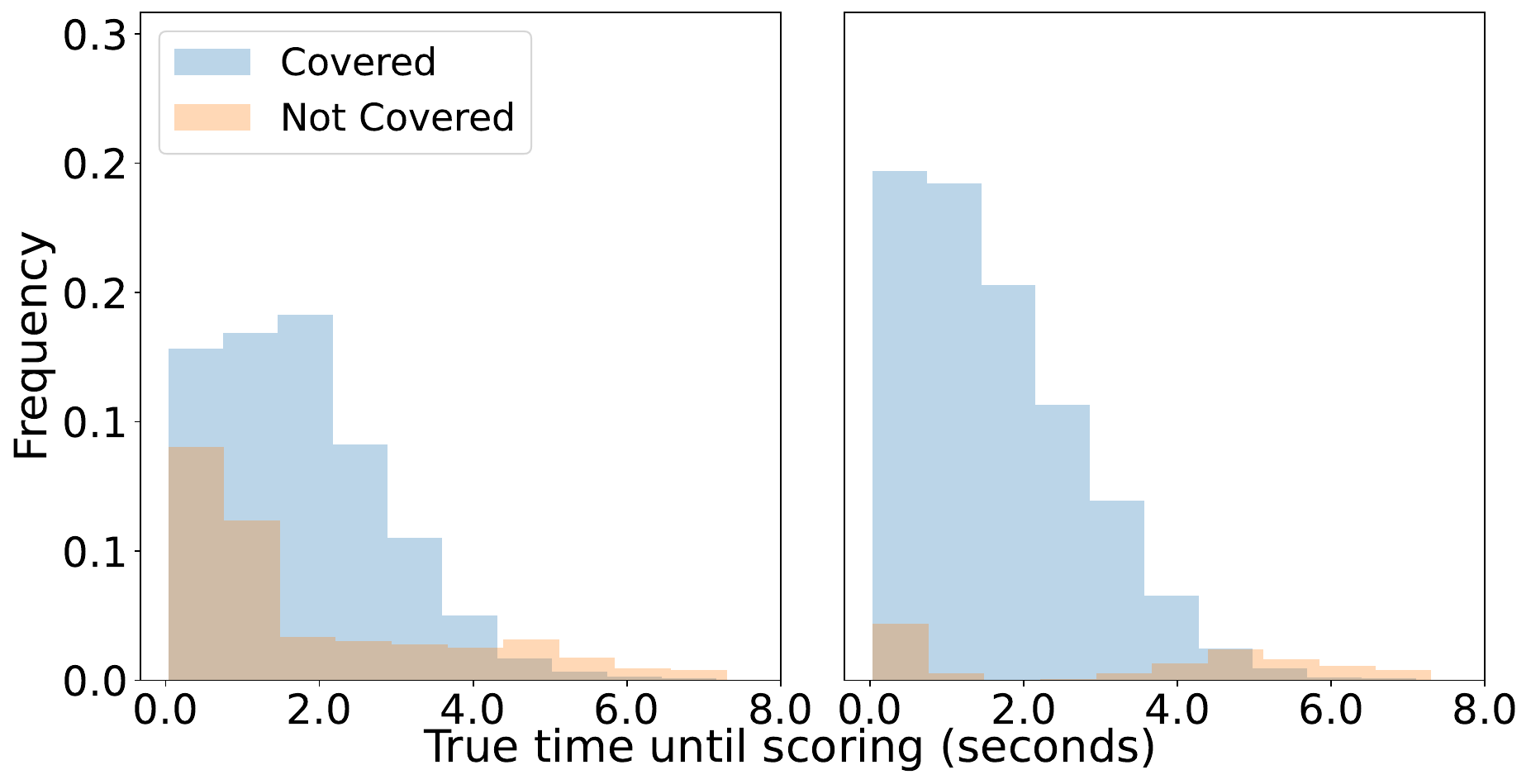}
\hspace{6.8mm}

\caption{\textbf{Confidence intervals for time-to-event prediction and coverage probability.}
The upper panel of each subplot shows heatmaps of the length of 0.9 confidence intervals for time-to-event prediction using the proposed foCus framework without and with time-dependent regularization. The histograms below show the frequency of intervals containing the true times, as a function of the true times. Unregularized foCus produces short intervals with poor coverage, whereas regularization yields intervals that tend to be larger when the ground-truth times are larger, and are much better calibrated. The subplots correspond to the River Raid, H.E.R.O., Road Runner, and Bank Heist datasets.
}
\label{fig:extra_coverage_plots}
\end{figure}

\Cref{fig:extra_coverage_plots}, complementing \Cref{fig:CI_analysis}, presents heatmaps of the confidence interval widths for different ground-truth time-to-event (upper panel of each subplot) for River Raid, H.E.R.O., Road Runner, and Bank Heist, as well as a histogram showing the fraction of intervals containing the ground-truth (lower panel of each subplot). Unregularized foCus produces very narrow confidence intervals with very poor coverage, whereas time-dependent regularization yields intervals that tend to be larger when the true time-to-event becomes greater (and hence generally more uncertain), achieving much better coverage. These finding further supports that time-dependent regularization performs better in confidence interval estimation. 

\section{Regularization Sensitivity Analysis}
\label{app:ablation}
In this section, we explore how regularization at a single entry affects model performance. Using the Seaquest game, we train four versions of foCus, each applying regularization to a different entry in the sequence, corresponding to 0, 3, 6, and 9 seconds. As shown in Tables~\ref{tab:last_epoch_score_varying_step} and \ref{tab:last_epoch_length_score_varying_step}, the model with regularization applied at the beginning (0 seconds) exhibits the most significant improvement in calibration, both for marginal probability estimation and time-to-event prediction confidence intervals. This suggests that applying regularization early in the sequence is crucial for maintaining proper calibration throughout the whole sequence. These results also validate our heuristic procedure to implement time-dependent regularization.

\begin{table}[h]
\small
\caption{\textbf{Marginal probability estimation performance sensitivity analysis on Seaquest.} We compare versions of foCus where regularization is applied to a single random variable which is 0, 3, 6, and 9 second(s) after the start of the sequence. Results are presented as mean ± standard error from three independent model realizations. Best performance is achieved when applying regularization to the random variable immediately after the sequence starts.}
\label{tab:last_epoch_score_varying_step}
\begin{center}
\begin{tabular}{ccccc}
\toprule
 Reg. Time & ECE $(\downarrow)$ & AUC $(\uparrow)$ & CE $(\downarrow)$ & BS $(\downarrow)$ \\
\midrule
0 second & \bf{0.0290 $\pm$ 0.0008} & 0.8748 $\pm$ 0.0033 & \bf{0.8073 $\pm$ 0.0120 }& \bf{0.1164 $\pm$ 0.0008} \\
3 seconds & 0.0424 $\pm$ 0.0019 & 0.8705 $\pm$ 0.0027 & 1.0227 $\pm$ 0.0150 & 0.1227 $\pm$ 0.0011 \\
6 seconds & 0.0424 $\pm$ 0.0016 & 0.8783 $\pm$ 0.0024 & 0.9599 $\pm$ 0.0208 & 0.1211 $\pm$ 0.0009 \\
9 seconds & 0.0409 $\pm$ 0.0020 & \bf{0.8784 $\pm$ 0.0023} & 0.9450 $\pm$ 0.0065 & 0.1212 $\pm$ 0.0013 \\
\bottomrule
\end{tabular}
\end{center}
\end{table}

\begin{table}[h]
\small
\caption{\textbf{Time-to-event prediction confidence interval performance sensitivity analysis on Seaquest.} The table presents metrics on time-to-event prediction confidence intervals from the same experiments as in \Cref{tab:last_epoch_score_varying_step}. Regularizing he random variable immediately after the sequence starts significantly improves 90\% CI ($I_{0.9}$) coverage.
}
\label{tab:last_epoch_length_score_varying_step}
\begin{center}
\begin{tabular}{cccc}
\toprule
 Reg. Time & Coverage Prob. of $I_{0.9}$ & Relative Width of $I_{0.9}$ & Relative MAE $(\downarrow)$  \\
\midrule
0 second & \bf{0.4972 $\pm$ 0.0208} & 1.6983 $\pm$ 0.0497 & 0.5162 $\pm$ 0.0125 \\
3 seconds & 0.3577 $\pm$ 0.0171 & 0.7056 $\pm$ 0.0320 & \bf{0.4466 $\pm$ 0.0036} \\
6 seconds & 0.3556 $\pm$ 0.0099 & 0.7338 $\pm$ 0.0126 & 0.4495 $\pm$ 0.0070 \\
9 seconds & 0.3578 $\pm$ 0.0016 & 0.7421 $\pm$ 0.0308 & 0.4522 $\pm$ 0.0138 \\
\bottomrule
\end{tabular}
\end{center}
\end{table}

\section{Conditional Probability Estimation}
\label{app:case_study}
\begin{figure*}
\begin{center}
\includegraphics[width=0.9\columnwidth]{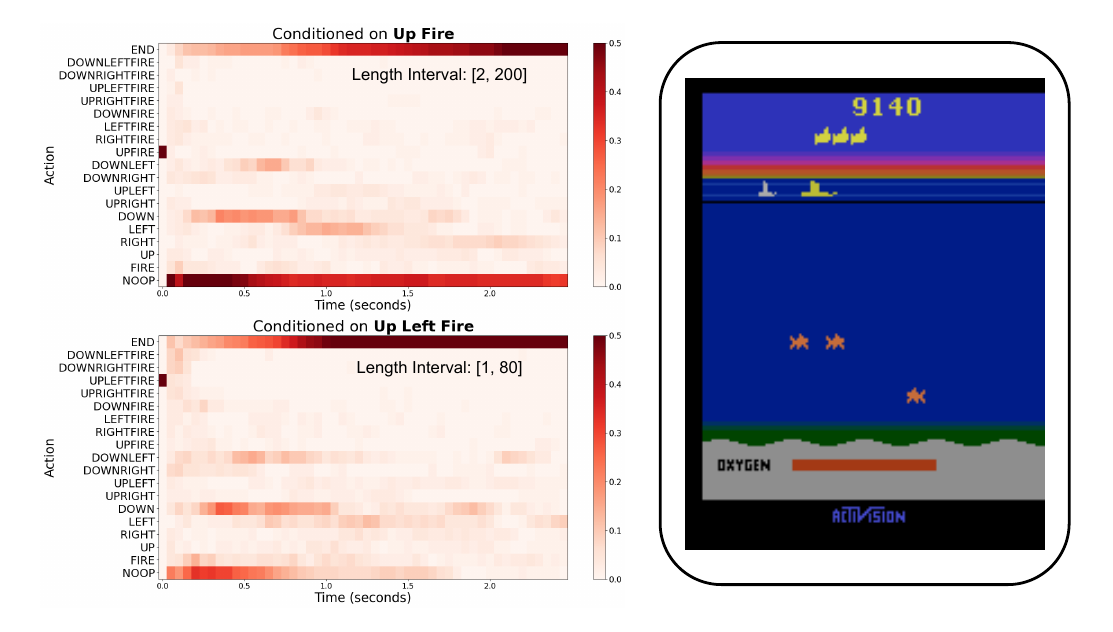}
\caption{\textbf{Action distribution conditioned on different first actions.} The figure shows empirical action distribution for a Seaquest game frame (shown on the right) given first condition actions: Up Fire and action Up Left Fire.}
\label{fig:seaquest_case_study}
\end{center}
\end{figure*}

In this section, we illustrate the conditional probability estimation task by showcasing how early actions substantially influence probability predictions at later stages, as shown in \Cref{fig:seaquest_case_study}. 

We analyze the probabilities estimated from a frame of the game Seaquest (right panel of \Cref{fig:seaquest_case_study}) by foCus using time-dependent regularization. %
We estimate the conditional probabilities %
given two different first-step actions: \textit{Up Fire} and \textit{Up Left Fire}. The left panel of \Cref{fig:seaquest_case_study} shows the resulting conditional probability estimation. 
When the first action is \textit{Up Left Fire} instead of \textit{Up Fire}, the frequency of \textit{NOOP} (no operation) decreases significantly, accelerating the player’s progress toward scoring. This makes intuitive sense, as \textit{Up Left Fire} moves the green submarine closer to the enemy blue submarine, allowing the torpedo to reach its target more quickly. This analysis illustrates that the model effectively learns meaningful conditional probability estimates.%

\section{Supplementary Figures}
\label{app:supp_figs}
\begin{figure}
\centering
\includegraphics[height=0.2\linewidth]{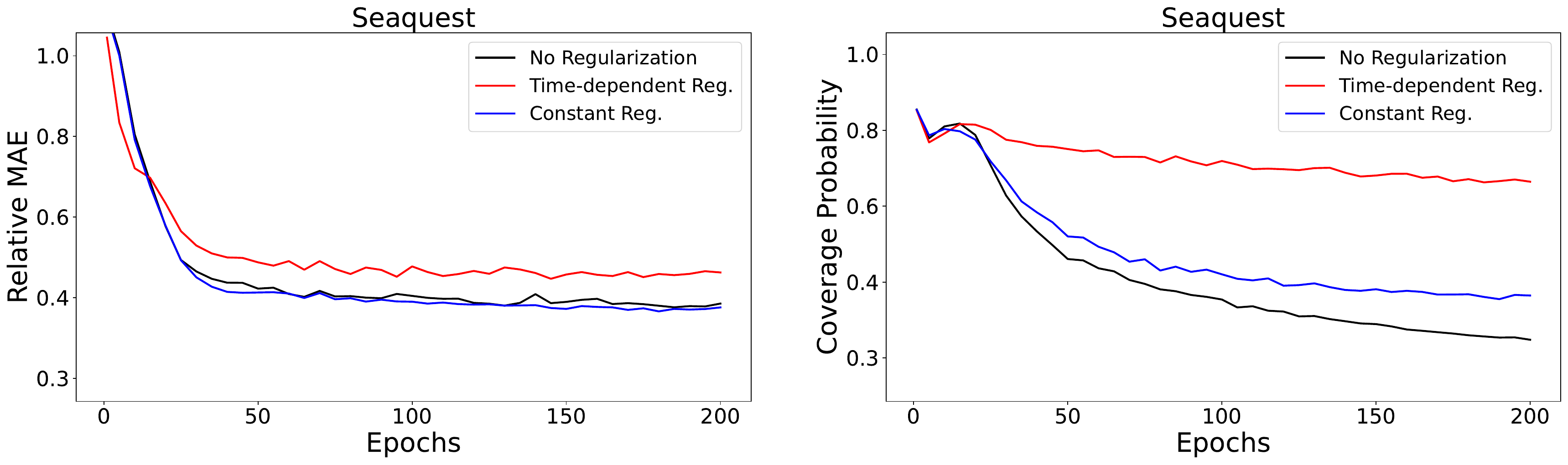} \\
\includegraphics[height=0.2\linewidth]{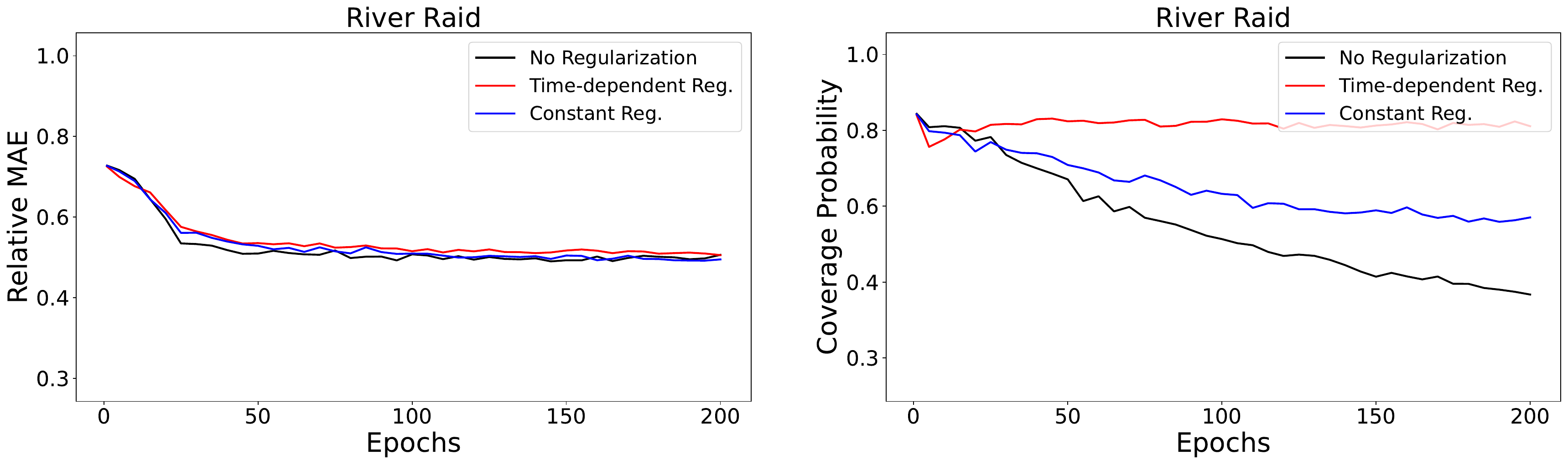} \\
\includegraphics[height=0.2\linewidth]{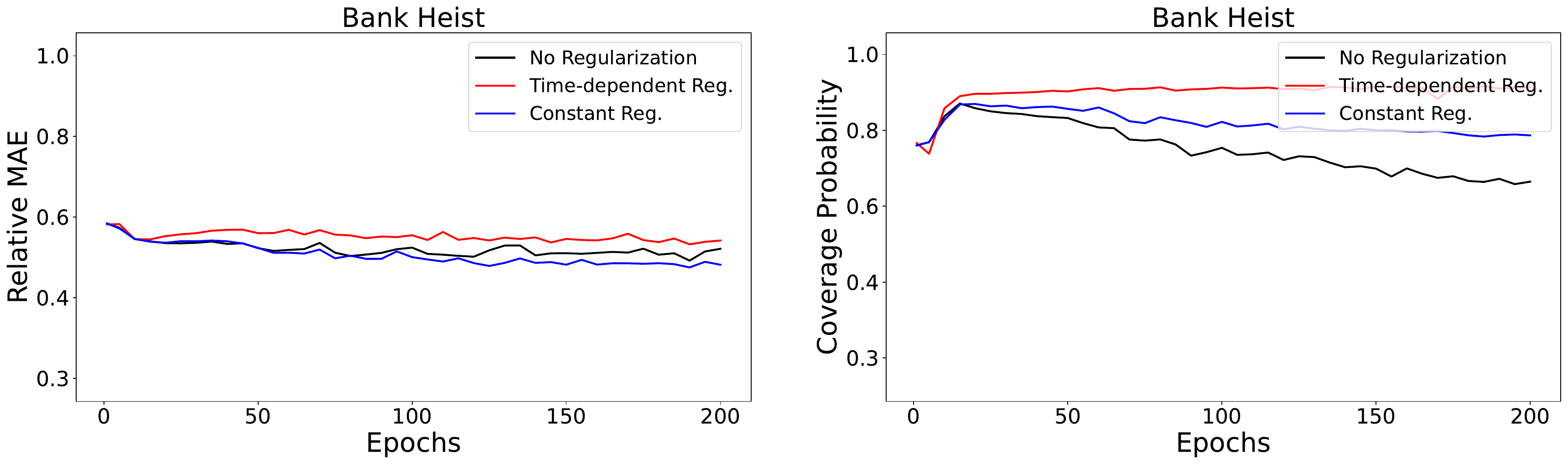} \\
\includegraphics[height=0.2\linewidth]{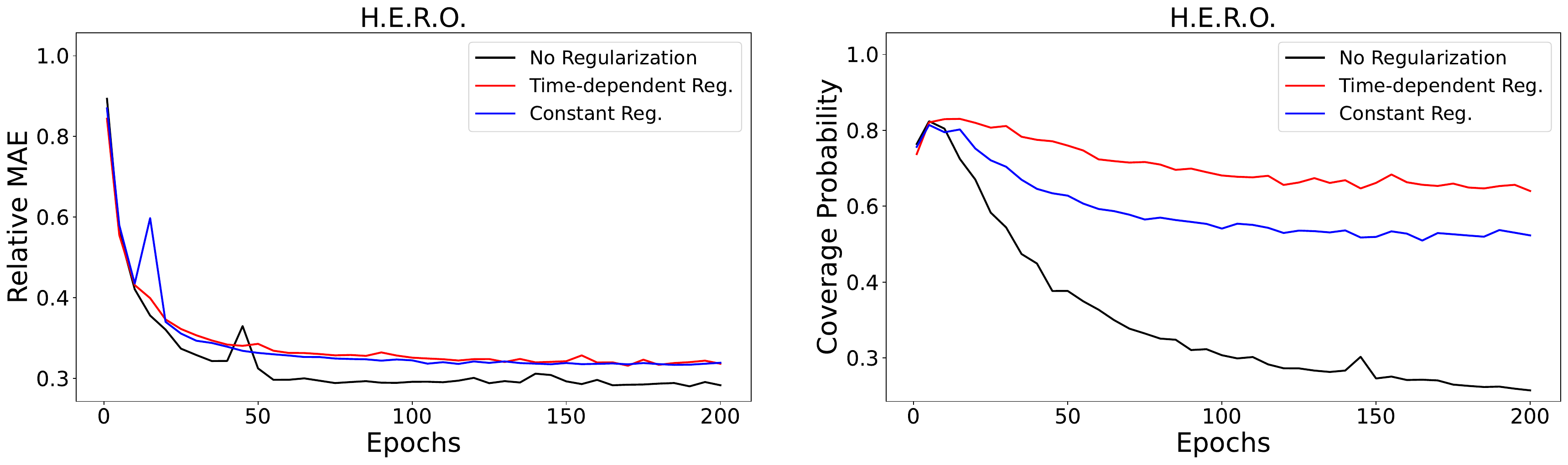} \\
\includegraphics[height=0.2\linewidth]{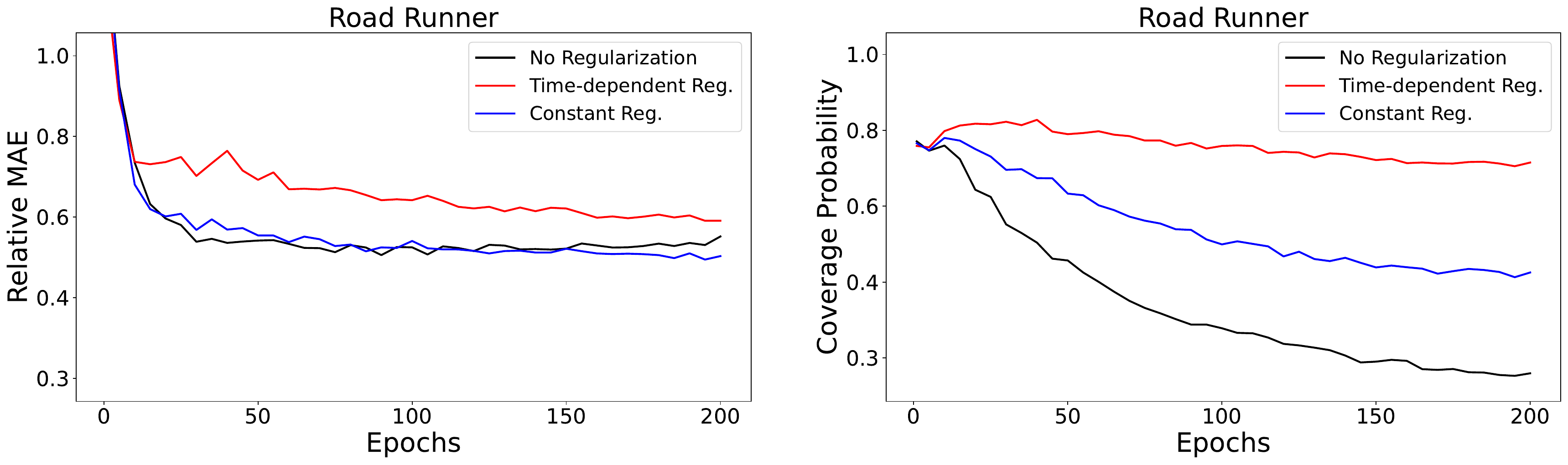} \\
\includegraphics[height=0.2\linewidth]{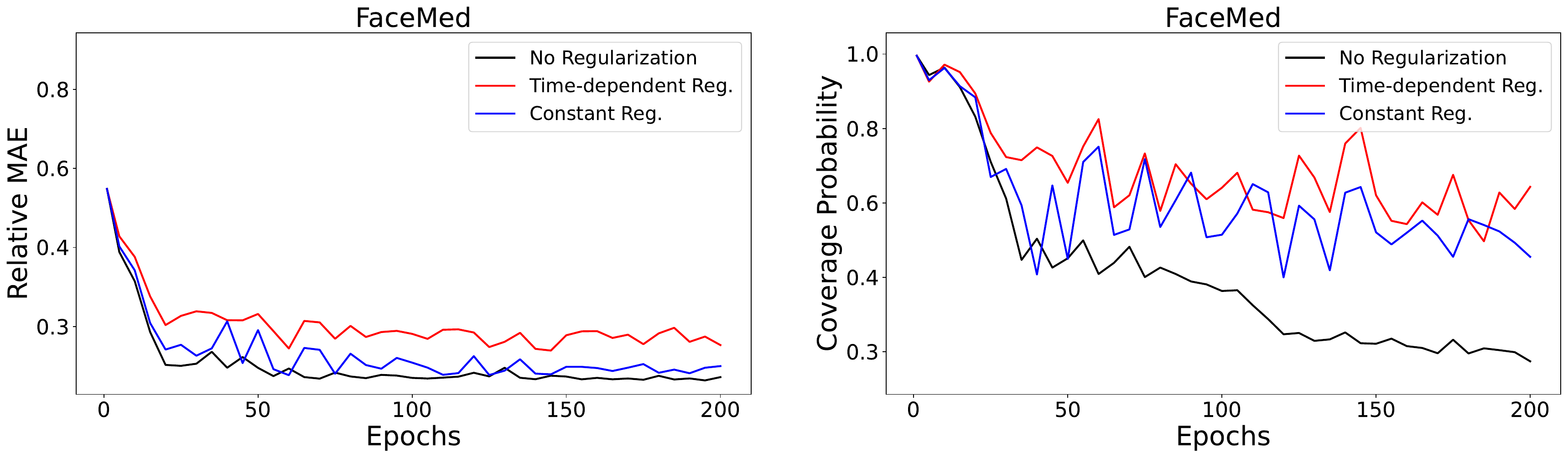} \\
\caption{\textbf{Time evolution of metrics for time-to-event confident intervals.} It plots how coverage probability and relative MAE of $I_{0.9}$ evolve along training epochs for three versions of foCus: without regularization (black), time-dependent regularization (red), and constant regularization (blue). The time-dependent regularization model achieves significantly better coverage probability.}
\label{fig:length_score_game}
\end{figure}

\begin{figure*}
\centering
    \includegraphics[height=0.2\linewidth]{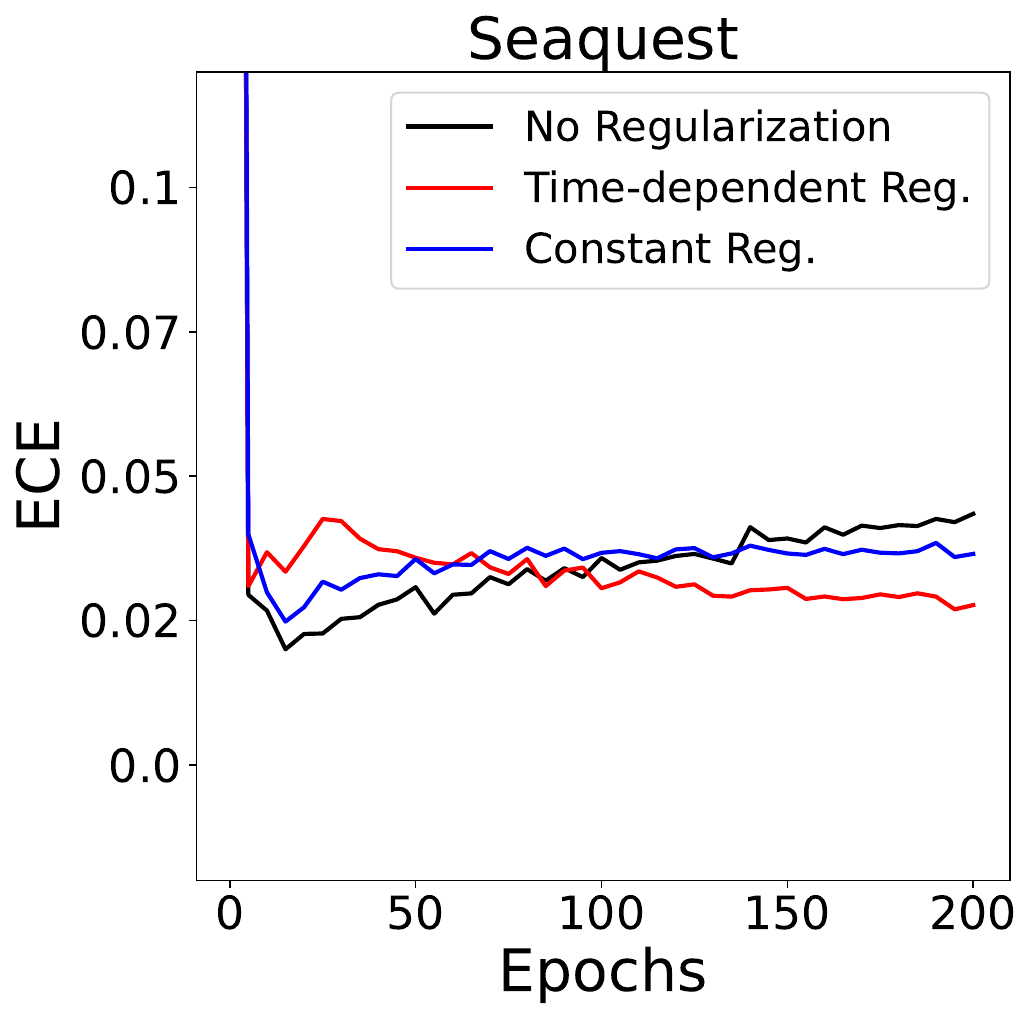}
    \includegraphics[height=0.2\linewidth]{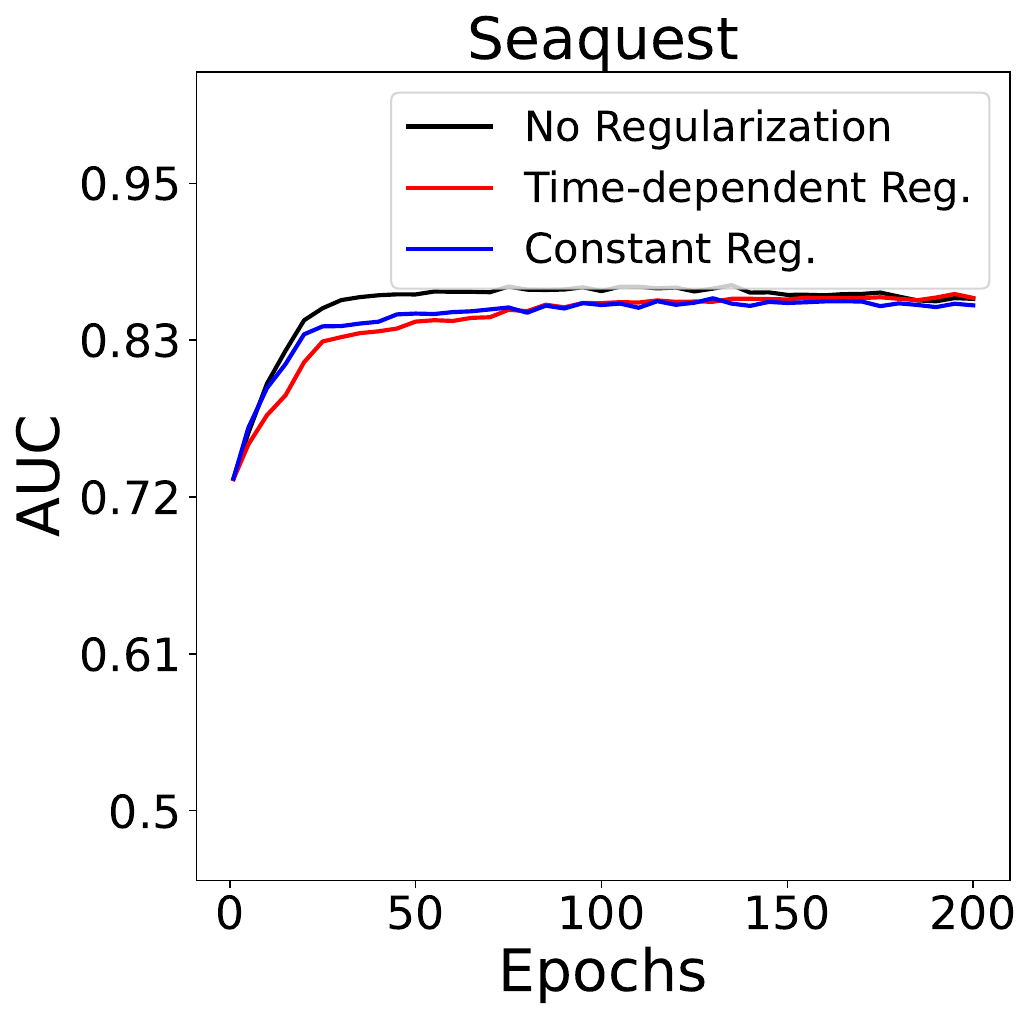}
    \includegraphics[height=0.2\linewidth]{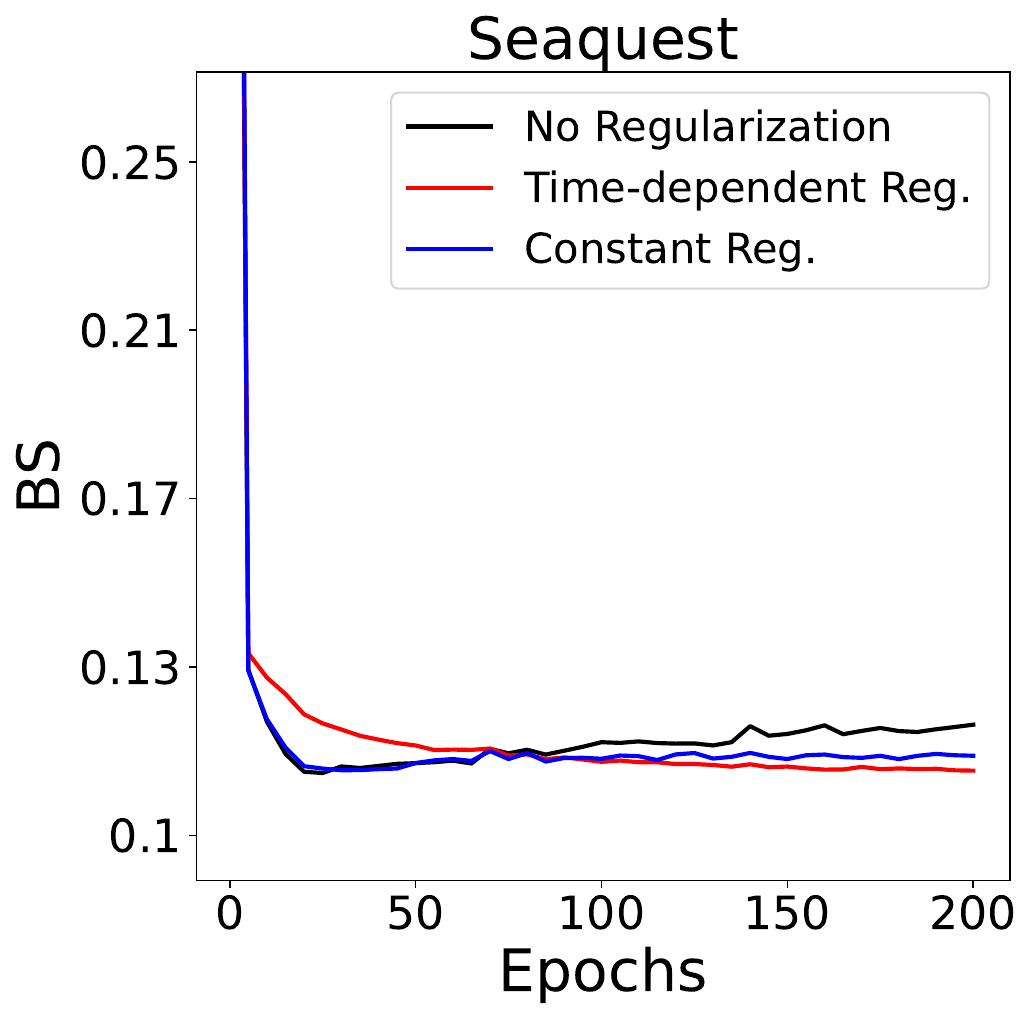}
    \includegraphics[height=0.2\linewidth]{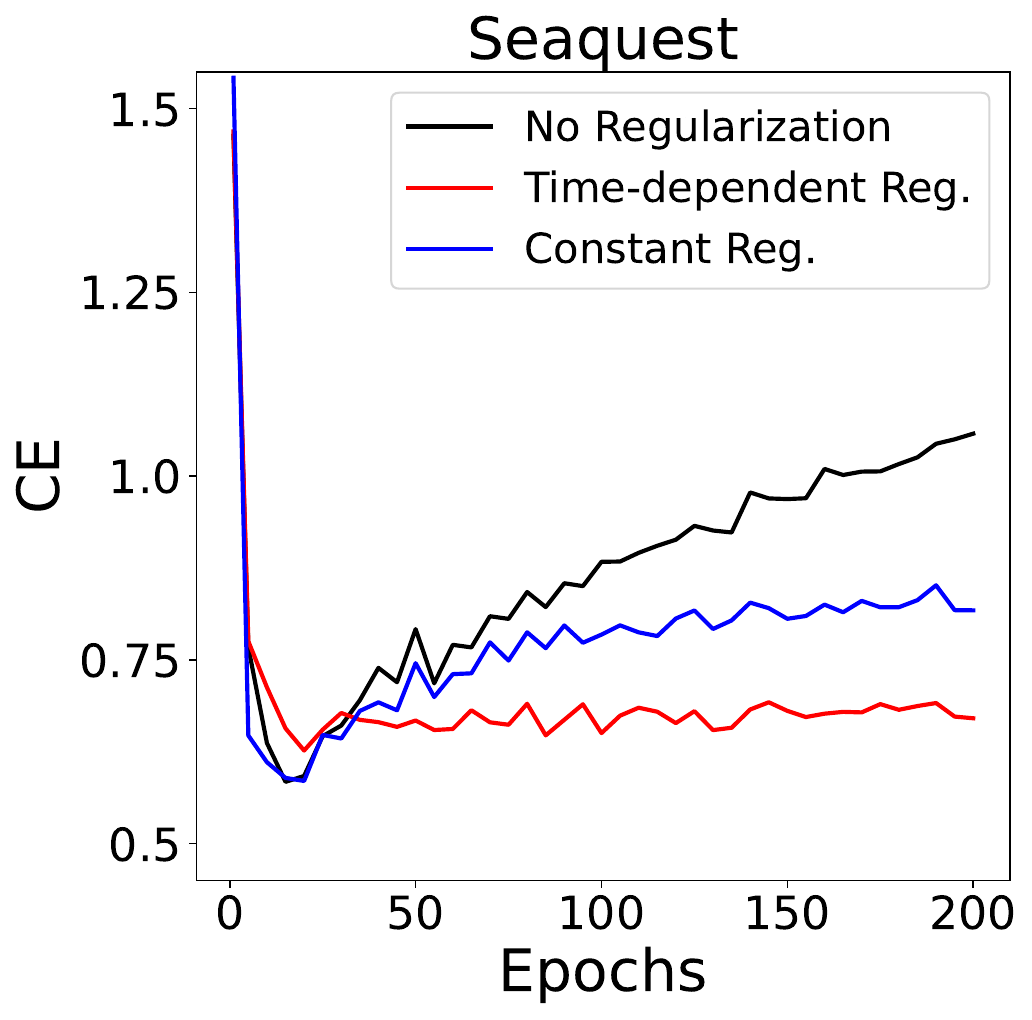} \\

    \includegraphics[height=0.2\linewidth]{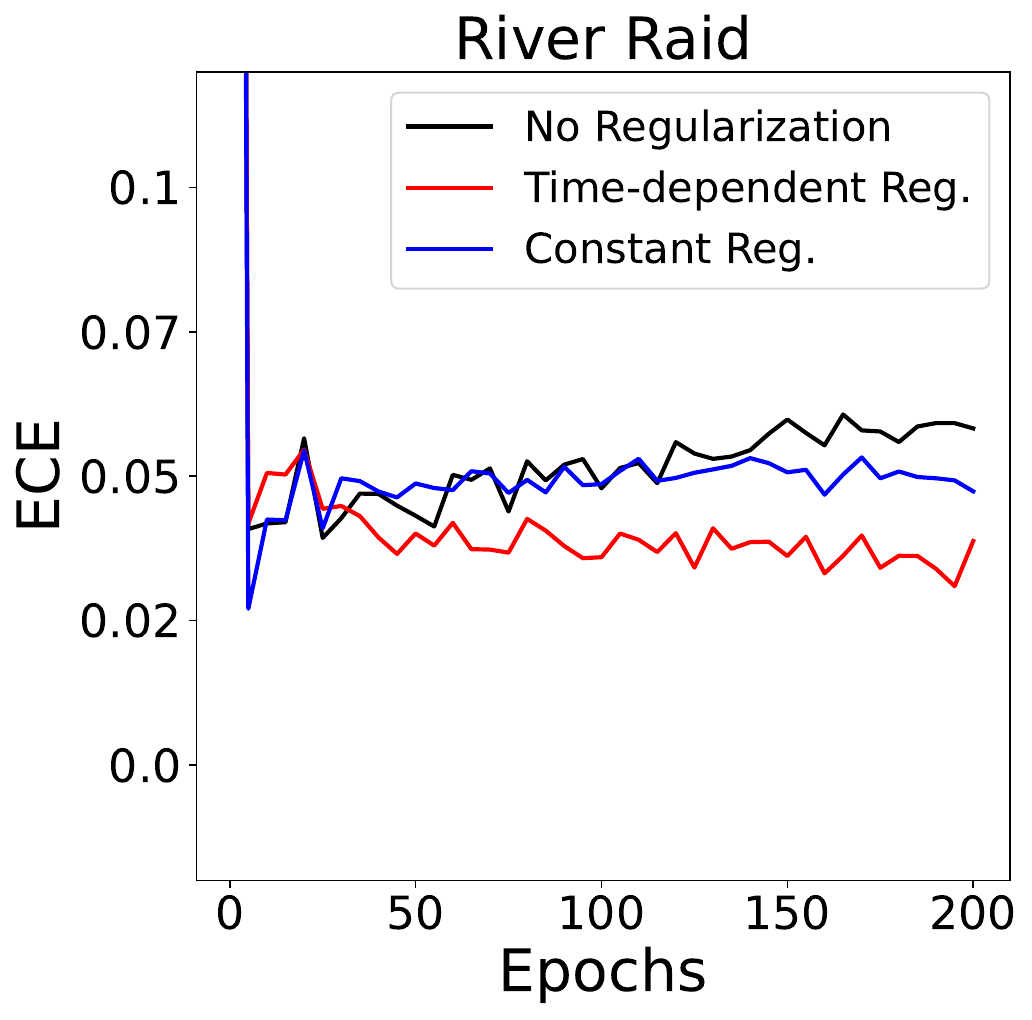}
    \includegraphics[height=0.2\linewidth]{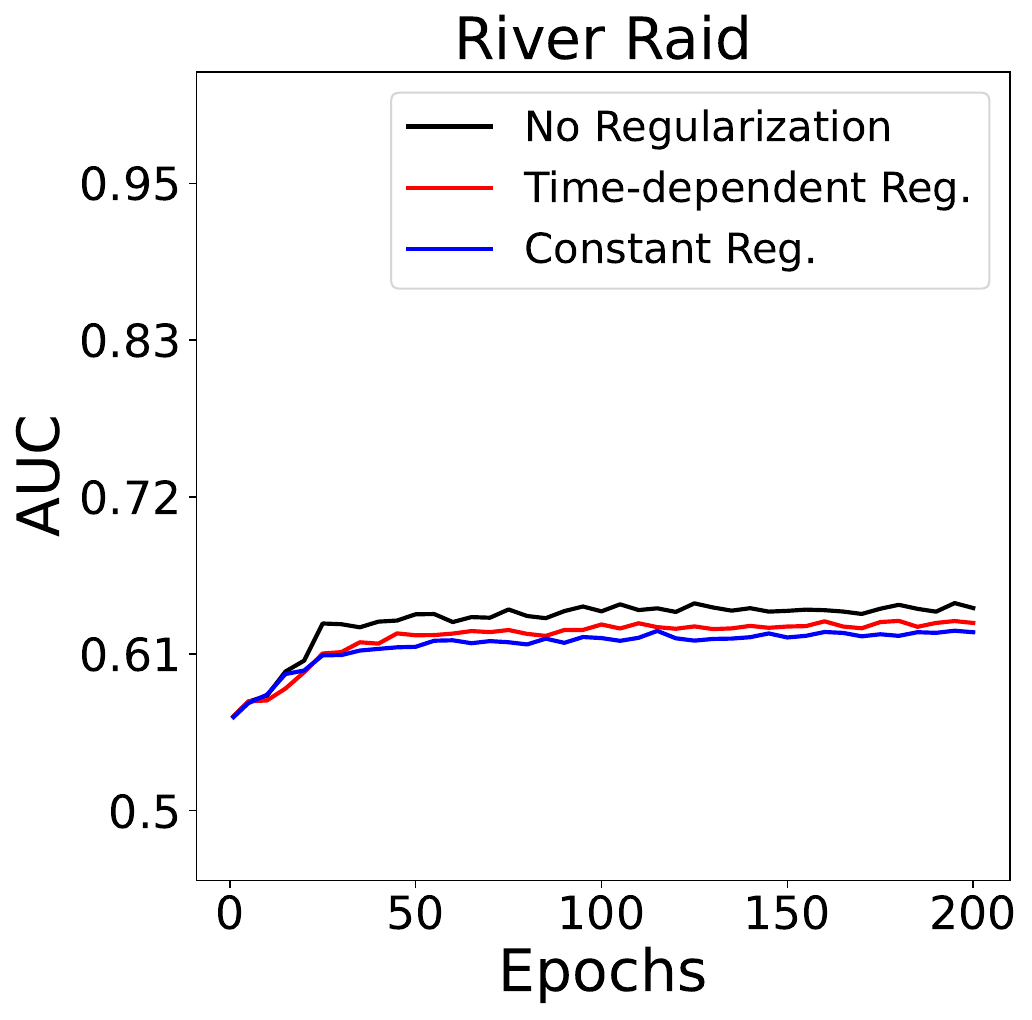}
    \includegraphics[height=0.2\linewidth]{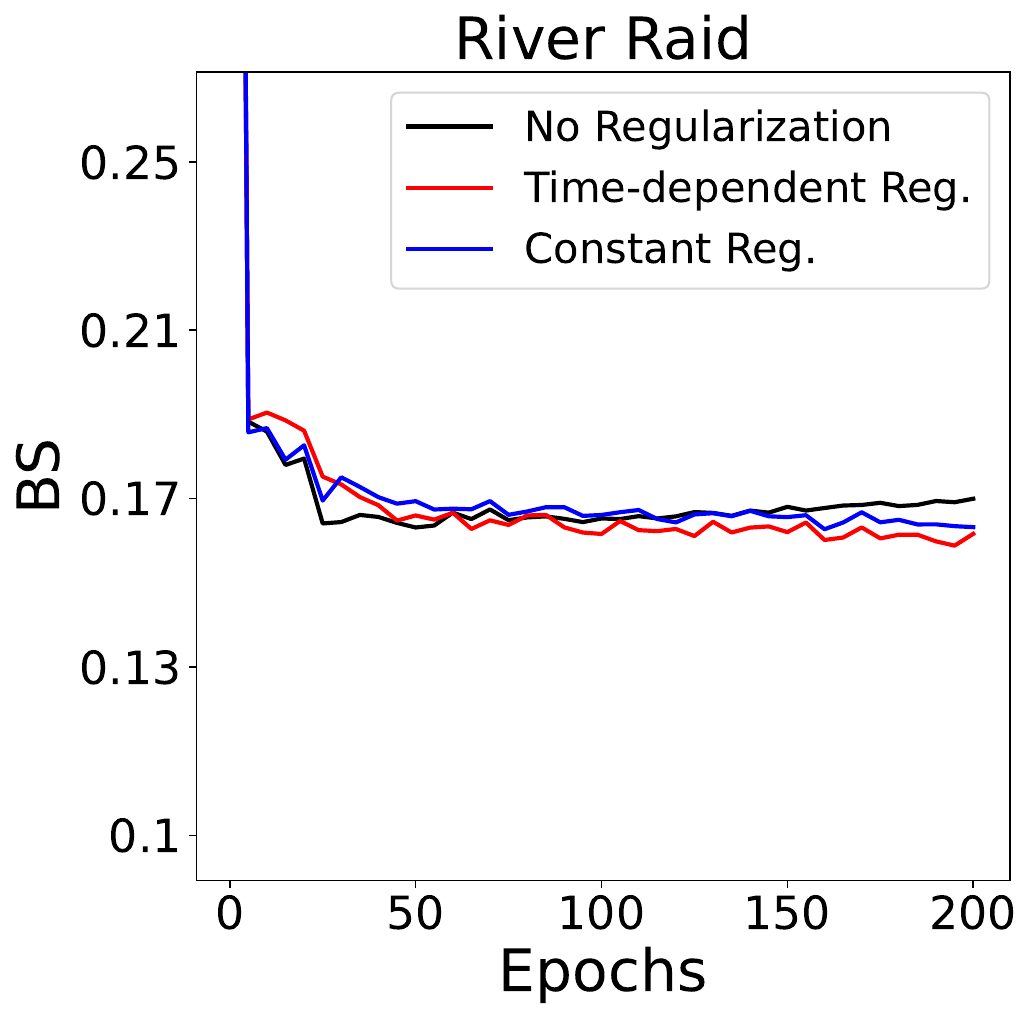}
    \includegraphics[height=0.2\linewidth]{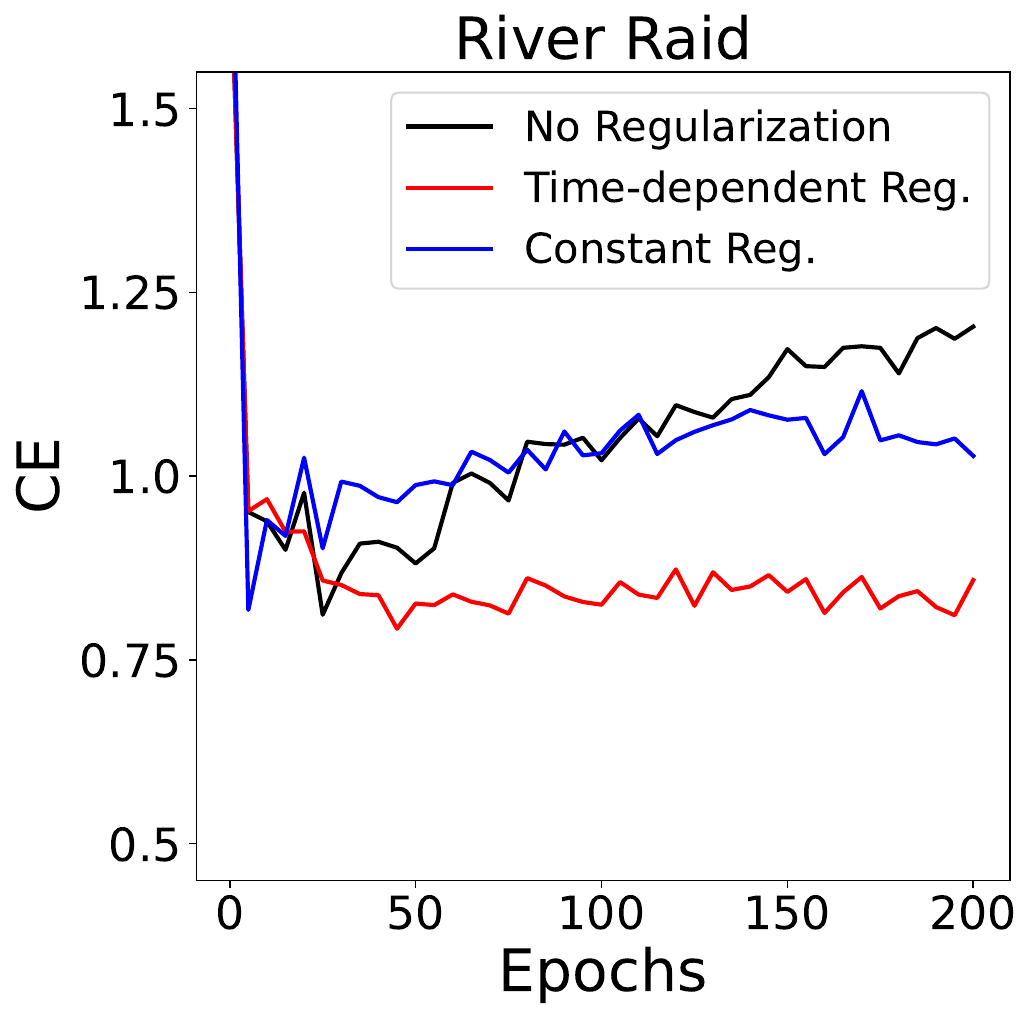} \\

    \includegraphics[height=0.2\linewidth]{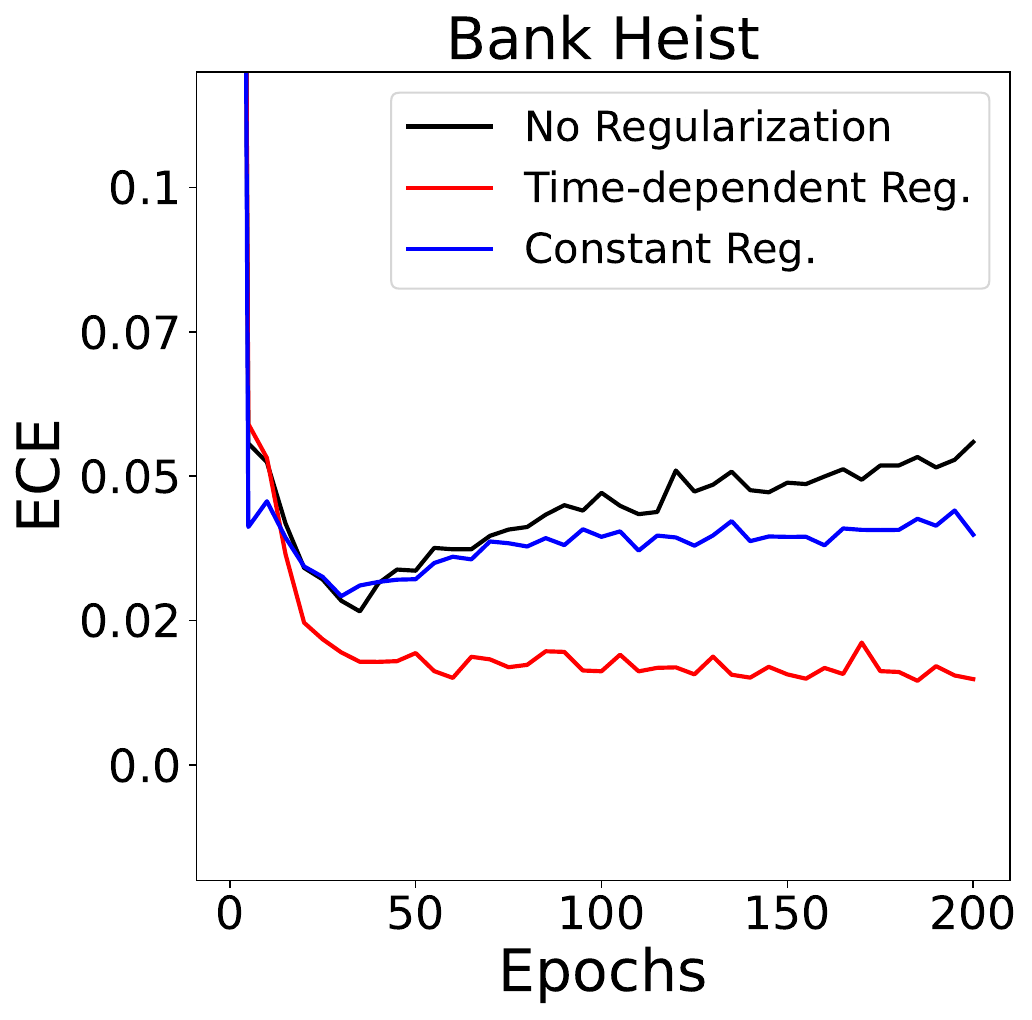}
    \includegraphics[height=0.2\linewidth]{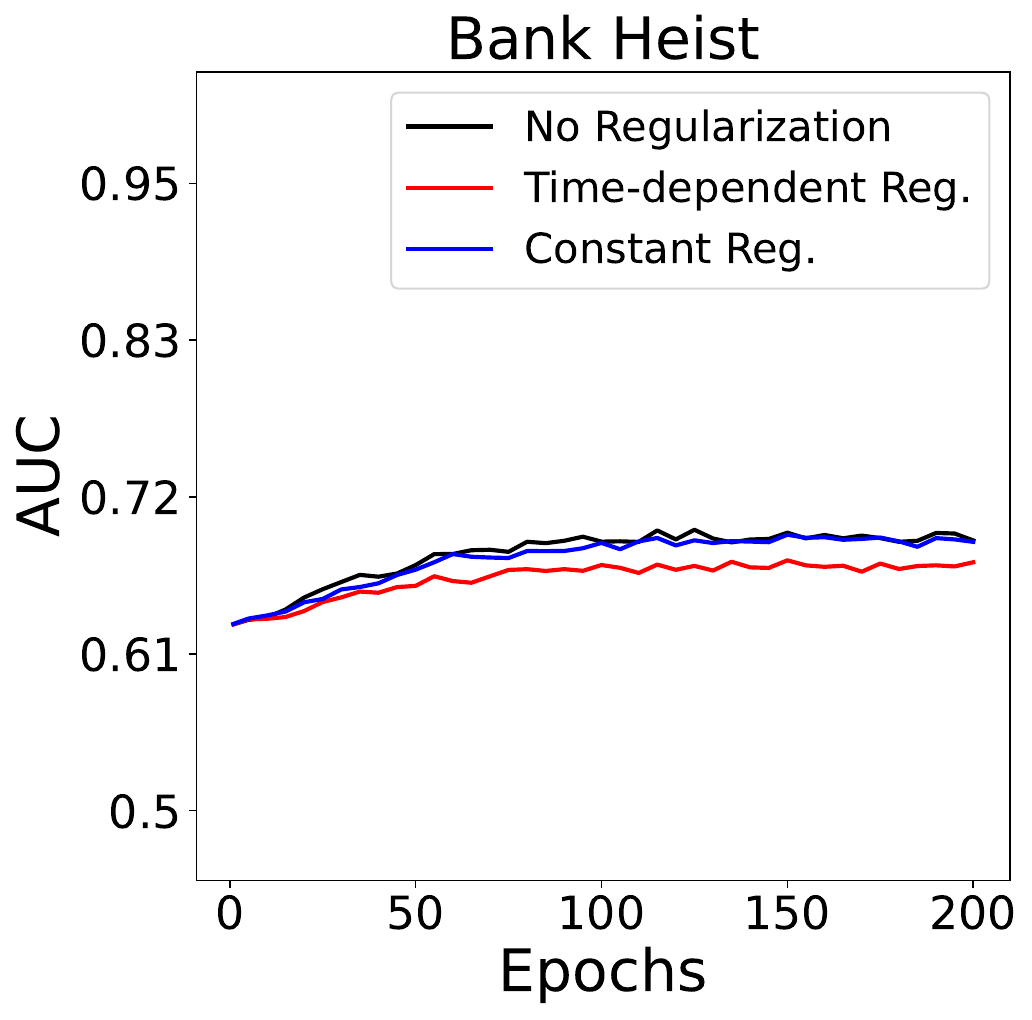}
    \includegraphics[height=0.2\linewidth]{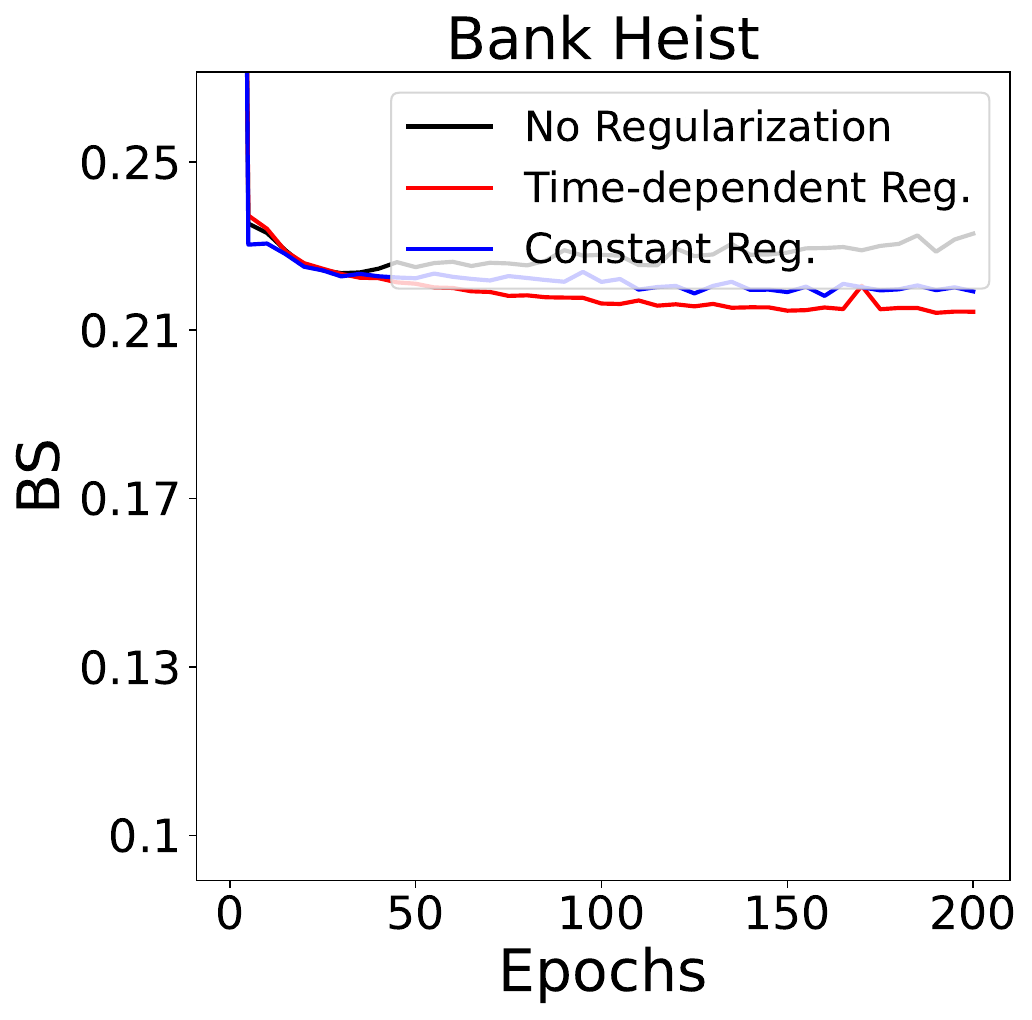}
    \includegraphics[height=0.2\linewidth]{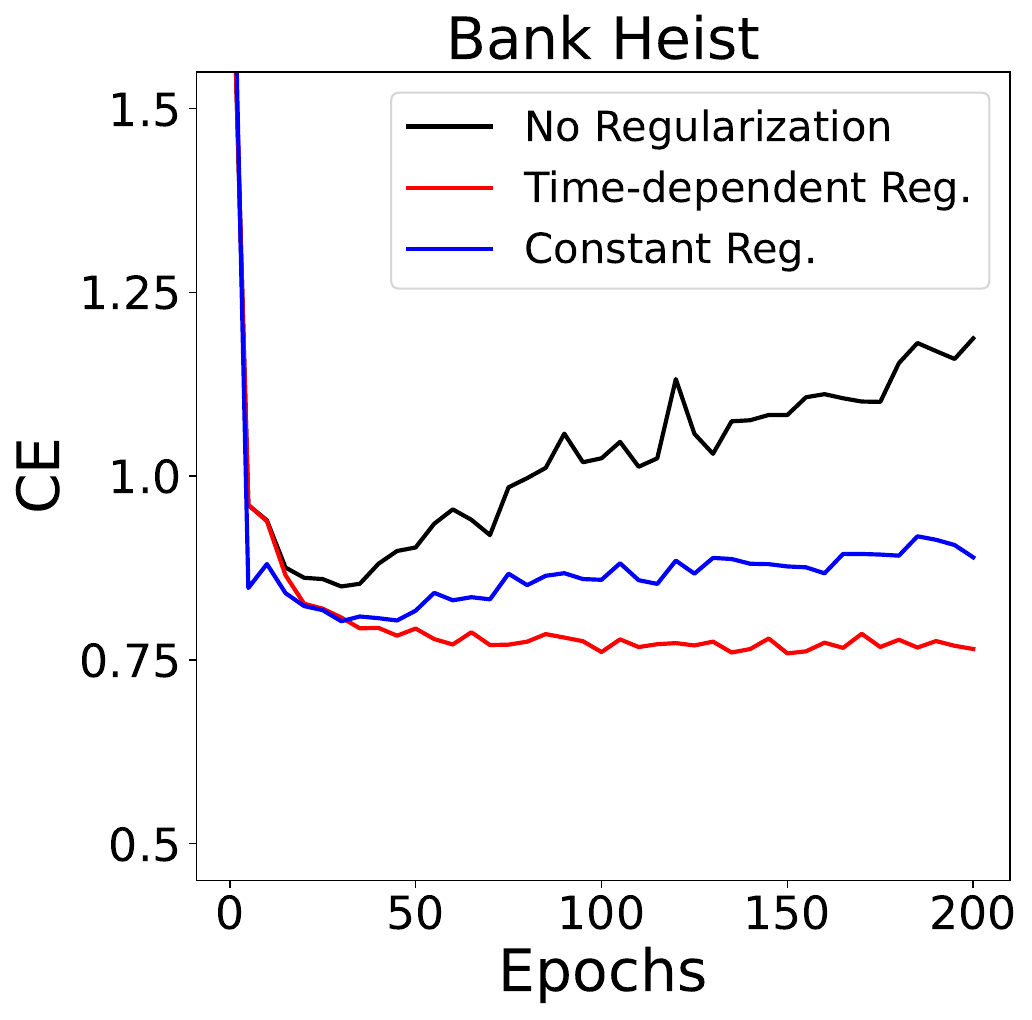} \\

    \includegraphics[height=0.2\linewidth]{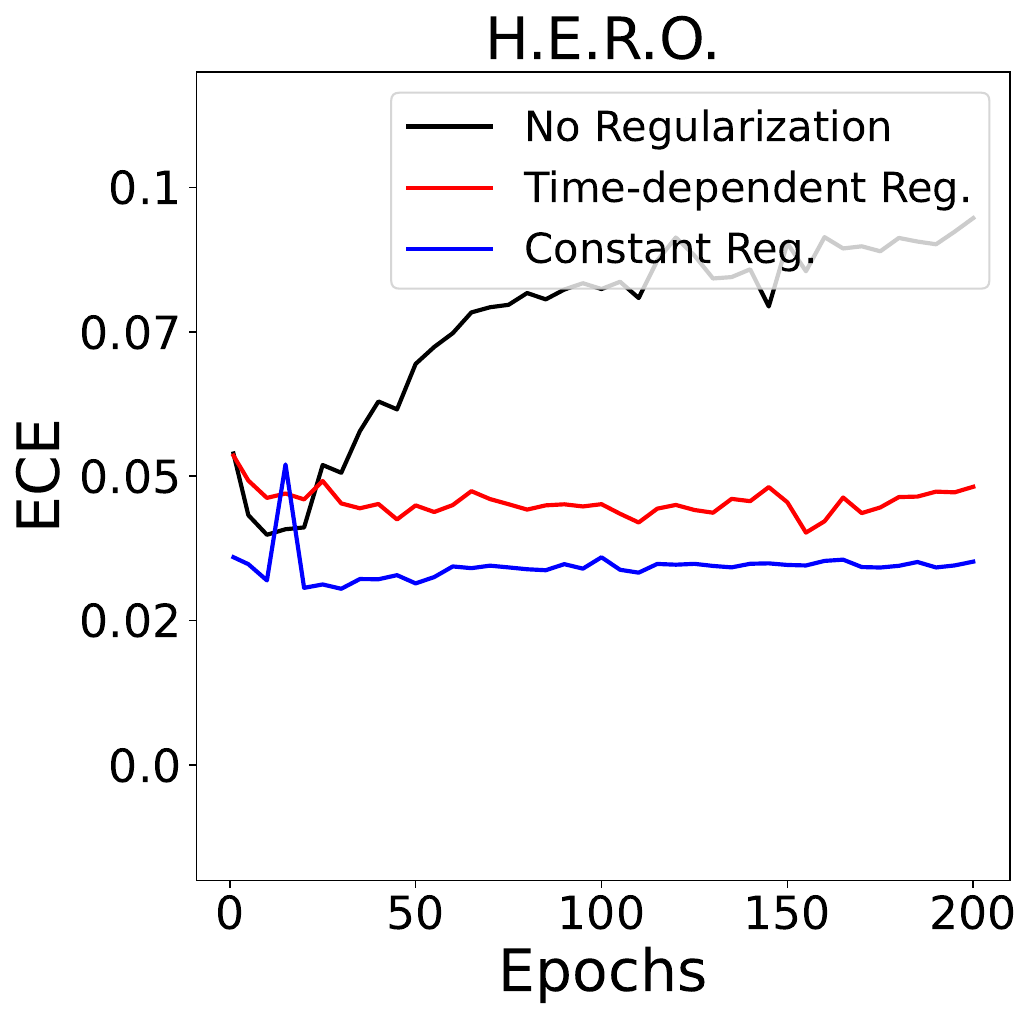}
    \includegraphics[height=0.2\linewidth]{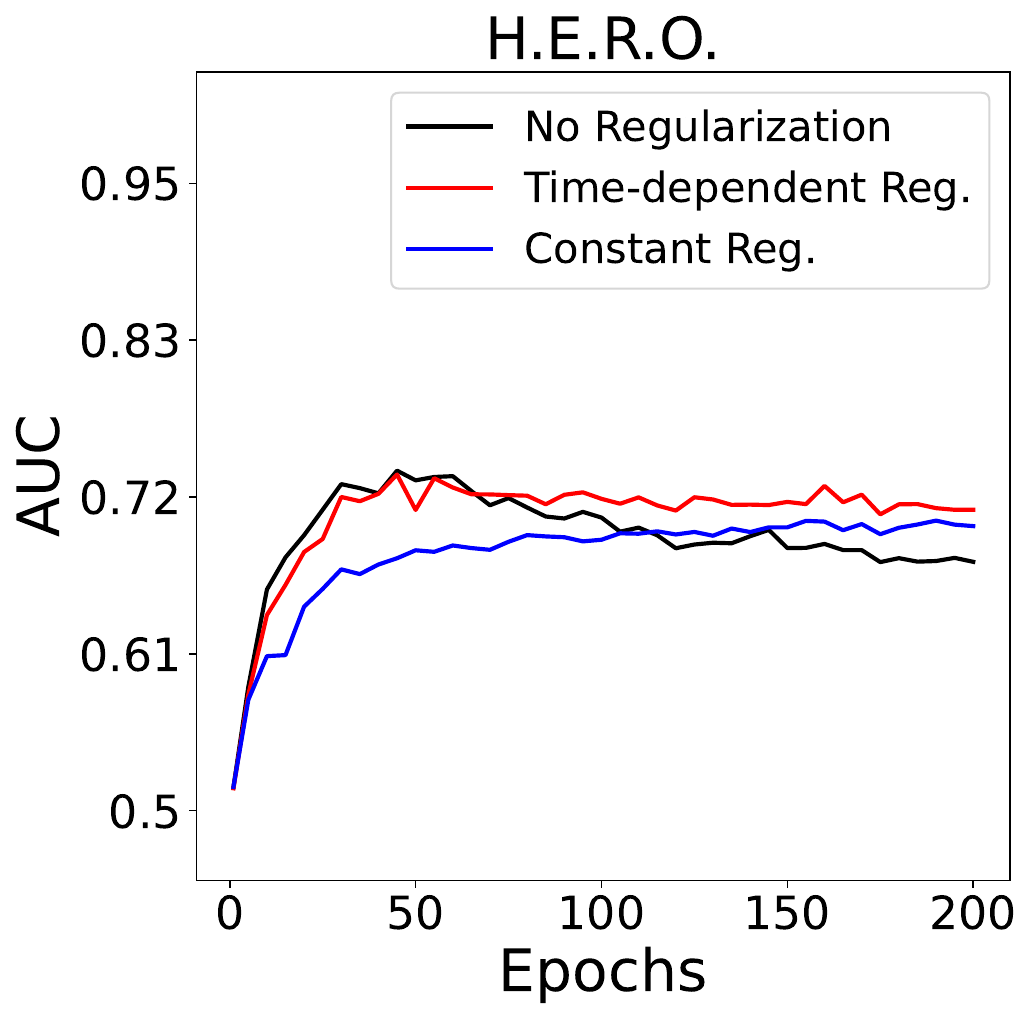}
    \includegraphics[height=0.2\linewidth]{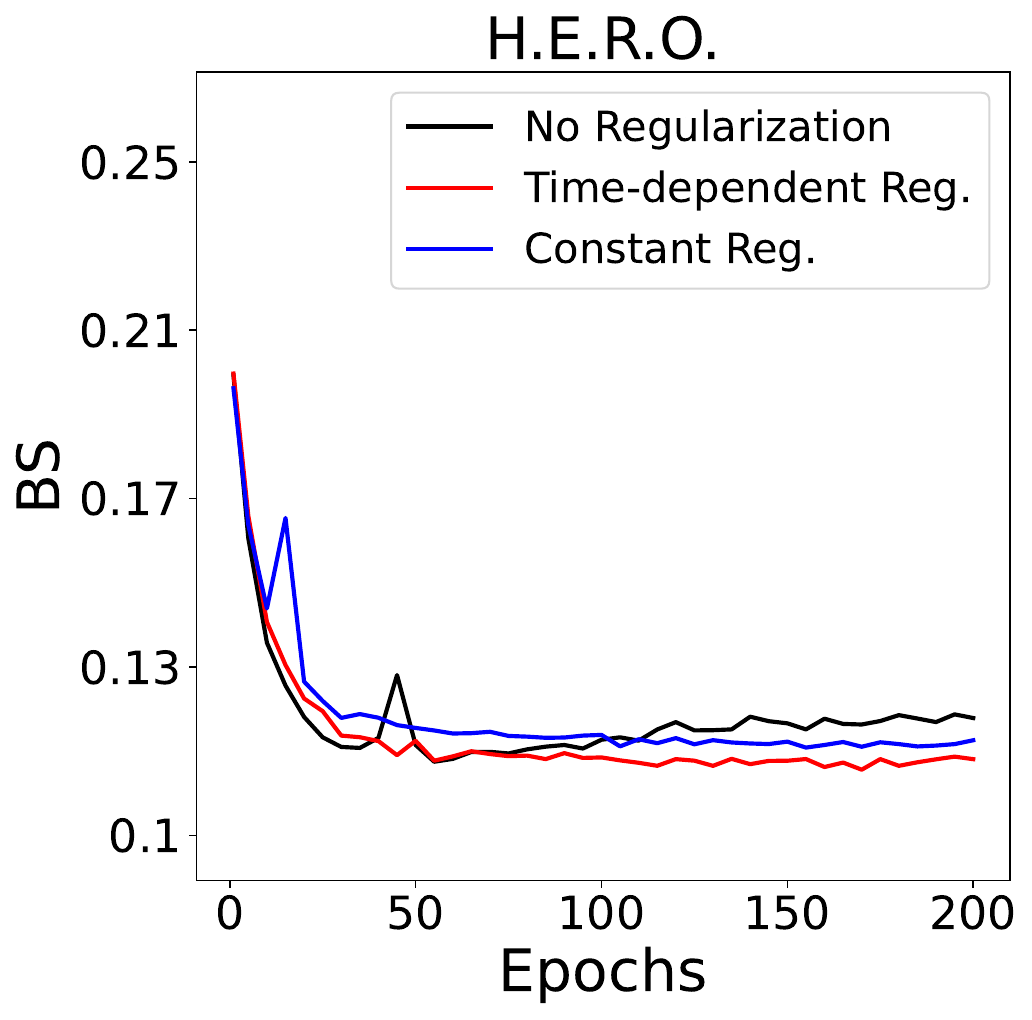}
    \includegraphics[height=0.2\linewidth]{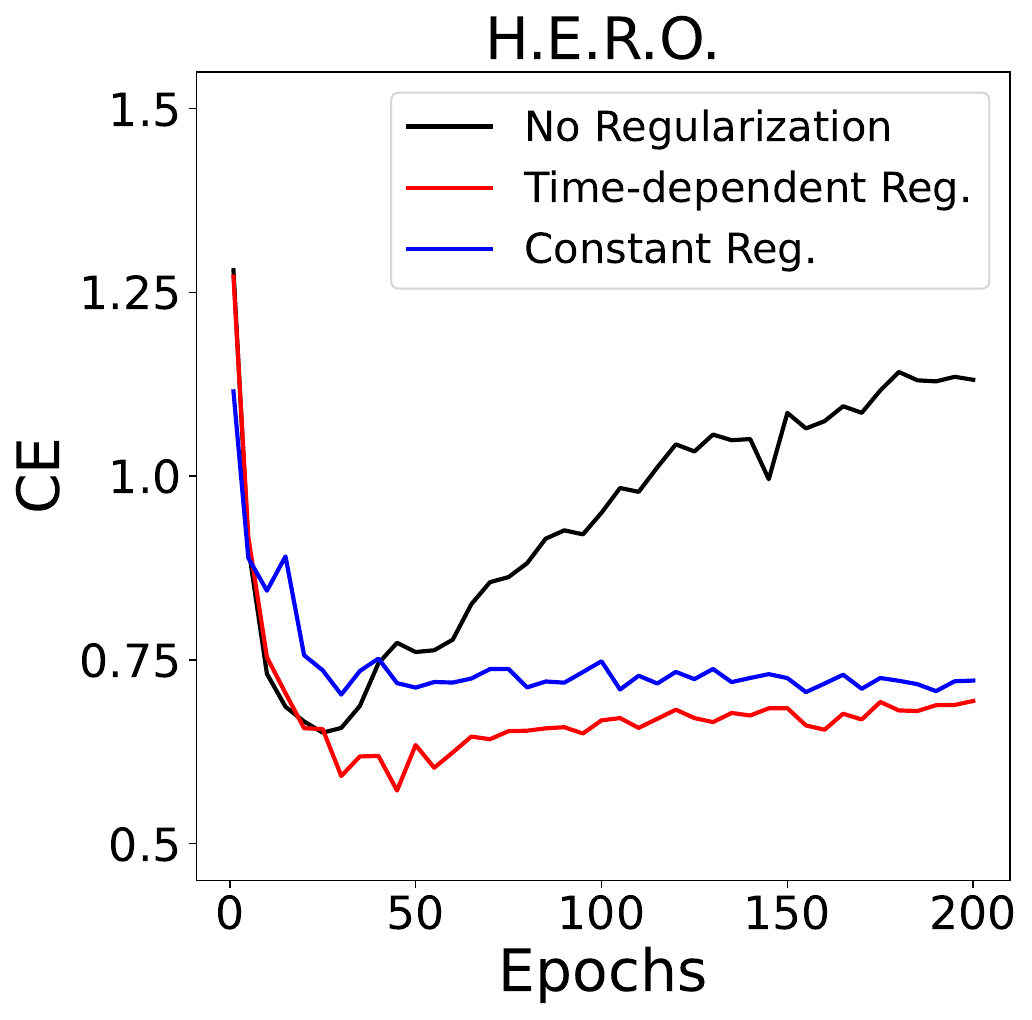} \\

    \includegraphics[height=0.2\linewidth]{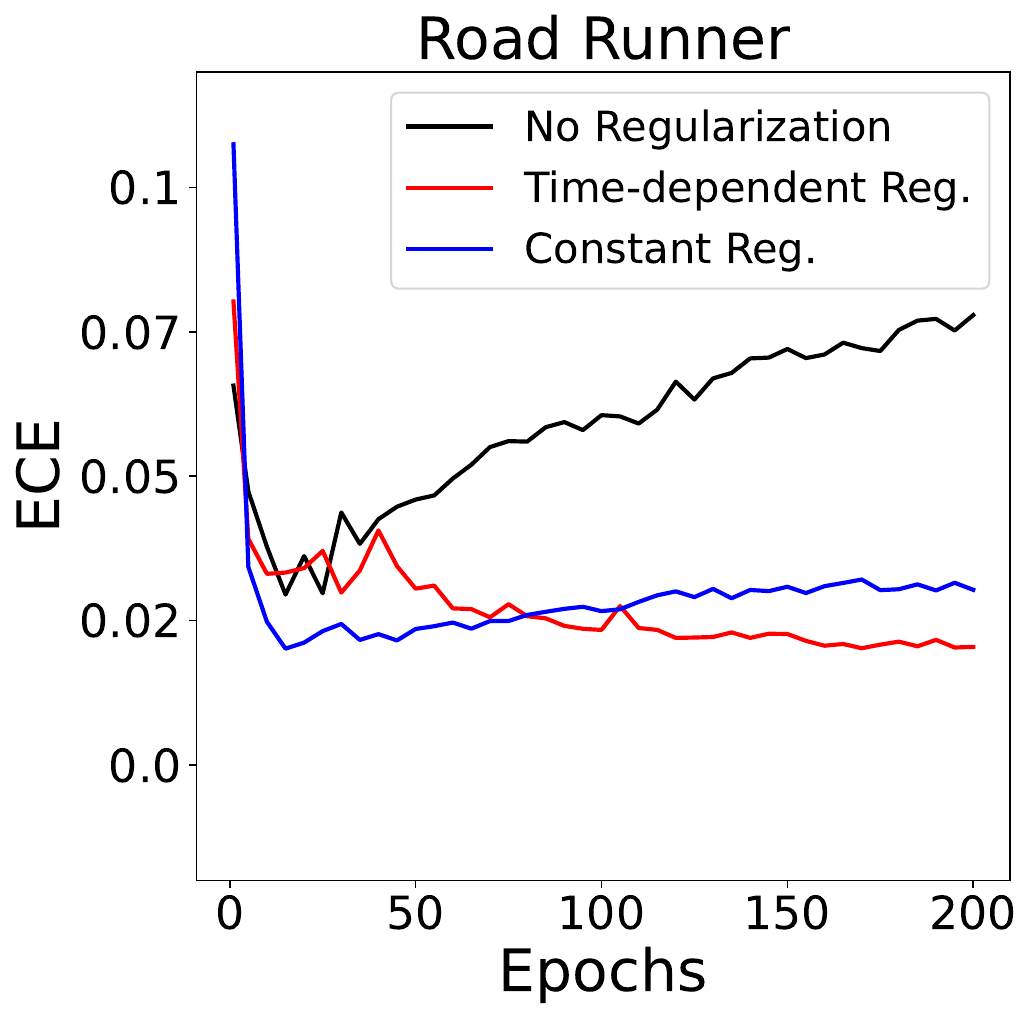}
    \includegraphics[height=0.2\linewidth]{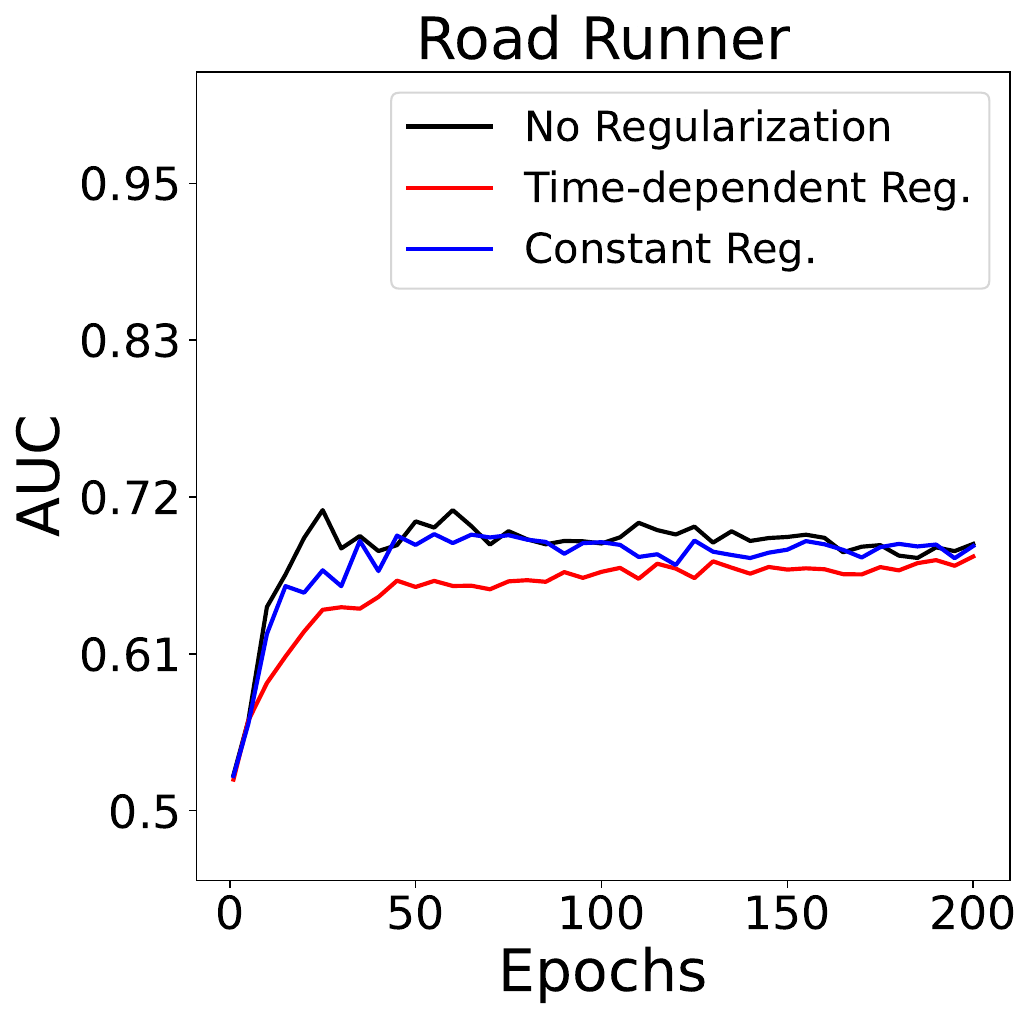}
    \includegraphics[height=0.2\linewidth]{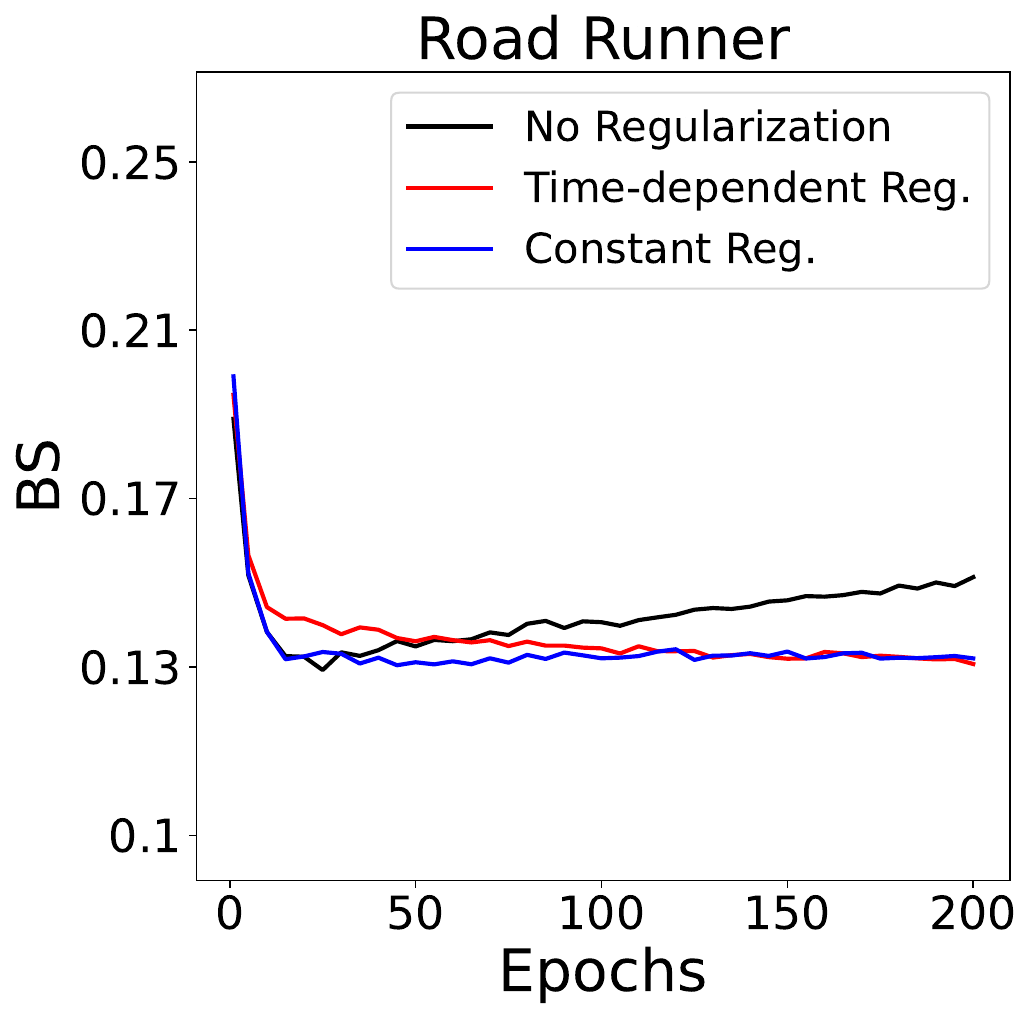}
    \includegraphics[height=0.2\linewidth]{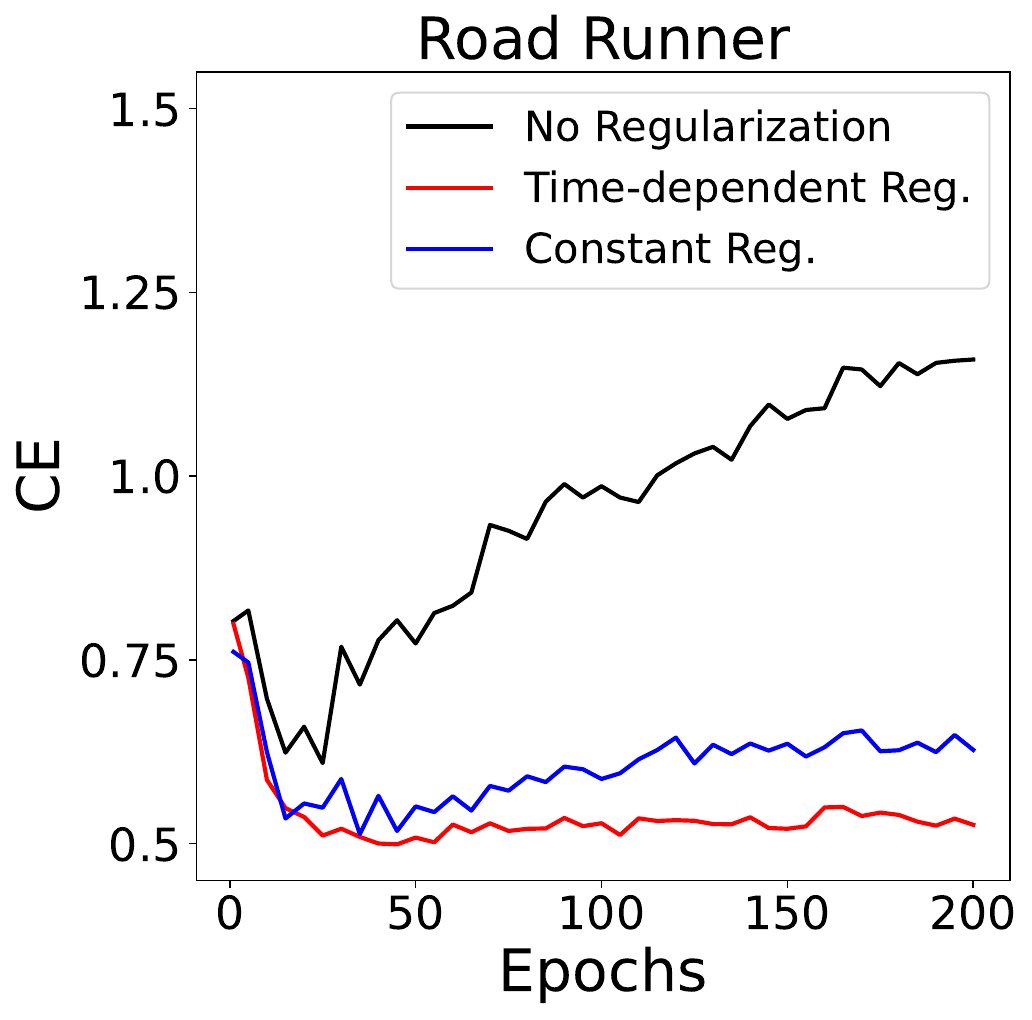} \\

    \includegraphics[height=0.2\linewidth]{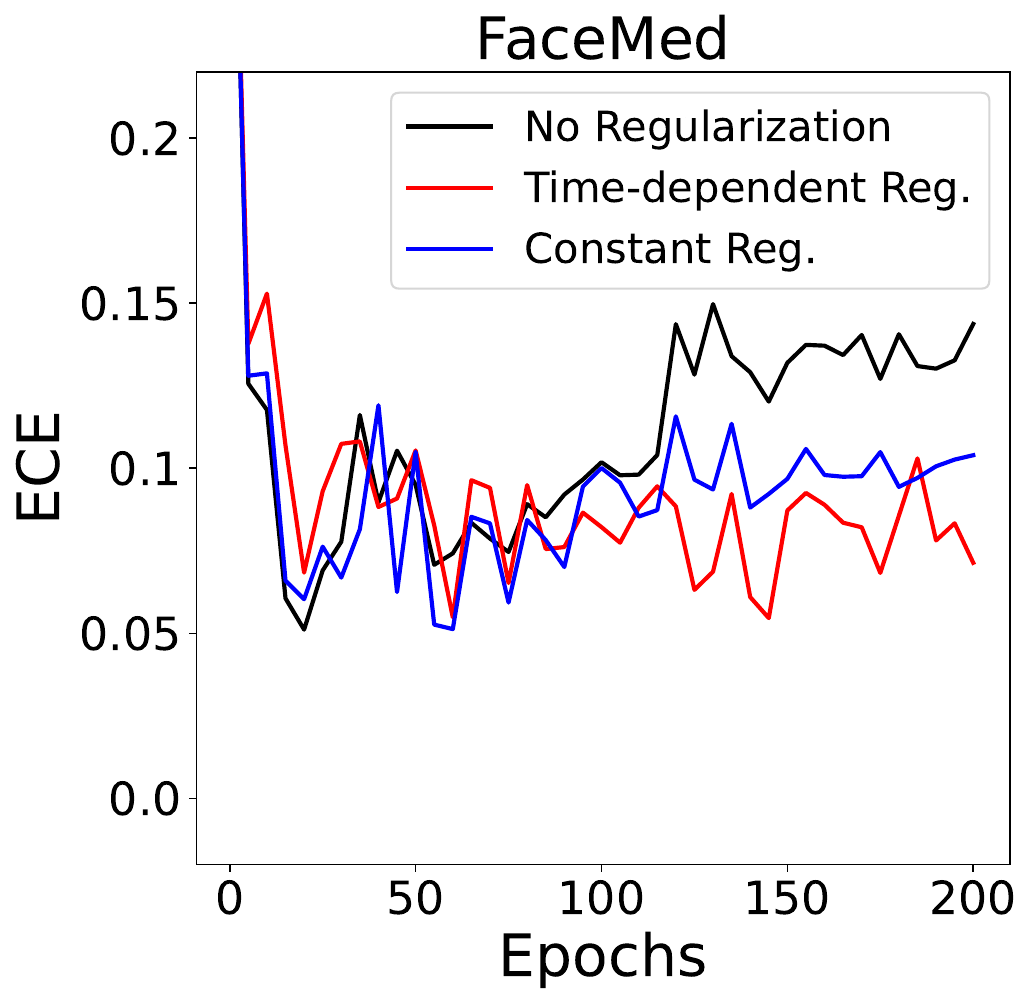}
    \includegraphics[height=0.2\linewidth]{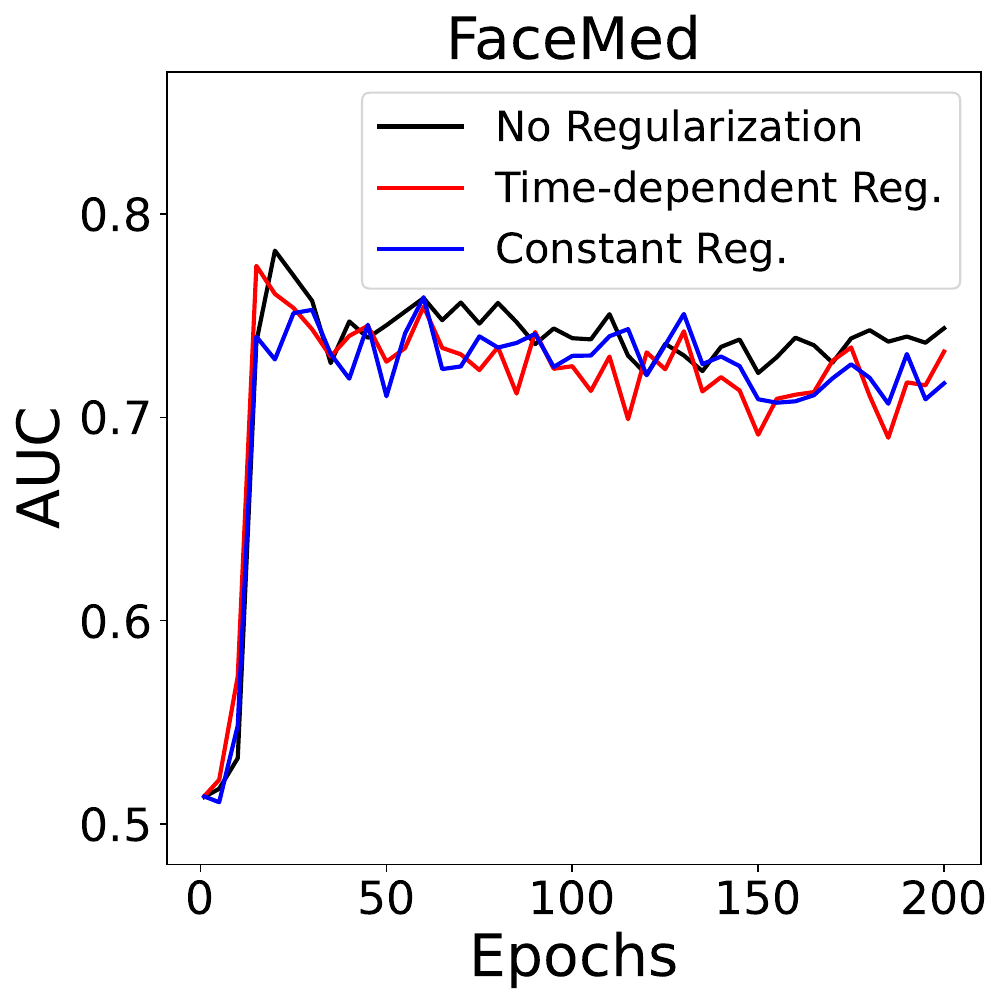}
    \includegraphics[height=0.2\linewidth]{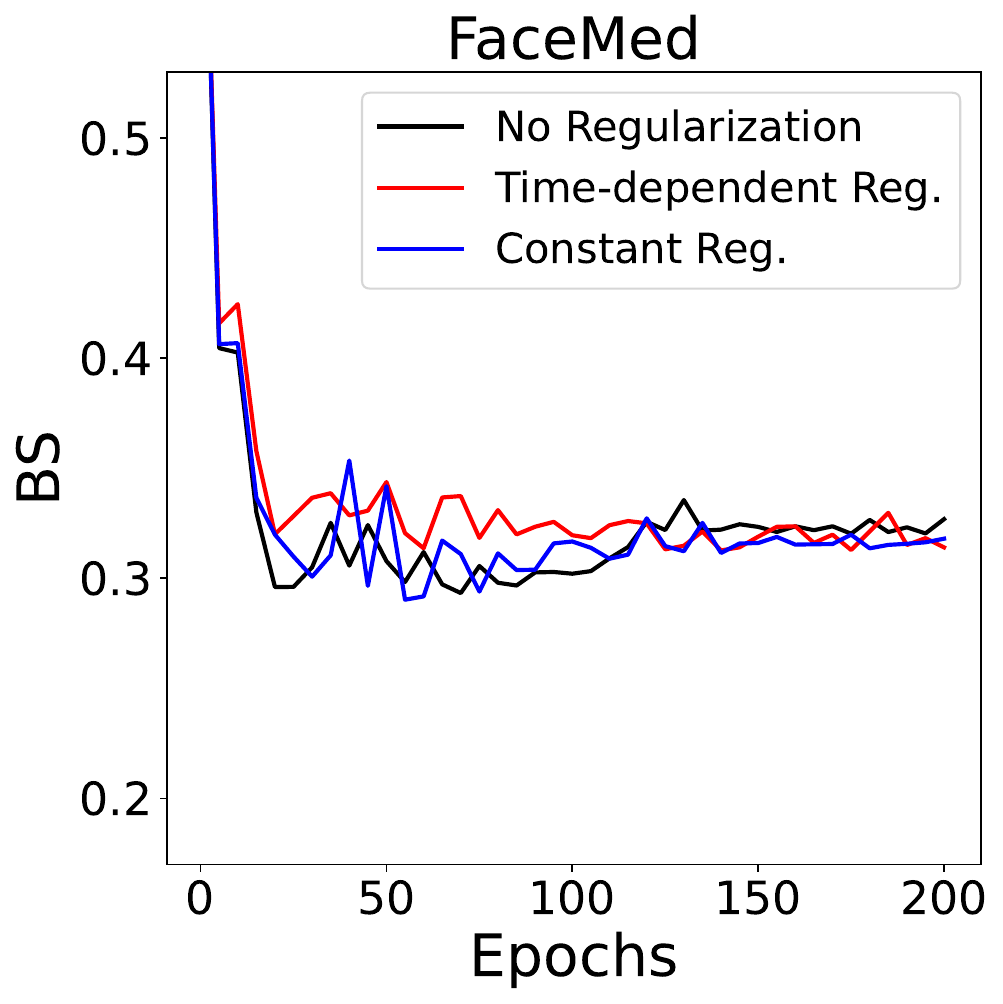}
    \includegraphics[height=0.2\linewidth]{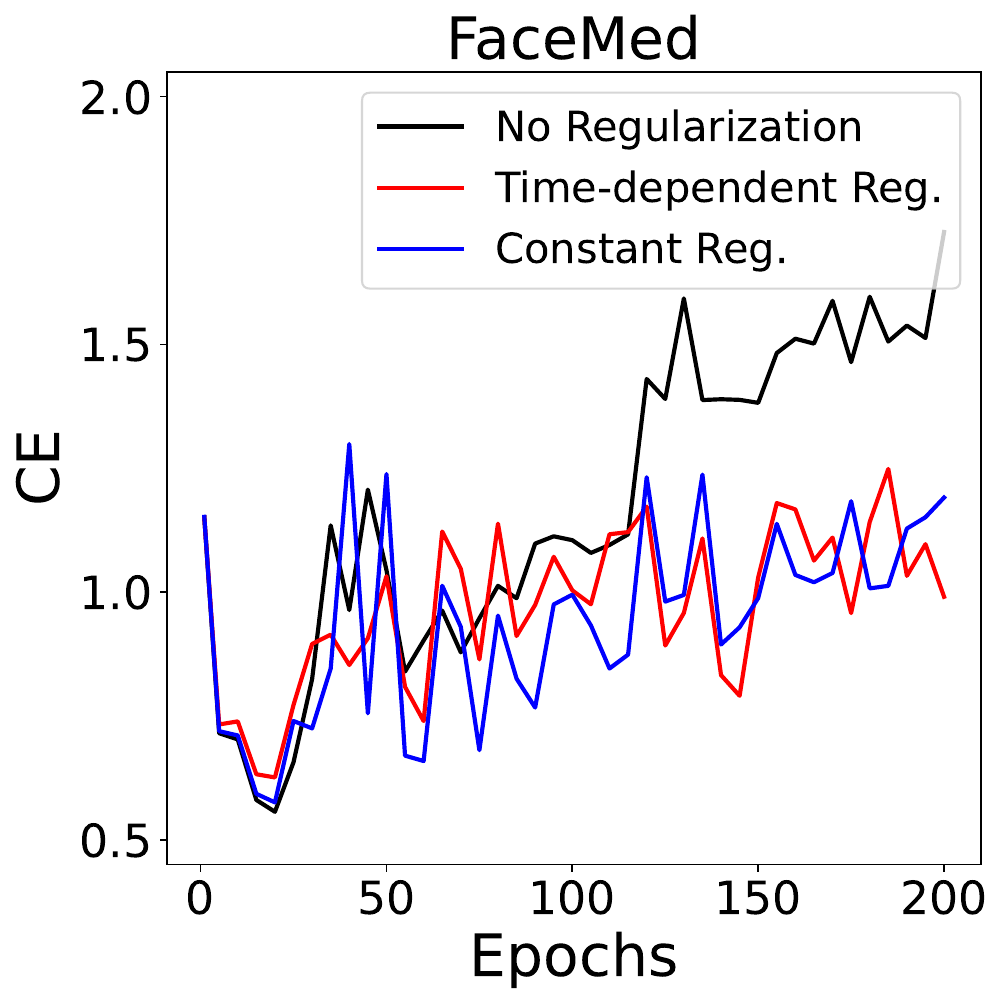} 
\caption{\textbf{Learning curves of sequence-level metrics for marginal probability estimation.} It plots how ECE, AUC, BS, and CE evolve along training epochs for three versions of foCus: without regularization (black), time-dependent regularization (red), and constant regularization (blue). The discriminability improves as the calibration decays. The time-dependent regularization model is able to keep calibrated while improving the discriminability.}
\label{fig:step_average_score_uncondition_game}
\end{figure*}

\begin{figure*}
\centering
    \includegraphics[height=0.2\linewidth]{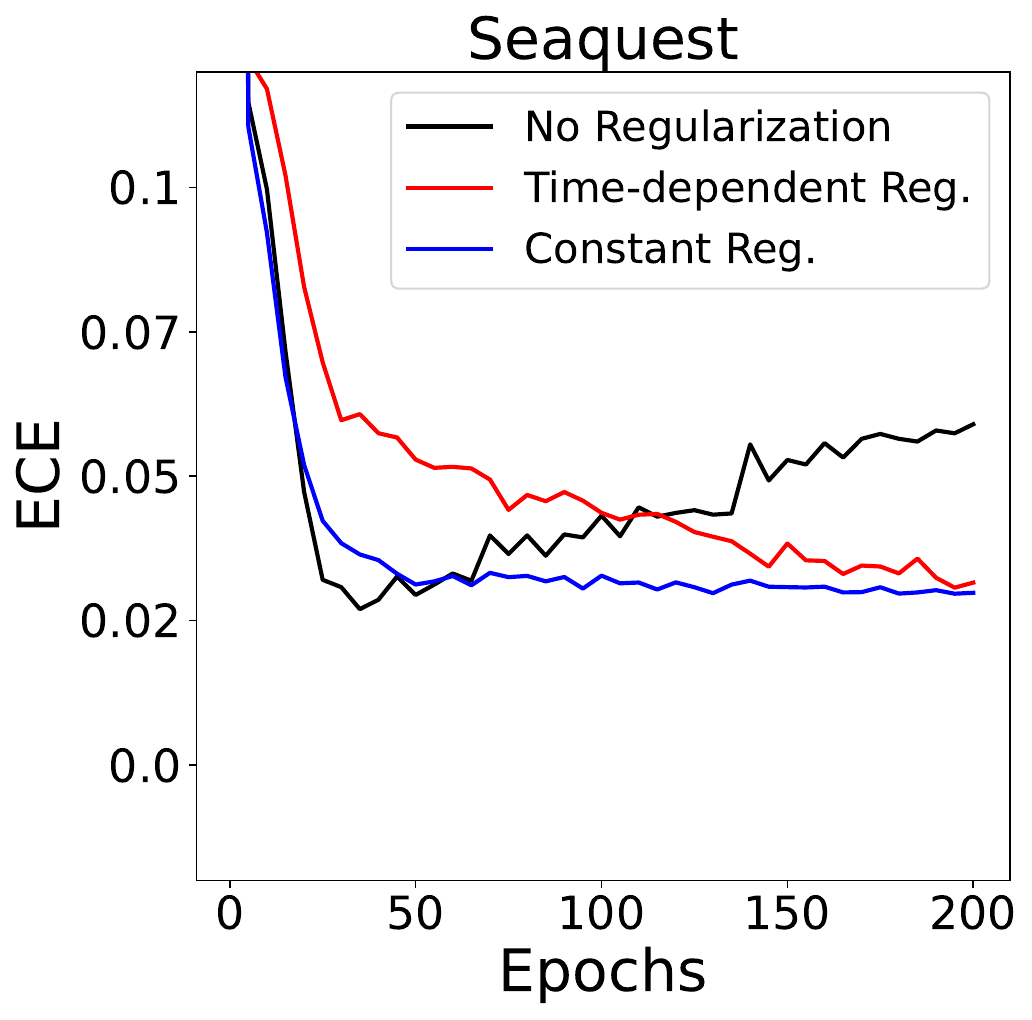}
    \includegraphics[height=0.2\linewidth]{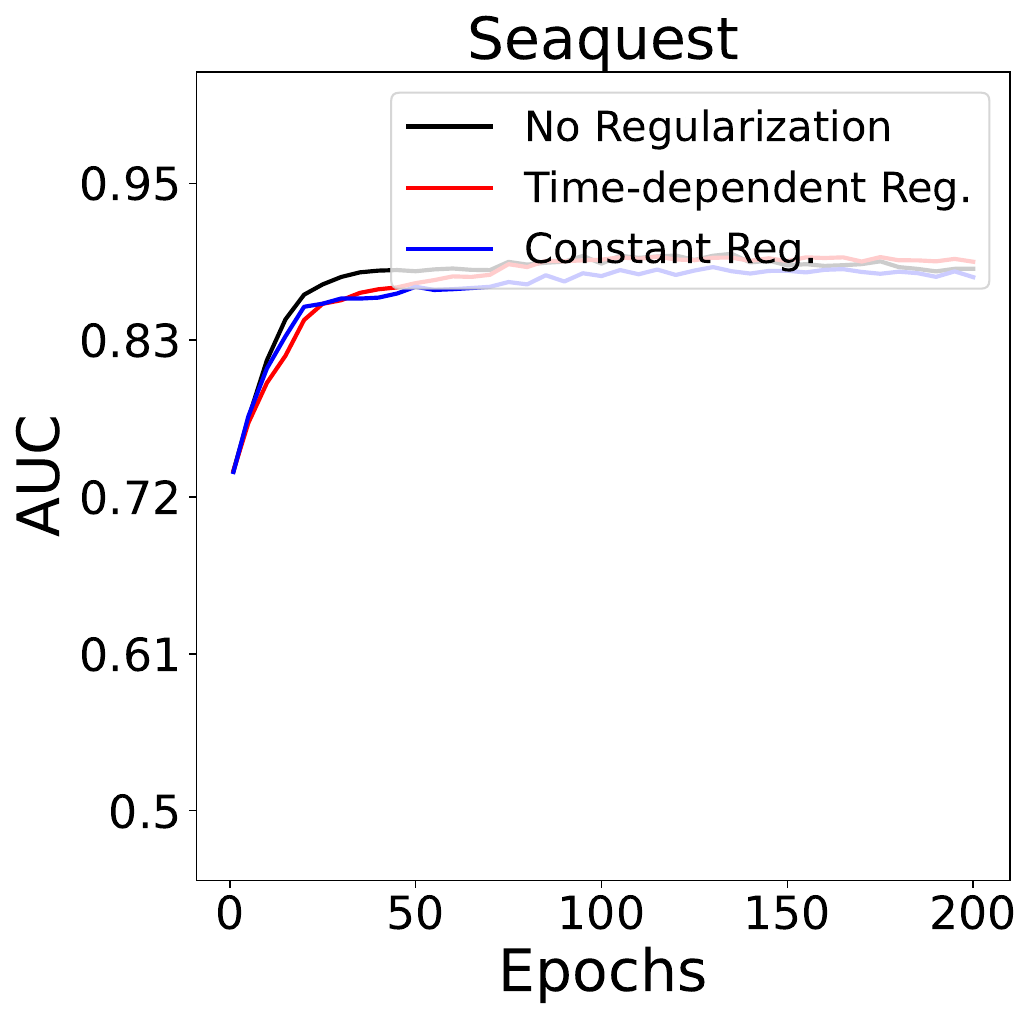}
    \includegraphics[height=0.2\linewidth]{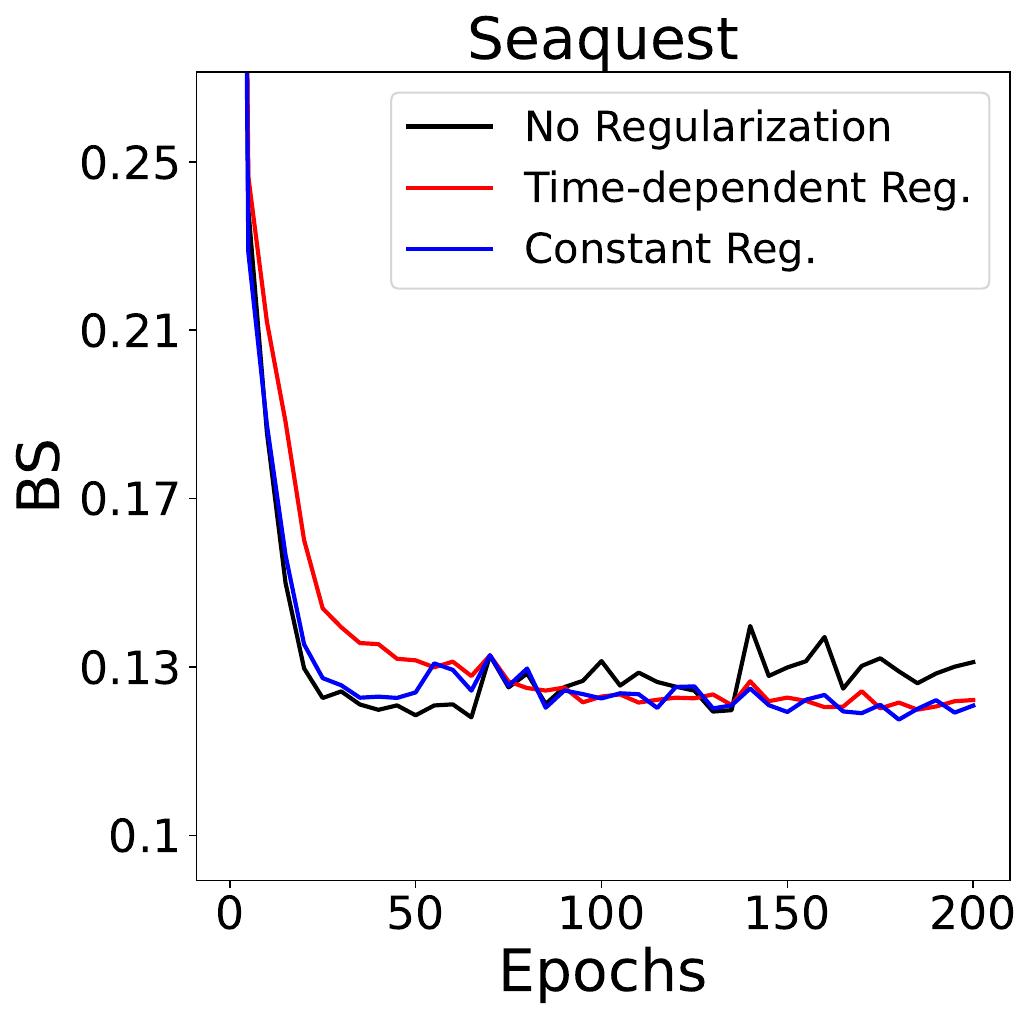}
    \includegraphics[height=0.2\linewidth]{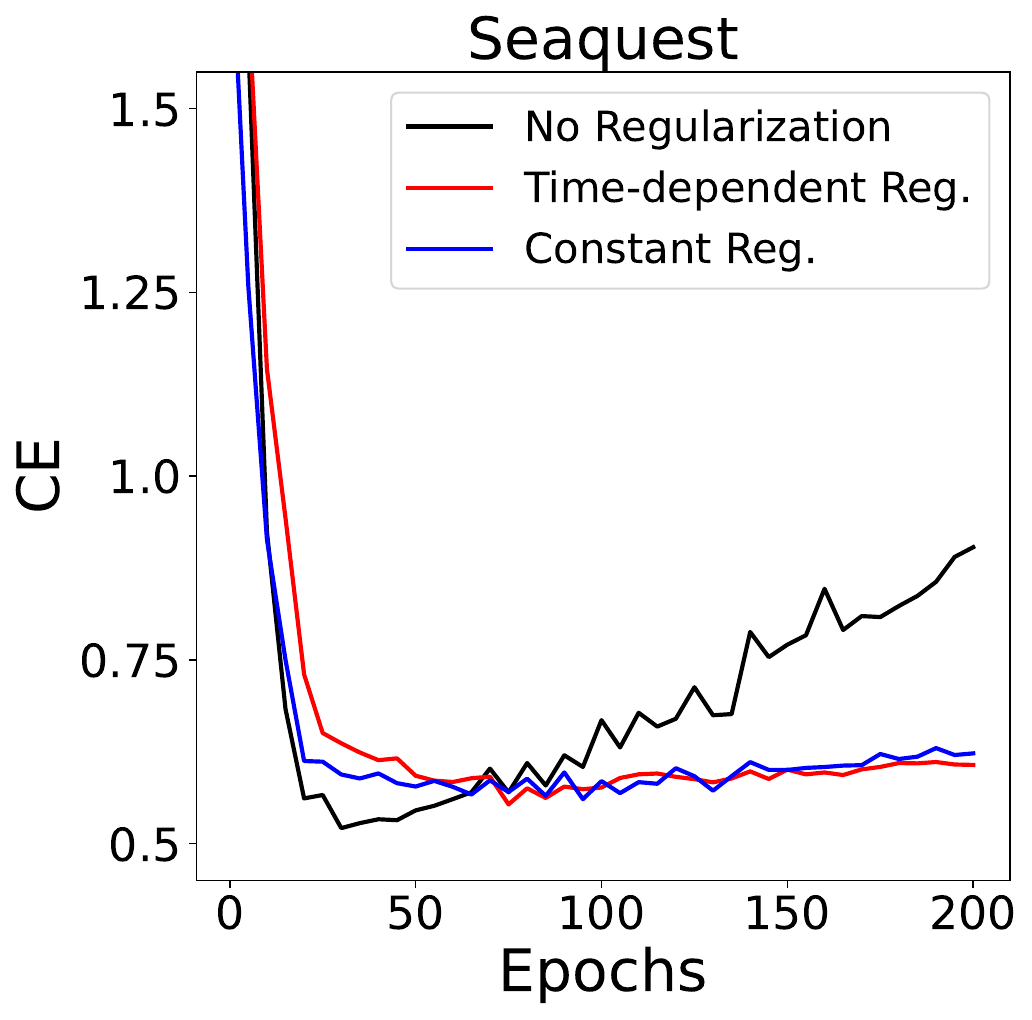} \\

    \includegraphics[height=0.2\linewidth]{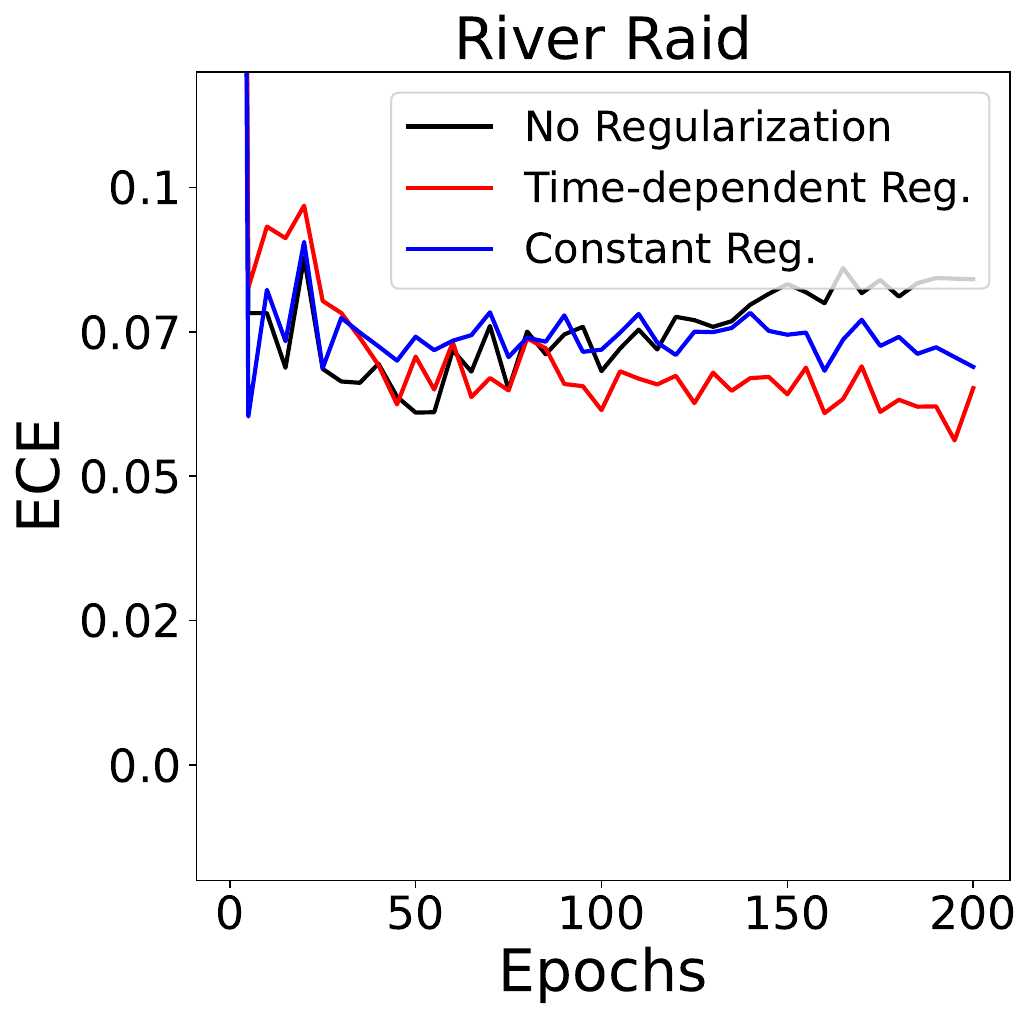}
    \includegraphics[height=0.2\linewidth]{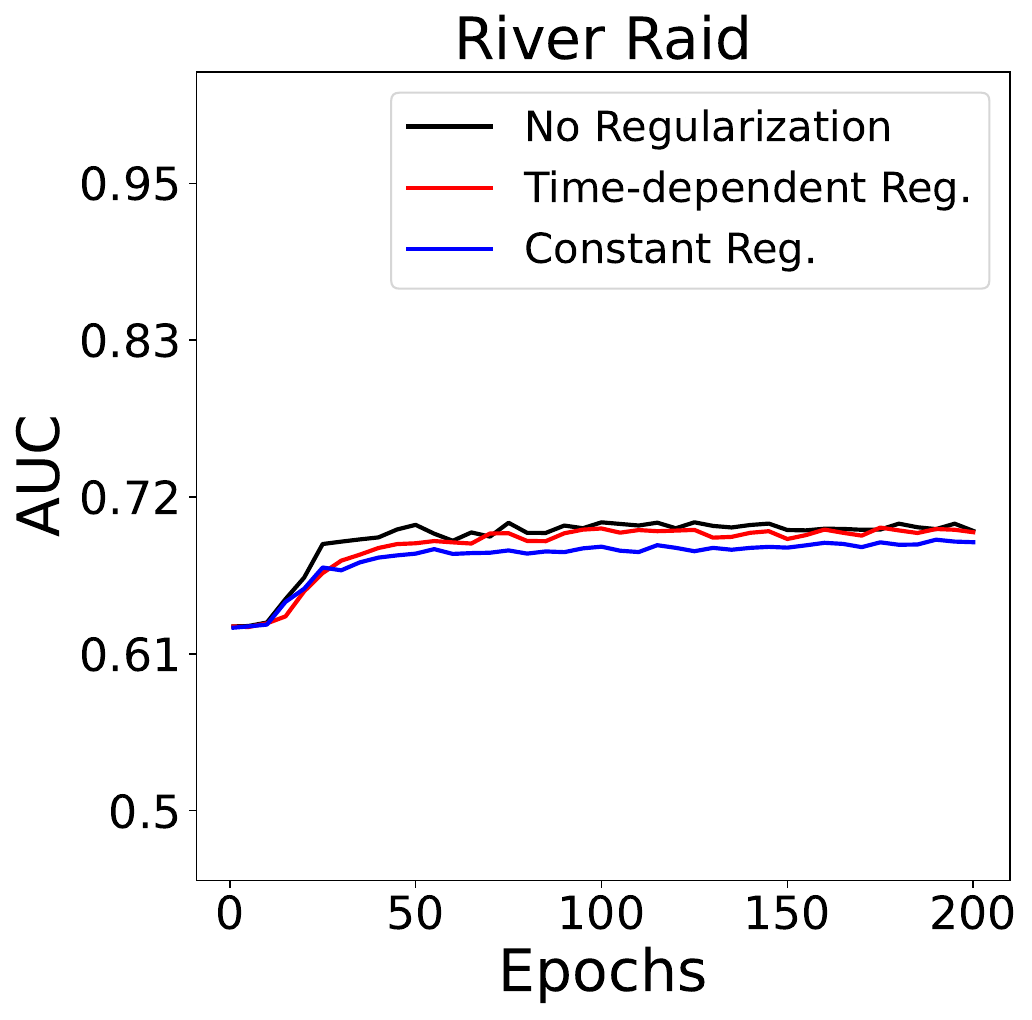}
    \includegraphics[height=0.2\linewidth]{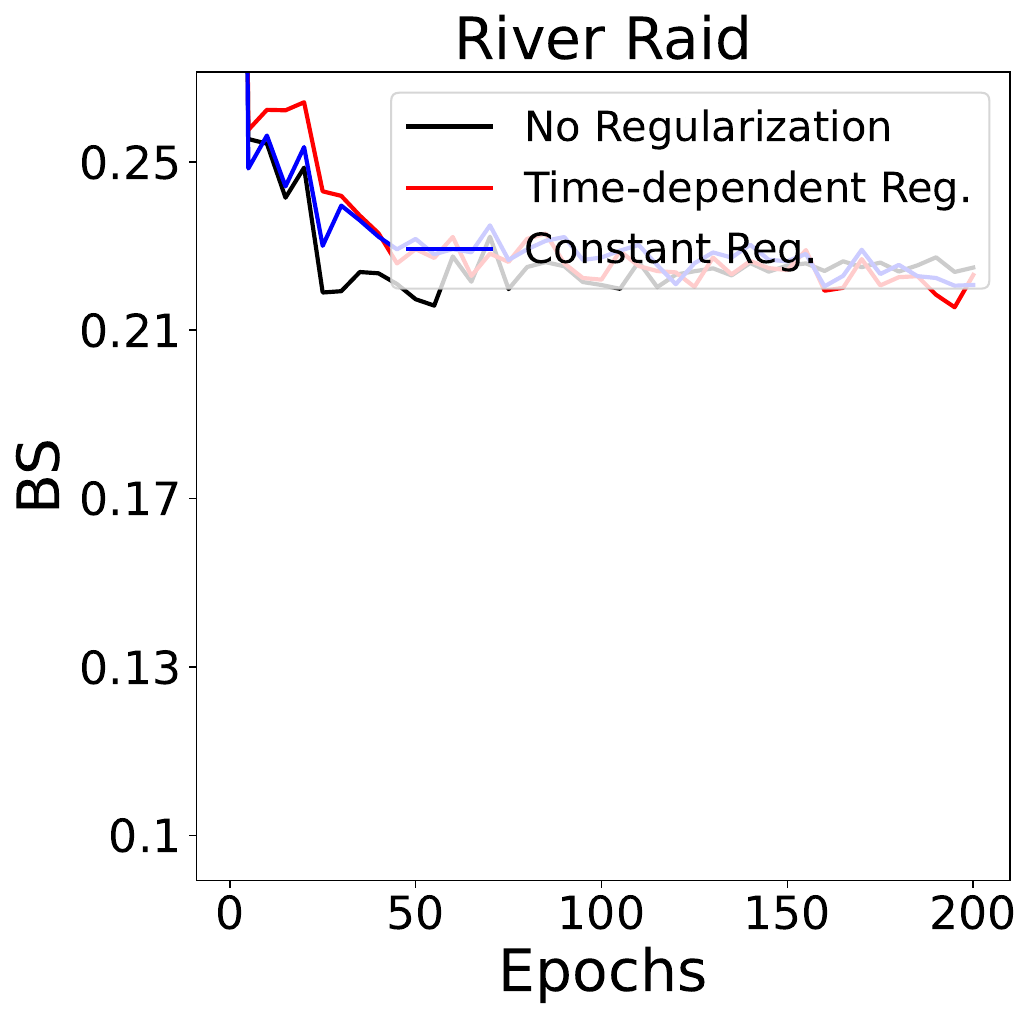}
    \includegraphics[height=0.2\linewidth]{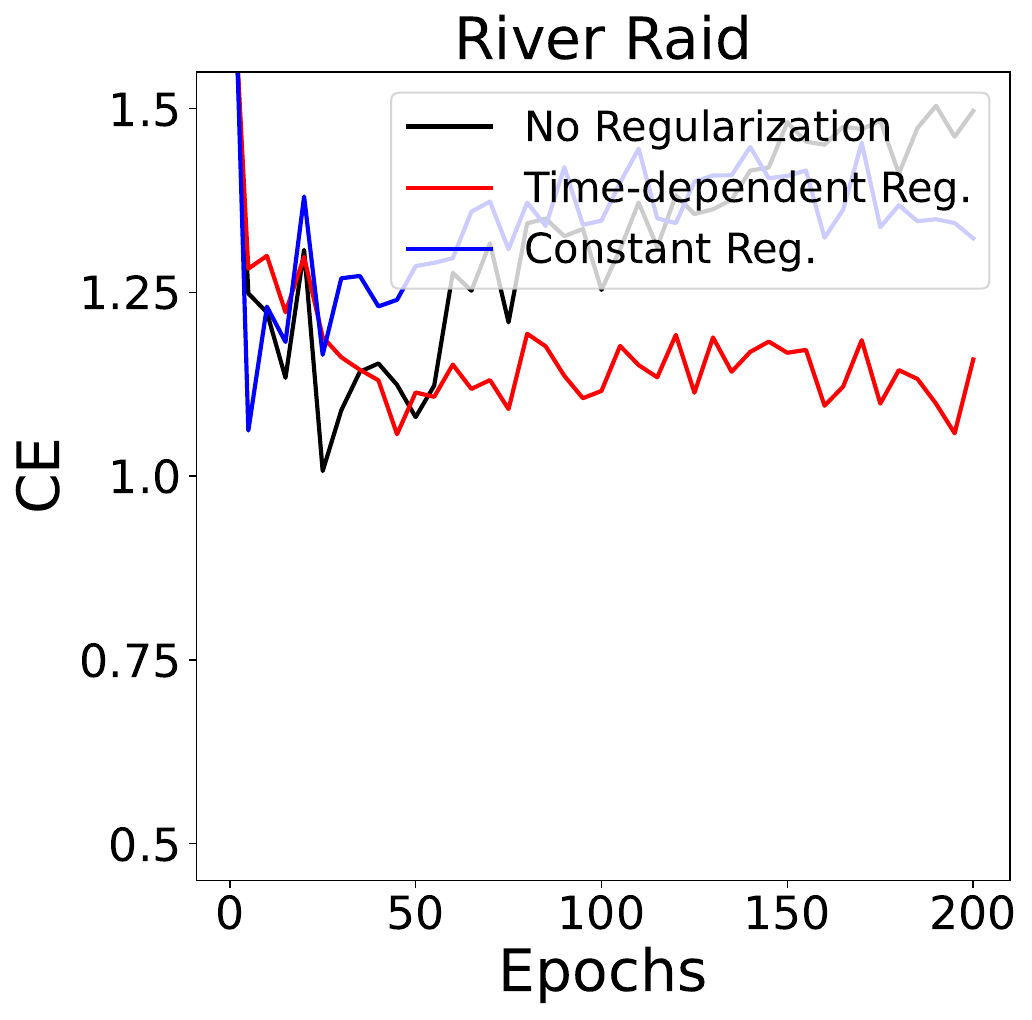} \\

    \includegraphics[height=0.2\linewidth]{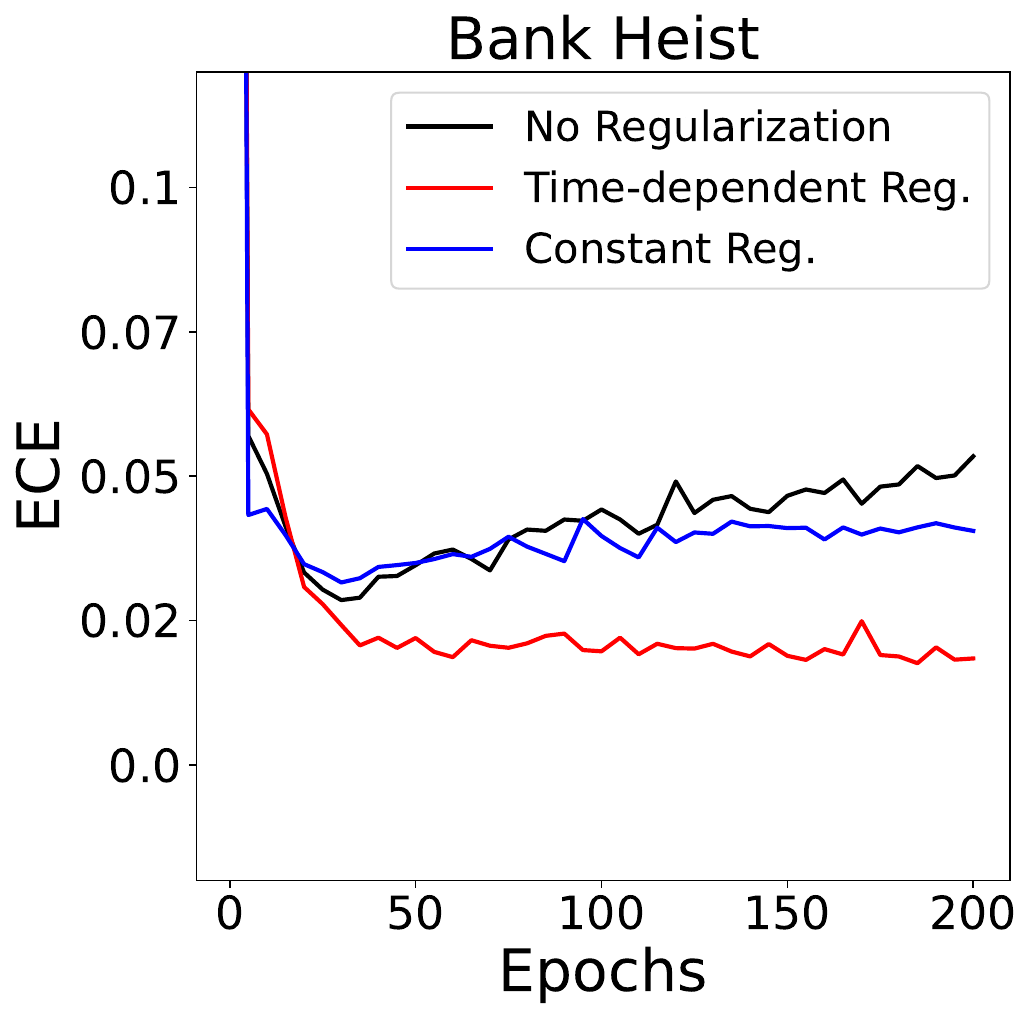}
    \includegraphics[height=0.2\linewidth]{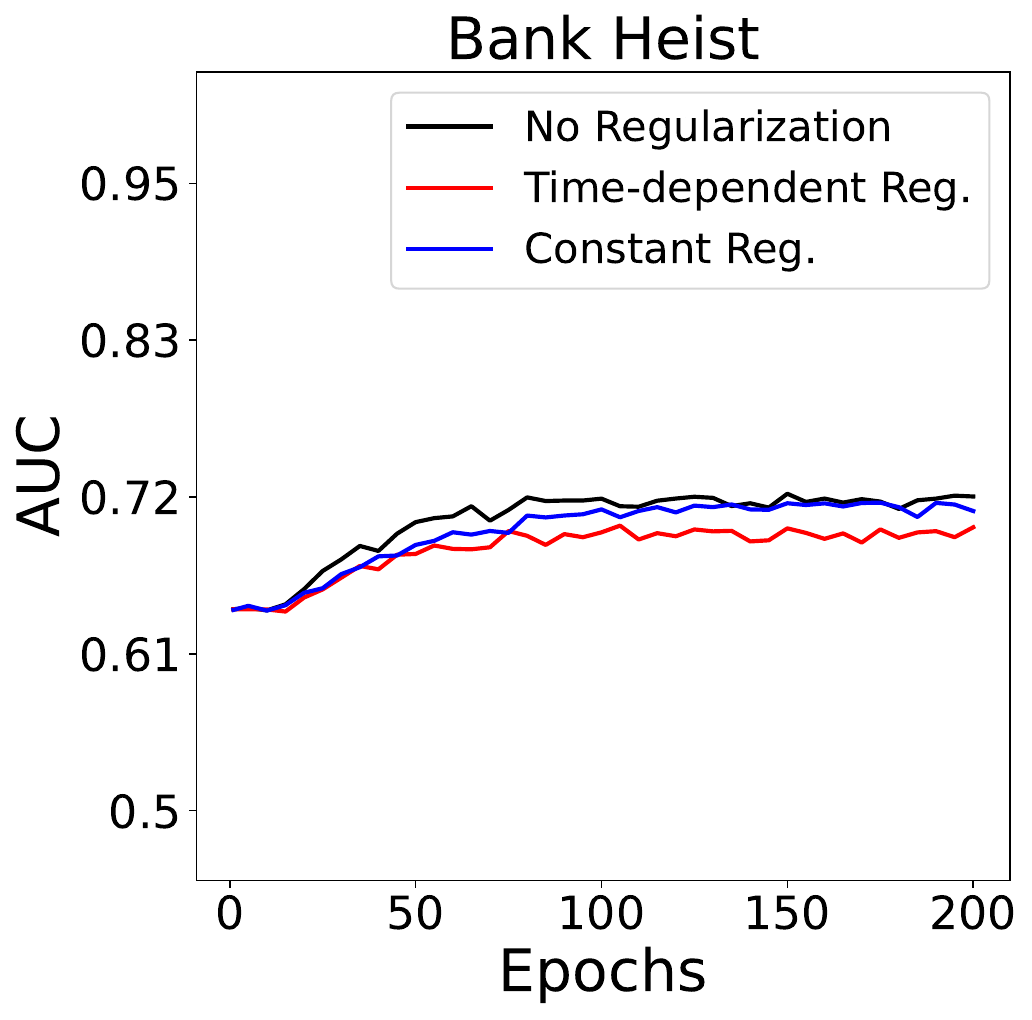}
    \includegraphics[height=0.2\linewidth]{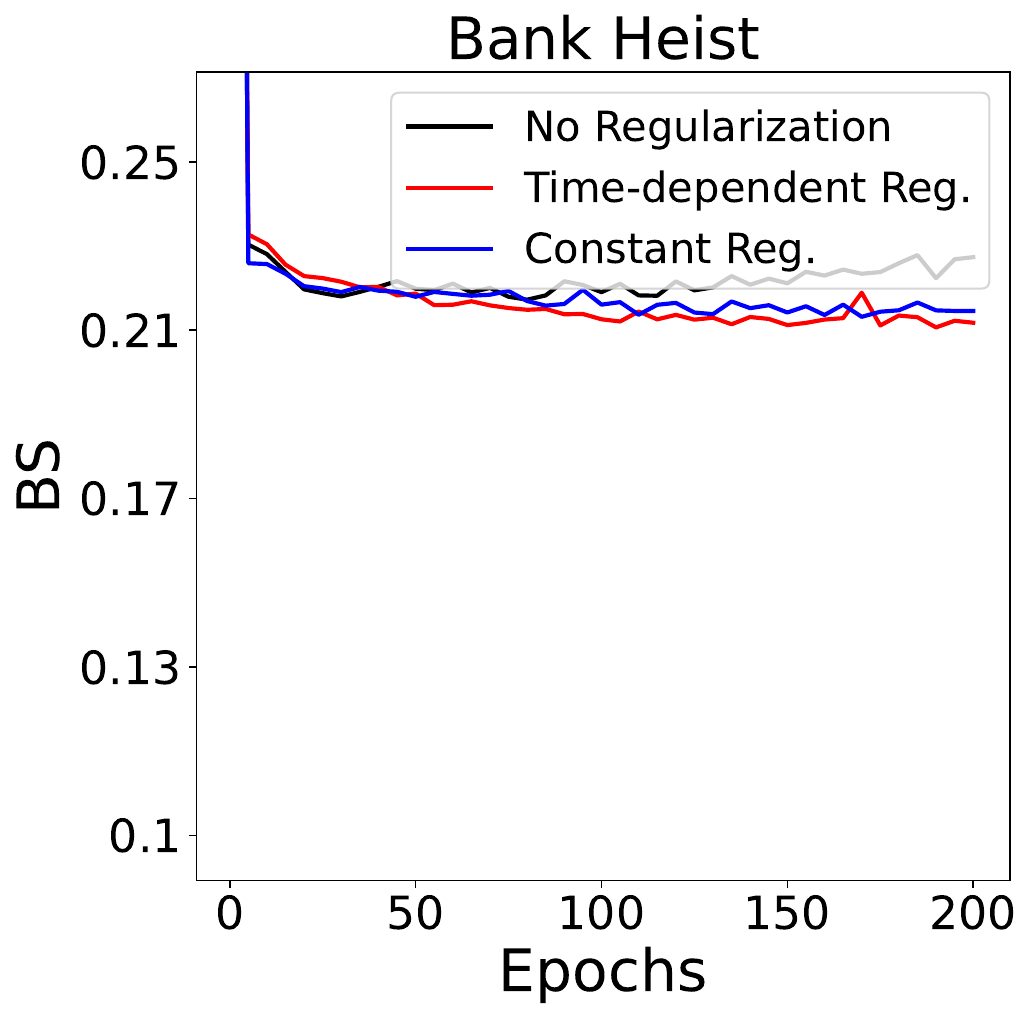}
    \includegraphics[height=0.2\linewidth]{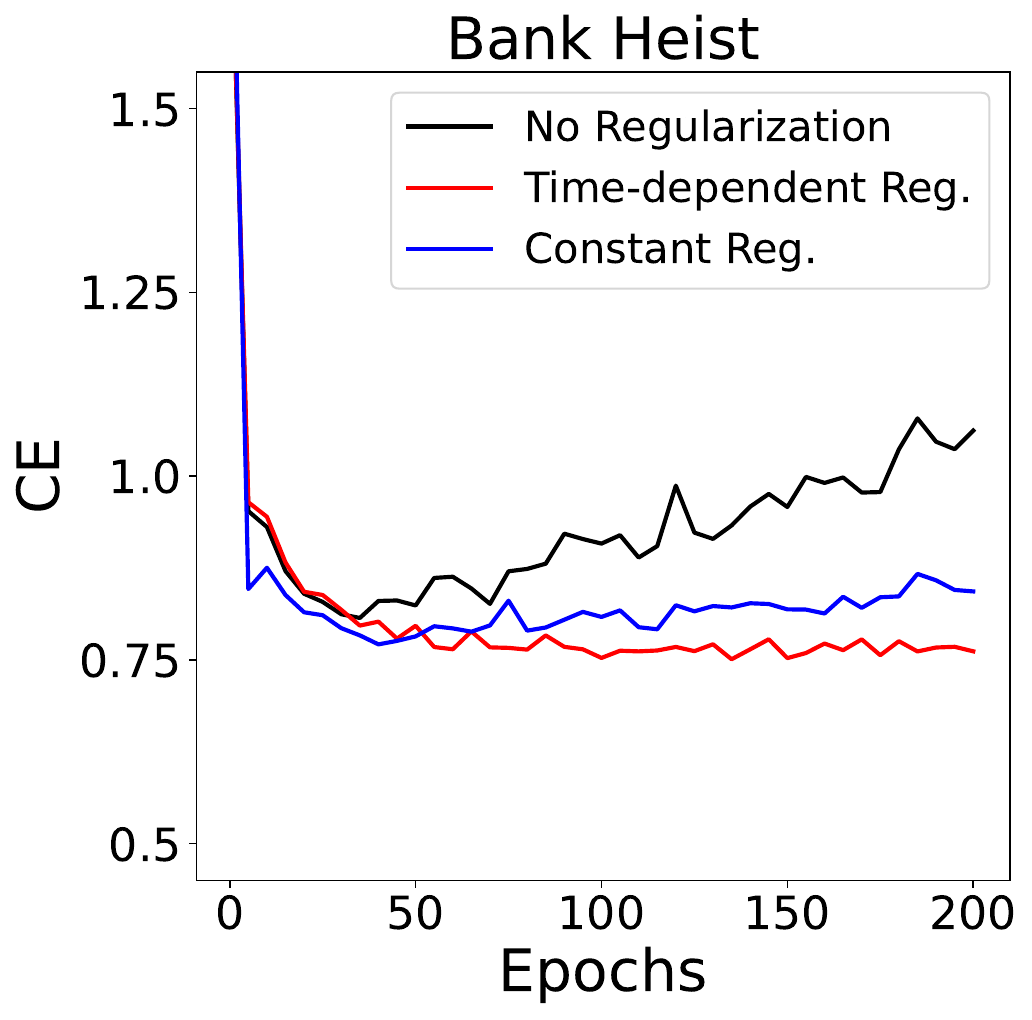} \\

    \includegraphics[height=0.2\linewidth]{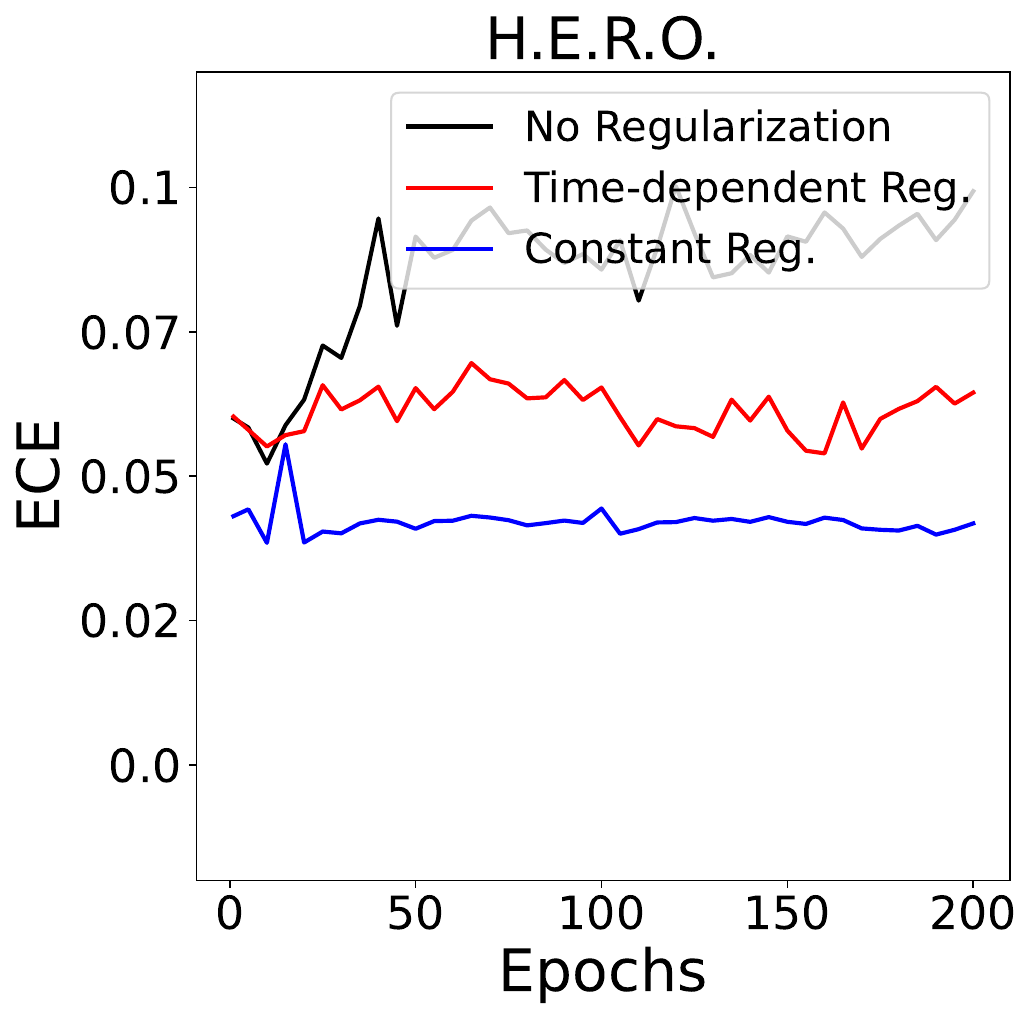}
    \includegraphics[height=0.2\linewidth]{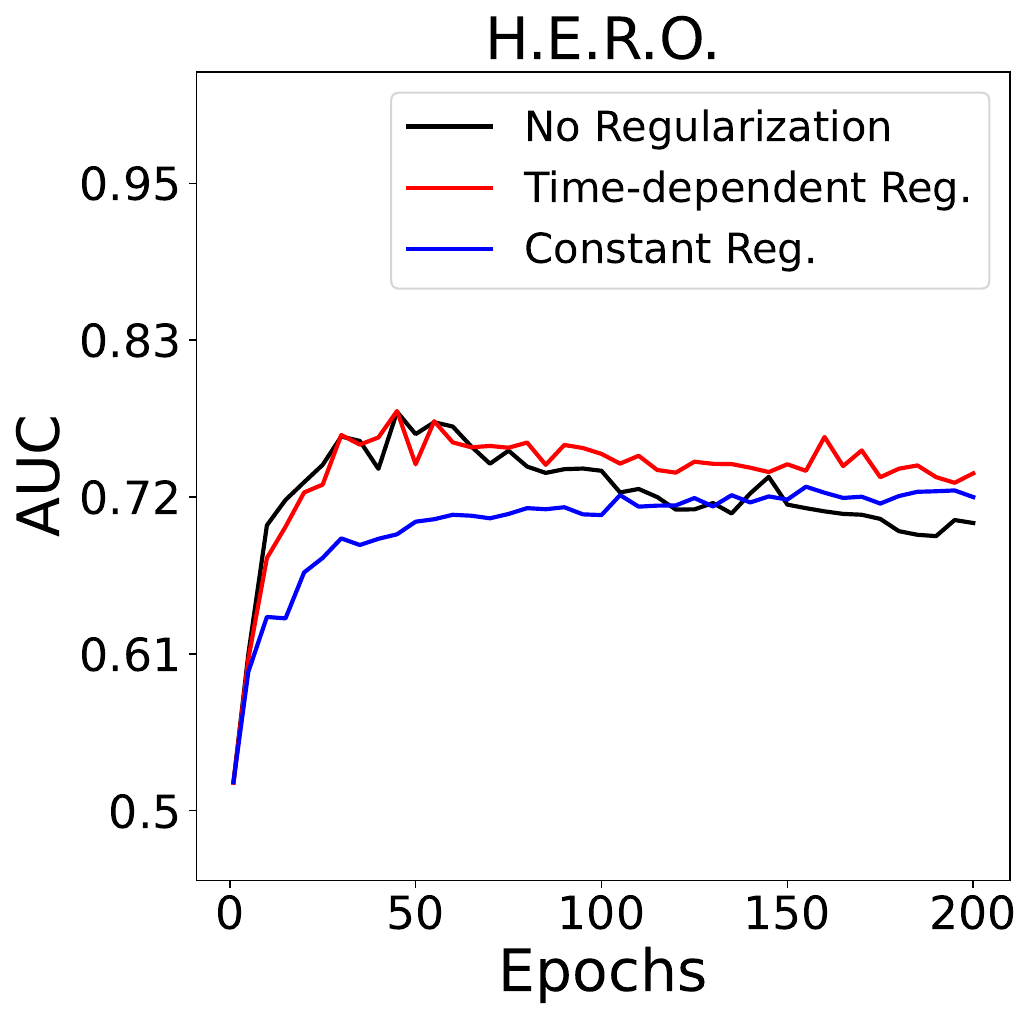}
    \includegraphics[height=0.2\linewidth]{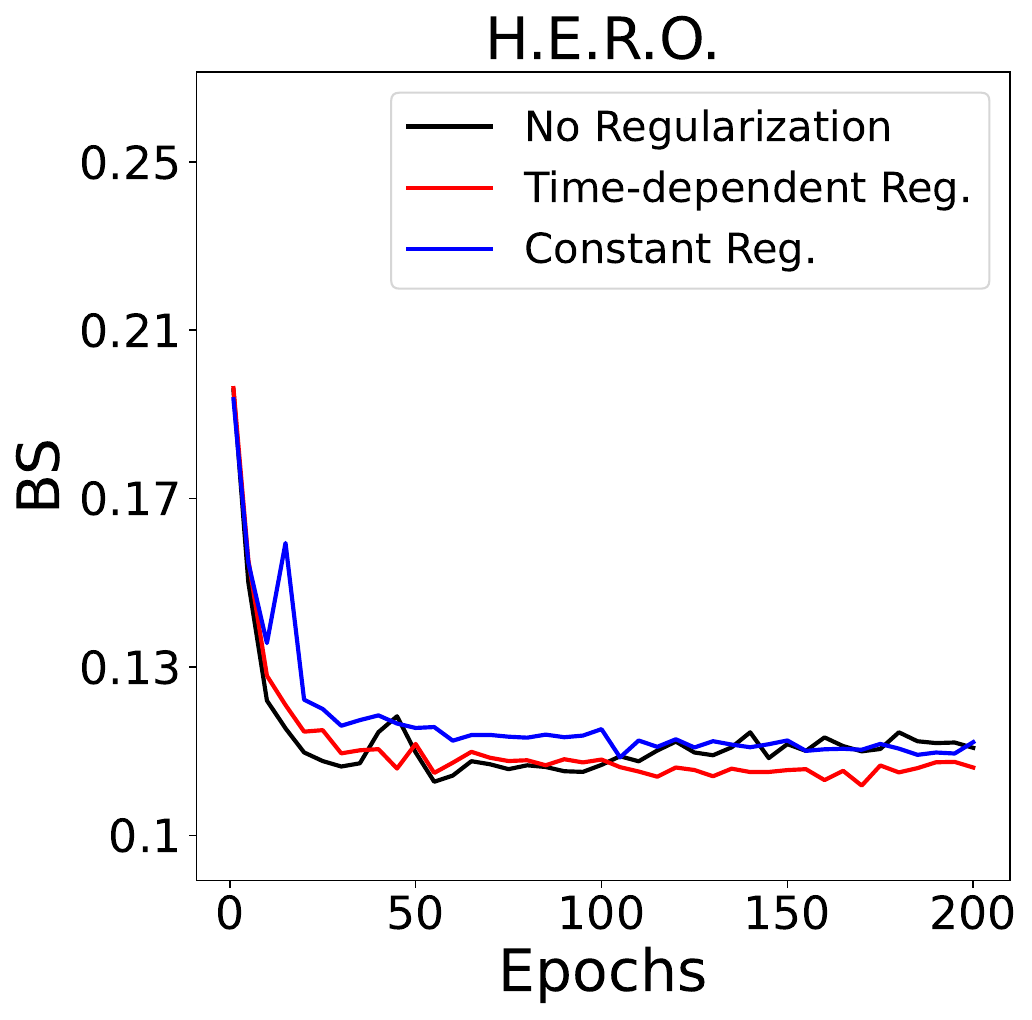}
    \includegraphics[height=0.2\linewidth]{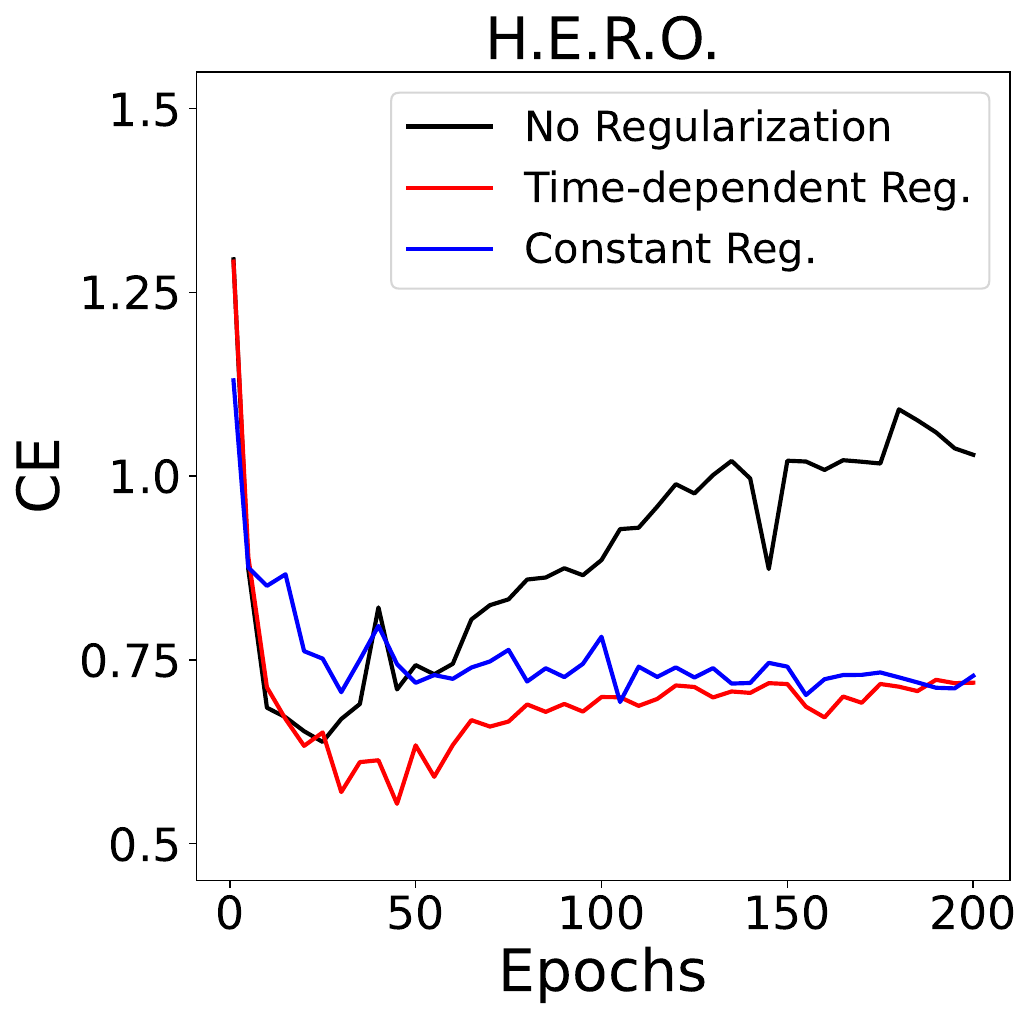} \\

    \includegraphics[height=0.2\linewidth]{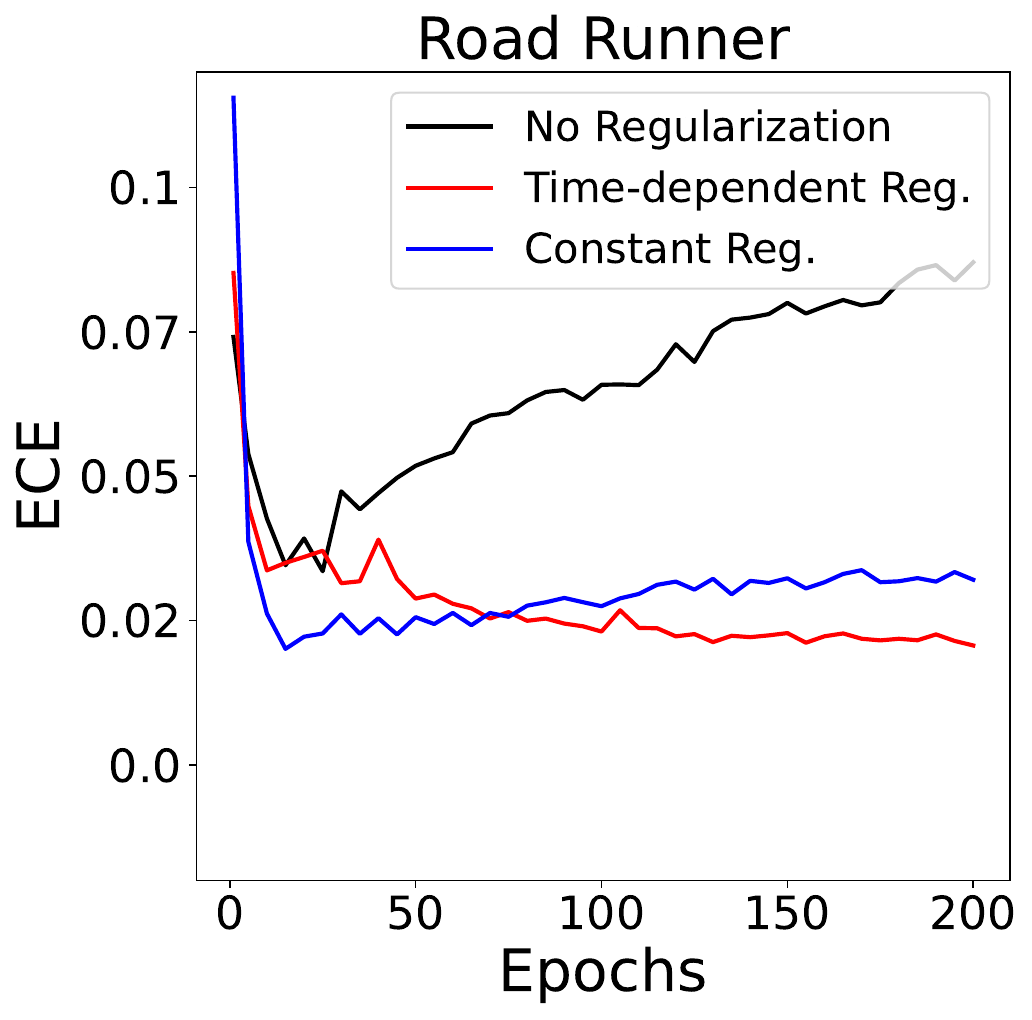}
    \includegraphics[height=0.2\linewidth]{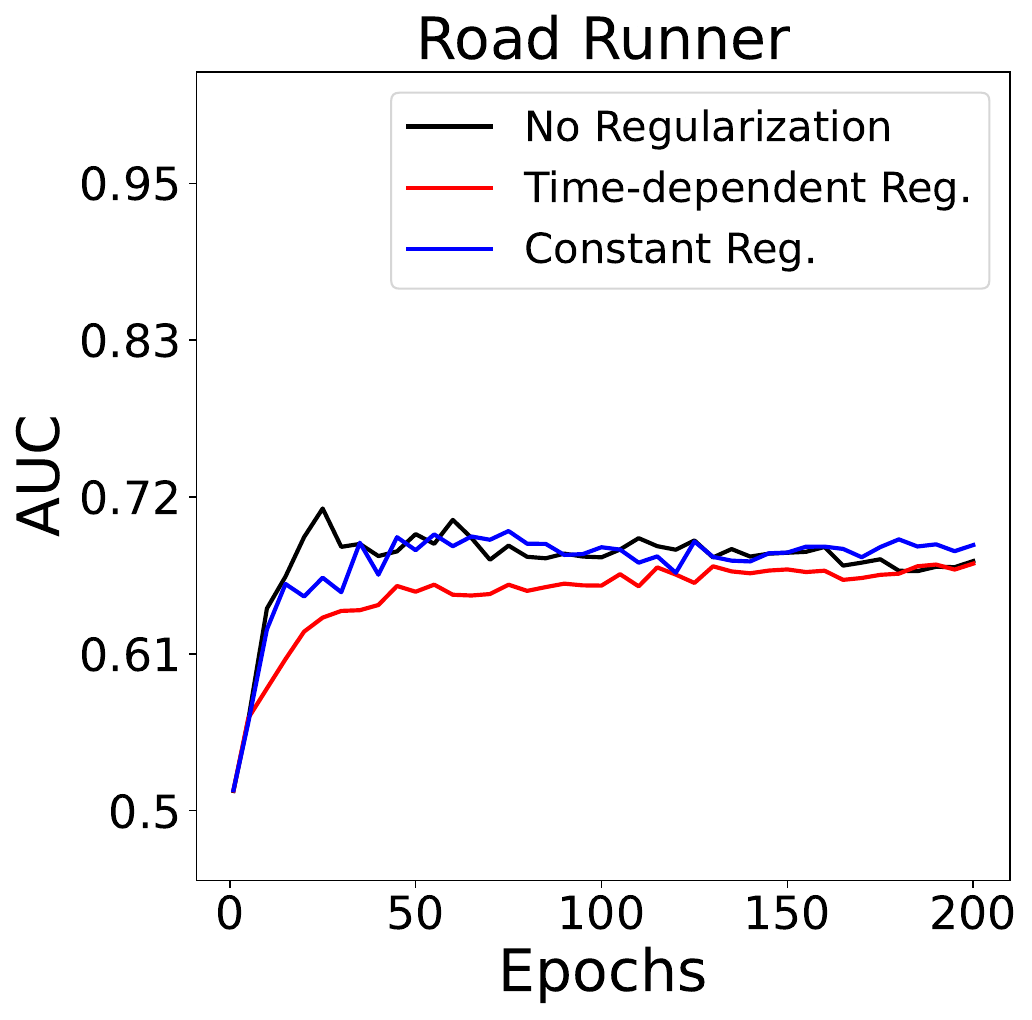}
    \includegraphics[height=0.2\linewidth]{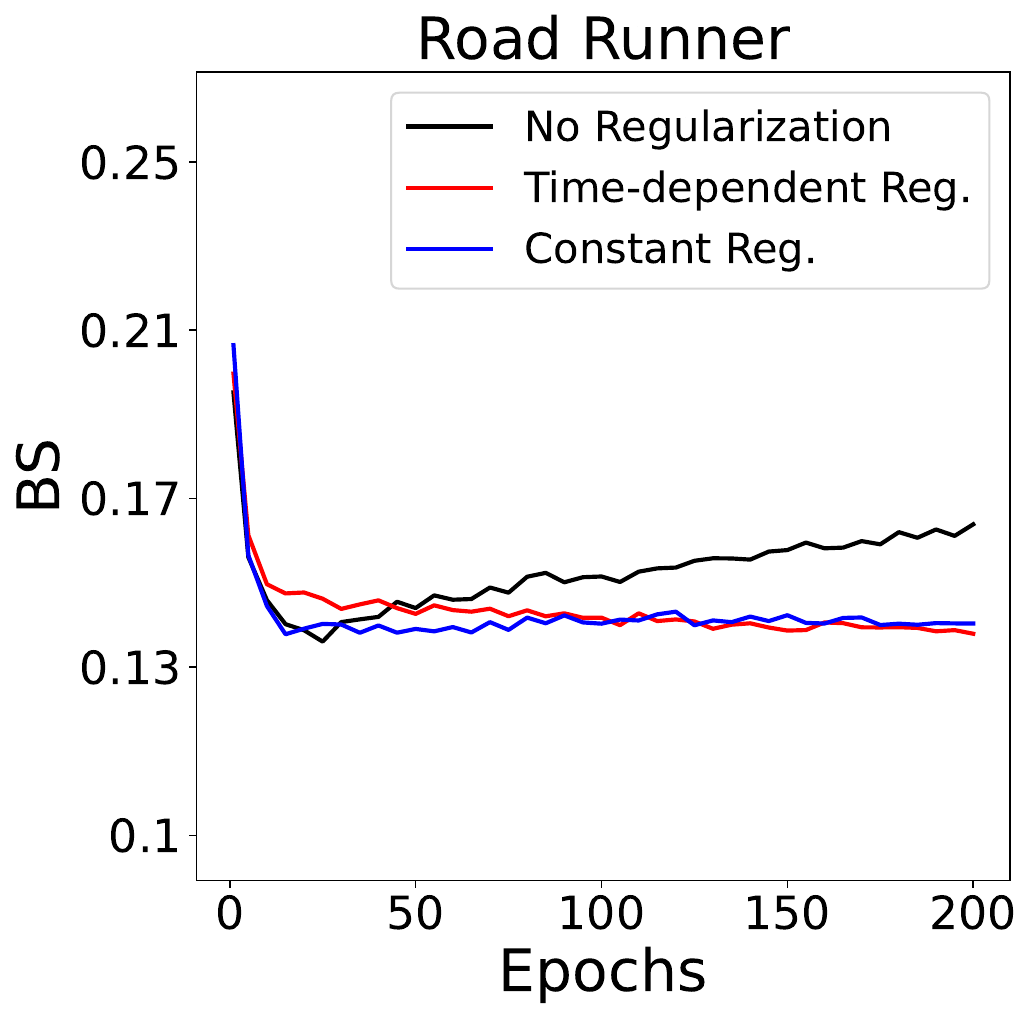}
    \includegraphics[height=0.2\linewidth]{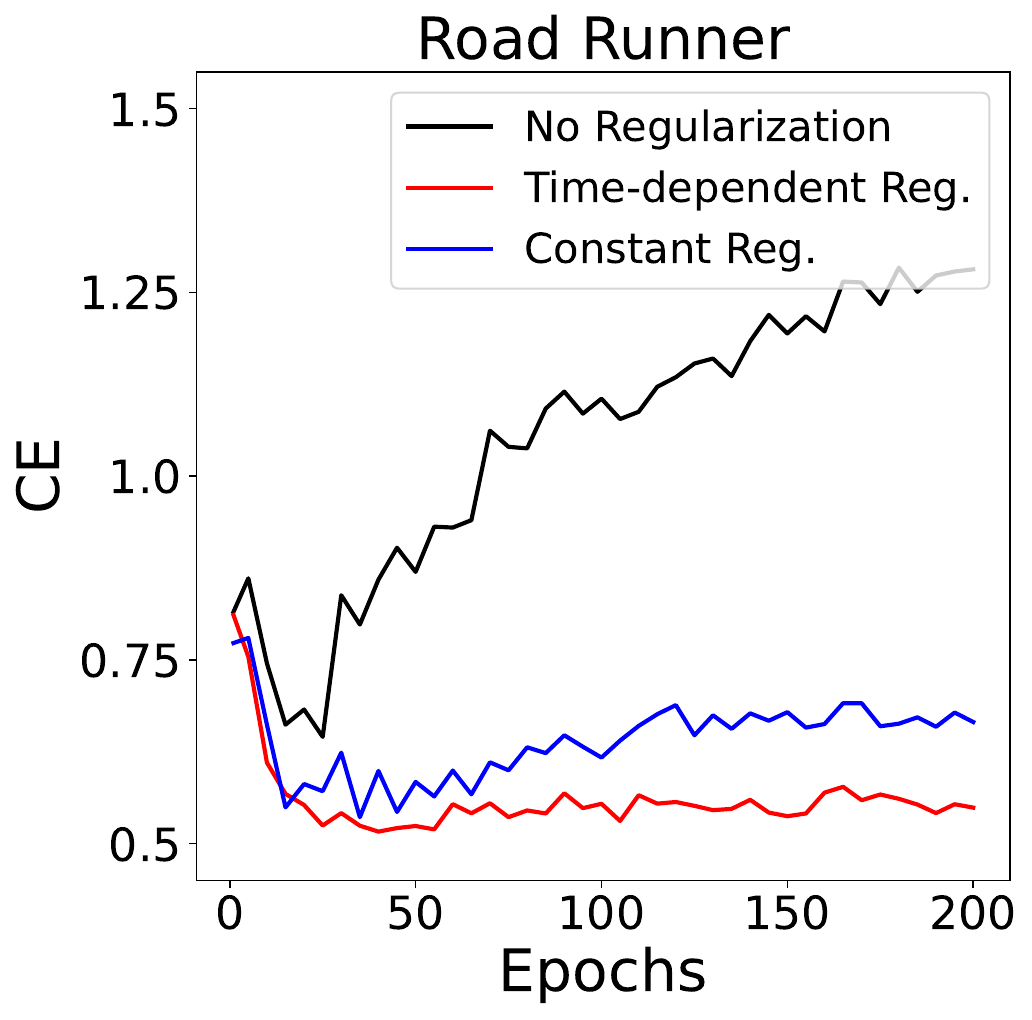} \\

    \includegraphics[height=0.2\linewidth]{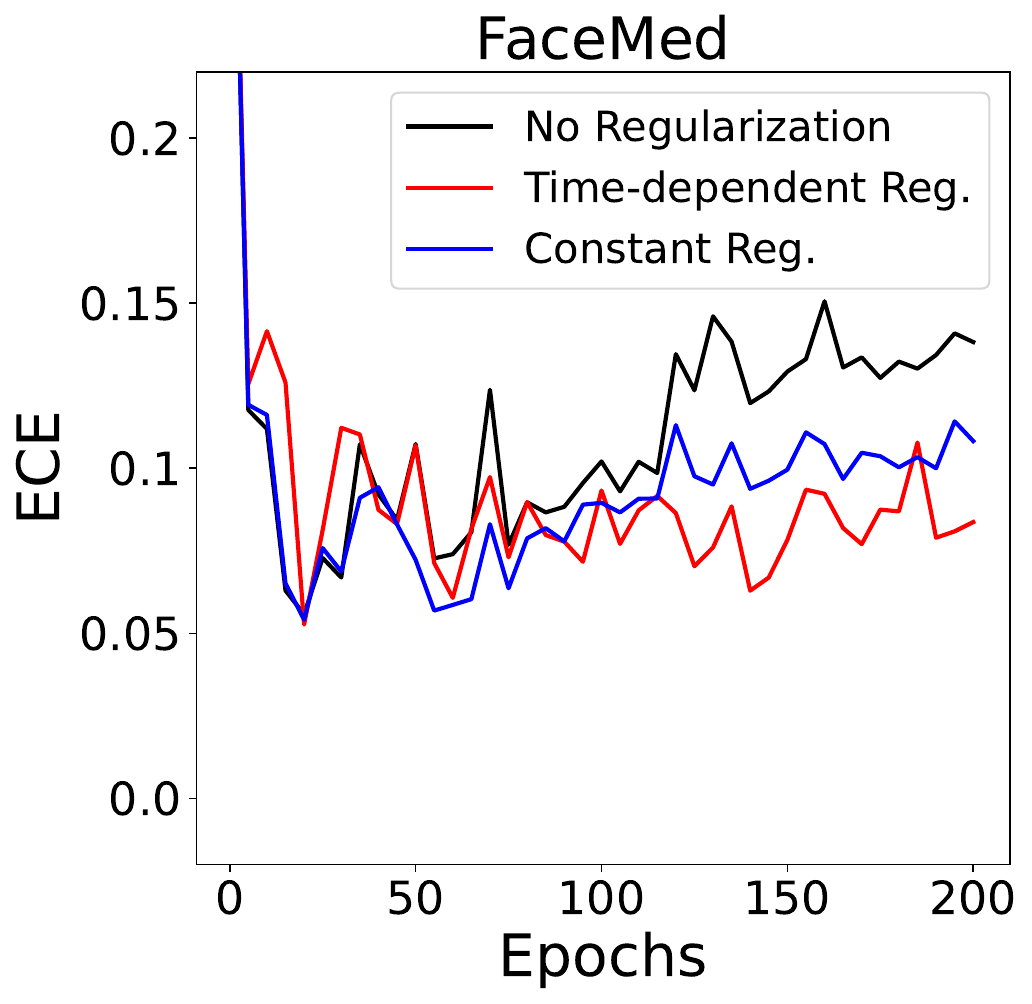}
    \includegraphics[height=0.2\linewidth]{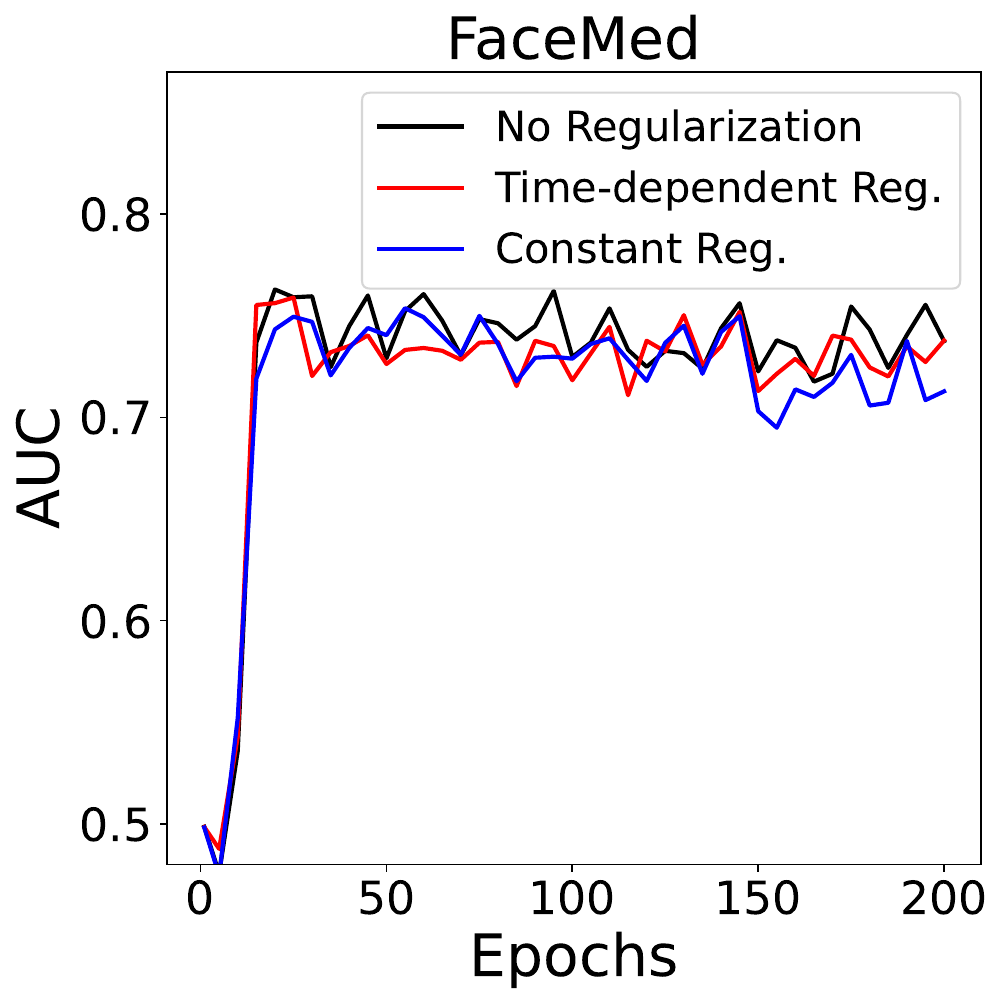}
    \includegraphics[height=0.2\linewidth]{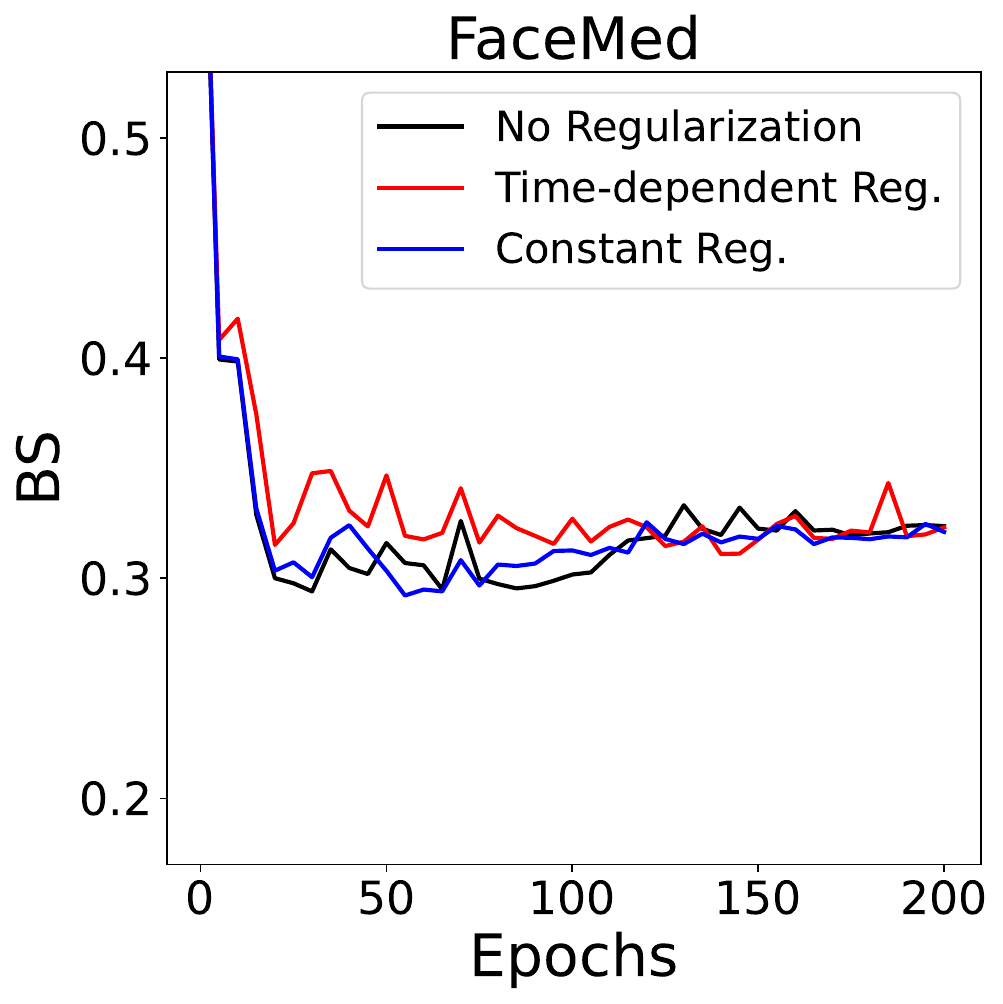}
    \includegraphics[height=0.2\linewidth]{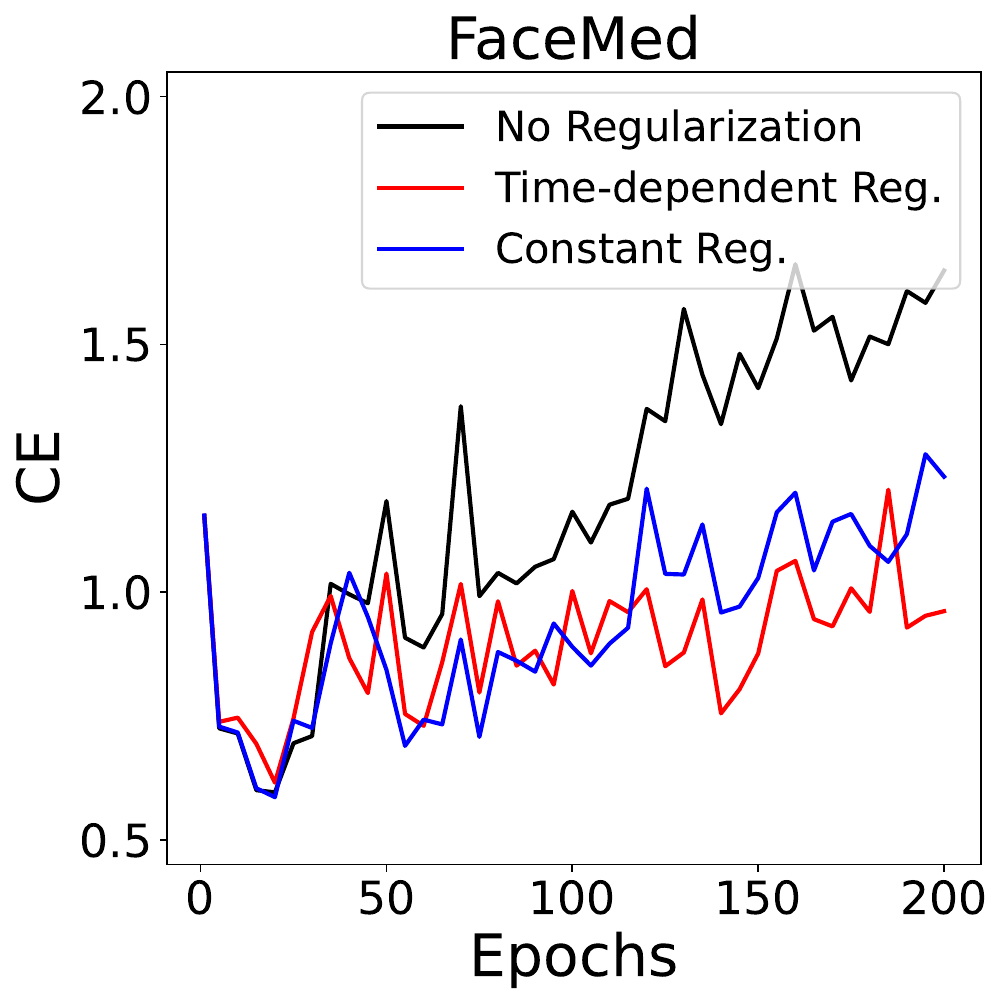}
\caption{\textbf{Learning curves of sequence-level metrics for conditional probability estimation.} A similar pattern in \Cref{fig:step_average_score_uncondition_game} is observed for conditional probability estimation from how ECE, AUC, BS, and CE evolve along training epochs for three versions of foCus.}
\label{fig:step_average_score_condition_game}
\end{figure*}

This section displays the learning curves for various sequence-level metrics on the test set at different epochs during training. All models were trained for 200 epochs, and the metrics were calculated separately for time-to-event prediction confidence intervals \Cref{fig:length_score_game}, marginal probability estimation \Cref{fig:step_average_score_uncondition_game}, and conditional probability estimation \Cref{fig:step_average_score_condition_game}. These curves highlight the trade-offs between discriminability and calibration over the course of training: lower relative MAE is associated with lower coverage probabilities, and higher AUC tends to come with higher ECE. Despite such trade-offs, the figures demonstrate that among the three foCus variants, the model with time-dependent regularization strikes the best balance, maintaining calibration while improving discriminability throughout training.

\end{appendix}